\documentclass[runningheads]{llncs}

\usepackage[mobile]{eccv}

\usepackage{eccvabbrv}
\usepackage{colortbl}
\usepackage{xcolor}
\definecolor{methodgray}{RGB}{245,247,250} 
\usepackage{graphicx}
\usepackage{booktabs}
\usepackage{rotating}
\usepackage[ruled,linesnumbered]{algorithm2e}
\PassOptionsToPackage{margin=0.5in}{geometry}
\usepackage[normalem]{ulem}
\usepackage{booktabs}
\PassOptionsToPackage{margin=0.5in}{geometry}

\usepackage[export]{adjustbox} % Required for the 'ClipBox' feature
\usepackage{pgffor} % Required for loops
\usepackage[percent]{overpic}
\usepackage{pgffor}
\usepackage{etoolbox} %

\usepackage[accsupp]{axessibility}  % Improves PDF readability for those with disabilities.

\usepackage{hyperref}

\usepackage{orcidlink}
\usepackage{multirow}
\usepackage{graphicx}
\usepackage{array}
\usepackage{tikz}
\usetikzlibrary{calc}

\usepackage{wrapfig}

\begin{document}

\def\method{SEM-ROVER}
\title{\method: Semantic Voxel-Guided Diffusion for Large-Scale Driving Scene Generation} 

\titlerunning{SEM-ROVER}
\authorrunning{H.~Dahmani et al.}
\author{Hiba Dahmani\orcidlink{0009-0008-0426-9919}\inst{1,2} \and
Nathan Piasco\inst{1}\orcidlink{0000-0001-7952-6643} \and 
Moussab Bennehar\inst{1}\orcidlink{0000-0002-6566-6132} \and
\\ Luis Rold{\~a}o\inst{1}\orcidlink{0000-0003-0482-3584} \and
Dzmitry Tsishkou\inst{1}\orcidlink{0009-0002-9798-3316} \and
Laurent Caraffa\inst{3}\orcidlink{0000-0002-8676-8058} \and
\\Jean-Philippe Tarel\inst{2}\orcidlink{0000-0002-9241-5347} \and
Roland Br{\'e}mond\inst{2}\orcidlink{0000-0003-3150-7624}
}

\institute{Noah's Ark, Huawei Paris Research Center, France \and
COSYS, Gustave Eiffel University, France \and
LASTIG, IGN-ENSG, Gustave Eiffel University, France}

\maketitle
\begin{figure}[htb]

\centering
\includegraphics[width=.95\linewidth,height=.3\textheight,keepaspectratio]{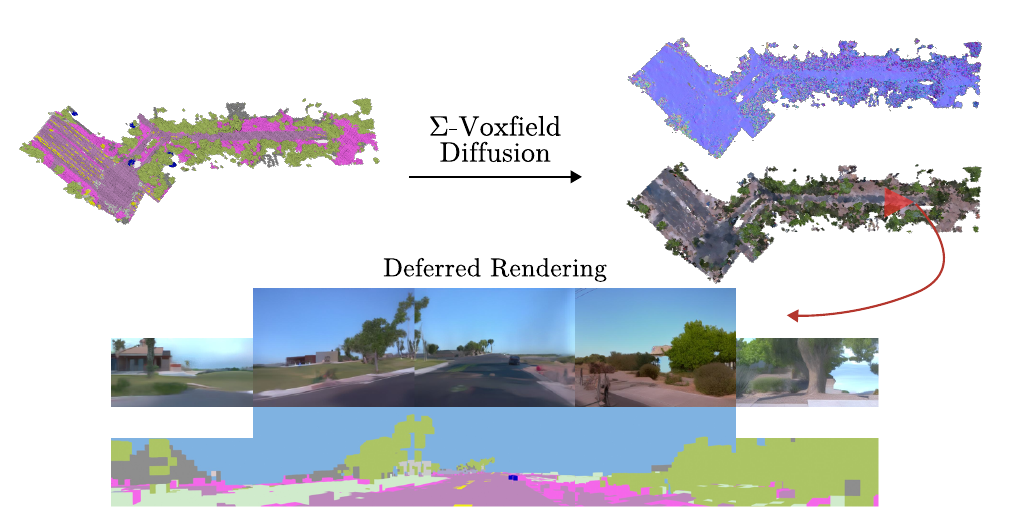}
\caption{Our model generates large-scale 3D driving scenes given a coarse semantic voxel grid. Here we show a generated large-scale driving scene spanning $\approx$100 000 m2}
\label{fig:teaser}
\end{figure}

\begin{abstract}
Scalable generation of outdoor driving scenes requires 3D representations that remain consistent across multiple viewpoints and scale to large areas. Existing solutions either rely on image or video generative models distilled to 3D space, harming the geometric coherence and restricting the rendering to training views, or are limited to small-scale 3D scene or object-centric generation. In this work, we propose a 3D generative framework based on $\Sigma$-Voxfield grid, a discrete representation where each occupied voxel stores a fixed number of colorized surface samples. To generate this representation, we train a semantic-conditioned diffusion model that operates on local voxel neighborhoods and uses 3D positional encodings to capture spatial structure. We scale to large scenes via progressive spatial outpainting over overlapping regions. Finally, we render the generated $\Sigma$-Voxfield grid with a deferred rendering module to obtain photorealistic images, enabling large-scale multiview-consistent 3D scene generation without per-scene optimization. Extensive experiments show that our approach can generate diverse large-scale urban outdoor scenes, renderable into photorealistic images with various sensor configurations and camera trajectories while maintaining moderate computation cost compared to existing approaches.
\end{abstract}
\section{Introduction}
\label{sec:intro}

Generation of 3D outdoor driving scenes is central to simulation, data synthesis, and controllable scene editing. These applications require a scene representation that remain consistent across viewpoints, scale to large spatial extents, and enable free rendering rather than fixed views tied to a predefined camera trajectory.

Existing approaches only partially satisfy these requirements. Occupancy- or layout-based methods capture coarse structures but often miss surface details and realistic appearance. 3D convolutional backbones scale poorly to large scenes because they require processing dense (or near-dense) voxel volumes, and their compute and memory grow rapidly with 3D grid size, making large-extent, high-resolution generation impractical.

Another solution consists of distilling image or video generative models into a common 3D representation, bypassing the scalability limitation of aforementioned approaches. Image- and video-based diffusion models \cite{gen3c,magicdrive,infinicube,urbanarchitect} can generate photorealistic observations, but their outputs do not form a persistent 3D scene and, therefore, provide limited viewpoint consistency and poor editability. Large-scale world models~\cite{cosmos} improve coverage, yet commonly rely on implicit or highly compressed representations that are difficult to render efficiently and to manipulate in a structured way. Consequently, jointly achieving 3D consistency, scalability, and photorealistic rendering in one framework remains challenging.

We address this problem by generating scenes \emph{directly in 3D} using a discrete surface representation, the \emph{$\Sigma$-Voxfield grid}. Each occupied voxel stores a fixed number of colorized surface samples, yielding discrete tokens that jointly encode local geometry and appearance while remaining aligned in world coordinates. To generate this representation, we train a \emph{semantic-conditioned diffusion model} that operates on spatially localized neighborhoods of $\Sigma$-Voxfield tokens and uses 3D positional encoding to capture spatial structure. Since the model is applied to local neighborhoods, computation stays bounded. To synthesize large scenes, we progressively expand the grid via \emph{spatial outpainting} over overlapping regions, enforcing continuity across neighborhood boundaries.

Finally, the generated $\Sigma$-Voxfield grid provides a persistent 3D scene buffer that can be rendered from arbitrary viewpoints without per-scene optimization. We render this buffer efficiently and produce photorealistic images using a deferred rendering module conditioned on the rendered $\Sigma$-Voxfield output, which compensates for surface discretization and missing content such as sky and distant background. Extensive experiments show that our approach scales to large scenes with moderate computation cost compare to competitors.

\noindent In summary, our contributions are as follows:
\begin{itemize}
    \item We introduce a novel 3D representation tailored for 3D generative modeling, \textbf{$\Sigma$-Voxfield}, a fixed-cardinality discrete surface approximation representing the geometric and photometric field within local voxel.
    \item To jointly generate the photometric and geometric characteristics of urban scenes, we convert $\Sigma$-Voxfield grids to unordered tokens with additional 3D positional encoding for training a semantically conditioned transformer-based diffusion model.
    % \item We propose a \textbf{semantic-conditioned Diffusion model} that generates local $\Sigma$-Voxfield neighborhoods using 3D positional encoding for coherent geometry and appearance.
    \item We scale synthesis via \textbf{progressive spatial outpainting} in $\Sigma$-Voxfield space, enabling large-scene generation while maintaining a constant computation budget.
    \item We couple our 3D generation with \textbf{deferred rendering} to obtain photorealistic images conditioned on a persistent 3D buffer, without per-scene optimization.
\end{itemize}
\section{Related Work}

\subsection{Diffusion Models for Driving Scene Generation}

Recent works in driving scene generation largely rely on diffusion models in image or video space. While methods like DreamDrive~\cite{dreamdrive} and MagicDrive~\cite{magicdrive} synthesize realistic, temporally coherent videos conditioned on text prompts, trajectories, or layouts, their outputs remain tied to specific inference trajectories and do not provide a persistent scene representation in world coordinates.

Several approaches extend this paradigm by reconstructing the underlying 3D structure from intermediate view generation. MagicDrive3D~\cite{magicdrive3d} combines multi-view video diffusion with 3D Gaussian Splatting (3DGS) reconstruction, while GEN3C~\cite{gen3c} performs video diffusion conditioned on a 3D cache and decodes latent videos into RGB frames, relying on precomputed geometry. InfiniCube~\cite{infinicube} scales this design through a pipeline that integrates voxel-level generation, video synthesis, and feed-forward 3DGS reconstruction. More recently, ScenDi~\cite{scendi} proposes a 3D-to-2D diffusion cascade where coarse latent 3D Gaussians are generated and subsequently refined through view-conditioned 2D diffusion. Despite strong visual quality, these methods typically obtain 3D structure through intermediate rendering, reconstruction, or cascaded refinement, rather than directly generating a persistent surface representation in 3D space.

Complementary work explores structured spatial priors for large-scale environments. LSD-3D~\cite{lsd3d} leverages layout and point cloud conditioning but couples generation with optimization of large Gaussian sets, leading to high memory usage and long per-scene runtimes. Similarly, Urban Architect~\cite{urbanarchitect} generates scenes from semantic layouts and reconstructs geometry via optimization of implicit fields guided by 2D diffusion model SDS distillation~\cite{sds}. Overall, these approaches tightly couple view synthesis with reconstruction or optimization, which can limit scalability, editability, and consistent novel-view rendering. In contrast, our method generates a $\Sigma$-Voxfield grid directly in 3D with a semantic-conditioned diffusion model, and scales to large scenes via voxel-space outpainting without per-scene optimization. In the~\Cref{tab:fair_memory_comparison}, we provide a comparison of major characteristics of relevant methods for large-scale driving scenes generation.

\subsection{Diffusion over Structured 3D Representations}

Extending diffusion models to 3D remains challenging due to the high dimensionality and sparsity of 3D spatial data, especially in urban outdoor scenes. Prior works explore diffusion over point clouds, voxels, and sparse grids~\cite{pointdiffusion, ren2024scube, xiong2024octfusion}, but computation often scales with spatial resolution, making large-scale scene synthesis expensive. Primitive-based formulations, such as GaussianCube~\cite{zhang2024gaussiancube} and DiffusionGS~\cite{cai2024diffusiongs}, provide explicit and renderable representations, yet are often demonstrated on object-centric or spatially bounded settings.

Latent 3D diffusion improves efficiency by operating in compressed spaces. For instance, L3DG~\cite{l3dg} models vector-quantized 3D Gaussian representations using latent diffusion with sparse convolutional encoders, enabling efficient room-scale generation. 
Relevant to our method, TRELLIS~\cite{trelis} uses a 1-D transformer to diffuse object-centric sparse 3D latent space that can be decoded into various 3D representations. Notably, the 3D conditioning is provided through positional embedding rather than using an explicit 3D operator, such as a 3D convolution block. Their method is limited to small-scale scene or object generation.

Our work differs by diffusing discrete surface tokens directly in 3D space and scaling generation through progressive outpainting, keeping per-step computation bounded while producing a persistent, renderable scene representation.

% \subsection{Large-Scale Driving Scene Reconstruction}

% \NP{I think we can remove this subseciton as we are not doing reconstruction} Reconstructing large-scale driving scenes has been dominated by implicit neural representations such as NeRF~\cite{nerf} and its large-scale variants like Block-NeRF~\cite{blocknerf}, which can model long trajectories \MB{we don't model trajectories!} but require dense observations and significant optimization. Explicit point-based methods, notably 3D Gaussian Splatting~\cite{kerbl20233dgs}, substantially improve scalability and enable real-time novel-view rendering. Extensions such as Scaffold-GS~\cite{scaffoldgs} introduce spatial organization of primitives, highlighting the importance of structured representations for robustness at scale. This emphasis on spatial decomposition also motivates our design: a world-aligned, local surface representation that supports efficient rendering and is well-suited for scalable generative modeling.

\begin{table}[t]
\centering
\tiny
\setlength{\tabcolsep}{1.0pt}
\resizebox{\columnwidth}{!}{
\begin{tabular}{l c c c c}
\toprule
Method & Prior & Pipeline & Feed Forward \\
\midrule
Urban Architect & 3D Layout & 2D Diff + NeRF & $\times$ \\
LSD-3D & PC + Boxes & 2D Diff + 3DGS  & $\times$  \\
GEN3C & Text/Image + PC & Video Diff & $\checkmark$  \\
MagicDrive3D & HDMap + Text + Traj & Video Diff. + 3DGS & $\times$ \\
InfiniCube & HDMap + Text + Traj. & Vox Diff + Video Diff + FF 3DGS & $\checkmark$ \\
\midrule
\rowcolor{methodgray}
Ours & Semantic Voxels & Vox Diff. + Deferred rendering & $\checkmark$ \\
\bottomrule
\end{tabular}}
\caption{\textbf{Driving-scene generation methods.} We review relevant works in terms of required priors and computation pipelines. We explicitly distinguish per-scene optimization from feed-forward diffusion pipelines.}
\label{tab:fair_memory_comparison}
\end{table}
\section{Method}

% \subsection{Method overview}

\begin{figure*}[t]
\centering
\includegraphics[width=\textwidth]{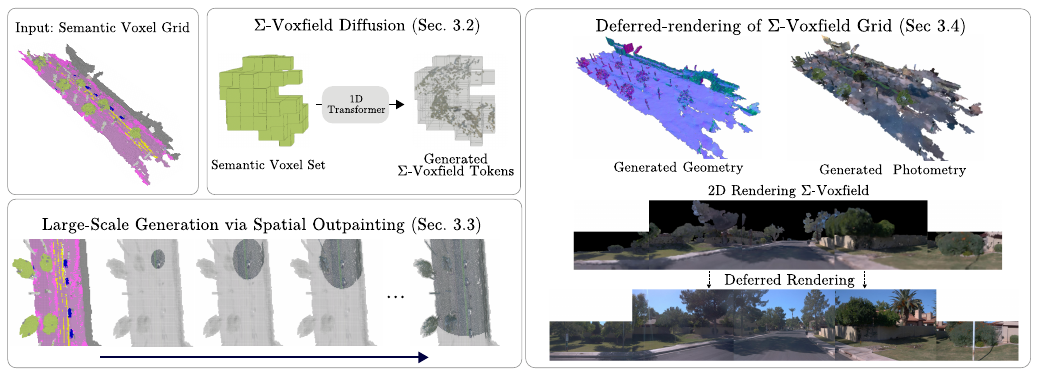}
\caption{Overview of \method. Our framework performs generation directly in 3D to ensure consistent geometry and appearance. It represents the scene with a $\Sigma$-Voxfield grid and applies transformer-based diffusion over 3D tokens. To scale to large environments efficiently, we use an iterative outpainting strategy. Finally, a deferred rendering engine converts the generated 3D scene into photorealistic views.}
\label{fig:method_pipeline}
\end{figure*}

We design our generative framework to meet three key criteria: 3D consistency, scalability, and photorealistic rendering. To ensure the 3D coherence and consistency of our generation, we perform the generation process in the 3D space, rather than distilling 2D generated information into a 3D model. We introduce in~\Cref{subsec:voxfield} our 3D representation, $\Sigma$-Voxfield grid, a local and discrete representation of a colorized surface field designed to be diffused as 3D tokens with a transformer, as explained in~\Cref{subsec:diffusion}. To enable large-scale synthesis while maintaining a reasonable computational budget, we introduce an iterative outpainting method in~\Cref{subsec:outpainting}. Finally, we couple our 3D generation pipeline with a deferred rendering engine to produce photorealistic images as explained in~\Cref{subsec:deferred_rendering}. The overall architecture of our framework is illustrated in~\Cref{fig:method_pipeline}.

% The overall architecture of our framework is illustrated in Fig.~\ref{fig:method_pipeline}. We generate large-scale driving scenes directly in structured 3D space by conditioning on a semantic voxel scaffold that encodes the global layout. Local geometry and appearance are represented using Semantic Surface Voxel Tokens (SSVT), which store surface-aligned points within each voxel. Diffusion is performed over local voxel neighborhoods in world coordinates, producing spatially coherent surface content anchored to the semantic structure.
% To extend generation beyond the fixed diffusion context, we progressively expand the scene via voxel-space spatial outpainting while preserving global consistency through the shared scaffold. For efficient rendering without per-scene optimization, the generated SSVT tokens are deterministically converted into 2D Gaussian splats (2DGS), leveraging their surface alignment. These splats form a structured 3D scene buffer that enables arbitrary viewpoint rendering. Final images are obtained through a render-conditioned image-space diffusion renderer operating on splatted views, producing high-fidelity visuals while remaining fully grounded in the generated 3D representation.

\subsection{$\Sigma$-Voxfield: a joint geometric and photometric representation}

\begin{wrapfigure}{t}{0.50\textwidth}
  \centering
  \vspace{0.2cm}
  \begin{subfigure}[t]{0.32\linewidth}
    \centering
    \includegraphics[width=\linewidth]{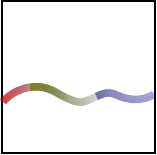}
    \caption{}
    \label{fig:abc-a}
  \end{subfigure}\hfill
  \begin{subfigure}[t]{0.32\linewidth}
    \centering
    \includegraphics[width=\linewidth]{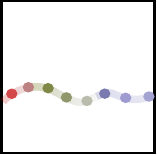}
    \caption{}
    \label{fig:abc-b}
  \end{subfigure}\hfill
  \begin{subfigure}[t]{0.32\linewidth}
    \centering
    \includegraphics[width=\linewidth]{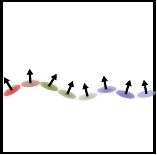}
    \caption{}
    \label{fig:abc-c}
  \end{subfigure}
  \caption{\textbf{Simplified illustration of 2D $\Sigma$ Voxfield.} a) A 2D voxel containing a continuous colorized surface field. b) 2D $\Sigma$ Voxfield points, uniformly sampled on the surface. c) Rendering method of the $\Sigma$ Voxfield, where each point is replaced by a 2D Gaussians aligned with the implicit surface.}
  \label{fig:voxfield}
\end{wrapfigure}

\label{subsec:voxfield}
\paragraph{Definition.} To describe a large outdoor driving scene, we propose to use $\Sigma$-Voxfield grid. A $\Sigma$-Voxfield is a local and discrete representation of a colorized surface field. It is defined at a voxel level and is parametrized by the voxel size $v_s$ and the $\Sigma$-Voxfield cardinality $n$. Each $\Sigma$-Voxfield is composed of $n$ 3D points with associated RGB color, sampled on the surface of the scene lying within the boundary of the voxel. Formally, a $\Sigma$-Voxfield $v_{\Sigma}$ is defined as:
\begin{equation}
    v_{\Sigma} = \left\{(x^i, y^i, z^i), (r^i, g^i, b^i) \right\}_{i \in [1, n]},
\end{equation}
with $(x^i, y^i, z^i)$ the 3D position of the sampled point $i$, defined relative to the voxel center, and $(r^i, g^i, b^i)$ its color.

\paragraph{Properties.} A $\Sigma$-Voxfield represents both the local geometry and the appearance of the scene with a fixed number of points, making this representation an ideal choice for generative 3D modeling. Indeed, geometry and photometry are entangled properties of a scene; it is fundamental to generate them jointly to capture the complexity of outdoor scenes. Moreover, given the fixed cardinality of $\Sigma$-Voxfield, it is straightforward to consider each $\Sigma$-Voxfield as a token within a transformer architecture.

\paragraph{2D Rendering.} $\Sigma$-Voxfield is a point-based representation that can be easily rendered into 2D images. However, like any point cloud rendering, the completeness of the 2D rendering will depend a lot on the density of the point cloud. In order to maintain the cardinality of $\Sigma$-Voxfield sufficiently low to be tractable for a transformer model, we propose a conversion method that will increase the completeness of 2D rendering of $\Sigma$-Voxfield without increasing the number of sampled points. For each point in the $\Sigma$-Voxfield, we create a 2D Gaussian aligned with the surface implicitly present in our representation. Formally, we compute a local normal $(n_x^i, n_y^i, n_z^i)$ via PCA over spatial neighbors and use this normal to initialize rotation matrices $R^i \in SO(3)$ that align each Gaussian with the local tangent plane to the surface. We fix the scale factor along the plane axis to ensure optimal coverage for the 2D rendering. A simplified illustration of $\Sigma$-Voxfield definition and properties is shown in \Cref{fig:voxfield}.

\paragraph{$\Sigma$-Voxfield grid conversion.} Given a textured mesh of a scene, we can easily obtain the counterpart $\Sigma$-Voxfield grid representation. We first voxelize the 3D scene with voxel size $v_s$, then discard all the empty voxels. For each remaining voxel, we uniformly sample $n$ points on the textured mesh surface lying within the voxel to obtain the $\Sigma$-Voxfield grid.

\subsection{$\Sigma$-Voxfields Diffusion}
\label{subsec:diffusion}

% \paragraph{Diffusion over local sets of $\Sigma$-Voxfields.}
We perform diffusion over \emph{local sets} of $\Sigma$-Voxfields. We denote by $\mathcal{G}$ the set of all non-empty $\Sigma$-Voxfields in a scene. We consider a local subsets of adjacent voxels $\mathcal{X}_{\xi}\subset\mathcal{G}$ containing at most $N_{\xi}$ $\Sigma$-Voxfields.
% by sliding a fixed-size window over the scene; each window defines a local set  
% (typically covering $\sim$4$\times$4\,m  $N_{\xi}$ \in [50, 150]) 

Each $\Sigma$-Voxfield $v_{\Sigma}\in\mathcal{X}_{\xi}$ is represented by channel-wise stacking of its $n$ surface samples:
\begin{equation}
\psi(v_{\Sigma})=\big[x^1,y^1,z^1,r^1,g^1,b^1,\dots,x^n,y^n,z^n,r^n,g^n,b^n\big]\in\mathbb{R}^{6n}.
\end{equation}
We order the points stacked in $v_{\Sigma}$ by increasing distance to the $\Sigma$-Voxfield center, so that $\psi(v_{\Sigma})$ is defined deterministically. 

\noindent \textbf{Semantic conditioning.} We associate for each $\Sigma$-Voxfield $v_{\Sigma}$ a semantic label $s_{v_\Sigma}$ (e.g., road, sidewalk, building, vegetation, etc.). The model is conditioned on the semantic labels $\{s_{v_\Sigma}\}_{v_{\Sigma}\in\mathcal{X}_{\xi}}$.

\noindent \textbf{3D positional embedding.} We also denote by $\mathbf{x}_{v_\Sigma}$ the center location of each $v_{\Sigma}$. The center locations are used through 3D positional encodings to expose the 3D structure of the scene to the transformer. 
Formally, we compute a sinusoidal positional encoding from the 3D coordinates of each $v_{\Sigma}\in\mathcal{X}_{\xi}$, project it through a learnable layer and then sum it with the corresponding noisy token.
%\NP{Formaly, the positional encodings is computing using ... and added/concatenated to the token list $\{\psi(v_{\Sigma})\mid v_{\Sigma}\in\mathcal{X}_{\xi}\}$}.
% , and uses 3D positional encodings computed from $\{\mathbf{x}_{v_\Sigma}\}_{v_{\Sigma}\in\mathcal{X}_{\xi}}$ to capture the 3D structure of the local set.

We diffuse the set $\{\psi(v_{\Sigma})\mid v_{\Sigma}\in\mathcal{X}_{\xi}\}$ with a 1D Diffusion Transformer~\cite{dit} architecture by applying the standard forward diffusion process:
\begin{equation}
q(\psi(\mathcal{X}_{\xi,t}) \mid \psi(\mathcal{X}_{\xi,0}))=
\sqrt{\bar{\alpha}_t}\,\psi(\mathcal{X}_{\xi,0})+
\sqrt{1-\bar{\alpha}_t}\,\boldsymbol{\epsilon},\quad
\boldsymbol{\epsilon}\sim\mathcal{N}(\mathbf{0},\mathbf{I}),
\end{equation}
where $t$ denotes the time step and $\bar{\alpha}_t=\prod_{s=1}^{t}(1-\beta_s)$ is the cumulative product of the noise schedule. We train our model $f_\theta$ with \emph{sample prediction}:
\begin{equation}
% \widehat{\psi(\mathcal{X}_{\xi,0})}=f_\theta\!\left(\psi(\mathcal{X}_{\xi,t}),t,\mathbf{S}_{\xi}\right),
% \qquad
% \mathbf{S}_{\xi}=\{(s_{v_\Sigma},\mathbf{x}_{v_\Sigma}) \mid v_{\Sigma}\in\mathcal{X}_{\xi}\}.
\widehat{\psi}(\mathcal{X}_{\xi,0})=f_\theta\!\left(\psi(\mathcal{X}_{\xi,t}),t,\mathbf{S}_{\xi}\right),
\qquad
\mathbf{S}_{\xi}=\{(s_{v_\Sigma},\mathbf{x}_{v_\Sigma}) \mid v_{\Sigma}\in\mathcal{X}_{\xi}\}.
\end{equation}
We minimize an $\ell_2$ loss between $\widehat{\psi}(\mathcal{X}_{\xi,0})$ and $\psi(\mathcal{X}_{\xi,0})$. At inference, we run the reverse process to sample $\mathcal{X}_{\xi,0}$ from noise.
\subsection{Large-Scale Scene Generation via Spatial outpainting}
\label{subsec:outpainting}

% \textbf{Progressive outpainting in $\Sigma$-Voxfield space.}
Our diffusion model operates on local sets, keeping computational cost constant but constraining the spatial extent generated per denoising pass. To synthesize larger scenes, we progressively generate the full $\Sigma$-Voxfield grid by \emph{outpainting} new regions while conditioning on the existing neighborhood using the Repaint~\cite{repaint} diffusion scheduler.

Given a local set $\mathcal{X}_{\xi}$ containing denoised and noisy $\Sigma$-Voxfield tokens, we partition it into a \emph{known} part and a \emph{target} part:
\[
\mathcal{X}_{\xi}=\{\mathcal{X}_{\xi}^{\text{known}},\mathcal{X}_{\xi}^{\text{target}}\}.
\]
We keep $\mathcal{X}_{\xi}^{\text{known}}$ fixed and diffuse only $\mathcal{X}_{\xi}^{\text{target}}$. During reverse diffusion, each denoising step overwrites $\mathcal{X}_{\xi}^{\text{known}}$ with its fixed values and updates only $\mathcal{X}_{\xi}^{\text{target}}$, propagating geometry and appearance consistently across overlapping local sets. Details of partitioning an initial scene $\mathcal{G}$ into overlapping subsets $\mathcal{X}_{\xi}$ are provided in the supplementary materials.

\noindent\textbf{Scalability.} This progressive formulation decouples generation cost from scene size: the model is always applied to local sets of maximum size $N_{\xi}$ $\Sigma$-Voxfields, yet can be iterated to expand to arbitrarily large extents. Thus, we synthesize scenes with tens of thousands of $\Sigma$-Voxfields with linearly growing inference-time while keeping memory and compute comparable to a single denoising process.

\subsection{Deferred-rendering of $\Sigma$-Voxfield grid}
\label{subsec:deferred_rendering}

The model introduced in~\Cref{subsec:diffusion}, coupled with our outpainting strategy described in~\Cref{subsec:outpainting} generates a $\Sigma$-Voxfield grid representing a large outdoor scene. The representation can be rendered efficiently into 2D images using 2DGS~\cite{huang2dgs2024}, as explained in~\Cref{subsec:voxfield}. The rendered frames are by nature 3D consistent because $\Sigma$-Voxfield represents a discretized version of the scene surfaces, it cannot be used as is for most downstream applications requiring high-fidelity rendering. We propose using the rendered images from the $\Sigma$-Voxfield grid as the 3D-buffer input of a deferred-rendering module.

\paragraph{Rendering engine.} We denote $I_\Sigma(w)$ the rendered image from the $\Sigma$-Voxfield grid at pose $w$ and $I(w)$ the real image of the scene at the same pose. Our rendering module can be defined as a function $R$ that outputs an image $I_{DR}(w)$ conditioned on $I_\Sigma(w)$: $I_{DR}(w)=R\!\left(I_\Sigma(w)\right)$. Notice that $I_\Sigma$ does not necessarily contain all the necessary information to be decoded into $I_{DR}$, because: 1. $I_\Sigma(w)$ is a simplification through decimation of the real geometry and photometry of the scene and 2. some parts of the scene may not be covered by the $\Sigma$-Voxfield grid, such as distant background and sky region as shows \Cref{fig:method_pipeline} (2.4). For these reason, we propose to use a diffusion model with generative capability to implement our rendering engine $R$.

\paragraph{Diffusion rendering.}
We use a modified version of Stable Diffusion as our generative 2D deferred rendering engine. Stable Diffusion is a latent diffusion UNet model~\cite{stablediffusion} that iteratively denoises a Gaussian variable $x_{T}$ to reconstruct the original data sample $x_{0}$. $x_{0}$ is obtained by computing the latent representation of the original image $I$ with a Variational Autoencoder (VAE)~\cite{vae}. The denoising network $R_{\phi}$, is trained to predict noise conditioned on input signals $x_\Sigma$ (the latent representation of $I_\Sigma$) by minimizing:
\begin{equation}
    \mathcal{L}(\phi)=\mathbb{E}_{t, \epsilon}\|R_{\phi}(x_t, x_\Sigma)-\epsilon\|^{2}_{2},
\end{equation}
where $\epsilon\sim \mathcal{N}(0, I)$ is the additive Gaussian noise, $t\sim\mathcal{U}(0, T)$ is the time step, $x_t$ is the noisy latent at $t$.

\noindent\textbf{Sky and background modeling.} To avoid hallucinating geometry in areas not covered by our 3D buffer, we use an additional visibility mask as conditioning to indicate sky and background regions. During training, this mask is computed by segmenting the sky region in the real images, while at inference, we use a binary mask derived from the 3D buffer $I_\Sigma$, indicating area without any 2D Gaussians.

\noindent\textbf{Temporal consistency.} Area covered by our 3D buffer can be decoded almost deterministically by our diffusion model, as $I_\Sigma$ contain the coarse shape and colors of objects in the scene. However, high frequency details and area not covered by our 3D buffer are generated stochastically and can change subtly from one pose to another. To ensure temporal consistency of a generated sequence of images, we implement two variants for our diffusion renderer:
\begin{itemize}
    \item Autoregressive Stable Diffusion (ASD): inspired by GameNGen~\cite{gamengen}, we use as an additional conditioning to our diffusion model the previously predicted frame to ensure temporal consistency. 
    \item Video Stable Diffusion (VSD): we train a VSD~\cite{vsd} to generate 12 frames at each rendering step, ensuring a better temporal coherence compared to ASD at a cost of higher memory consumption.
\end{itemize}
More details about these models can be found in our supplementary materials.

\subsection{Data processing}

\paragraph{Large scale textured mesh computation.} We build the $\Sigma$-Voxfield grid from multi-view driving sequences using a geometry-based preprocessing pipeline. We reconstruct each scene with OmniRe~\cite{omnire}, a 3DGS method for urban dynamic scenes with an additional normal supervision regularizer obtained via DepthAnything~\cite{depthanything} monocular prior. From this reconstruction, we extract optimized poses and depth maps of the static background, fuse depths into an SDF, and extract a surface mesh $\mathcal{M}$ via Marching Cubes. We texture $\mathcal{M}$ by aggregating multi-view RGB observations using OpenMVS~\cite{openmvs}. Example of computed mesh and additional details about our data pre-processing pipeline can be found in supplementary. 

\paragraph{Semantic $\Sigma$-Voxfield tokens computation.} $\Sigma$-Voxfield grids are obtained from the textured mesh as explained in~\Cref{subsec:voxfield}. We also compute an aligned semantic voxel grid for our diffusion model conditioning by back-projecting and aggregating per-frame semantic segmentation with a modified TSDF fusion method.

\paragraph{Dataset computation for deferred rendering.} To obtain pairs of images $\{I_\Sigma, I \}$ necessary to train our deferred rendering diffusion module, we render for each camera and at each pose $w$ the $\Sigma$-Voxfield image $I_\Sigma(w)$. Because the original corresponding image $I(w)$ may contain dynamic objects, we render using the 3DGS static field of the trained model a static image $\hat{I}_{s}(w)$.
% We then voxelize the scene with $v_s{=}0.6m$, assign semantic labels by back-projecting and aggregating per-frame semantic segmentation, and keep only voxels intersecting $\mathcal{M}$. For each retained voxel, we sample $n$ points on $\mathcal{M}$ within the voxel bounds and store their local offsets (relative to the voxel center), colors, and voxel semantic label, yielding one $\Sigma$-Voxfield token per occupied voxel.

\section{Experiments}
\subsection{Experimental setup}
\noindent\textbf{Datasets.}
We evaluate on two large-scale autonomous driving datasets: Waymo Open Dataset (WOD)~\cite{waymo} and PandaSet~\cite{pandaset}. Training scene splits are detailed in the supplementary materials.

\noindent\textbf{$\Sigma$-Voxfield parameters.} In all our experiment We use $\Sigma$-Voxfield with voxel size $v_s=0.6m$, $n=20$ sampled points and local sets of $\Sigma$-Voxfields $\mathcal{X}_\xi$ composed of $N_{\xi} \in [50, 150]$ $\Sigma$-Voxfields. $\mathcal{X}_\xi$ typically cover a scene of $\sim$4$\times$4$m^2$, a good tradeoff between number of 3D points and scene context. Given these parameter, we choose $r=0.04m$ as fixed splat radius for 2DGS  $\Sigma$-Voxfield rendering.

\noindent\textbf{Model architecture and training details.}
For our $\Sigma$-Voxfield diffusion backbone, we use a 1-Dimensional DiT diffusion backbone with masked attention over voxel tokens, restricting each token to attend only to voxels within a 3-meter neighborhood. The model is trained with 1{,}000 denoising steps, ADAM optimizer and learning rate of $5e-4$. During training, we randomly drop the semantic conditioning with a probability of 10\% to enable classifier-free guidance. At inference time, we apply classifier-free guidance with a scale of 4.0. All models are trained for 4 days on 2×24GB GPUs with performance comparable to an RTX 4090. ASD deferred renderer is finetuned from SD 1.5 with ADAM optimizer and a learning rate of $5e-5$ on 1 GPU for approximately 4 days. More training hyperparameters can be found in our supplementary materials.

\noindent\textbf{Competitors.} We compare our proposal to two SOTA methods for large scale scene generation with different generation paradigm. GEN3C~\cite{gen3c} is a video diffusion based on the powerful COSMOS diffusion backbone~\cite{cosmos} with additional point cloud conditioning to guide the generation. Similar to LSD-3D~\cite{lsd3d}, we condition GEN3C on our initial rendering. InfiniCube~\cite{infinicube} is a multi-step pipeline that start to generate a fine voxel grid from an HDMap, followed by a video diffusion model conditioned on the generated geometry. The generated frames are used by a feedforward 3DGS network to obtain the final reconstruction.

\begin{figure*}[htbp]
\centering
\scriptsize
\setlength{\tabcolsep}{0pt}
\renewcommand{\arraystretch}{1.0}

% ---- knobs ----
\newcommand{\imgW}{0.19\textwidth}  % slightly smaller to make room for left label
\newcommand{\pairgap}{2pt}
\newcommand{\tightwithin}{-2pt}

% left vertical label (spans the two rows)
\newcommand{\vscenelabel}[1]{%
  \multirow[t]{2}{*}{\rotatebox{0}{\textbf{#1}}\hspace{6pt}}%
}

% 5 images touching in one row
\newcommand{\tilefive}[5]{%
\includegraphics[width=\imgW]{#1}%
\includegraphics[width=\imgW]{#2}%
\includegraphics[width=\imgW]{#3}%
\includegraphics[width=\imgW]{#4}%
\includegraphics[width=\imgW]{#5}%
}

\begin{tabular}{@{}c@{}c@{}}
% col1 = vertical label, col2 = the 5-image strip

% ===================== Scene 1 =====================
\vscenelabel{(a)} &
\tilefive
{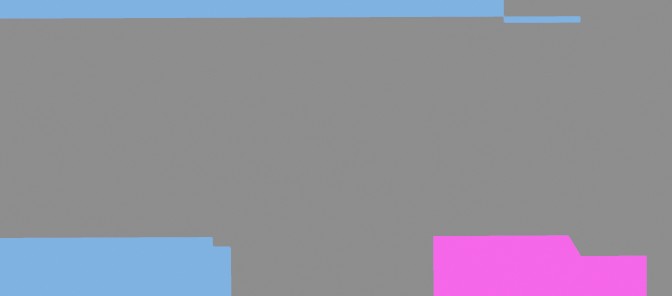}
{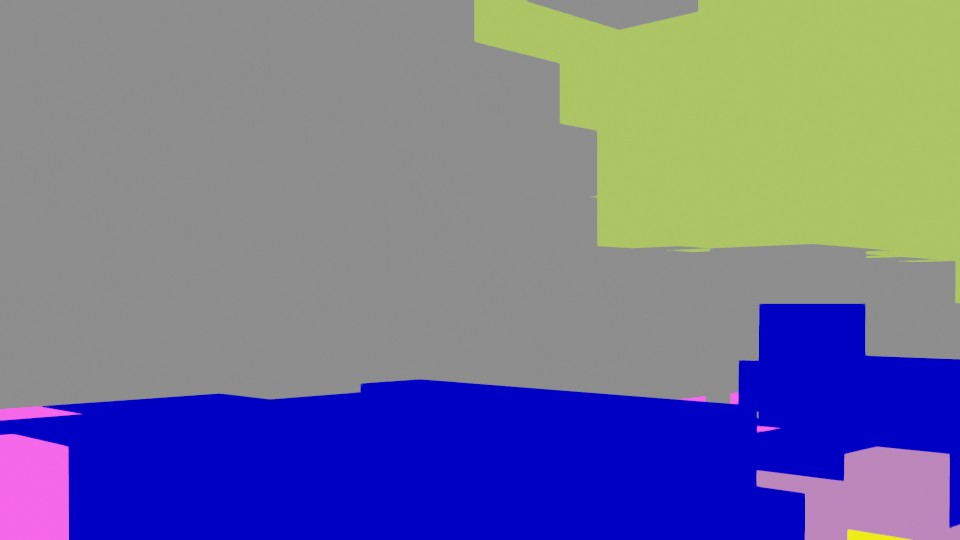}
{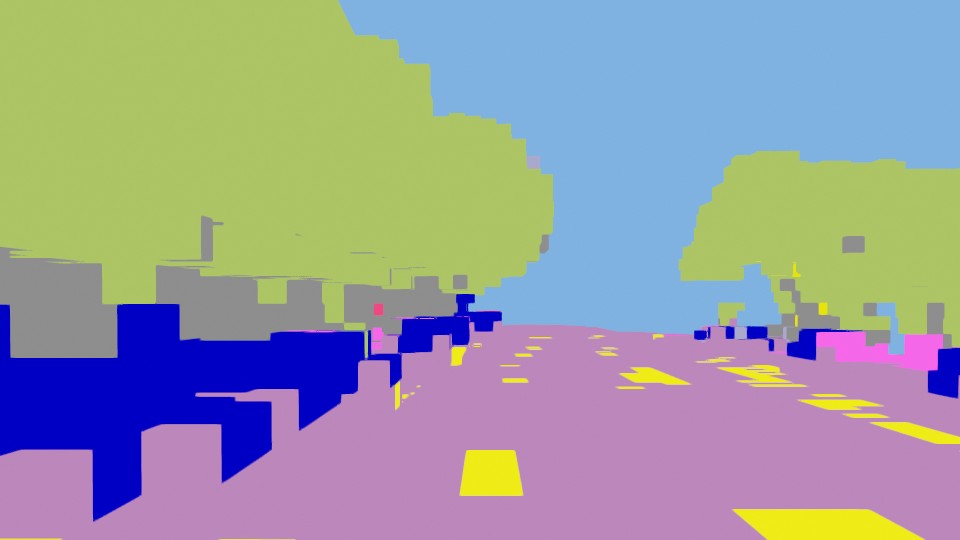}
{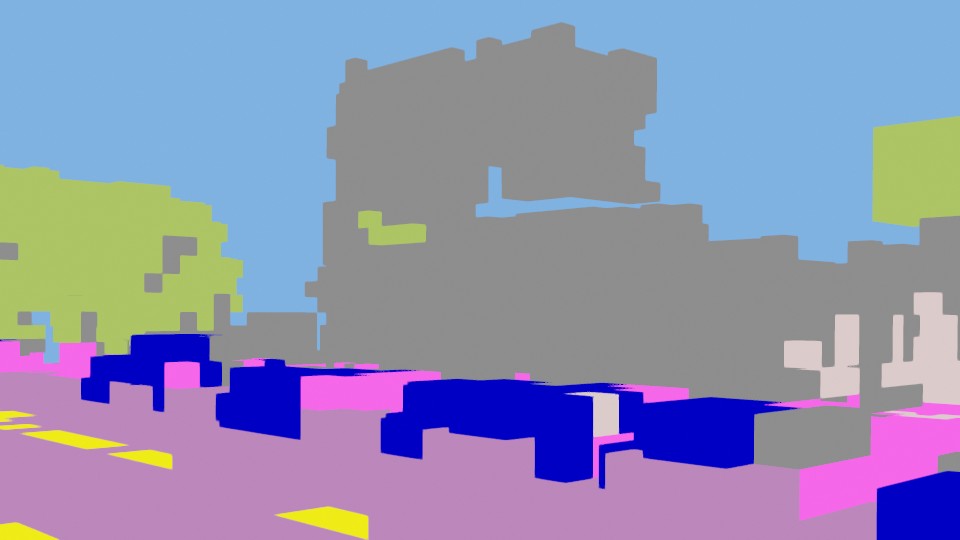}
{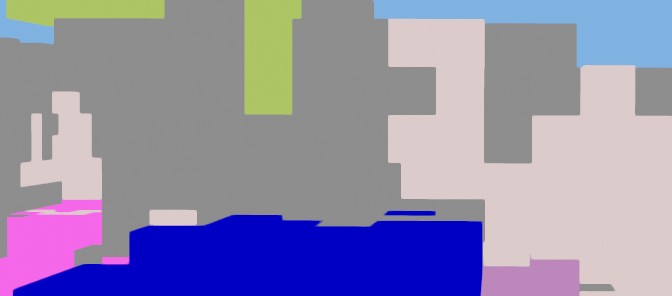}
\\[\tightwithin]
&
\tilefive
{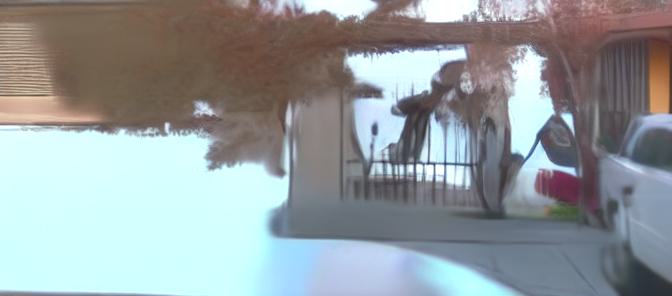}
{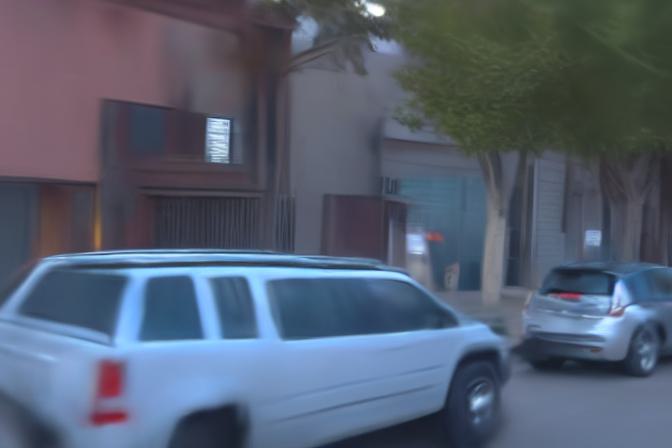}
{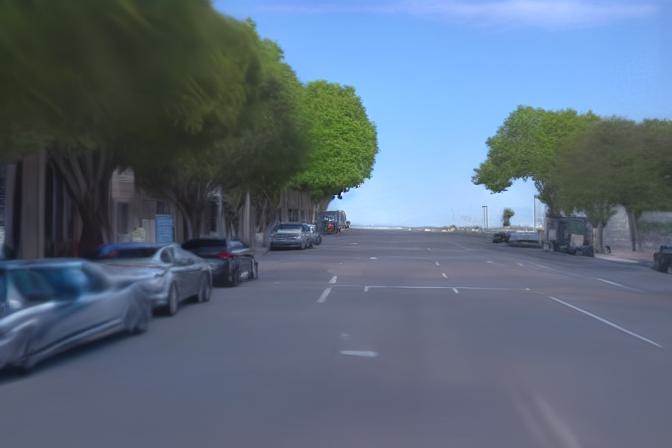}
{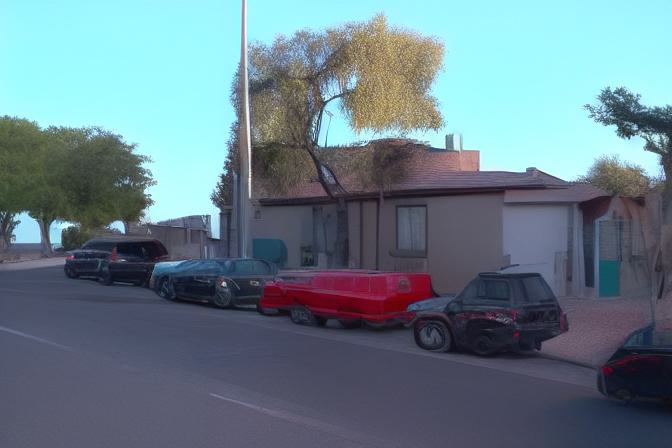}
{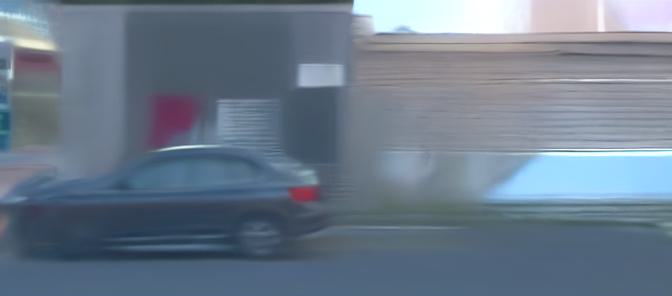}
\\[\pairgap]

% ===================== Scene 2 =====================
\vscenelabel{(b)} &
\tilefive
{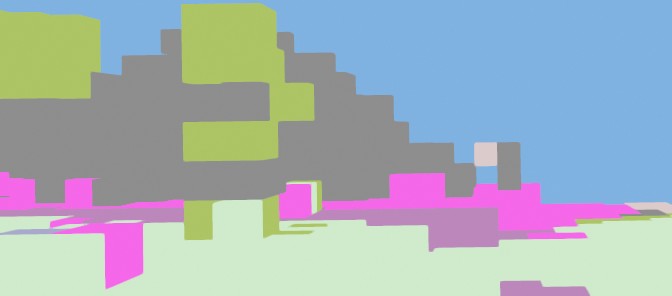}
{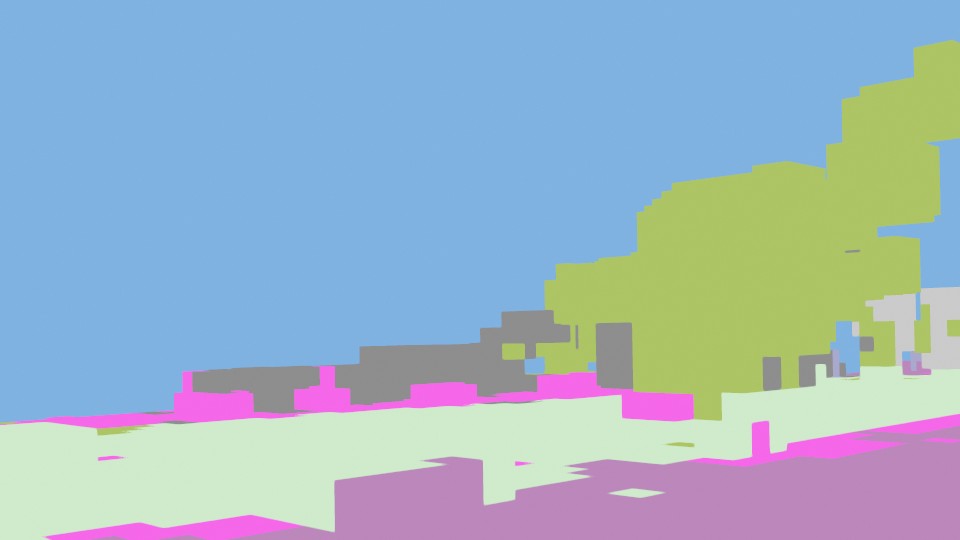}
{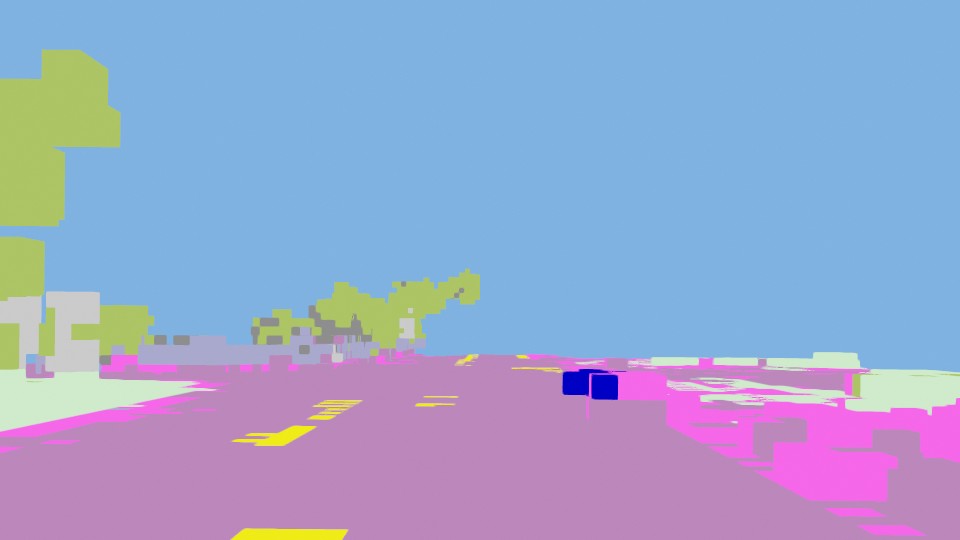}
{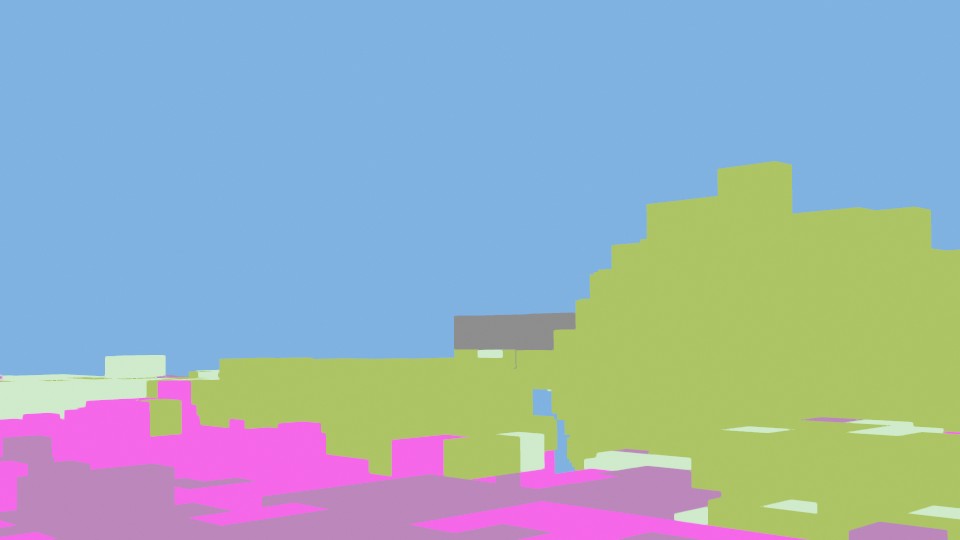}
{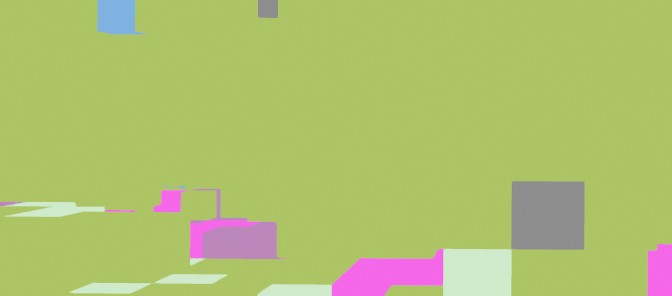}
\\[\tightwithin]
&
\tilefive
{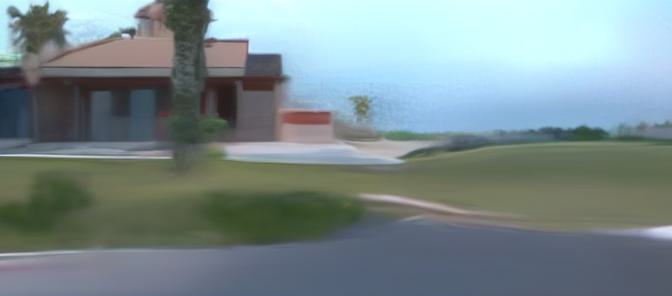}
{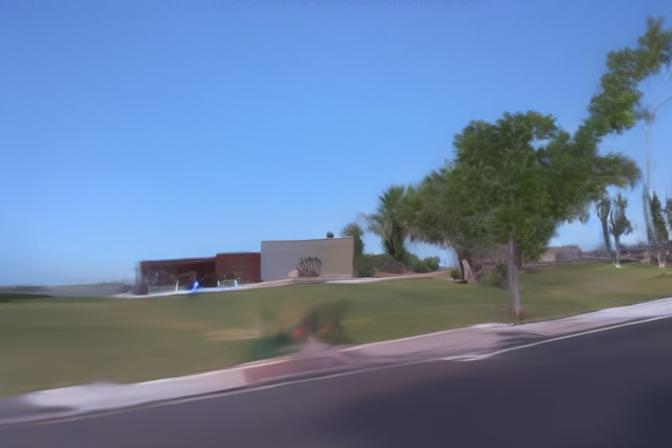}
{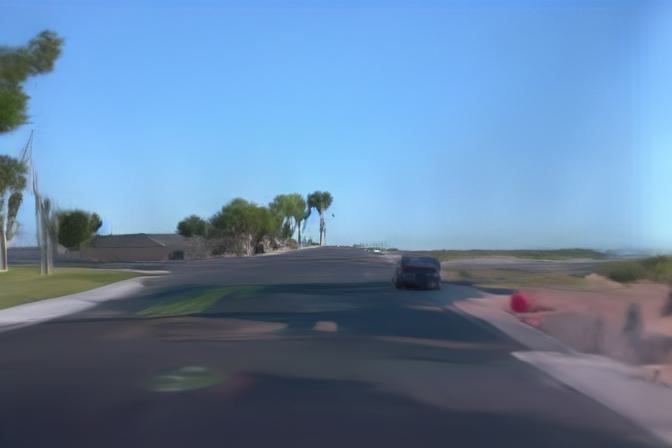}
{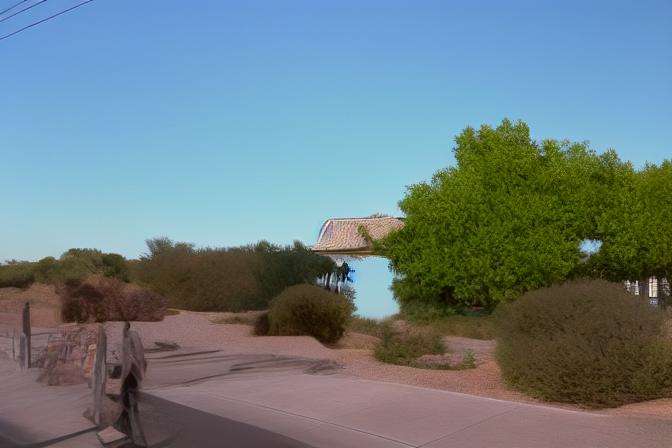}
{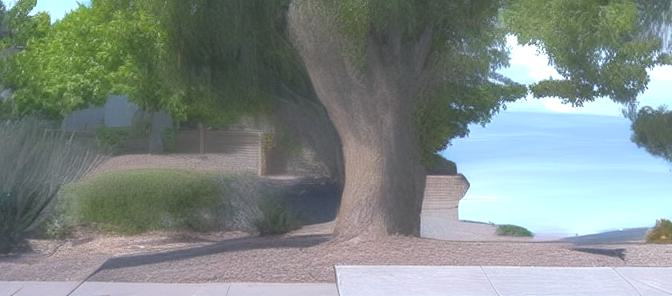}
\\[\pairgap]

% ===================== Scene 4 =====================
\vscenelabel{(c)} &
\tilefive
{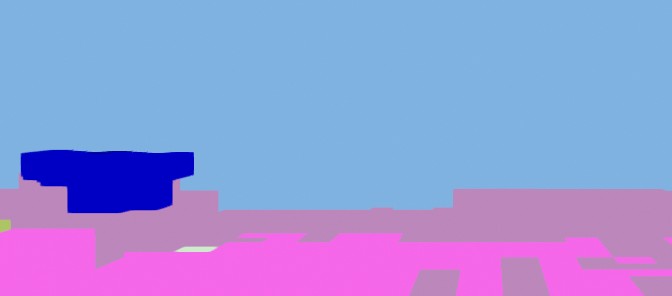}
{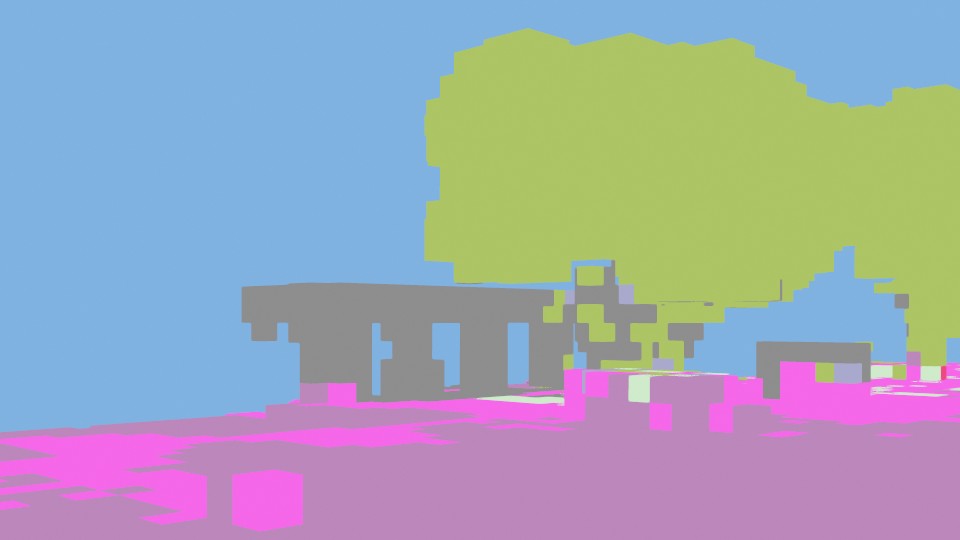}
{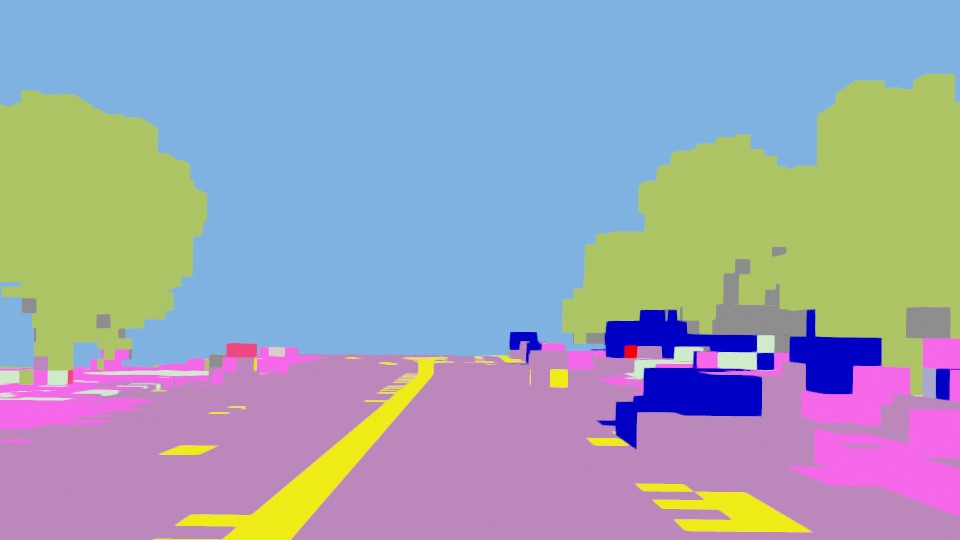}
{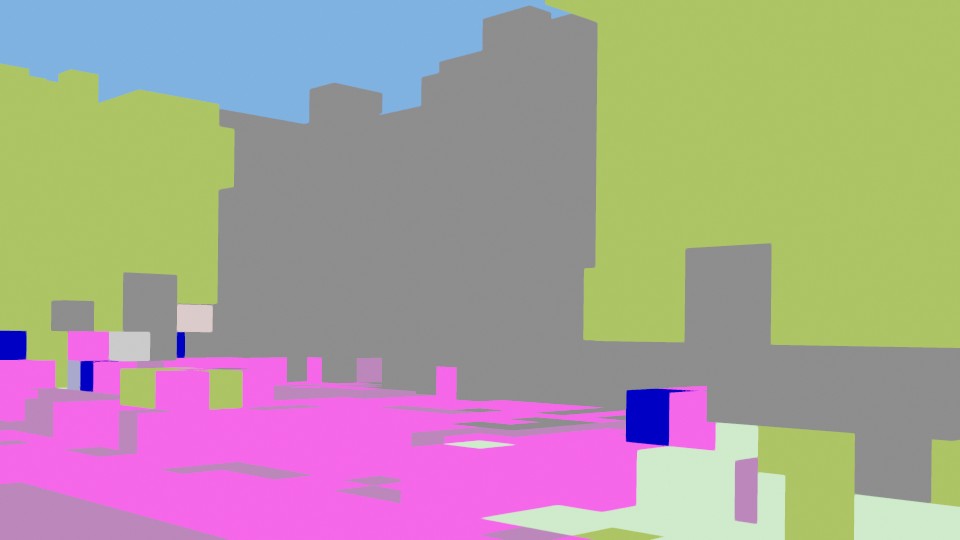}
{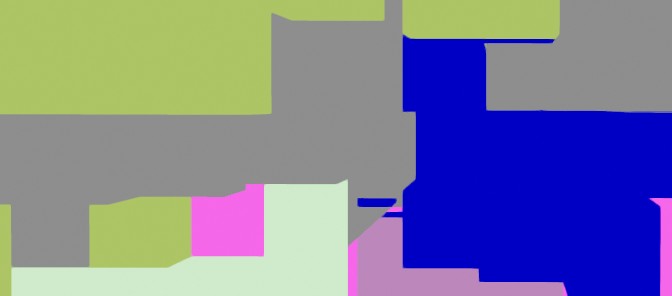}
\\[\tightwithin]
&
\tilefive
{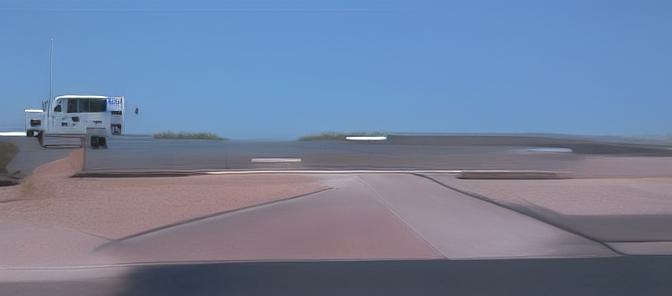}
{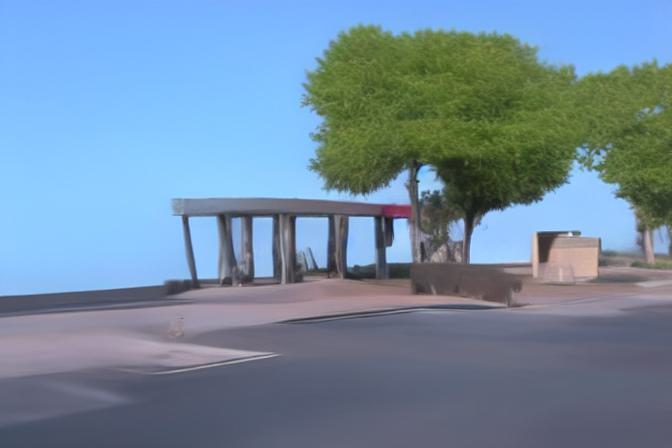}
{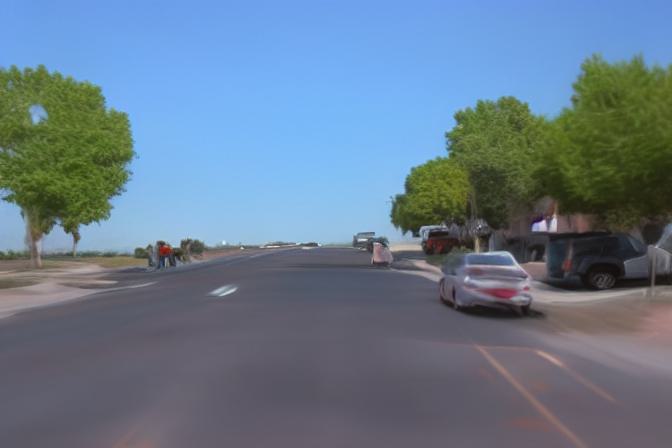}
{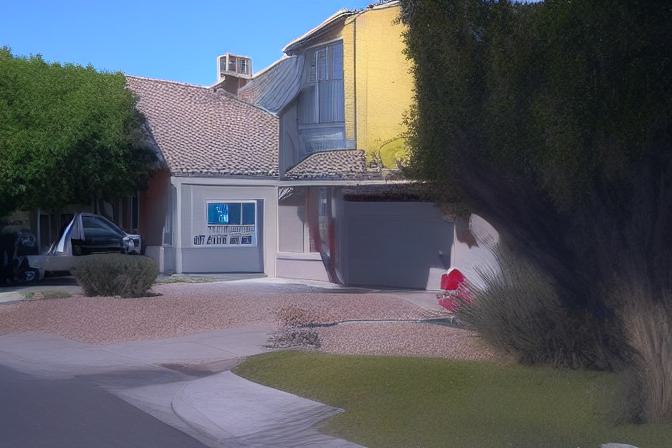}
{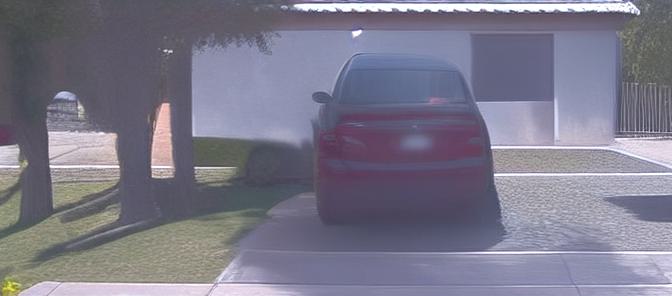}
\\

\end{tabular}

\caption{\textbf{Qualitative results on WOD.} We show three WOD scenes (a--c). For each scene, the \textbf{top} strip visualizes the semantic voxel rendering used for conditioning, and the \textbf{bottom} strip shows the corresponding generated scene from 5 camera views.}
\label{fig:results}
\end{figure*}
\begin{figure*}[htbp]
\centering
\scriptsize
\setlength{\tabcolsep}{0pt}
\renewcommand{\arraystretch}{0.0}

% ---- knobs ----
\newcommand{\imgW}{0.165\textwidth}  % smaller tiles
\newcommand{\pairgap}{1pt}
\newcommand{\tightwithin}{-3pt}

\newcommand{\vscenelabel}[1]{%
  \multirow[t]{2}{*}{\textbf{#1}\hspace{3pt}}%
}

\newcommand{\tilefive}[5]{%
\includegraphics[width=\imgW]{#1}%
\includegraphics[width=\imgW]{#2}%
\includegraphics[width=\imgW]{#3}%
\includegraphics[width=\imgW]{#4}%
\includegraphics[width=\imgW]{#5}%
}

\begin{tabular}{@{}c@{}c@{}}

% ===================== Scene 1 =====================
\vscenelabel{(a)} &
\tilefive
{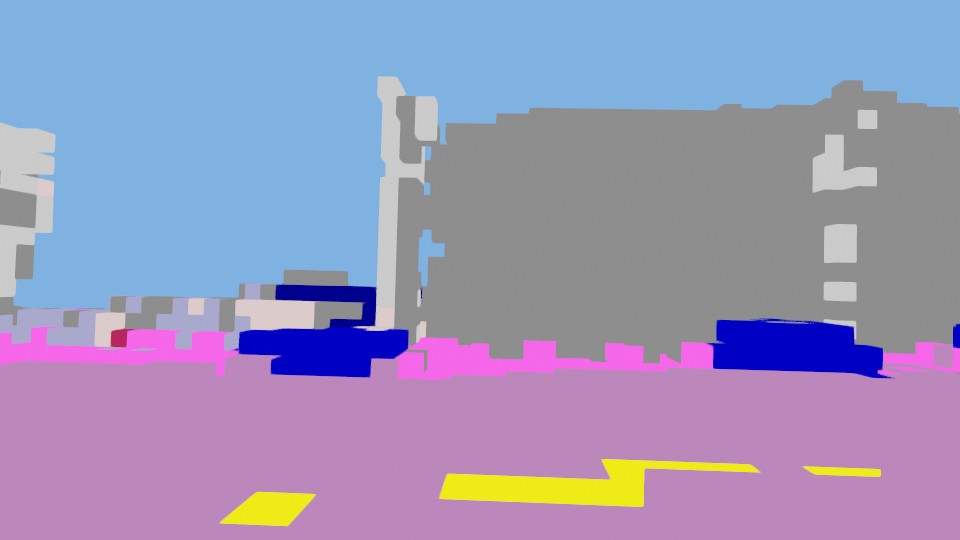}
{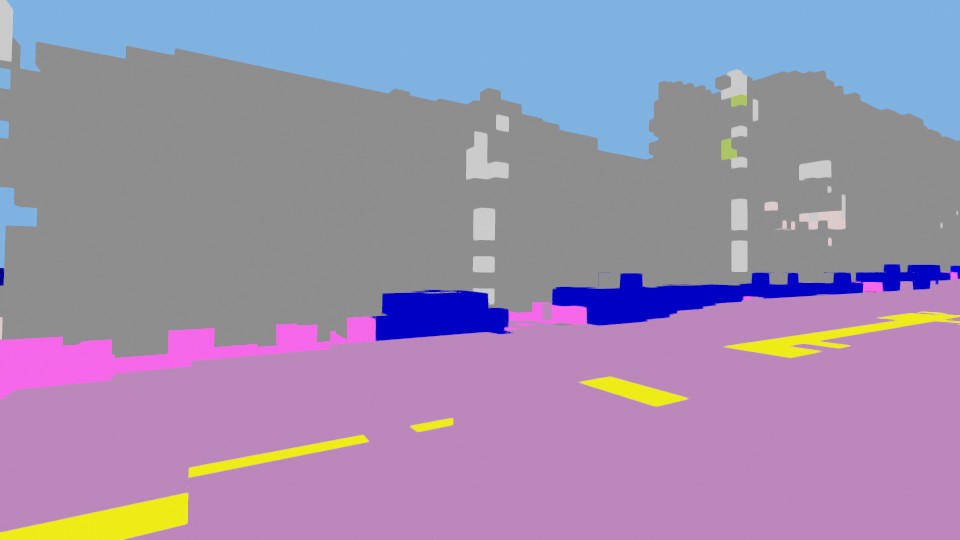}
{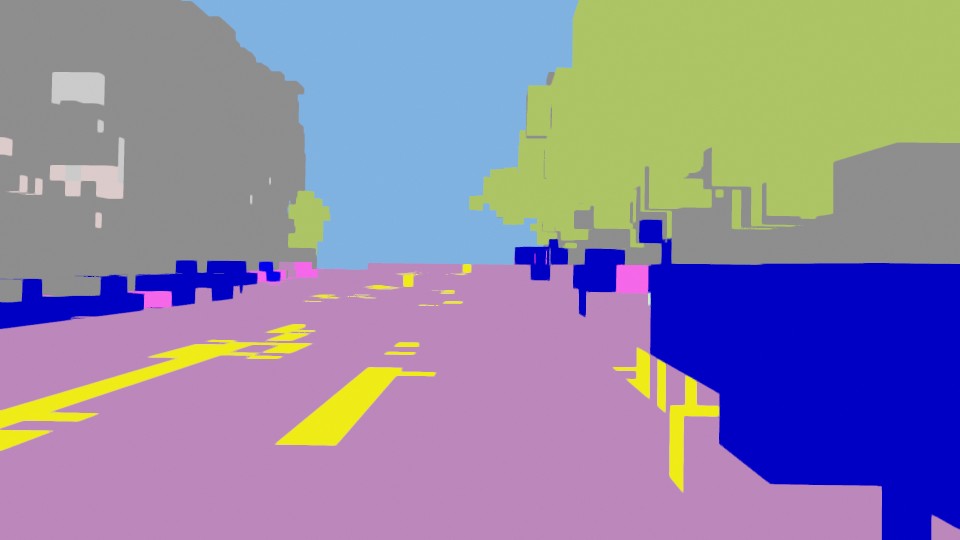}
{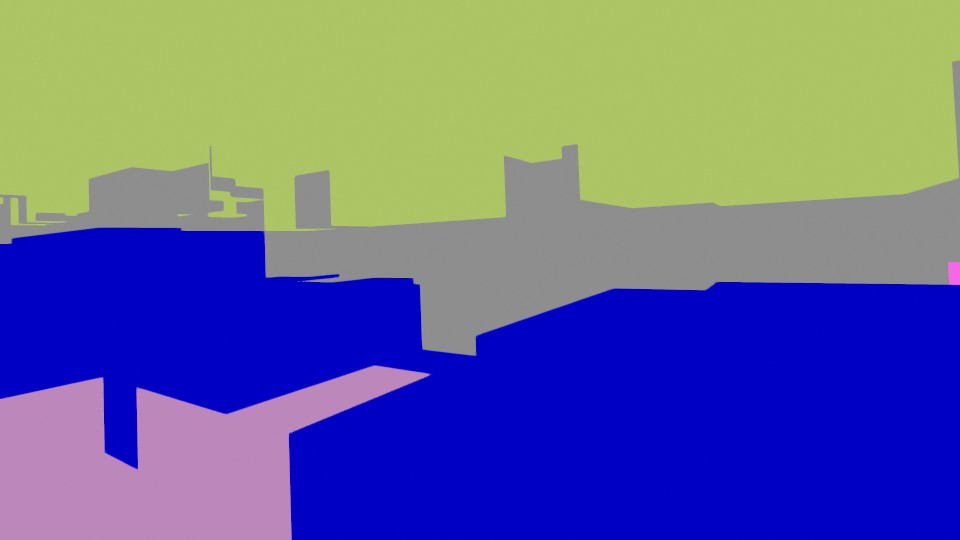}
{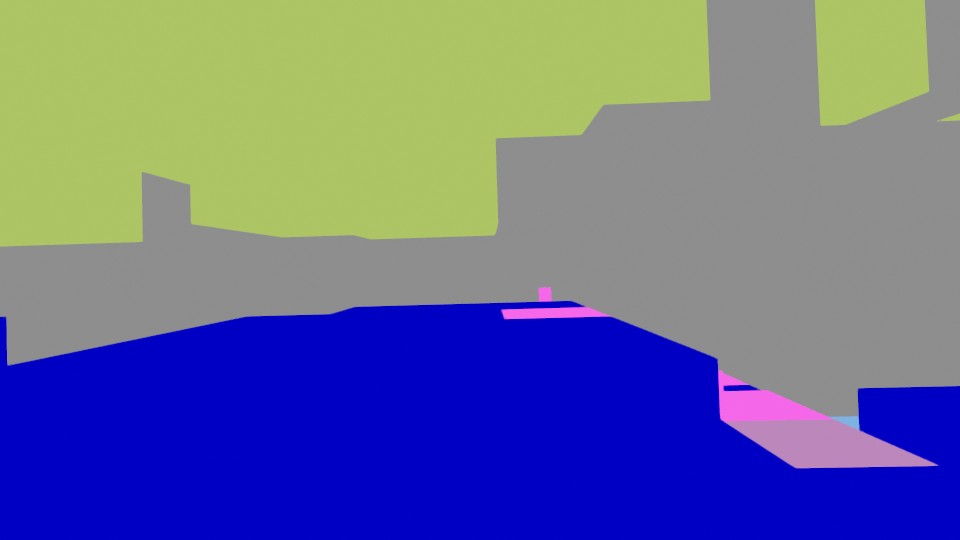}
\\[\tightwithin]
&
\tilefive
{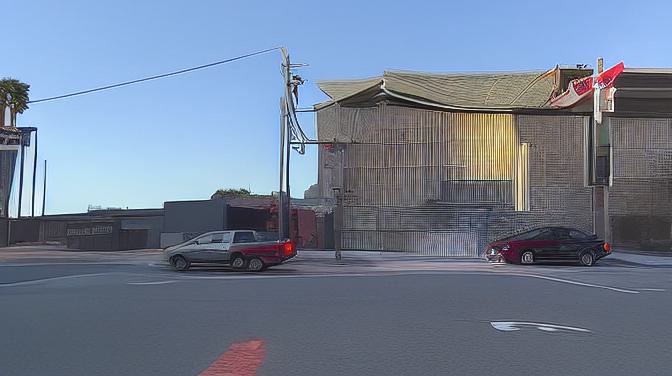}
{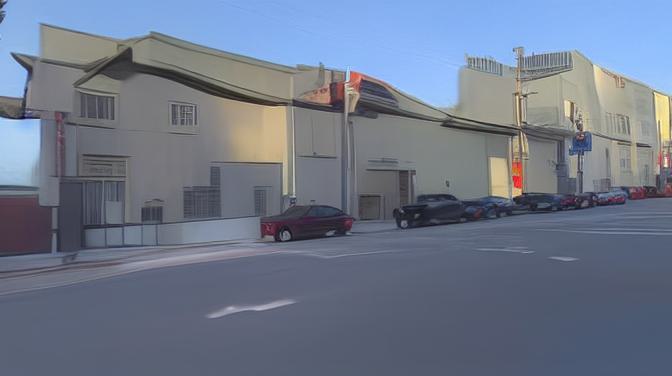}
{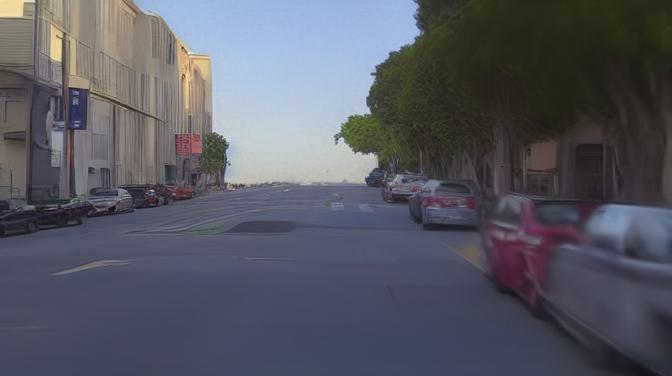}
{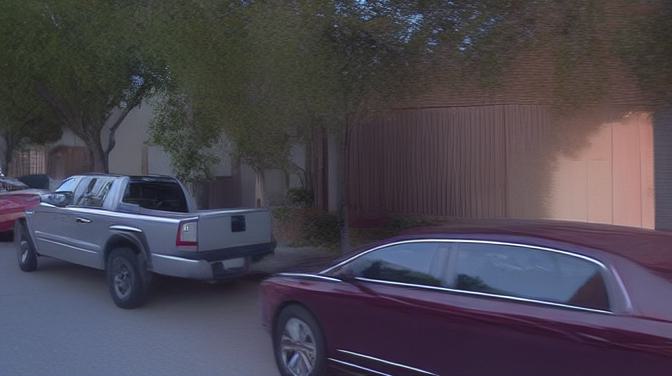}
{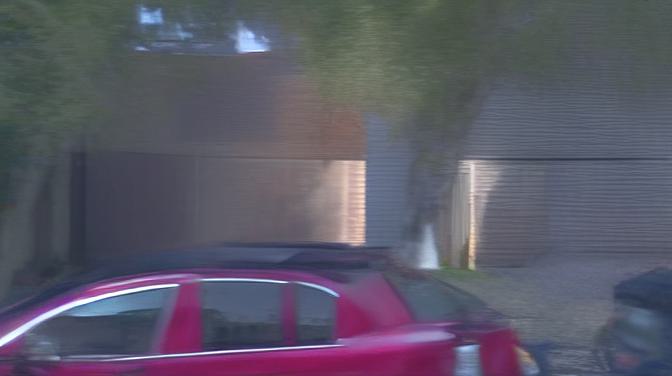}
\\[\pairgap]

% ===================== Scene 2 =====================
\vscenelabel{(b)} &
\tilefive
{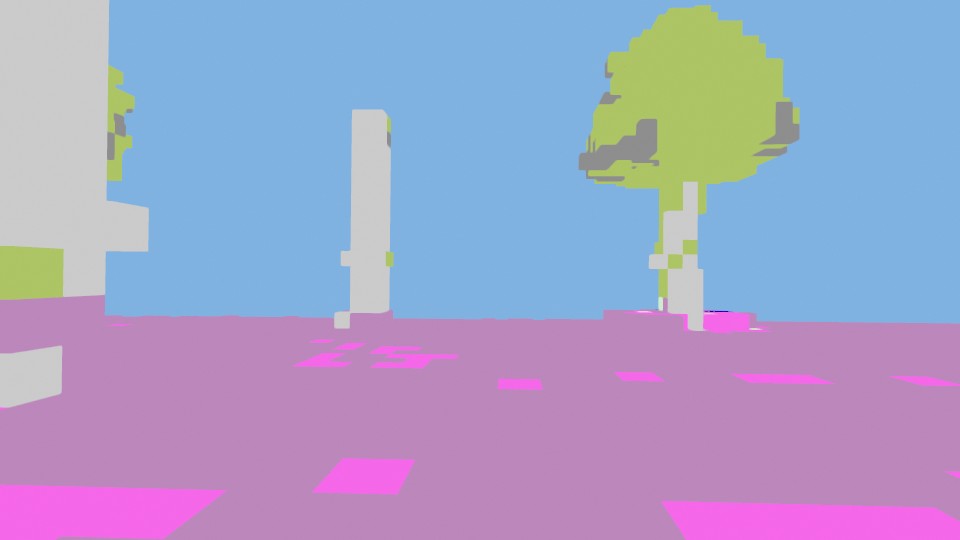}
{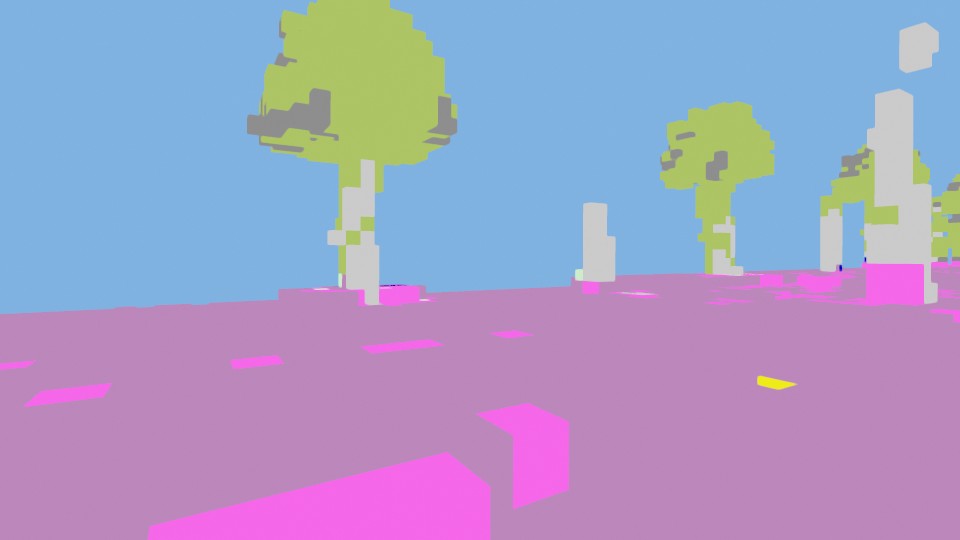}
{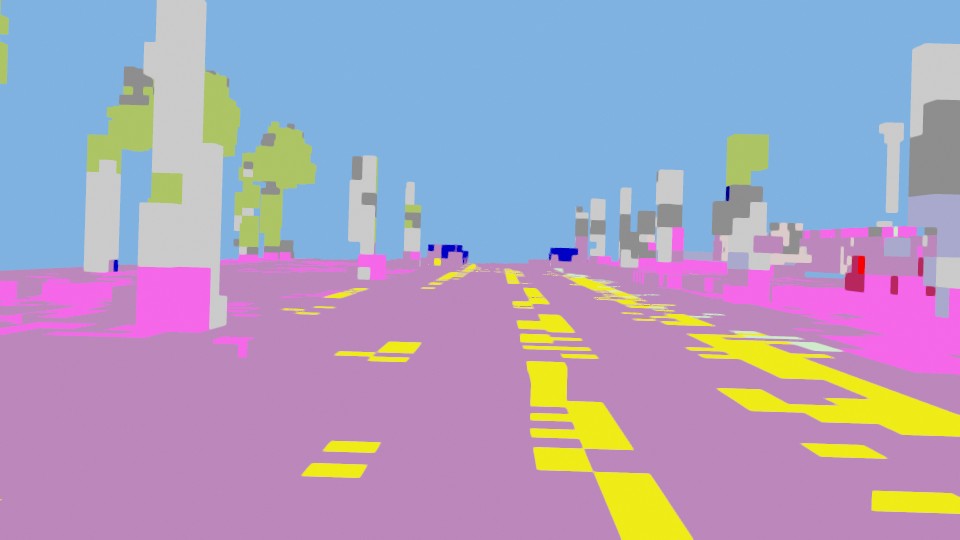}
{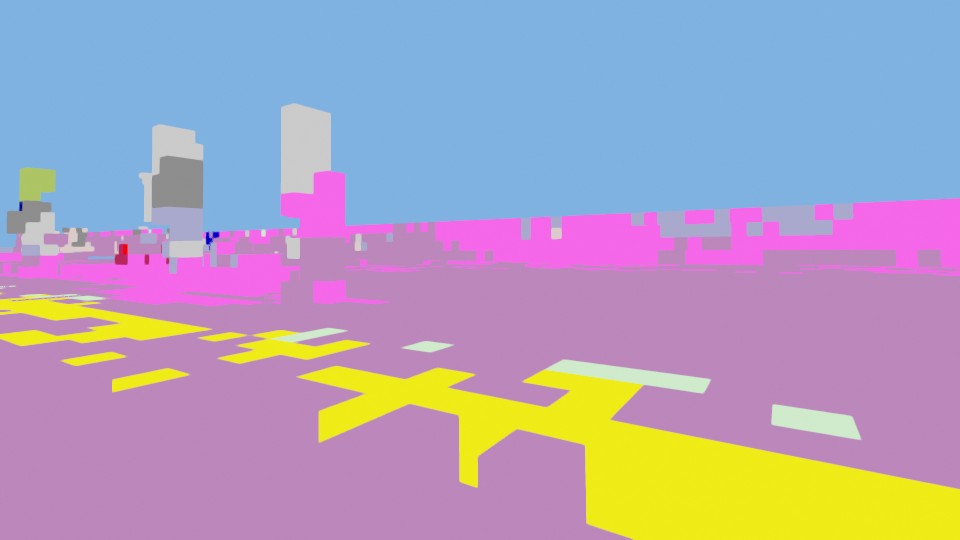}
{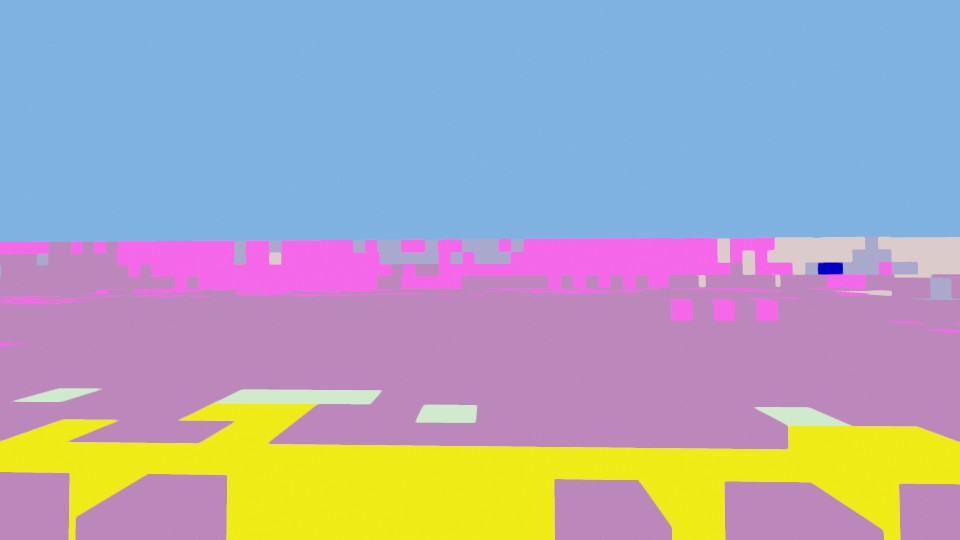}
\\[\tightwithin]
&
\tilefive
{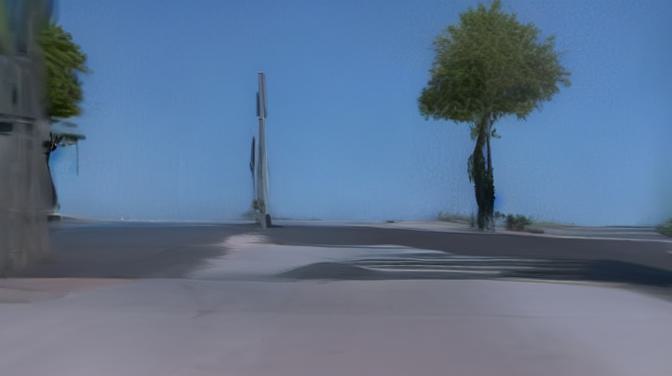}
{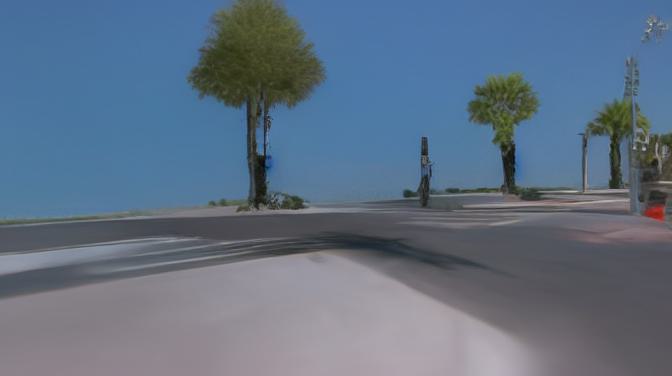}
{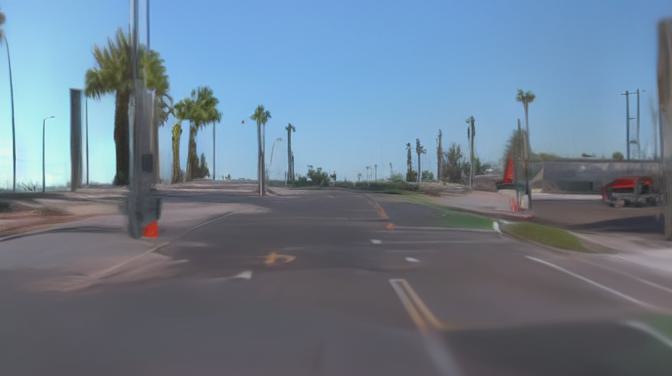}
{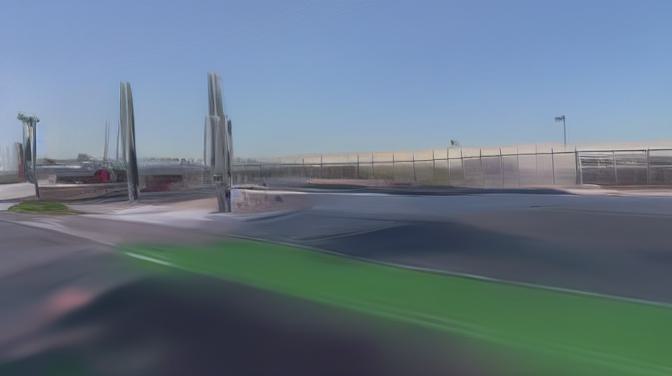}
{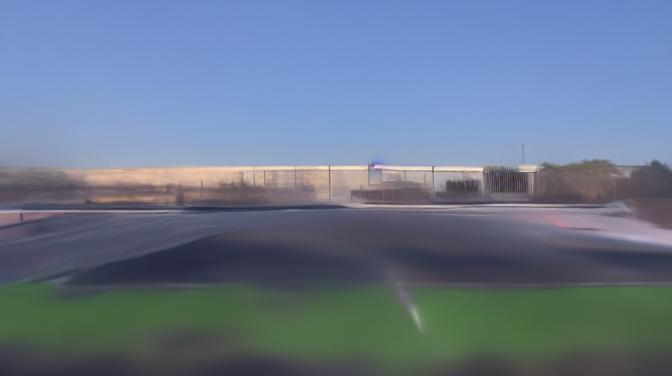}
\\[\pairgap]

% ===================== Scene 3 =====================
\vscenelabel{(c)} &
\tilefive
{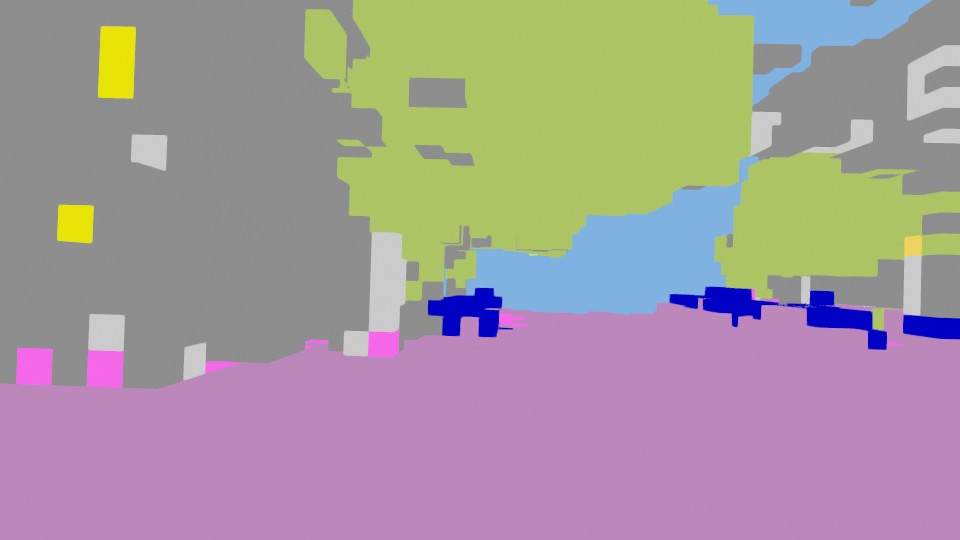}
{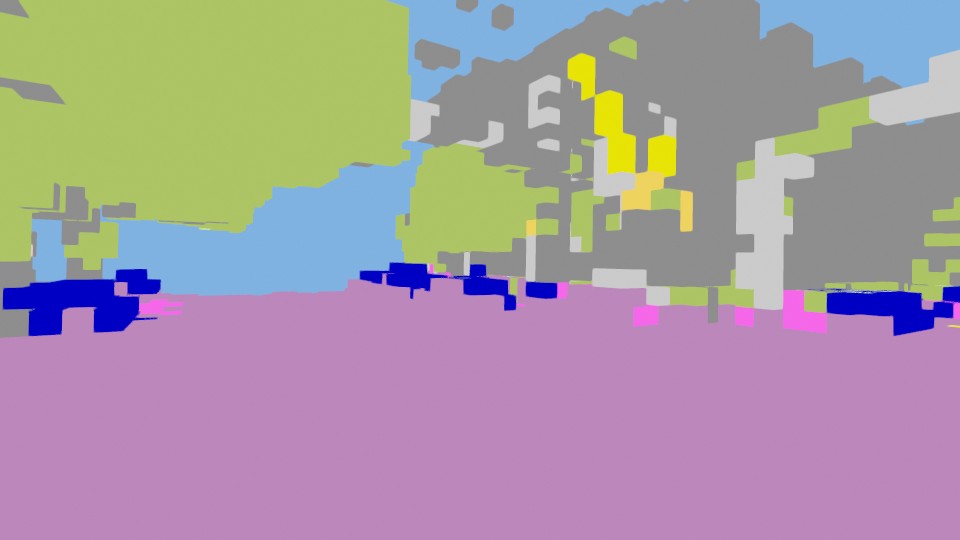}
{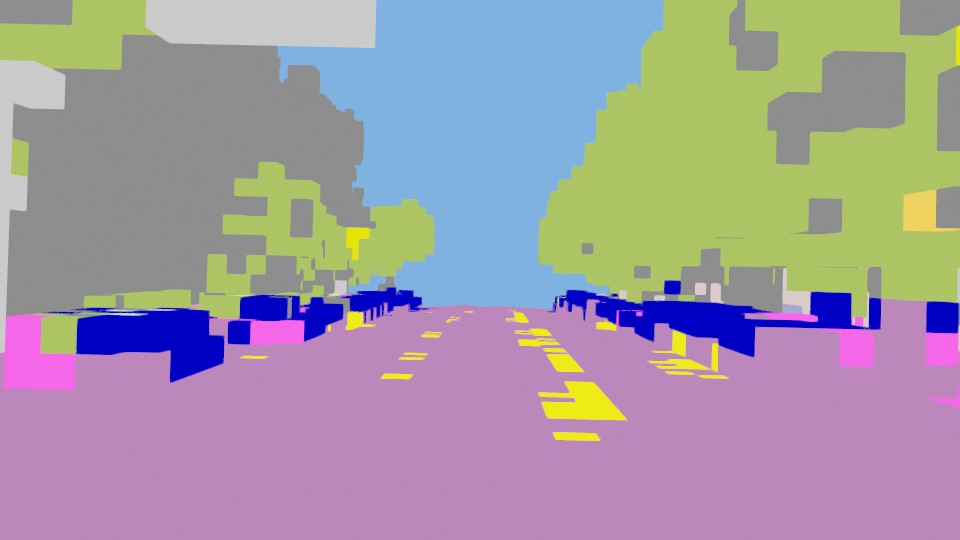}
{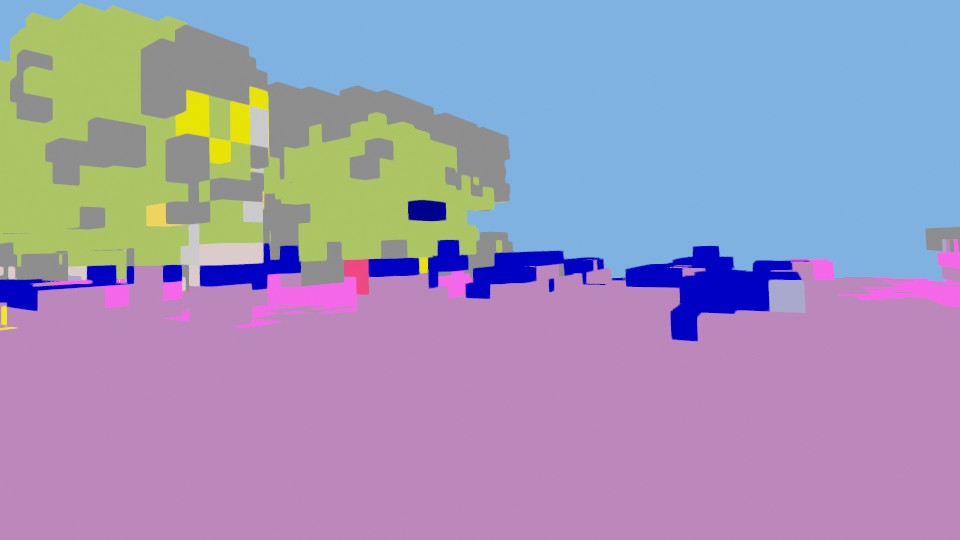}
{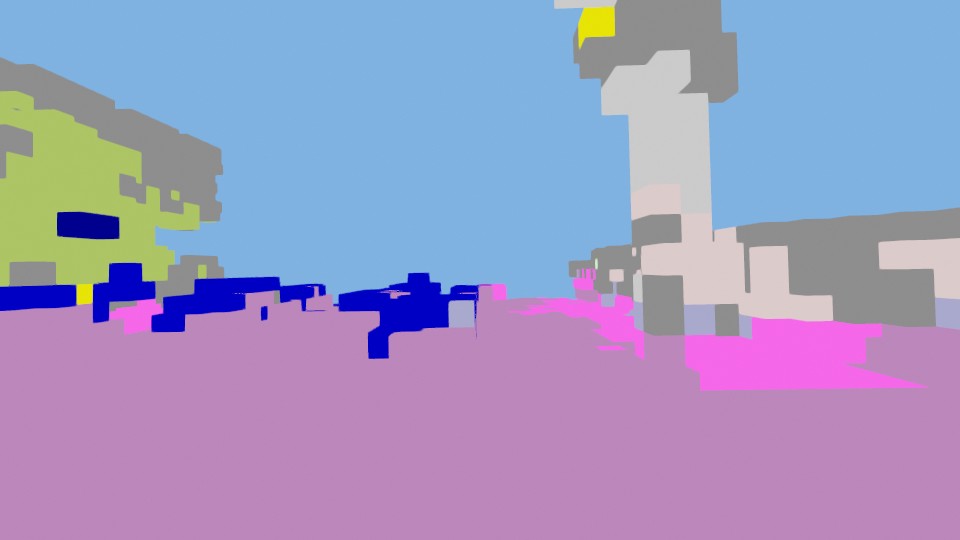}
\\[\tightwithin]
&
\tilefive
{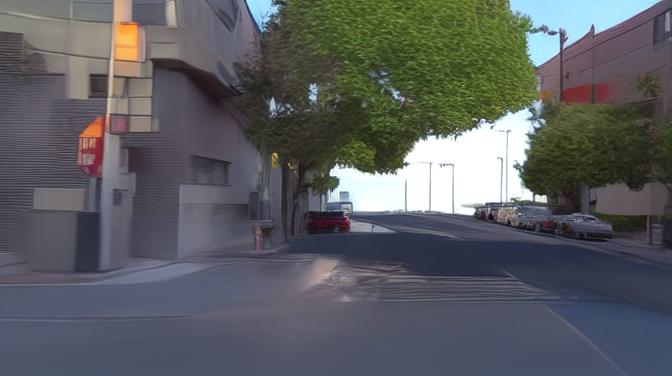}
{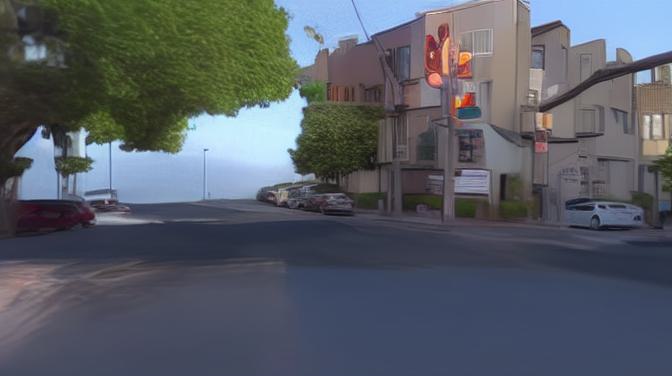}
{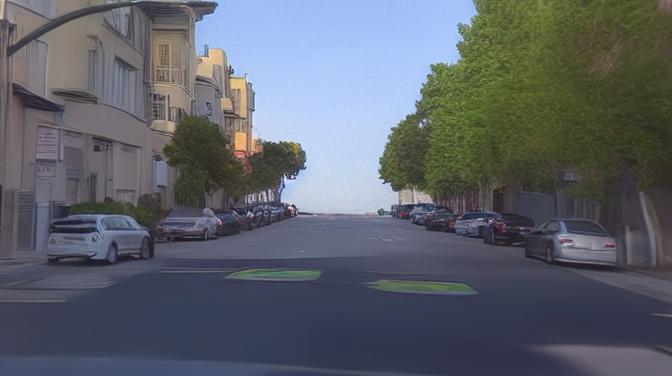}
{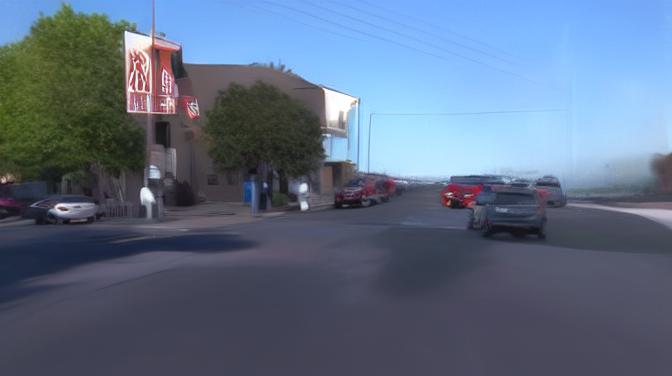}
{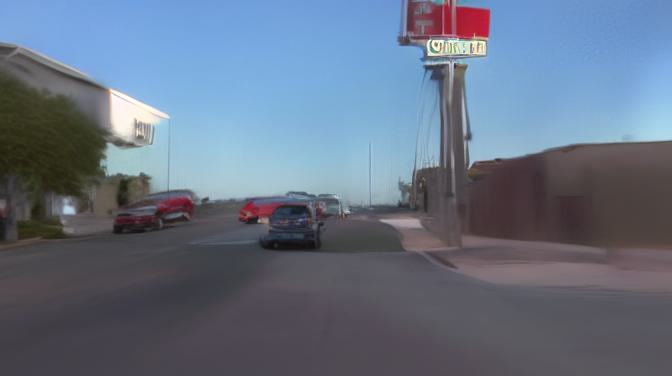}
\\[\pairgap]

% ===================== Scene 4 =====================
%\vscenelabel{(d)} &
%\tilefive
%{results_sel/pandaset_results/run_91/011_3_1.jpg}
%{results_sel/pandaset_results/run_91/011_1_1.jpg}
%{results_sel/pandaset_results/run_91/011_0_1.jpg}
%{results_sel/pandaset_results/run_91/011_2_1.jpg}
%{results_sel/pandaset_results/run_91/011_4_1.jpg}
%\\[\tightwithin]
%&
%\tilefive
%{results_sel/pandaset_results/run_91/011_3.jpg}
%{results_sel/pandaset_results/run_91/011_1.jpg}
%{results_sel/pandaset_results/run_91/011_0.jpg}
%5{results_sel/pandaset_results/run_91/011_2.jpg}
%{results_sel/pandaset_results/run_91/011_4.jpg}
%\\

\end{tabular}

\vspace{-1.5mm}
\caption{\textbf{Qualitative results on PandaSet.} We show three PandaSet scenes (a--c). For each scene, the \textbf{top} strip visualizes the semantic voxel rendering used for conditioning, and the \textbf{bottom} strip shows the corresponding generated scene from 5 camera views.}
\label{fig:results2}
\vspace{-2mm}
\end{figure*}
\begin{figure*}[htbp]
\centering
\scriptsize
\setlength{\tabcolsep}{0pt}
\renewcommand{\arraystretch}{1.0}

% ---- make it smaller ----
\newcommand{\imgW}{0.145\textwidth} % was 0.158

% tighter vertical spacing
\newcommand{\inpair}{-4pt}   % was -2pt
\newcommand{\betweens}{0pt}  % was 1pt
\newcommand{\scenegap}{0pt}  % was 2pt
\renewcommand{\arraystretch}{0.92} % was 1.0

% optional: remove extra whitespace around tabular
\setlength{\abovecaptionskip}{2pt}
\setlength{\belowcaptionskip}{0pt}

\newcommand{\five}[5]{%
\includegraphics[width=\imgW]{#1}&
\includegraphics[width=\imgW]{#2}&
\includegraphics[width=\imgW]{#3}&
\includegraphics[width=\imgW]{#4}&
\includegraphics[width=\imgW]{#5}%
}
% label cell WITHOUT multicolumn
\newcommand{\vlabel}[1]{\rotatebox{90}{\tiny\textbf{#1}}}

% keep semrow empty label cell as normal too
\newcommand{\semrow}[5]{%
 & \five{#1}{#2}{#3}{#4}{#5}\\[\inpair]
}

\newcommand{\methrow}[6]{%
\vlabel{#1} & \five{#2}{#3}{#4}{#5}{#6}\\
}

% now this spacing WILL apply
\begin{tabular}{@{}c@{\hspace{10pt}}ccccc@{}}

% ===================== Scene 1 =====================
% Ours pair: semantics -> ours
\semrow
{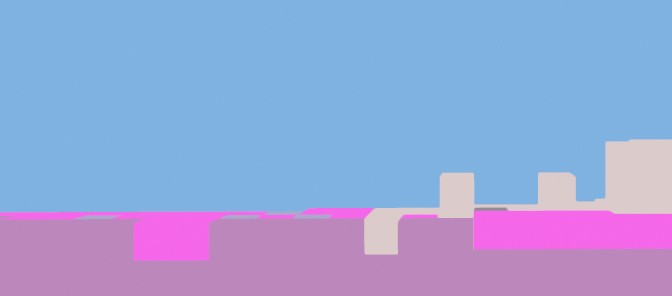}
{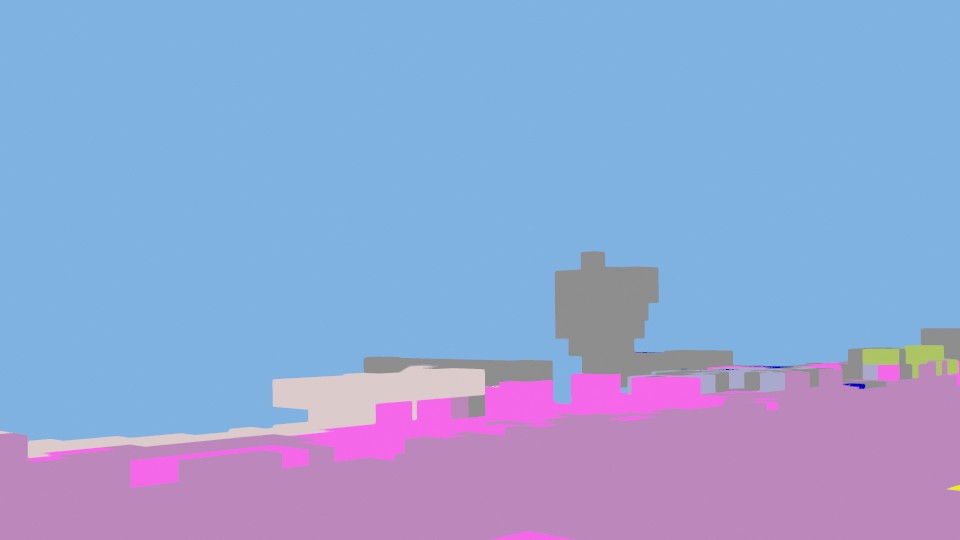}
{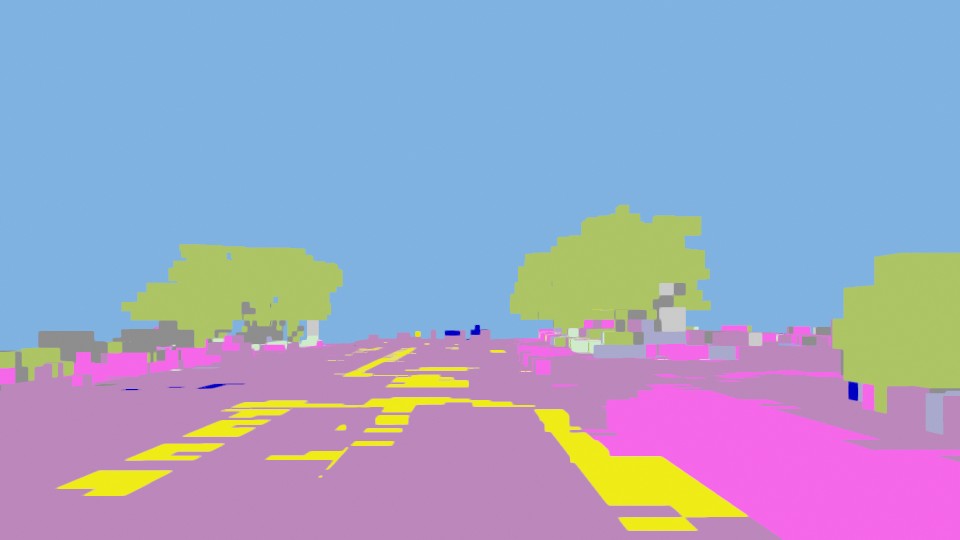}
{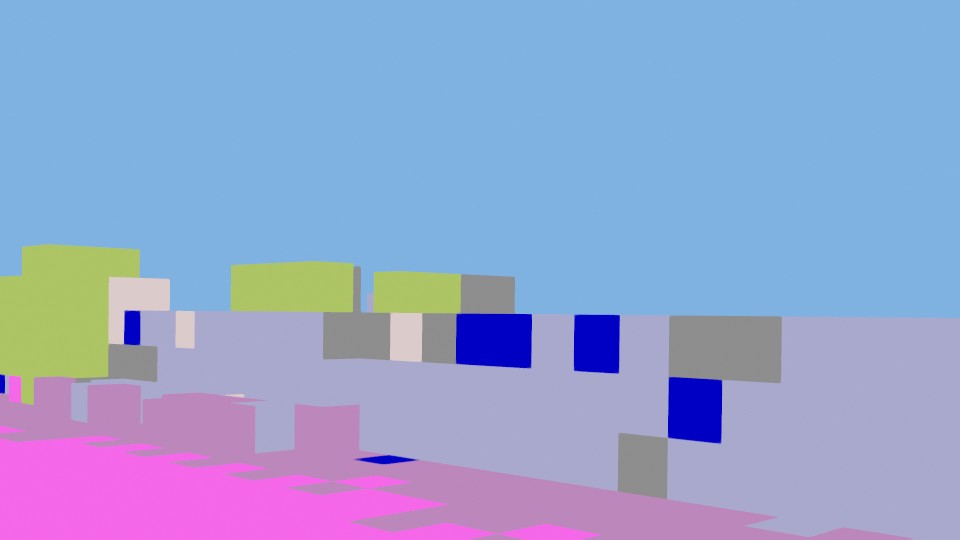}
{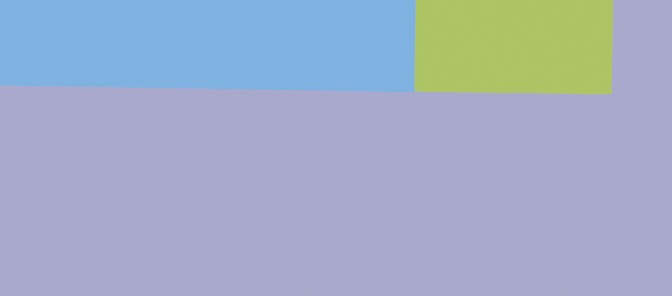}
\methrow{Ours(a)}
{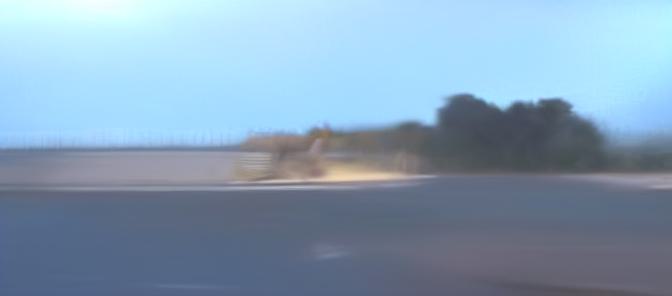}
{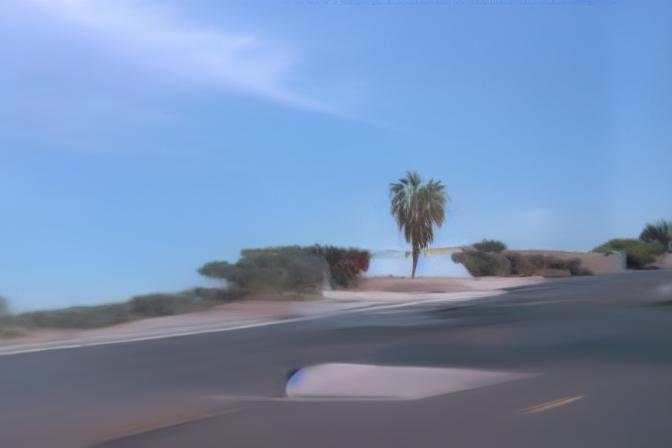}
{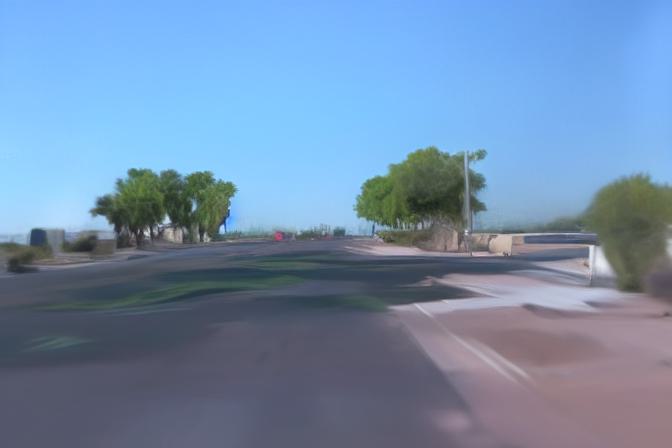}
{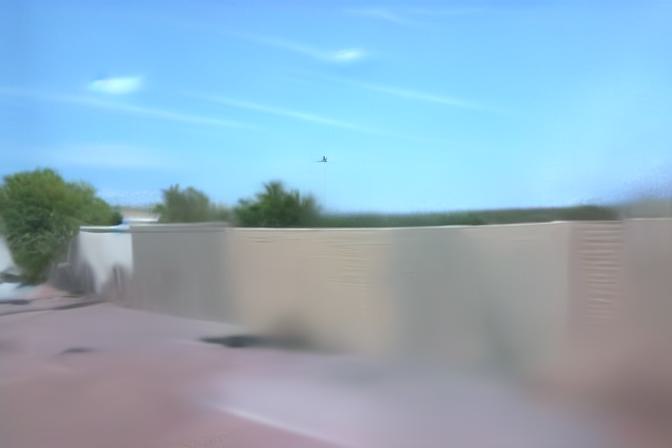}
{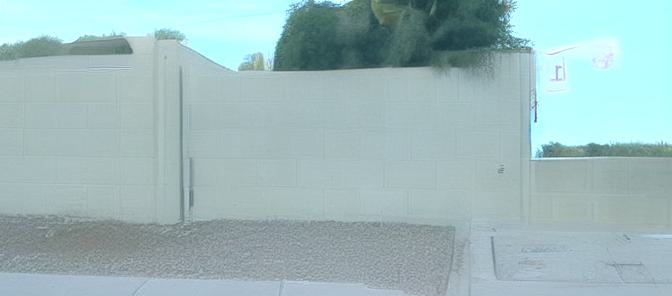}

% InfiniCube pair: semantics -> Infinicube
\semrow
{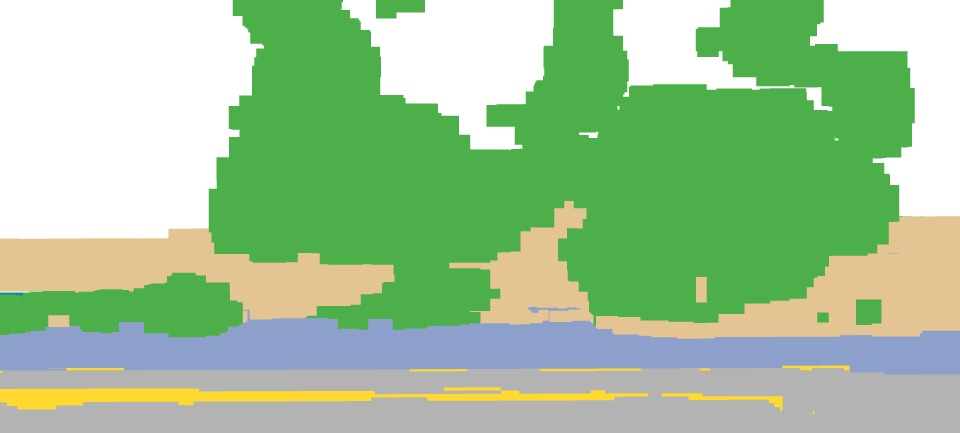}
{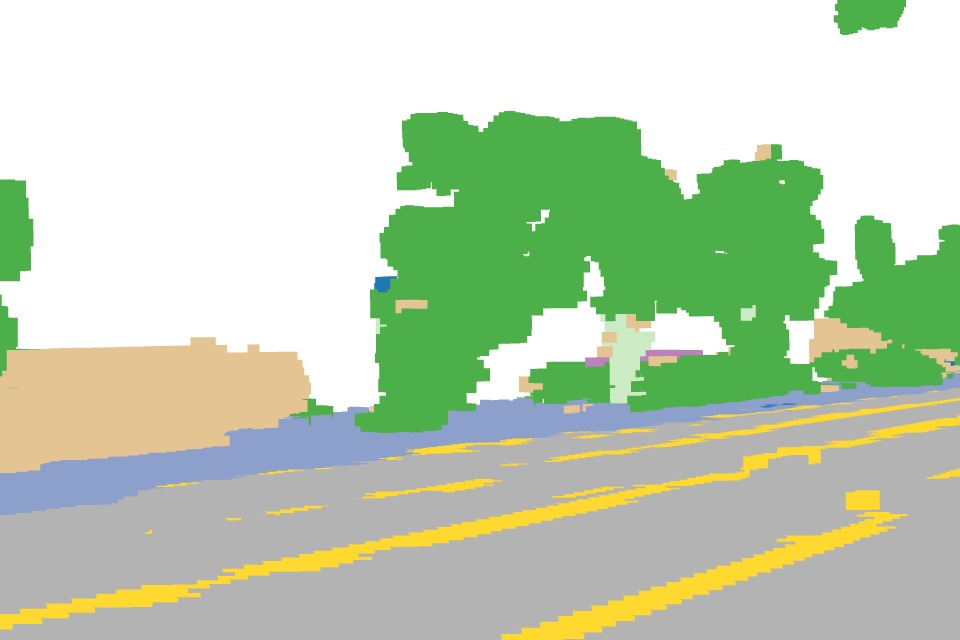}
{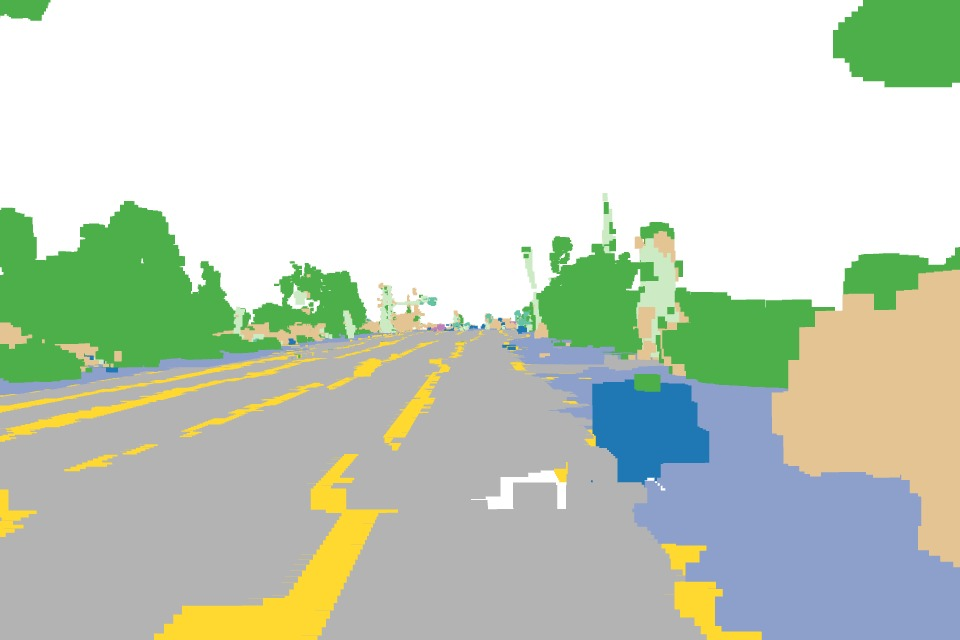}
{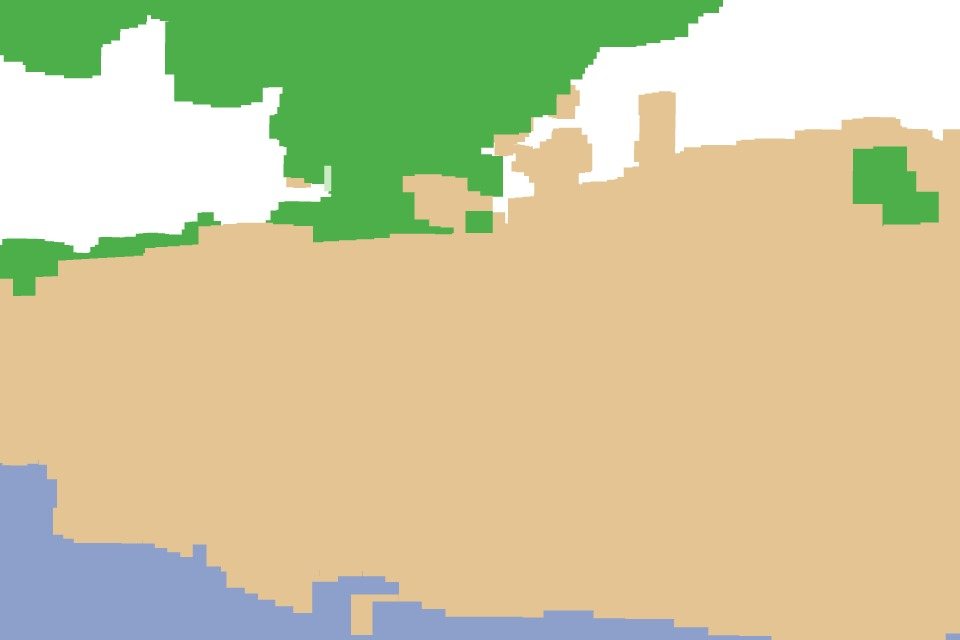}
{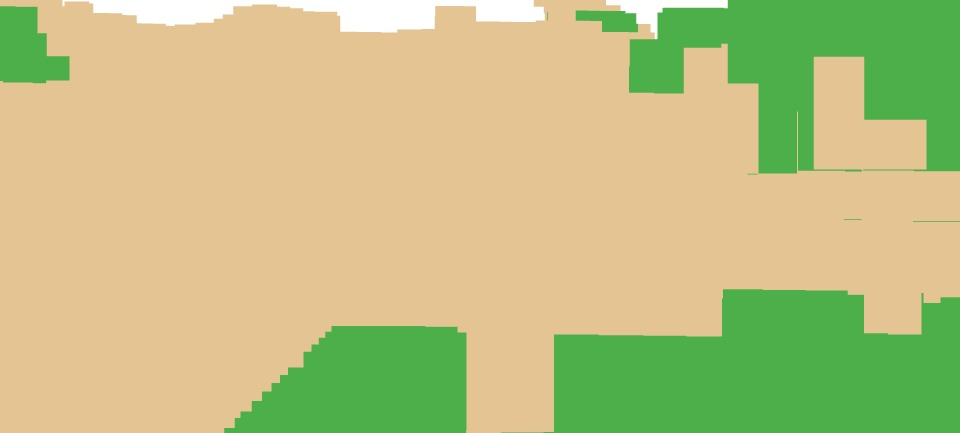}
\methrow{InfC(a)}
{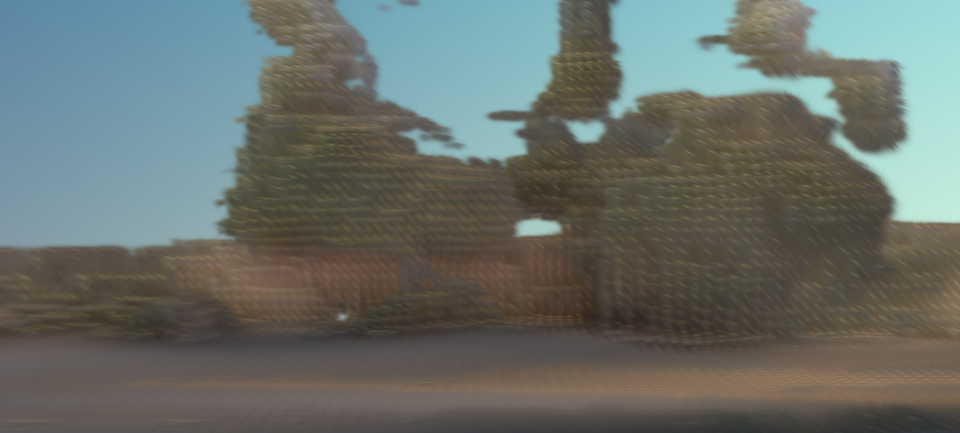}
{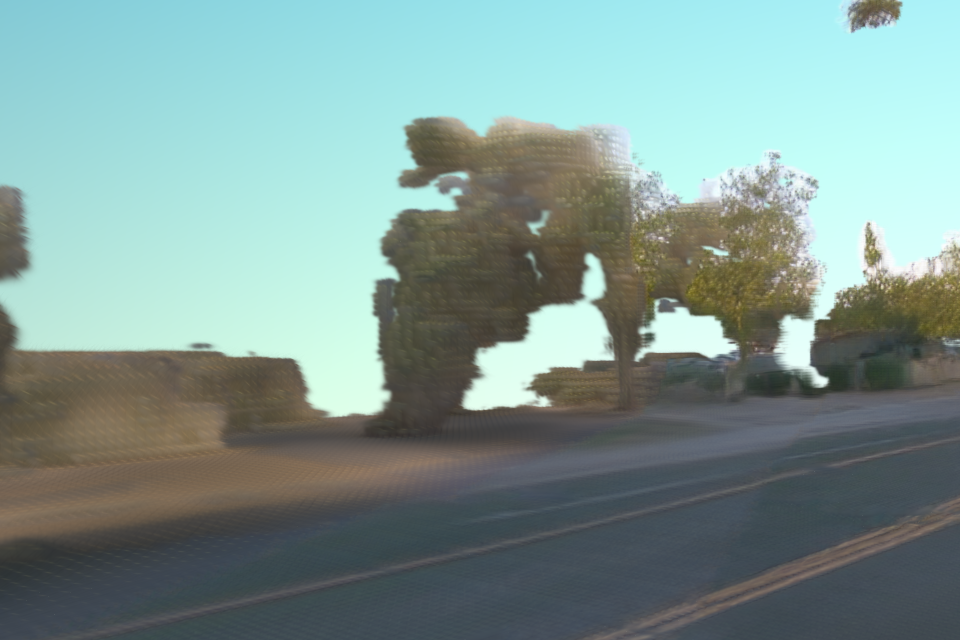}
{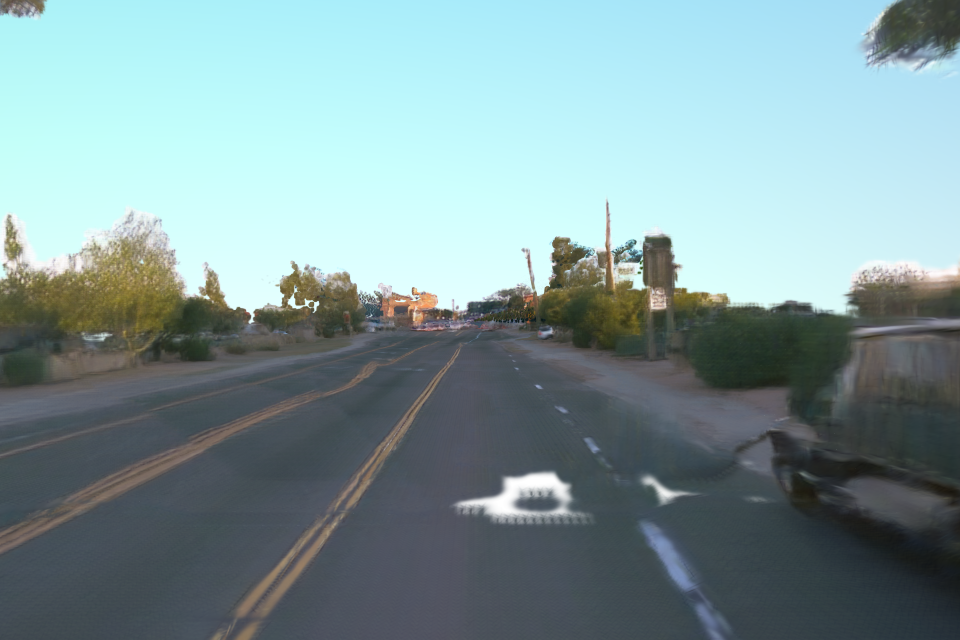}
{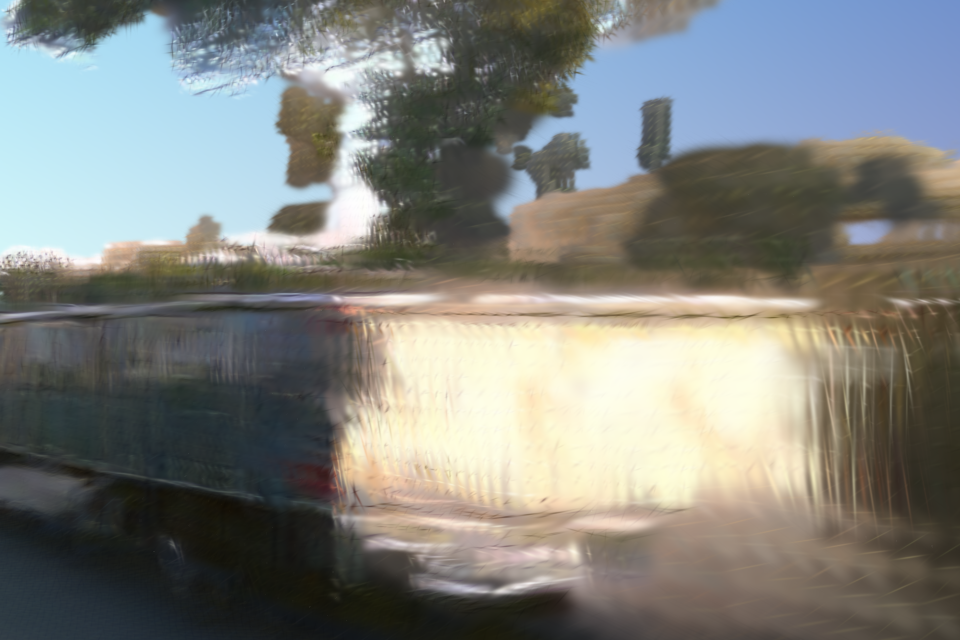}
{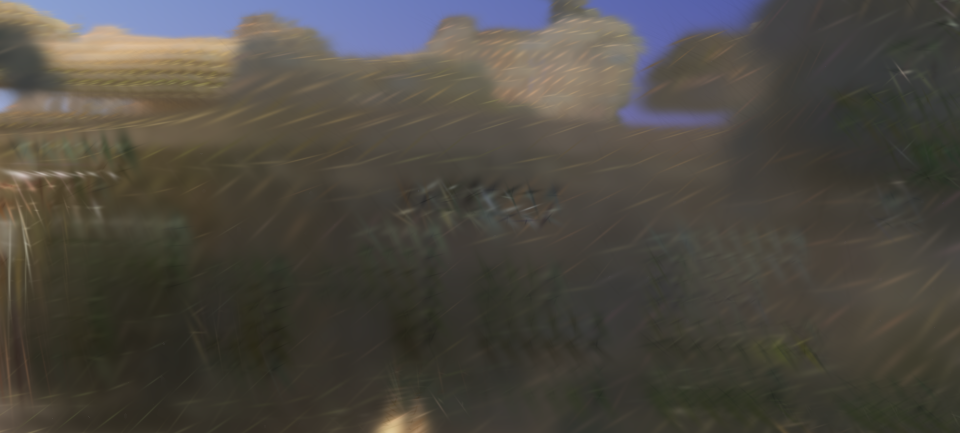}

% ===================== Scene 2 =====================
% Ours pair: semantics -> ours
\semrow
{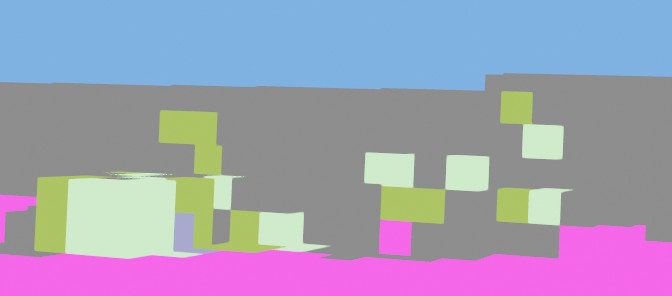}
{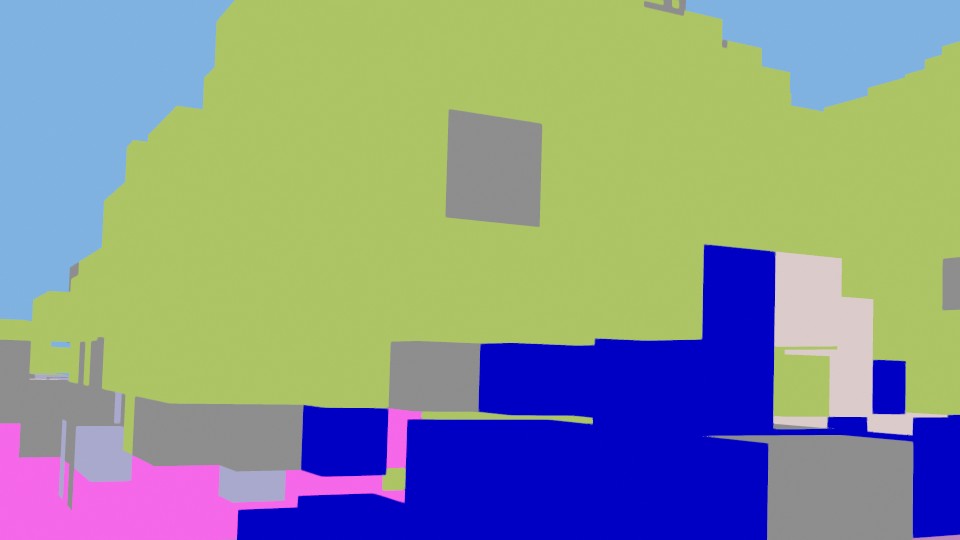}
{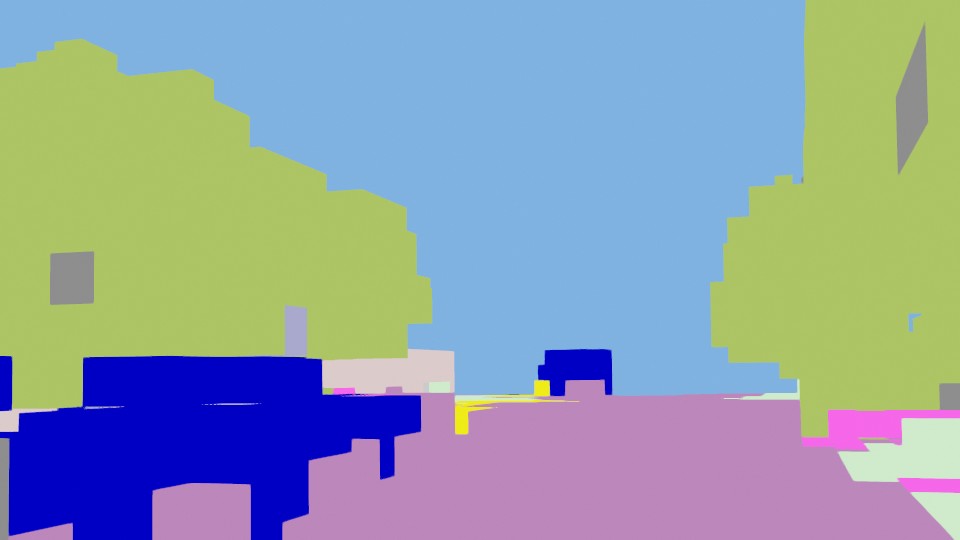}
{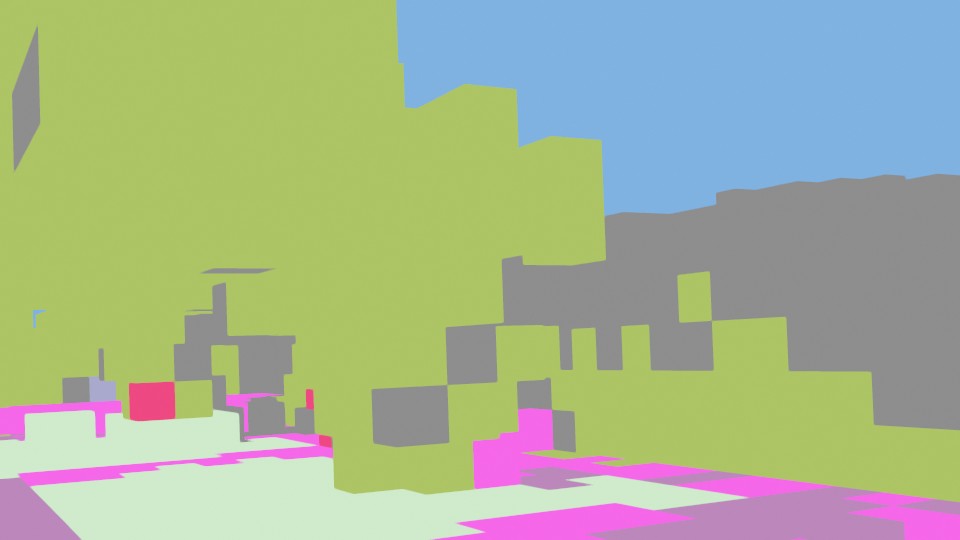}
{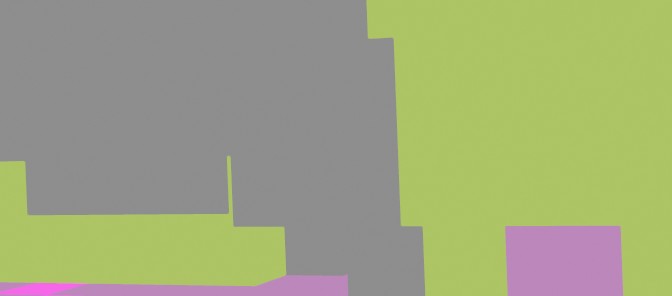}
\methrow{Ours(b)}
{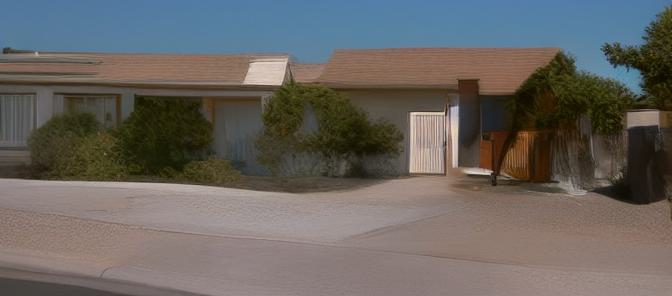}
{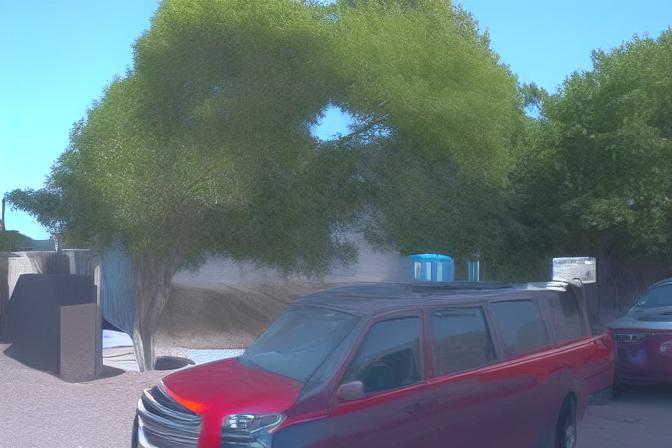}
{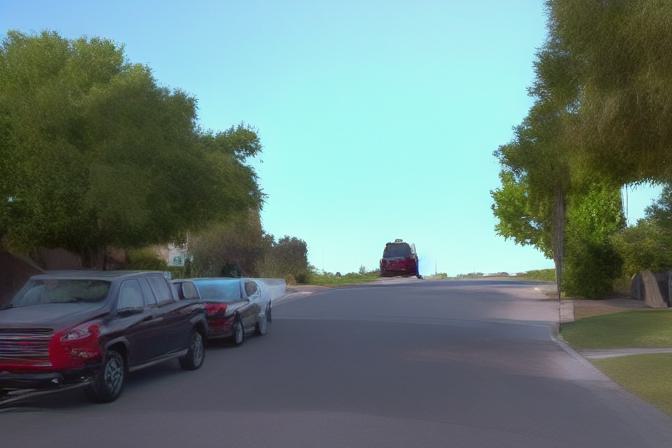}
{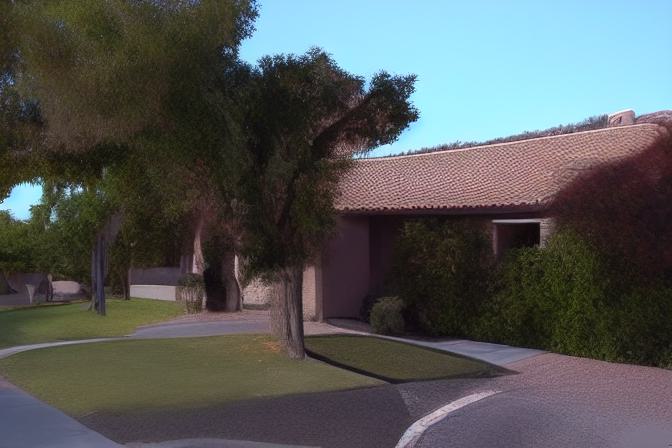}
{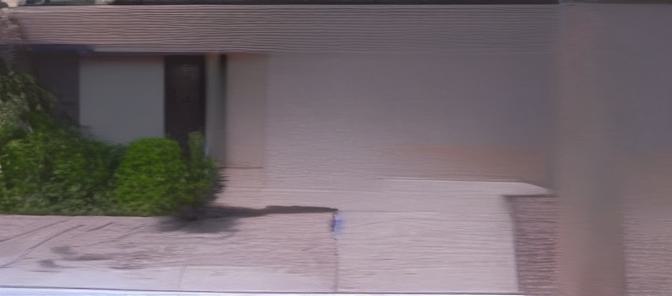}

% InfiniCube pair: semantics -> Infinicube
\semrow
{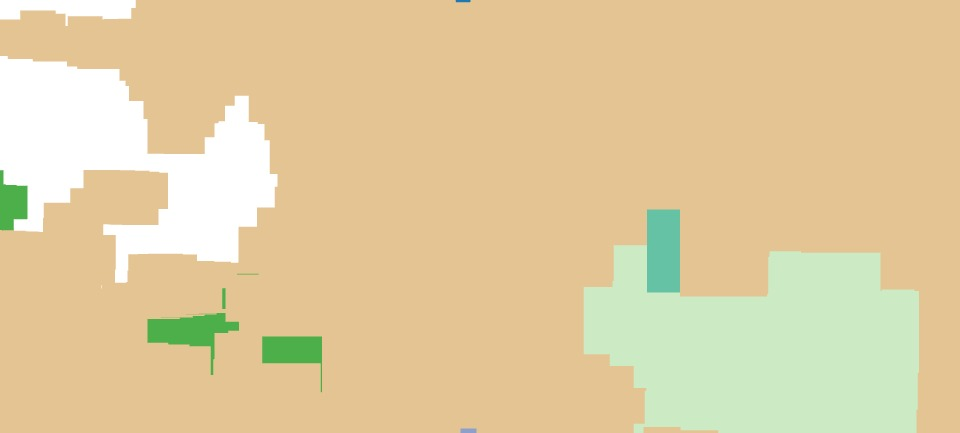}
{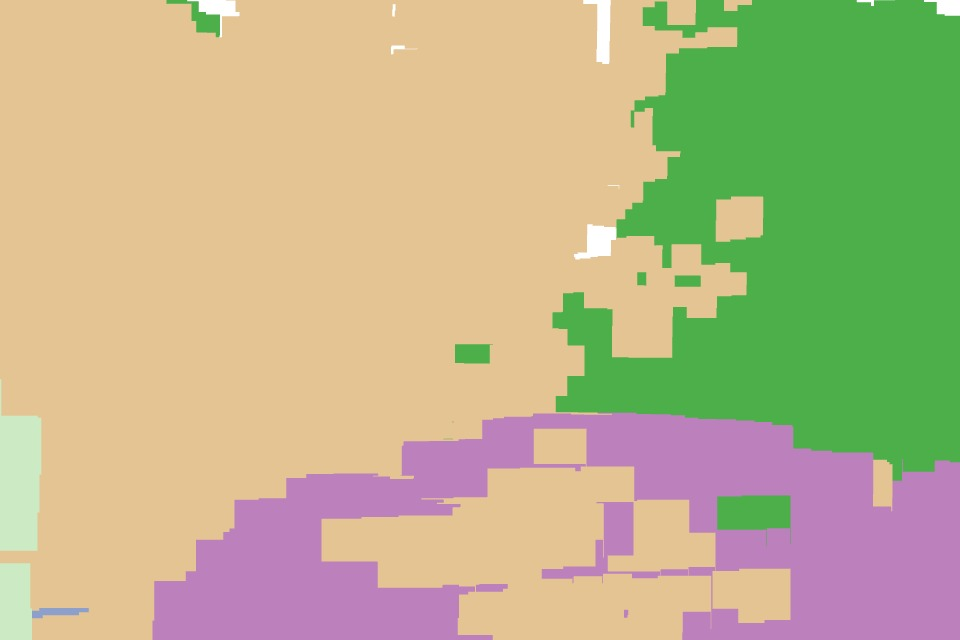}
{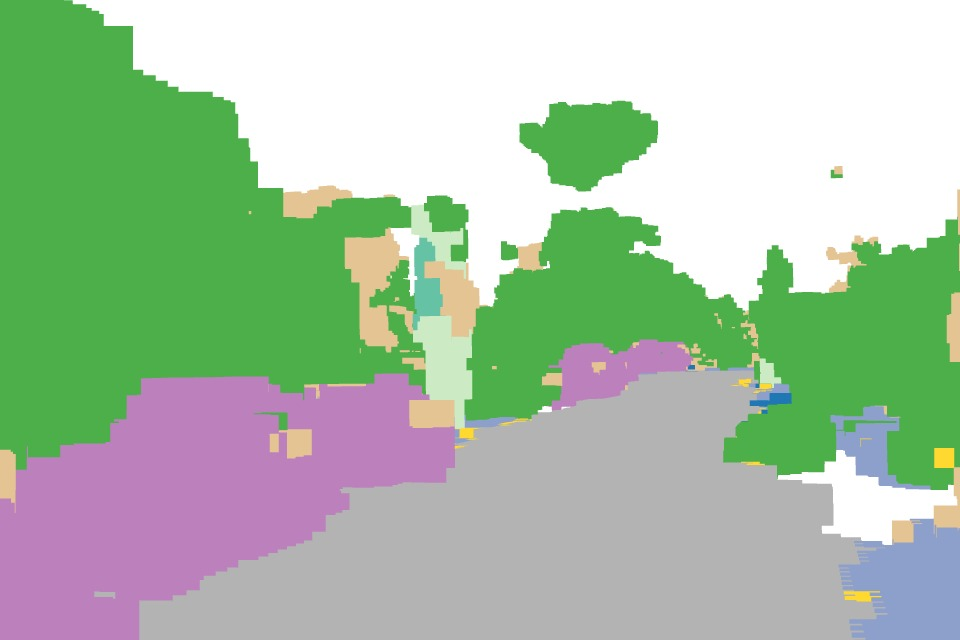}
{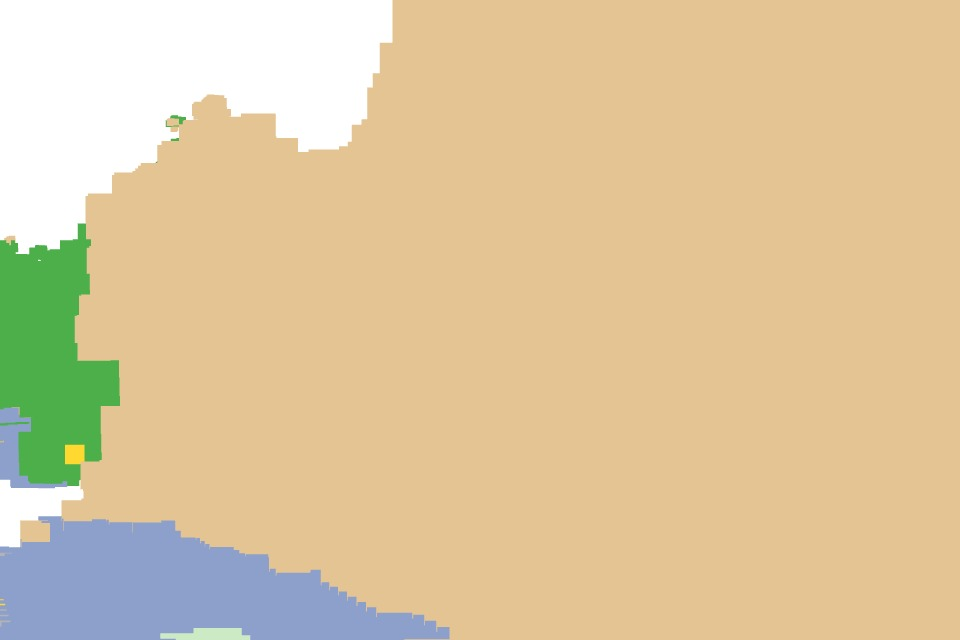}
{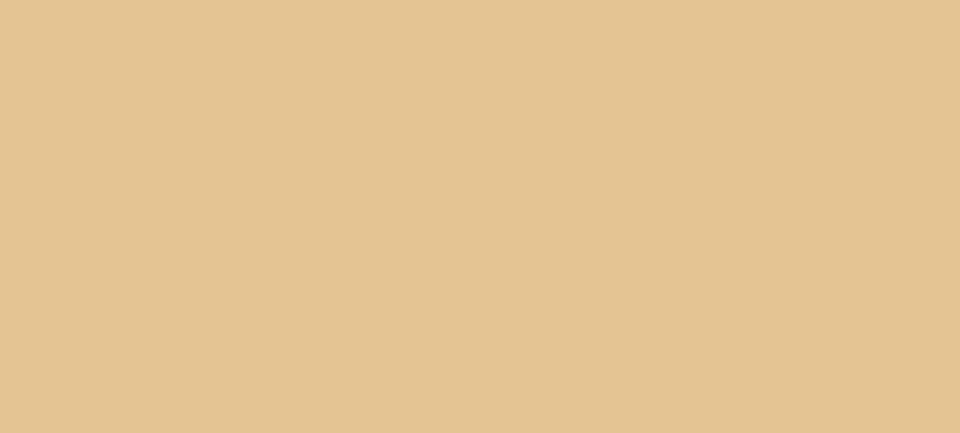}
\methrow{InfC(b)}
{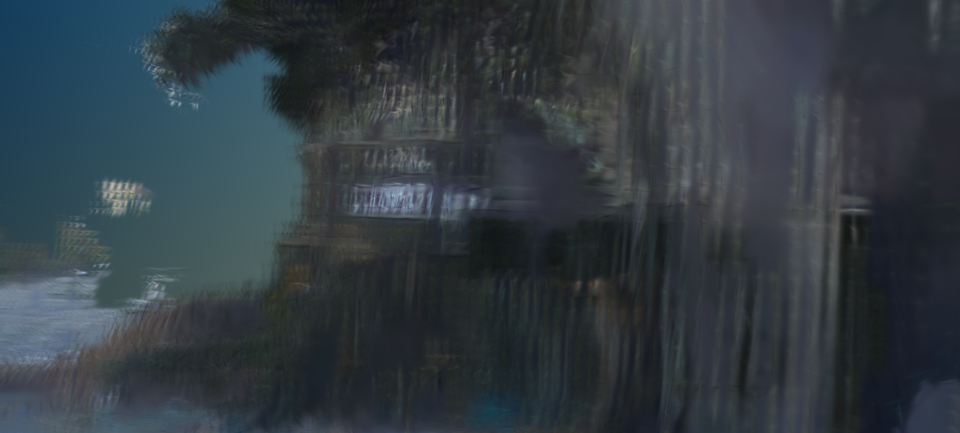}
{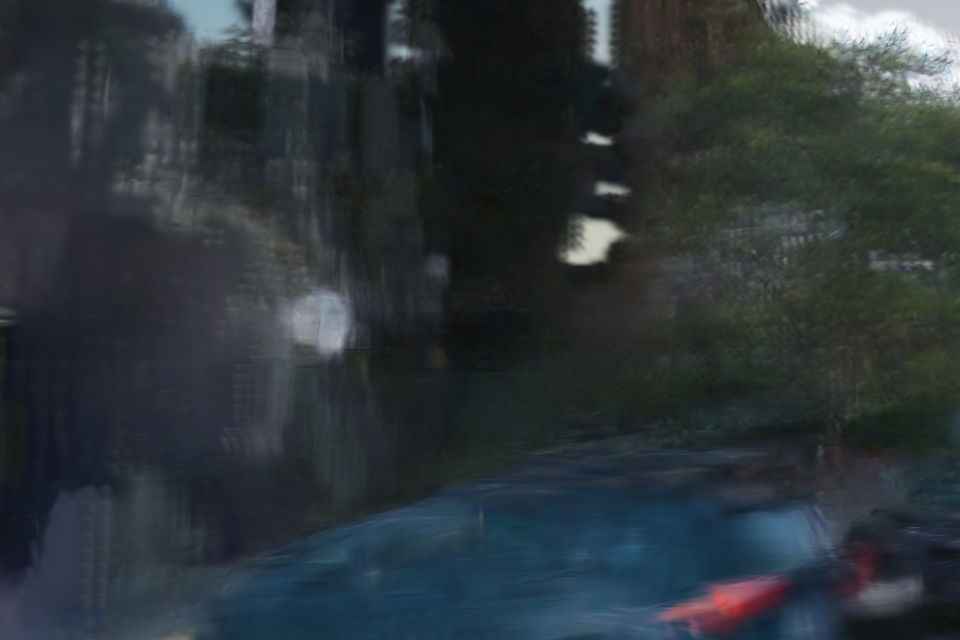}
{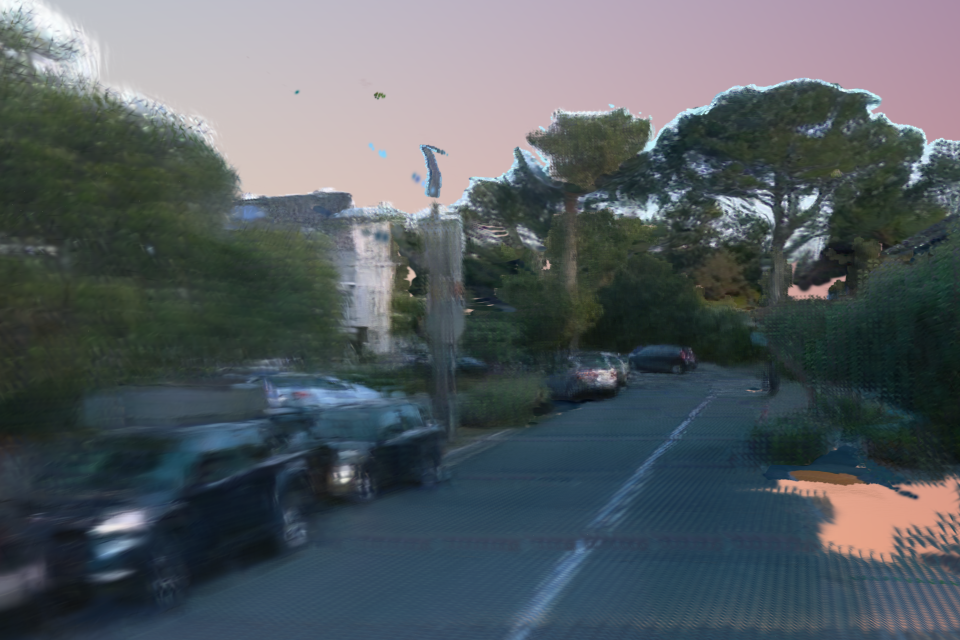}
{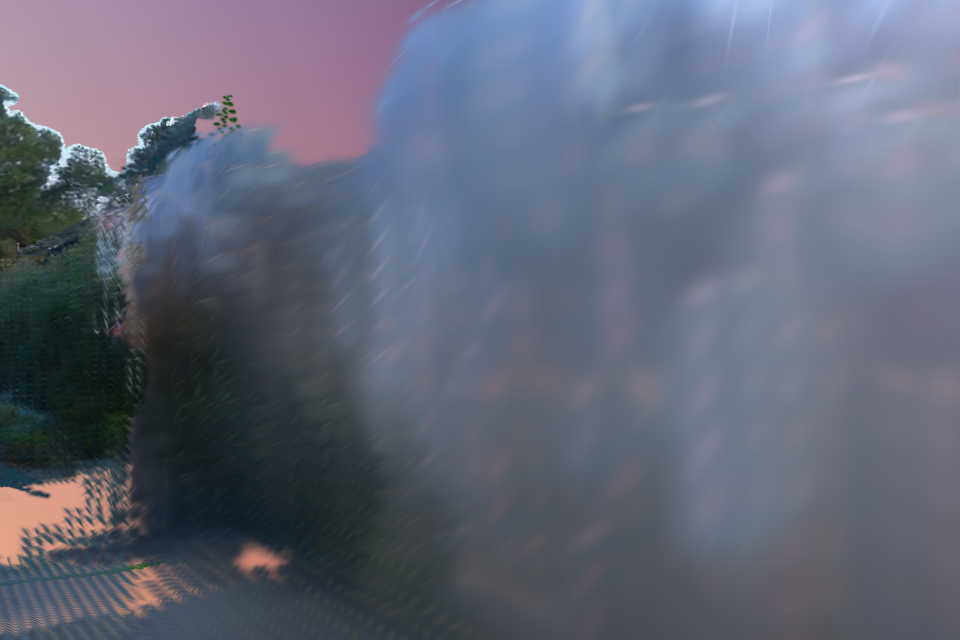}
{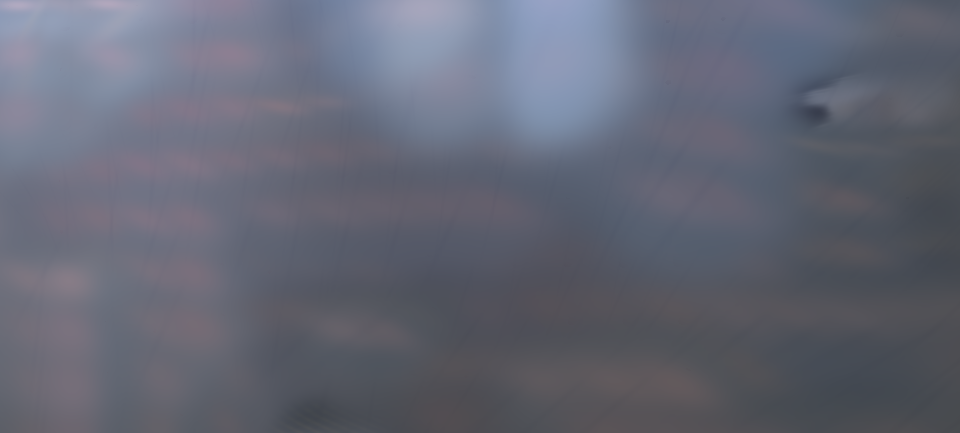}

\end{tabular}

\caption{\footnotesize \textbf{Qualitative comparison to InfiniCube~\cite{infinicube}.}
For two waymo scenes (a) and (b), we show five camera of the semantic conditioning and below the rendered generated scenes for Our method (Ours) and Infinicube (InfC).}
\label{fig:comparaison_baselines}
\vspace{-2mm}
\end{figure*}
\begin{figure*}[htbp]
\centering
\scriptsize
\setlength{\tabcolsep}{0pt}
\renewcommand{\arraystretch}{1.0}

% ---- knobs ----
\newcommand{\lblW}{0.12\textwidth}
\newcommand{\imgW}{0.176\textwidth}
\newcommand{\pairgap}{1pt}
\newcommand{\tightwithin}{-2pt}
\newcommand{\labelgap}{-1pt}

\newcommand{\rowlbl}[1]{%
\parbox[c]{\lblW}{{\tiny\textbf{#1}}}%
}
\newcommand{\fiveimgs}[5]{%
\includegraphics[width=\imgW]{#1}%
& \includegraphics[width=\imgW]{#2}%
& \includegraphics[width=\imgW]{#3}%
& \includegraphics[width=\imgW]{#4}%
& \includegraphics[width=\imgW]{#5}%
}

% <<< shrink factor (e.g., 0.9 = 90% size) >>>
\resizebox{0.9\textwidth}{!}{%
\begin{tabular}{@{}l@{}c@{}c@{}c@{}c@{}c@{}}
 &
\fiveimgs
{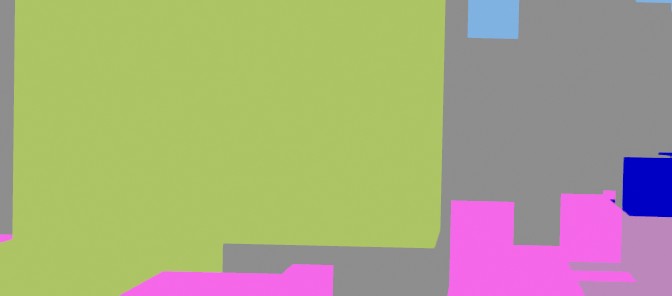}
{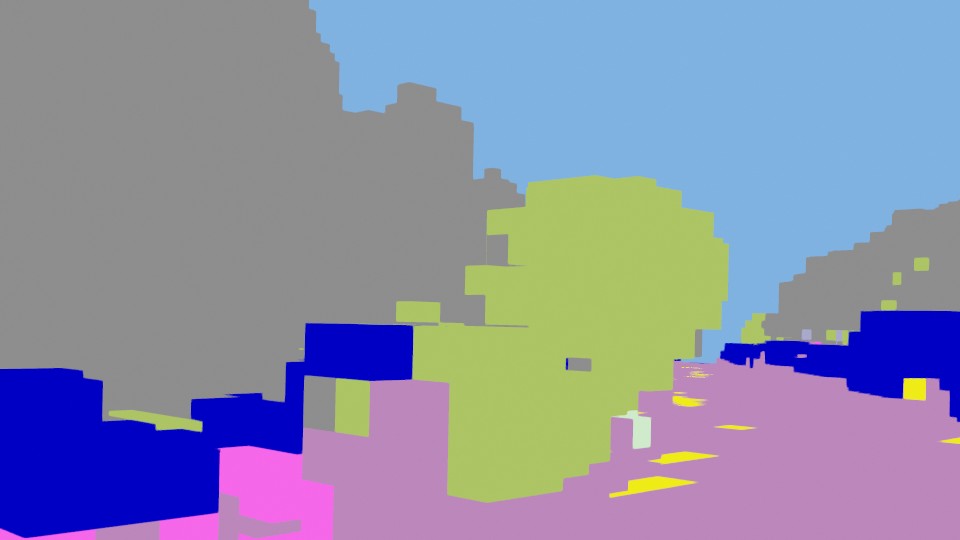}
{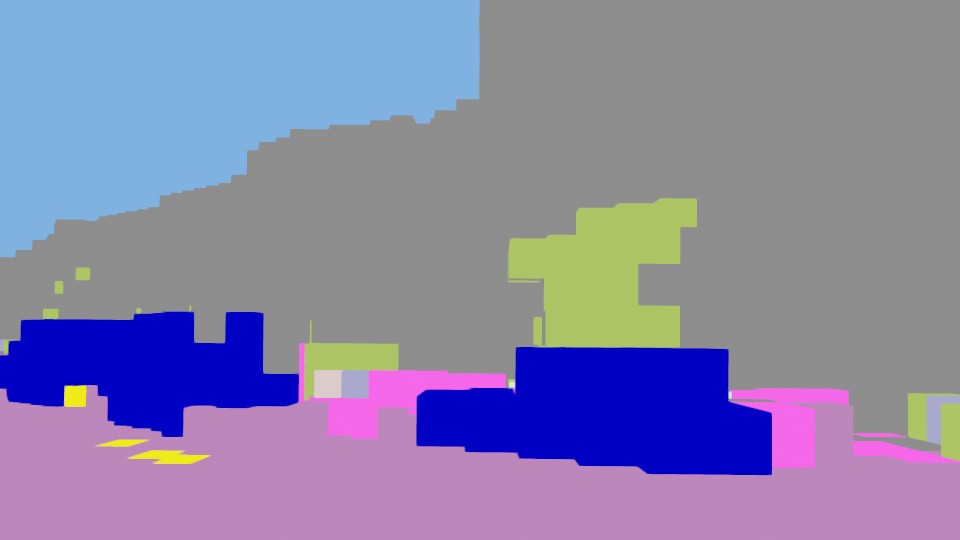}
{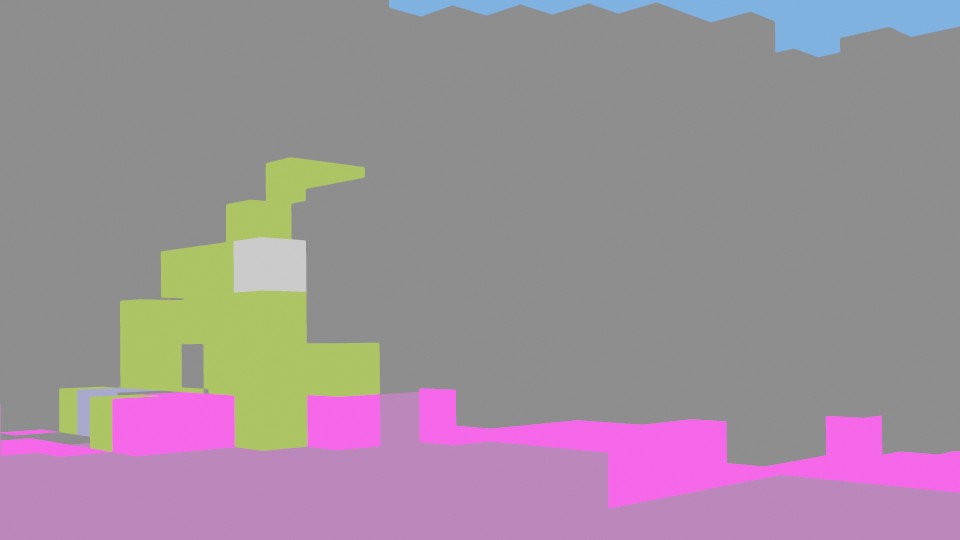}
{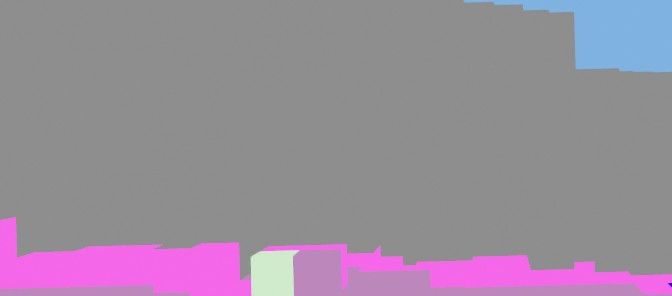}
\\[\tightwithin]

\rowlbl{(0)} &
\fiveimgs
{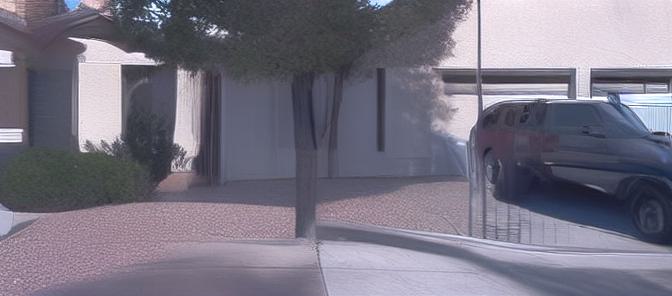}
{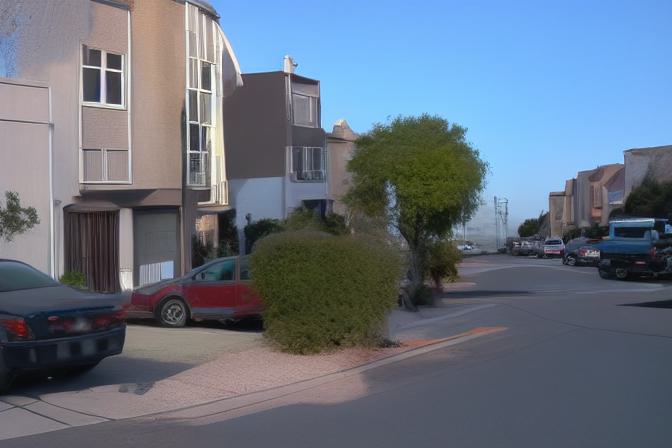}
{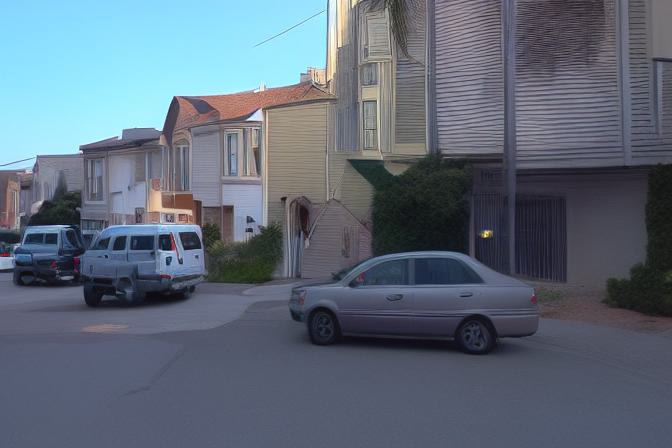}
{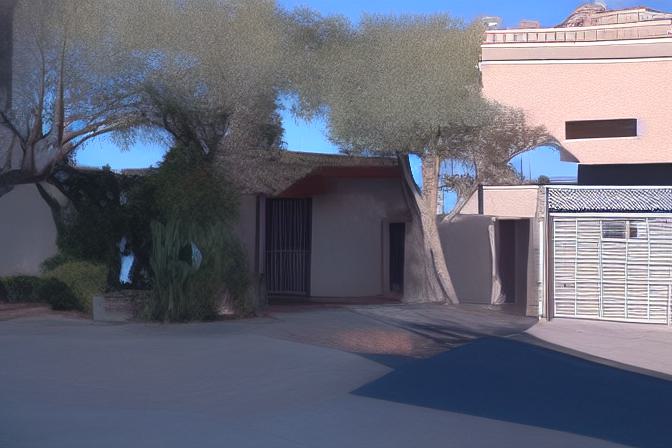}
{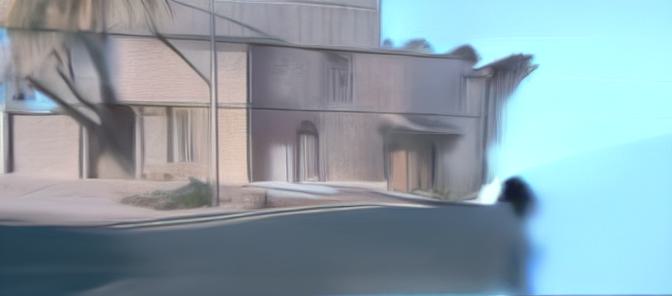}
\\[\labelgap]

\rowlbl{(1)} &
\fiveimgs
{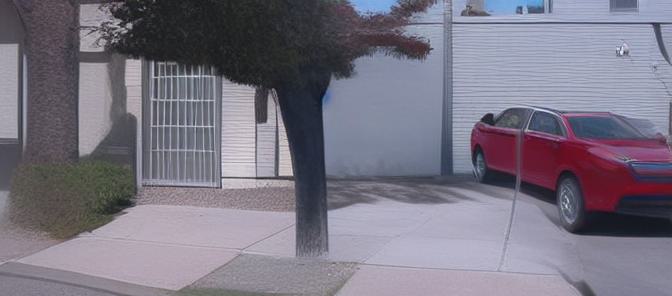}
{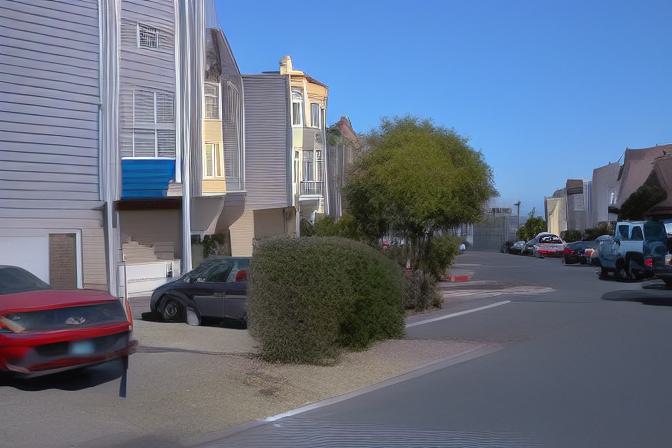}
{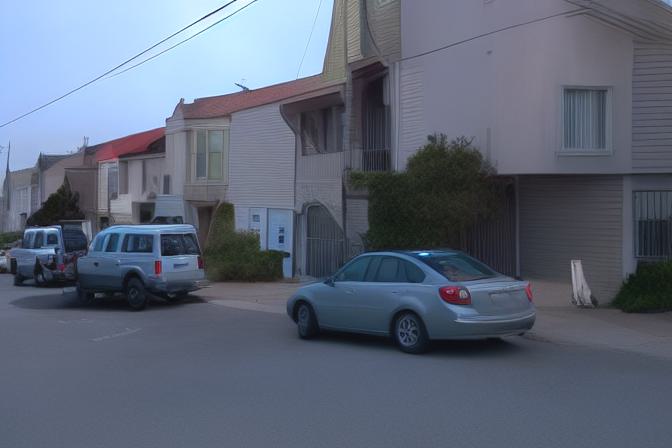}
{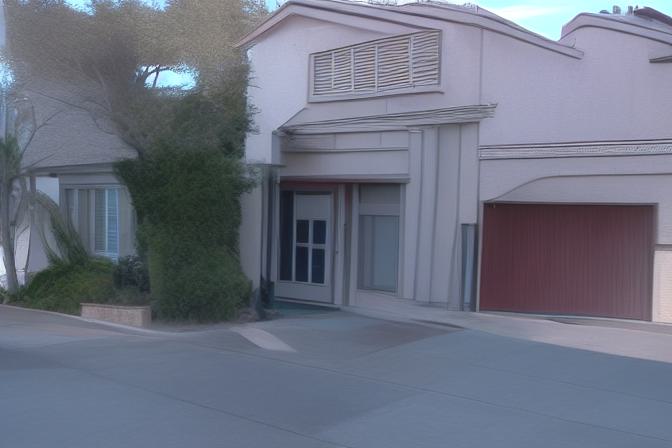}
{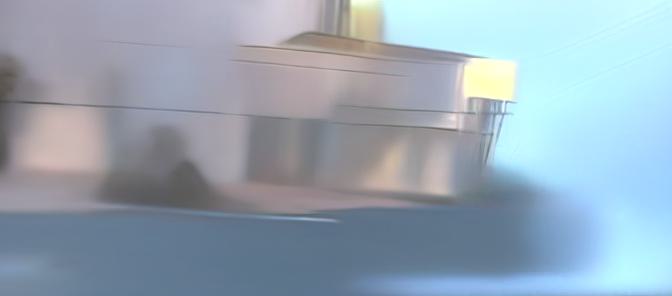}
\\[\pairgap]
\end{tabular}%
}

\caption{\textbf{Qualitative results: generative capabilities.}, the \textbf{top} row visualizes the conditioning signal, and the two rows below show \textbf{different generations} sampled from the same conditioning, highlighting output diversity under fixed structure.}
\label{fig:variance_gen}
\end{figure*}

\subsection{Qualitative Results}
We evaluate on \textbf{WOD} (\Cref{fig:results}) and \textbf{PandaSet} (\Cref{fig:results2}) datasets and present qualitative evidence that $\Sigma$-Voxfields Diffusion generates scenes that first adhere to the semantic voxel prior in global structure, secondly remain locally coherent near object regions and semantic boundaries, and finally are multi-view consistent under camera motion and multi-camera setup. On WOD, generations capture road topology and surrounding layout while maintaining stable appearance across consecutive views.
% ; under shifted camera poses, renderings remain compatible with a single underlying 3D scene rather than exhibiting view-dependent artifacts. 
We observe the same behavior on PandaSet, indicating that the method transfers well across datasets.
\Cref{fig:comparaison_baselines} further compares our method to InfiniCube~\cite{infinicube}. Across both scenes and multiple viewpoints, our generations better preserve scene geometry and layout consistency in a multi view setting. In particular, for side-camera viewpoints, our method produces more plausible and stable results, whereas InfiniCube degrades noticeably, consistent with its training primarily done with front-facing camera trajectories. As illustrated by the semantic voxel grids in \Cref{fig:comparaison_baselines}, InfiniCube uses a fine voxel conditioning (voxel size of 0.1m), whereas our coarser grid (voxel size of 0.6m) intentionally leaves more room for generative geometry, allowing greater shape variability.

Finally, \Cref{fig:variance_gen} illustrates the \textbf{generative capability} of our model by producing multiple plausible scenes from the same conditioning signal. While the semantic scaffold constrains the global layout, different samples vary in local geometry and appearance (e.g., surface details and textures) while remaining consistent with the conditioning and preserving multi-view coherence.

%\begin{figure*}[t]
%\centering
%\setlength{\tabcolsep}{2pt}

% ---- knobs ----
%\newcommand{\W}{0.30\linewidth}
%\newcommand{\Him}{0.10\textheight}
%\newcommand{\imgbox}[1]{\includegraphics[width=\W,height=\Him,keepaspectratio]{#1}}

% begin{tabular}{@{}ccc@{}}
%\imgbox{figures/zooming_out/run_11.png} &
%\imgbox{figures/zooming_out/run_3.png} &
%\imgbox{figures/zooming_out/run_16.png} \\
%\end{tabular}

%\vspace{-1.5mm}
%\caption{our 3D buffer }
%\label{fig:cameras_3d}
%\vspace{-2mm}
%\end{figure*}
\subsection{Quantitative Results}
We report \textbf{inference characteristics} and \textbf{rendered-view image quality} measured by FID/KID in Table~\ref{tab:fid_kid}. To compute these metrics, we render images from generated scenes and compute FID/KID against real frames from the corresponding ground-truth sequences, evaluated on \emph{seen} views (ground-truth poses) and \emph{novel} views (poses shifted from the original trajectory).

\paragraph{Rendered-view image quality.}
Table~\ref{tab:fid_kid} shows that our method is competitive on seen views and improves robustness under viewpoint shifts. Compared to GEN3C~\cite{gen3c}, \textbf{ours} yields substantially better FID/KID on both splits, consistent with more coherent 3D structure from our explicit 3D generation. Notably, our gains are most pronounced on \emph{novel} views, where viewpoint shifts expose geometric inconsistencies in methods that are trained with restricted view coverage (e.g., front views), such as InfiniCube. In contrast, our $\Sigma$-Voxfield representation better preserves geometry across pose perturbations, leading to improved shifted-view rendering.

\paragraph{Inference cost and scalability.}
Table~\ref{tab:fid_kid} summarizes inference memory footprints. Our method uses 8\,GB VRAM, versus 75\,GB for InfiniCube~\cite{infinicube} and 43\,GB for GEN3C~\cite{gen3c}. Runtime remains practical: we generate a scene in $\sim$20 minutes, comparable to InfiniCube at similar scale.
\begin{table}[htbp]
\centering
\scriptsize
\setlength{\tabcolsep}{1.6pt}
\renewcommand{\arraystretch}{1.02}
\begin{tabular}{lcccccc}
\toprule
Method &
FID (seen)$\downarrow$ &
FID (novel)$\downarrow$ &
KID (seen)$\downarrow$ &
KID (novel)$\downarrow$ &
Min. VRAM \\
\midrule
InfiniCube~\cite{infinicube} & 84.14 & 99.13 & \textbf{0.03} & \textbf{0.06} & 75 GB \\
GEN3C~\cite{gen3c}           & 113.27 & 117.63 & 0.08 & 0.09  & 43 GB \\
% \rowcolor{methodgray}
% \textbf{Ours (SVD)}           & 145.04 & --     & 0.12 & --     &  25 GB \\
\rowcolor{methodgray}
\textbf{Ours (ASD)}        & \textbf{81.98}  & \textbf{89.20}  & 0.05 & \textbf{0.06}   &  \textbf{8 GB} \\
\bottomrule
\end{tabular}

\caption{
Quantitative Evaluation of Generation Quality for 3D Scene Generation with our proposed method and existing approaches.
}
\label{tab:fid_kid}

\end{table}

\vspace{-5mm}

\vspace{-10mm}

% \begin{figure}[tbh]
% \centering
% \newlength{\abH}\newlength{\capH}
% \setlength{\abH}{0.18\textheight}
% \setlength{\capH}{1.4ex}

% \begin{minipage}[t][\abH][b]{0.495\linewidth}
%   \centering
%   % smaller plot to match table footprint
%   \includegraphics[width=0.68\linewidth,height=\dimexpr\abH-\capH\relax,keepaspectratio]{figures/ablation_waymo_N.png}
%   \parbox[t][\capH][t]{\linewidth}{\centering\scriptsize\textbf{(a)} Points per voxel.}
% \end{minipage}\hfill
% \begin{minipage}[t][\abH][b]{0.495\linewidth}
%   \centering
%   \scriptsize
%   \setlength{\tabcolsep}{3pt}
%   \renewcommand{\arraystretch}{1.06}
%   \vfill
%   \begin{tabular}{l c c}
%     \toprule
%     \textbf{Method} & \textbf{F3D$\downarrow$} & \textbf{MMD$\downarrow$} \\
%     \midrule
%     w/o sem. cond. & 3.927 & 0.105\\
%     w/o ordering   & 3.585 & 0.093\\
%     Ours           & 3.523 & 0.091\\
%     \bottomrule
%   \end{tabular}
%   \vfill
%   \parbox[t][\capH][t]{\linewidth}{\centering\scriptsize\textbf{(b)} Diffusion ablations.}
% \end{minipage}
% \vspace{-1.2mm}
% \caption{\textbf{Ablations.} (a) CD to Waymo mesh (voxel size $0.6\,\mathrm{m}$) saturates around $N{=}20$ ($\approx0.025\,\mathrm{m}$) and stays below the splat radius ($r{=}0.04\,\mathrm{m}$). (b) Feature-space ablation of our diffusion model.}
% \label{fig:ablation_combined}
% \vspace{-2mm}
% \end{figure}

\begin{table}[htbp]
\centering
\begin{tabular}{lcc}
\toprule
\textbf{Method} & \textbf{E3D$\downarrow$} & \textbf{MMDS$\downarrow$} \\
\midrule
w/o sem. cond. & 3.927 & 0.105 \\
w/o ordering   & 3.585 & 0.093 \\
Ours           & 3.523 & 0.091 \\
\bottomrule
\end{tabular}
\caption{Feature-space ablation of our 3D diffusion model.}
\label{fig:ablation_combined}
\end{table}

\subsection{Ablation Studies}
% \paragraph{Diffusion model ablations.}
We evaluate the $\Sigma$-Voxfield diffusion model in a learned 3D feature space, since image metrics (e.g., FID) do not apply to 3D generation. We train a PointNet++ semantic segmentation network on labeled point clouds used as a feature extractor to evaluated using F3D and MMD. Tab.~\ref{fig:ablation_combined} shows the full model performs best: removing semantic conditioning degrades F3D/MMD, and disabling point ordering also hurts performance, indicating both are important.

% \paragraph{Sampled Points per voxel.}
% We choose $N{=}20$ surface samples per voxel based on geometric fidelity. Using the Waymo mesh as reference, we measure Chamfer Distance between sampled points and the ground-truth surface as a function of $N$ (voxel size $0.6\,\mathrm{m}$). As shown in Fig.~\ref{fig:ablation_combined}(a), CD quickly decreases and saturates around $N{\approx}20$ at $\sim0.025\,\mathrm{m}$, which is below the splat radius $r{=}0.04\,\mathrm{m}$. Larger $N$ yields diminishing returns while increasing DiT input size linearly ($6N$ scalars per voxel), so we adopt $N{=}20$ as a good accuracy--efficiency trade-off.

\subsection{Applications}

\begin{figure}[htbp]
\centering
\setlength{\tabcolsep}{0pt}
\renewcommand{\arraystretch}{0}

\begin{tabular}{@{}c@{}c@{}c@{}c@{}}
% ---------------- Row 1 (Before) ----------------
\begin{tikzpicture}
  \node[inner sep=0,outer sep=0] (img) {\includegraphics[width=0.16\linewidth,height=0.09\linewidth]{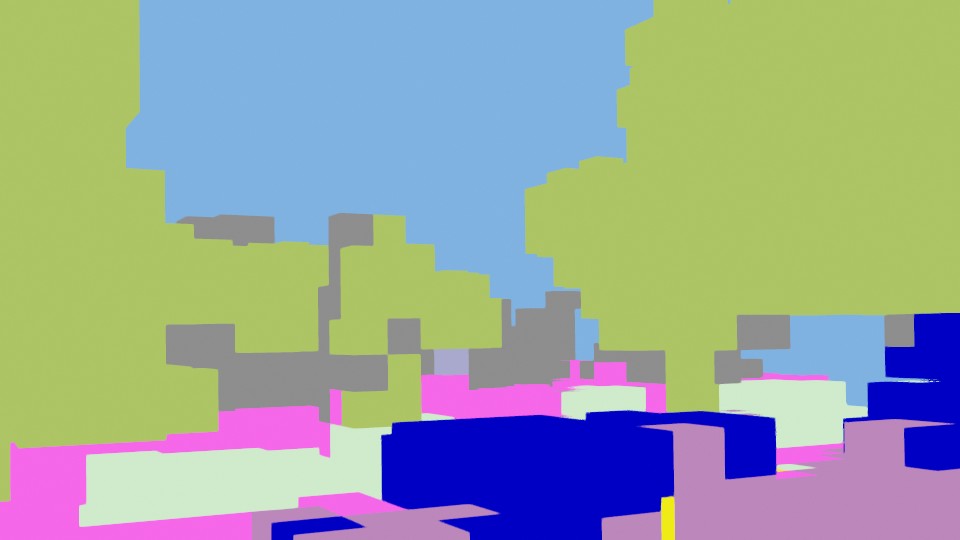}};
  \begin{scope}[shift={(img.south west)}, x={(img.south east)}, y={(img.north west)}]
    \draw[red,line width=0.3pt] (0.3,0) rectangle (0.8,0.4);
    \draw[red,line width=0.3pt] (0.85,0.1) rectangle (1,0.4);
  \end{scope}
\end{tikzpicture}
&
\begin{tikzpicture}
  \node[inner sep=0,outer sep=0] (img) {\includegraphics[width=0.16\linewidth,height=0.09\linewidth]{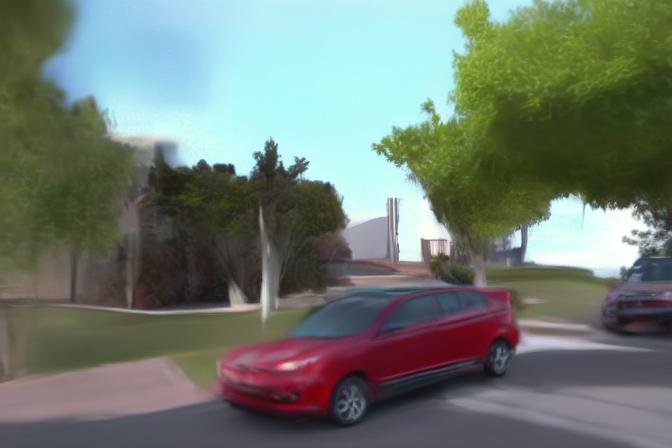}};
  \begin{scope}[shift={(img.south west)}, x={(img.south east)}, y={(img.north west)}]
    \draw[red,line width=0.3pt] (0.3,0) rectangle (0.8,0.4);
    \draw[red,line width=0.3pt] (0.85,0.1) rectangle (1,0.4);
  \end{scope}
\end{tikzpicture}

&
\begin{tikzpicture}
  \node[inner sep=0,outer sep=0] (img) {\includegraphics[width=0.16\linewidth,height=0.09\linewidth]{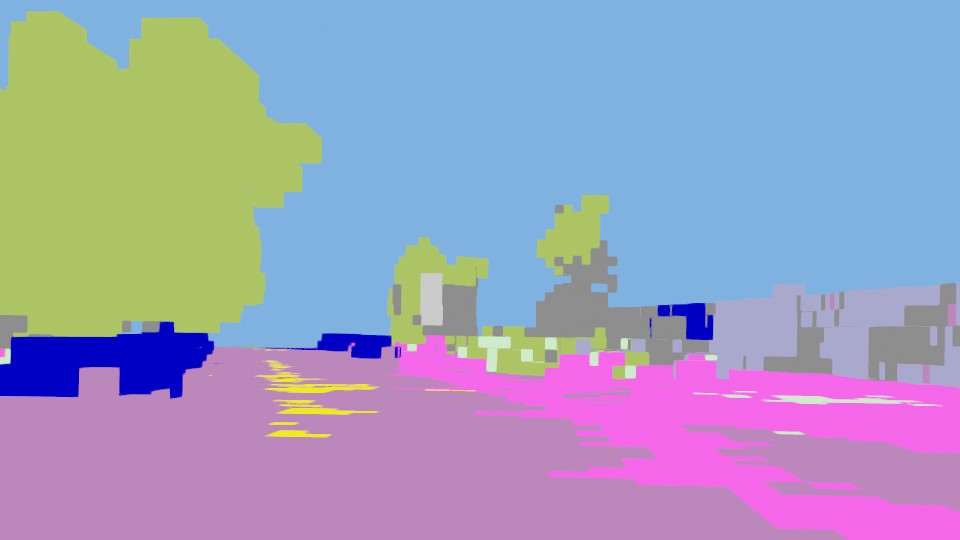}};
  \begin{scope}[shift={(img.south west)}, x={(img.south east)}, y={(img.north west)}]
    \draw[red,line width=0.3pt] (0.000,0.265) rectangle (0.222,0.404);
    \draw[red,line width=0.3pt] (0.309,0.339) rectangle (0.406,0.383);
    ;
  \end{scope}
\end{tikzpicture}
&
\begin{tikzpicture}
  \node[inner sep=0,outer sep=0] (img) {\includegraphics[width=0.16\linewidth,height=0.09\linewidth]{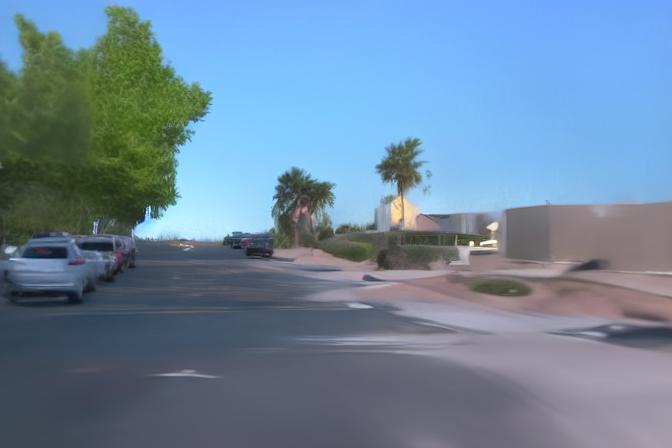}};
  \begin{scope}[shift={(img.south west)}, x={(img.south east)}, y={(img.north west)}]
    \draw[red,line width=0.3pt] (0.000,0.265) rectangle (0.222,0.6);
    \draw[red,line width=0.3pt] (0.309,0.4) rectangle (0.406,0.5);
    ;
  \end{scope}
\end{tikzpicture}

\\

\begin{tikzpicture}
  \node[inner sep=0,outer sep=0] (img) {\includegraphics[width=0.16\linewidth,height=0.09\linewidth]{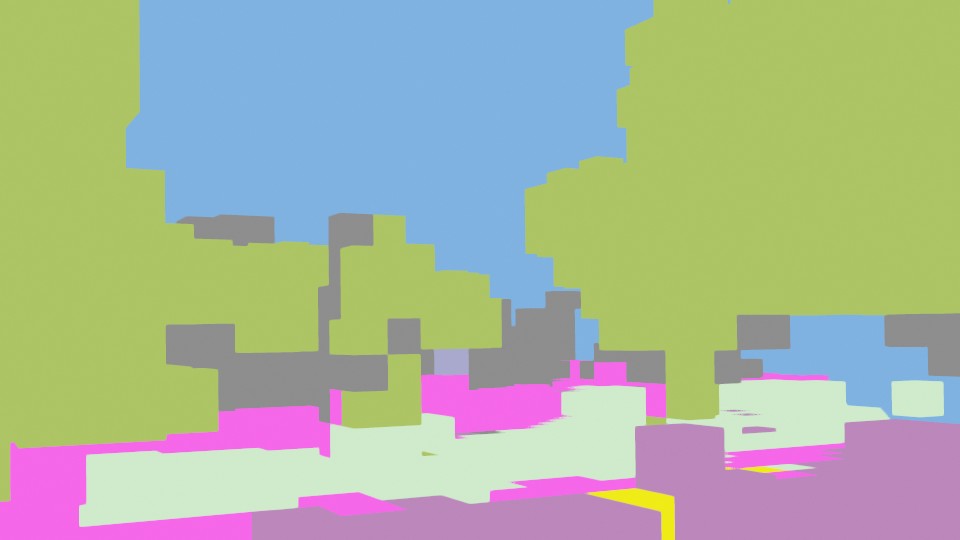}};
  \begin{scope}[shift={(img.south west)}, x={(img.south east)}, y={(img.north west)}]
    \draw[red,line width=0.3pt] (0.3,0) rectangle (0.8,0.4);
    \draw[red,line width=0.3pt] (0.85,0.1) rectangle (1,0.4);
  \end{scope}
\end{tikzpicture}
&
\begin{tikzpicture}
  \node[inner sep=0,outer sep=0] (img) {\includegraphics[width=0.16\linewidth,height=0.09\linewidth]{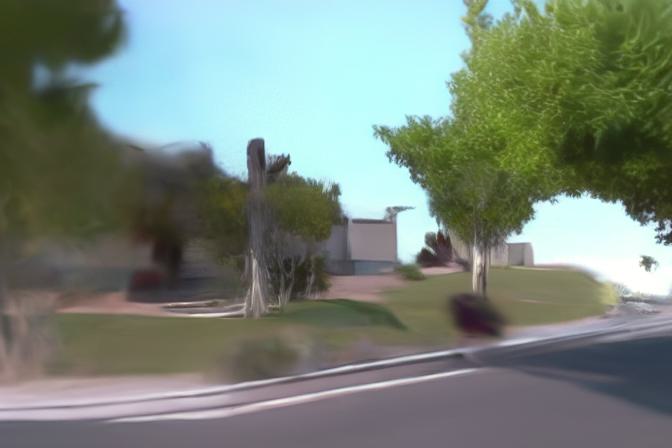}};
  \begin{scope}[shift={(img.south west)}, x={(img.south east)}, y={(img.north west)}]
    \draw[red,line width=0.3pt] (0.3,0) rectangle (0.8,0.4);
    \draw[red,line width=0.3pt] (0.85,0.1) rectangle (1,0.4);
  \end{scope}
\end{tikzpicture}
&
\begin{tikzpicture}
  \node[inner sep=0,outer sep=0] (img) {\includegraphics[width=0.16\linewidth,height=0.09\linewidth]{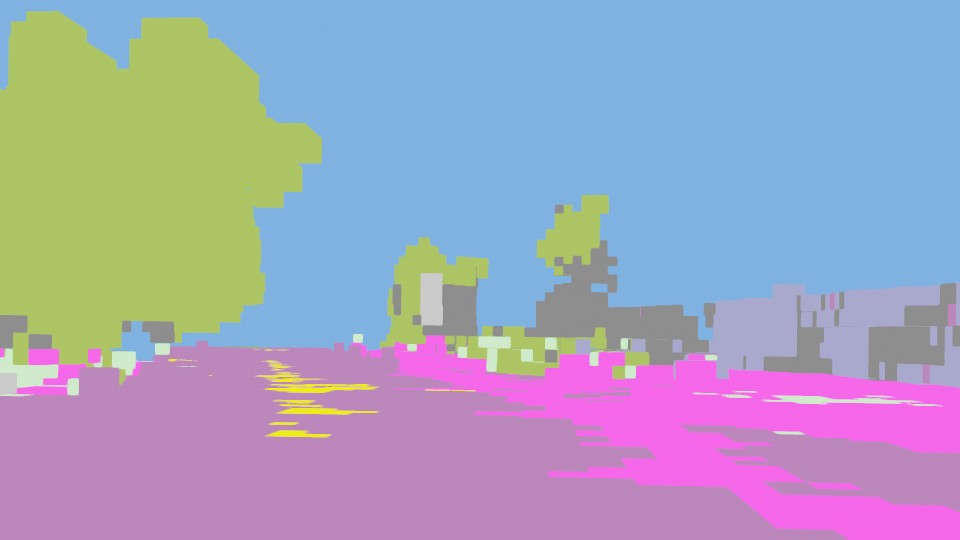}};
  \begin{scope}[shift={(img.south west)}, x={(img.south east)}, y={(img.north west)}]
    \draw[red,line width=0.3pt] (0.000,0.265) rectangle (0.222,0.404);
    \draw[red,line width=0.3pt] (0.309,0.339) rectangle (0.406,0.383);
    ;
  \end{scope}
\end{tikzpicture}
&
\begin{tikzpicture}
  \node[inner sep=0,outer sep=0] (img) {\includegraphics[width=0.16\linewidth,height=0.09\linewidth]{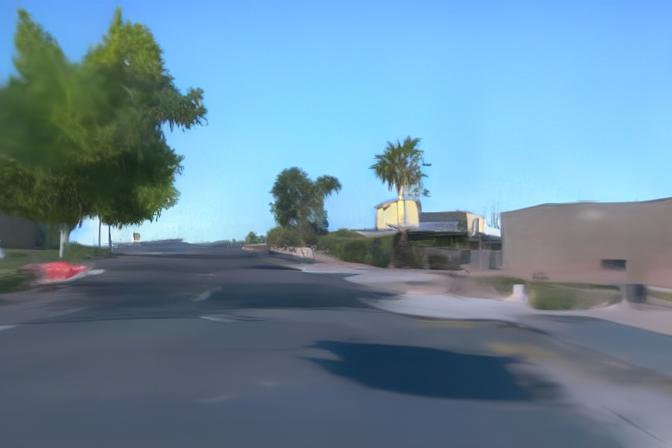}};
  \begin{scope}[shift={(img.south west)}, x={(img.south east)}, y={(img.north west)}]
    \draw[red,line width=0.3pt] (0.000,0.265) rectangle (0.222,0.6);
    \draw[red,line width=0.3pt] (0.309,0.4) rectangle (0.406,0.5);
    ;
  \end{scope}
\end{tikzpicture}

\\
\end{tabular}

\caption{\textbf{Semantic editing}: We remove existing cars by editing the semantic grid (red boxes), then regenerate the content conditioned on the updated semantics. The generated results remain coherent and consistent with the surrounding scene.  }
\label{fig:edit}
\end{figure}
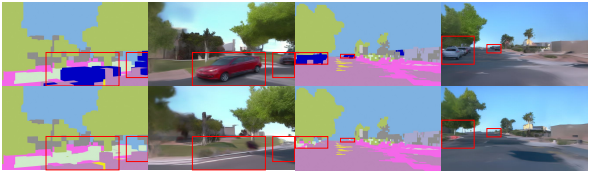

\noindent\textbf{Semantic editing.} We enable object-level editing directly on the semantic grid: vehicles can be removed or newly inserted, after which the model regenerates the scene content accordingly. Guided by the edited semantic grid, the resulting generations remain spatially consistent and coherent, as shown in \Cref{fig:edit}.
%\textbf{Simulation-to-Real Transfer}
%\input{carla_transfer}
%We use the CARLA simulator to obtain semantic voxel grids of simulated scenes and perform inference with our model to translate synthetic appearance into realistic outputs as shown in~\Cref{fig:CARLA_Transfer}
\newcommand{\boxLeft}{0.10}
\newcommand{\boxRight}{0.90}
\newcommand{\boxBottom}{0.02}
\newcommand{\boxTop}{0.52}

\newcommand{\fiveWithBox}[6]{%
\begin{tikzpicture}
\node[inner sep=0pt, outer sep=0pt, anchor=south west] (row) at (0,0) {%
  \includegraphics[width=#1]{#2}%
  \includegraphics[width=#1]{#3}%
  \includegraphics[width=#1]{#4}%
  \includegraphics[width=#1]{#5}%
  \includegraphics[width=#1]{#6}%
};
\begin{scope}[x={(row.south east)}, y={(row.north west)}]
  \draw[red, line width=0.9pt] (\boxLeft,\boxBottom) rectangle (\boxRight,\boxTop);
\end{scope}
\end{tikzpicture}%
}

% ---- layout helpers (measure left block height and reuse it on the right) ----
\newsavebox{\editLeftBox}
\newlength{\editLeftH}

\begin{figure*}[t]
\centering
\scriptsize
\setlength{\tabcolsep}{0pt}
\renewcommand{\arraystretch}{1.0}

\newcommand{\imgW}{0.17\textwidth}
\newcommand{\labelgap}{-1pt}

% -------- left block stored in a box (to measure its height) --------
\sbox{\editLeftBox}{%
\begin{minipage}{0.70\textwidth}
\centering
% (a) label at top-left of the left panel
\vspace{0.5ex}

\begin{tabular}{@{}r@{\hspace{3pt}}c@{}}
\tiny\textbf{(1)} &
\begin{tabular}{@{}c@{}}
\includegraphics[width=\imgW]{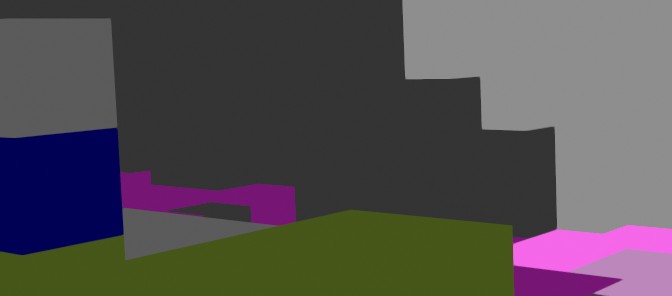}%
\includegraphics[width=\imgW]{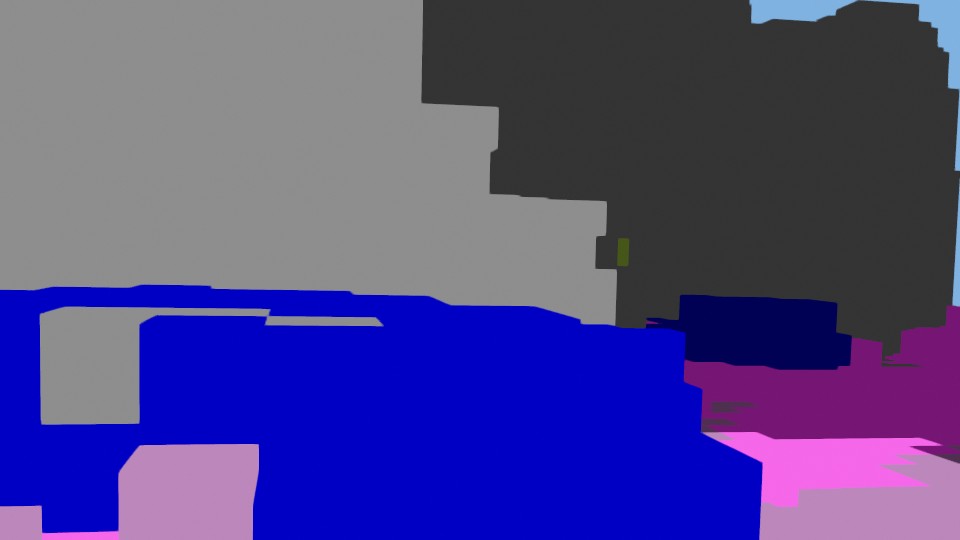}%
\includegraphics[width=\imgW]{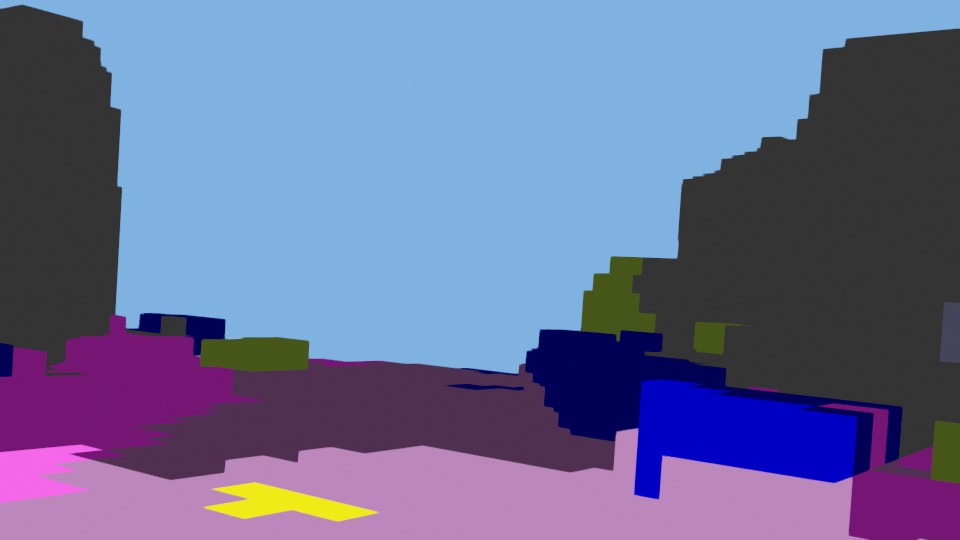}%
\includegraphics[width=\imgW]{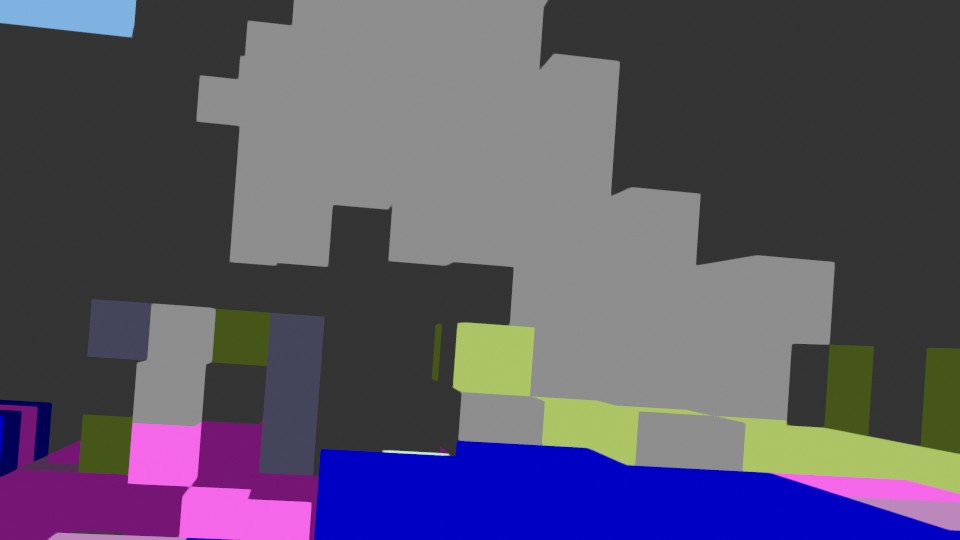}%
\includegraphics[width=\imgW]{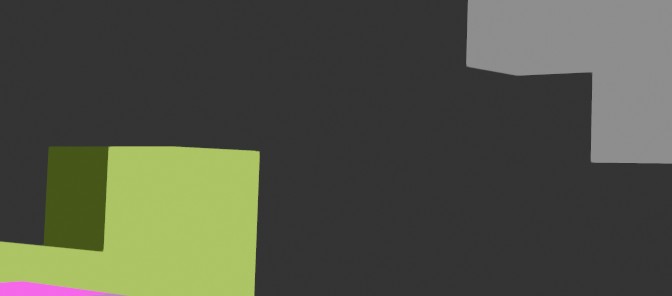}%
\end{tabular}
\\[\labelgap]

\tiny\textbf{(2)} &
\begin{tabular}{@{}c@{}}
\fiveWithBox{\imgW}
{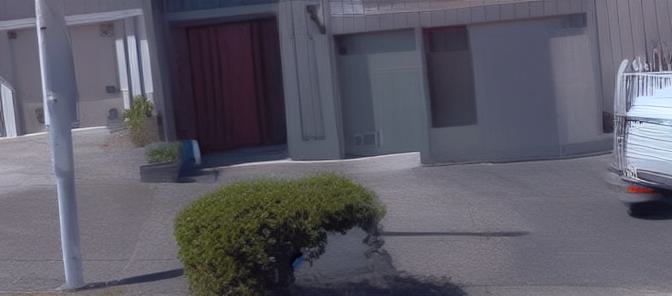}
{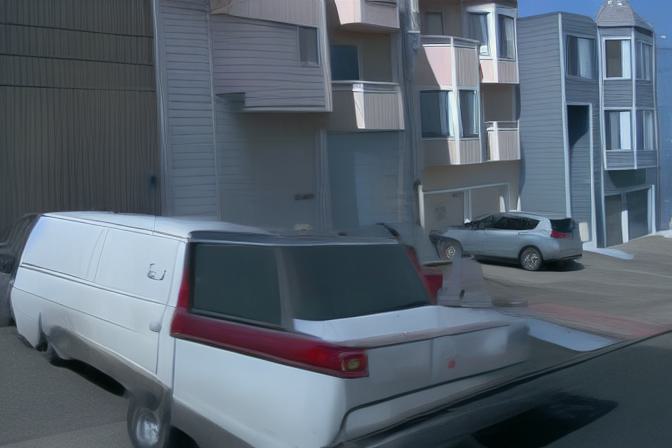}
{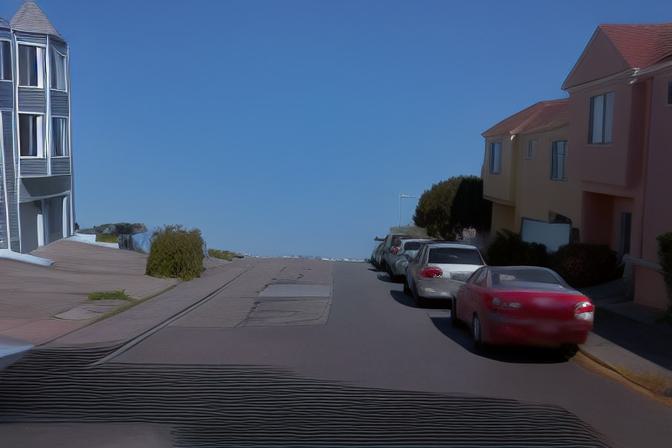}
{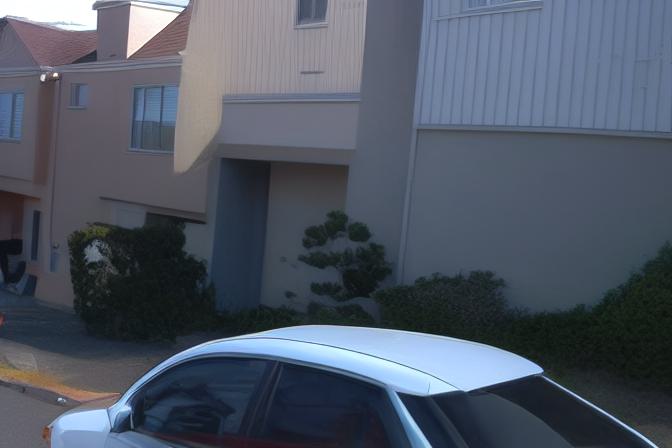}
{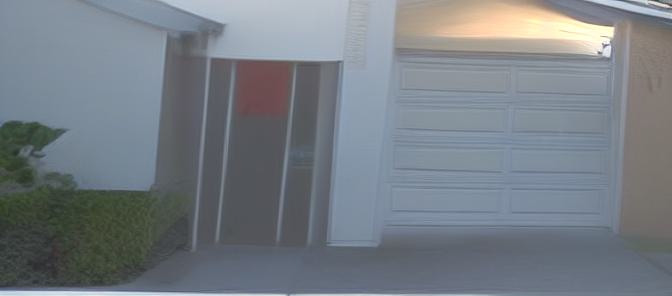}
\end{tabular}
\\[\labelgap]

\tiny\textbf{(3)} &
\begin{tabular}{@{}c@{}}
\fiveWithBox{\imgW}
{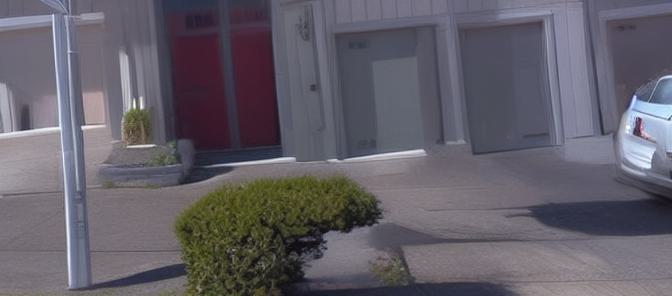}
{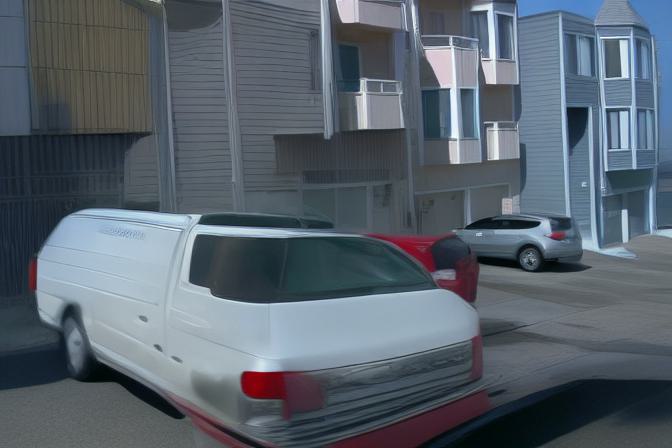}
{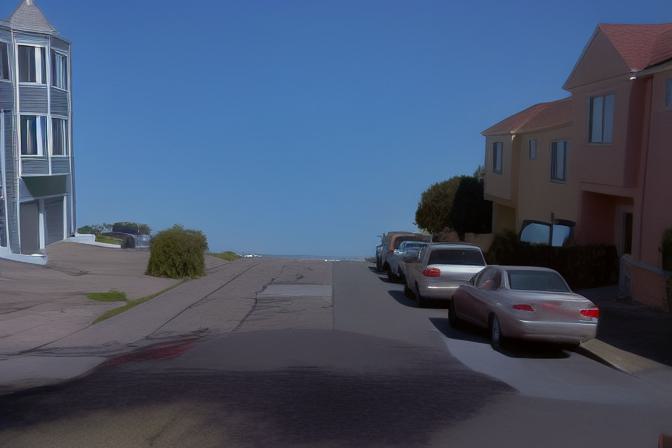}
{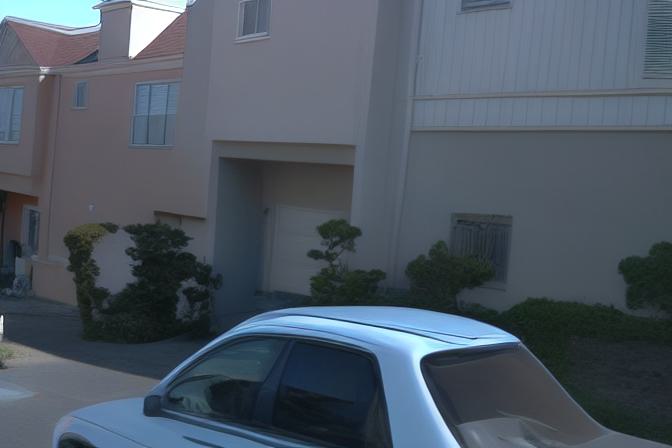}
{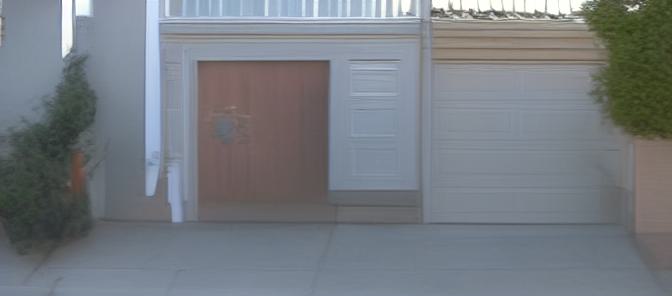}
\end{tabular}
\\
\end{tabular}
\end{minipage}%
}
\setlength{\editLeftH}{\dimexpr\ht\editLeftBox+\dp\editLeftBox\relax}

% -------- place left + right with aligned top/bottom --------
\noindent
\begin{tabular}{@{}p{0.70\textwidth}@{\hspace{0.02\textwidth}}p{0.28\textwidth}@{}}
\begin{minipage}[t][\editLeftH][t]{\linewidth}
  \centering
  \vspace{0pt}
  \usebox{\editLeftBox}
\end{minipage}

\end{tabular}
\vspace{-1.5mm}
\caption{\textbf{Scene inpainting} We mask a voxel region as shows (1) (darker voxel colors for unmasked) and re-generate only this region while keeping the rest of the $\Sigma$-Voxfield grid fixed. (2) and (3) show two inpainted results (red box) that remain coherent with the background and vary in both structure and appearance.}
\label{fig:variance_editing}
\vspace{-2mm}
\end{figure*}

\noindent\textbf{Scene inpainting.} \Cref{fig:variance_editing} demonstrates \textbf{local editing} enabled by voxel-space inpainting. Starting from a generated $\Sigma$-Voxfield grid, we define a foreground region as a subset of $\Sigma$-Voxfields and re-run diffusion only on this target subset while keeping the rest of the grid fixed. While different samples produce diverse foreground realizations that remain coherent with the surrounding context across viewpoints.

\noindent\textbf{Infinite scene generation.} Despite being trained on local voxel neighborhoods, our method produces continuous scenes over large spatial extents as shown in \Cref{fig:teaser}.

\subsection{Limitations}
While our method enables scalable and semantically structured 3D scene generation, several limitations remain. First, our model provides limited control over the generated appearance: conditioning is primarily semantic and geometric, so attributes such as texture style, lighting, and material properties cannot be specified explicitly. Second, our formulation focuses on static scene generation and does not model dynamic elements such as moving vehicles, pedestrians, or temporal scene evolution. Extending the representation and diffusion process to account for the dynamics and time-varying content remains an open challenge.

\section{Conclusion}
We introduce a scalable framework for semantically structured 3D driving-scene generation that operates directly in 3D. Our method generates a $\Sigma$-Voxfield grid with a semantic-conditioned diffusion model over local neighborhoods, and scales to large environments via voxel-space spatial inpainting. The resulting 3D buffer is rendered with a deferred diffusion engine to produce photorealistic images without per-scene optimization. Experiments on Waymo and PandaSet demonstrate competitive rendered-view quality, strong semantic coherence, and scalability to large scenes with moderate computation cost. This work points toward structured 3D scene generation and motivates future extensions such as richer appearance control and dynamic scene modeling.

\title{\method: Semantic Voxel-Guided Diffusion for Large-Scale Driving Scene Generation: Supplementary material} 
\titlerunning{SEM-ROVER}
\authorrunning{H.~Dahmani et al.}
\author{Hiba Dahmani\orcidlink{0009-0008-0426-9919}\inst{1,2} \and
Nathan Piasco\inst{1}\orcidlink{0000-0001-7952-6643} \and 
Moussab Bennehar\inst{1}\orcidlink{0000-0002-6566-6132} \and
\\ Luis Rold{\~a}o\inst{1}\orcidlink{0000-0003-0482-3584} \and
Dzmitry Tsishkou\inst{1}\orcidlink{0009-0002-9798-3316} \and
Laurent Caraffa\inst{3}\orcidlink{0000-0002-8676-8058} \and
\\Jean-Philippe Tarel\inst{2}\orcidlink{0000-0002-9241-5347} \and
Roland Br{\'e}mond\inst{2}\orcidlink{0000-0003-3150-7624}
}

\institute{Noah's Ark, Huawei Paris Research Center, France \and
COSYS, Gustave Eiffel University, France \and
LASTIG, IGN-ENSG, Gustave Eiffel University, France}

\maketitle
\section{Data Processing}
\begin{figure}[tbh]
\centering
\scriptsize
\setlength{\tabcolsep}{0pt}
\renewcommand{\arraystretch}{1.0}
\includegraphics[width=0.24\textwidth]{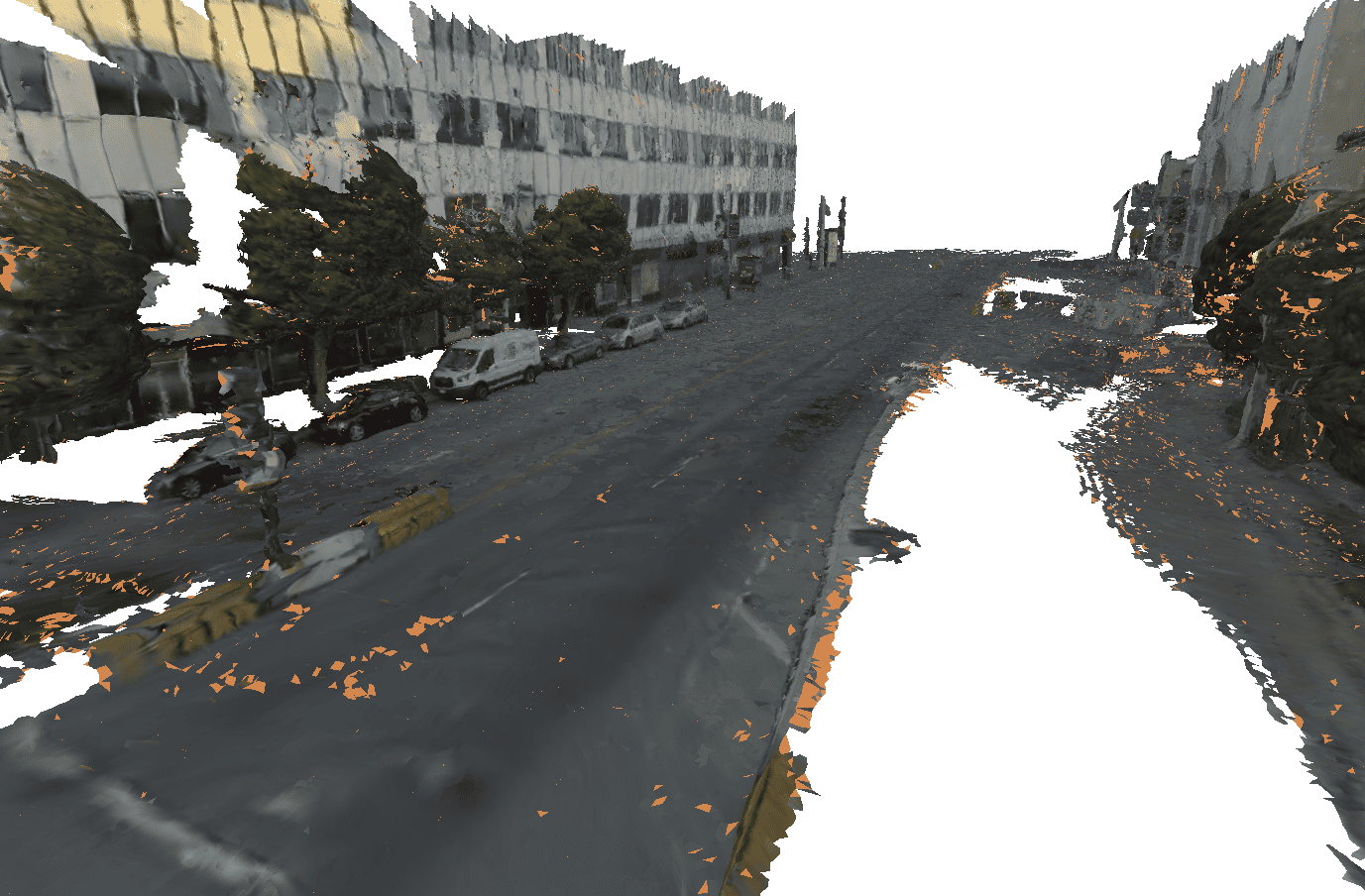}
\includegraphics[width=0.24\textwidth]{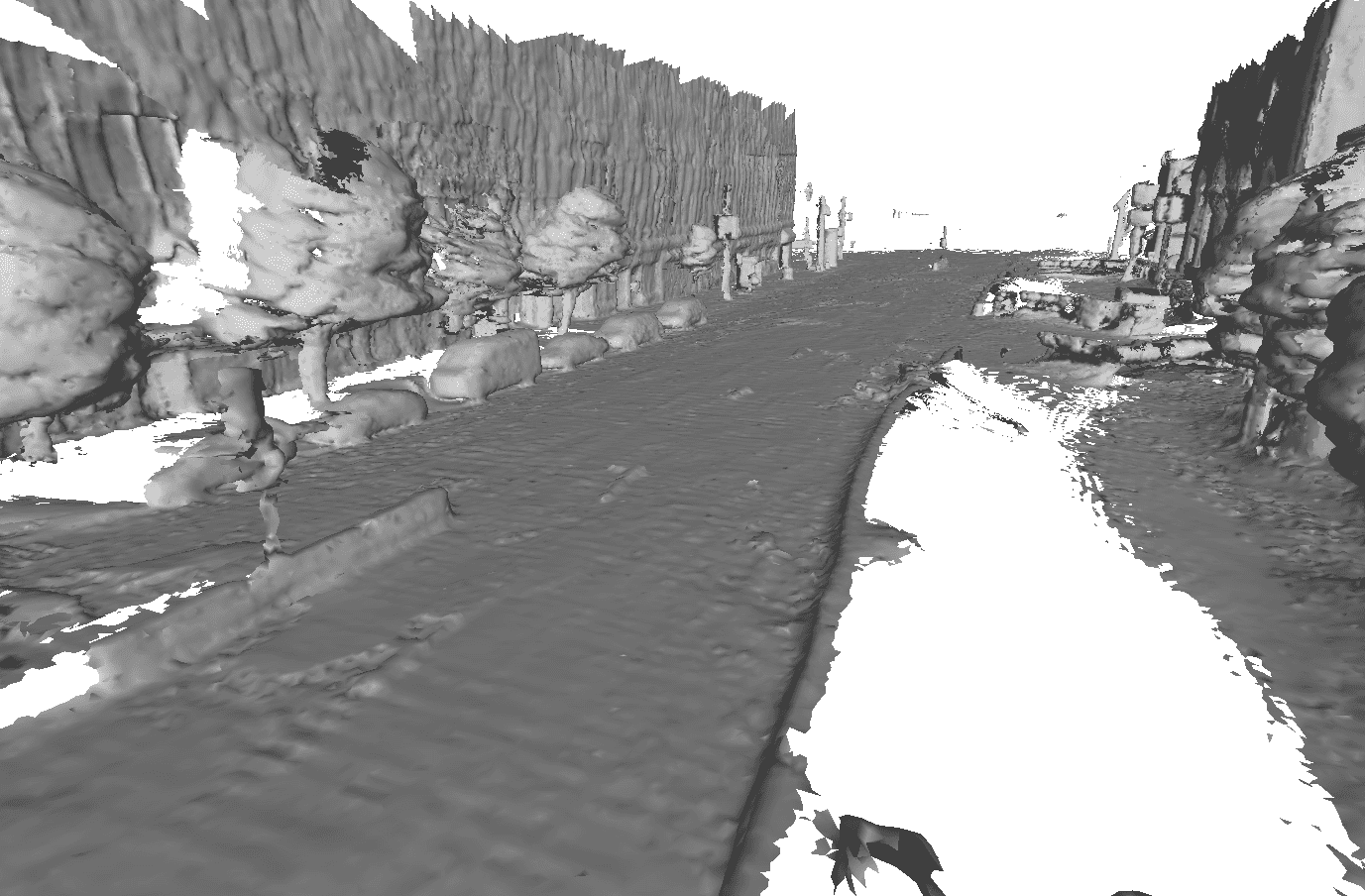}
\includegraphics[width=0.24\textwidth]{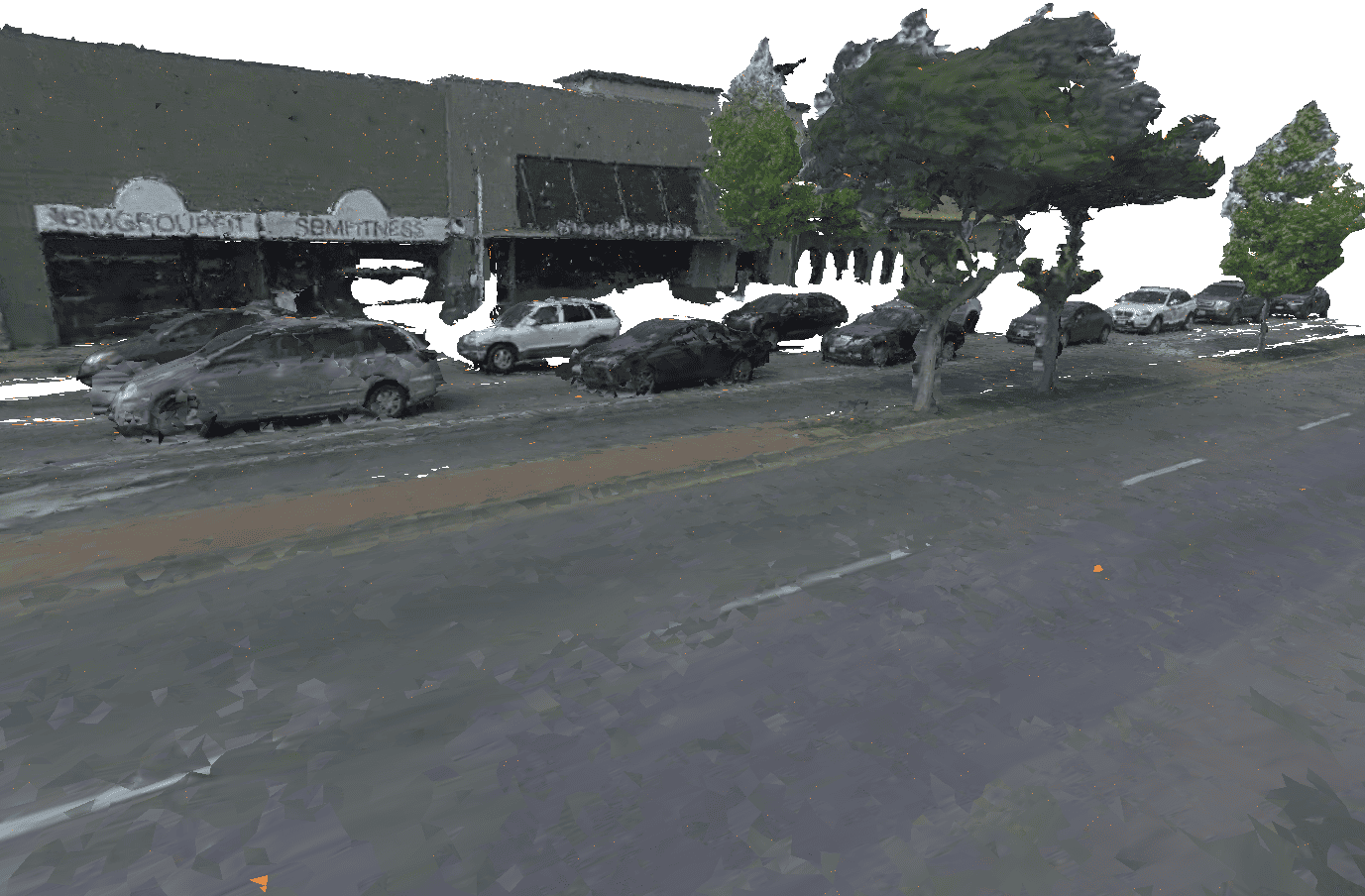}
\includegraphics[width=0.24\textwidth]{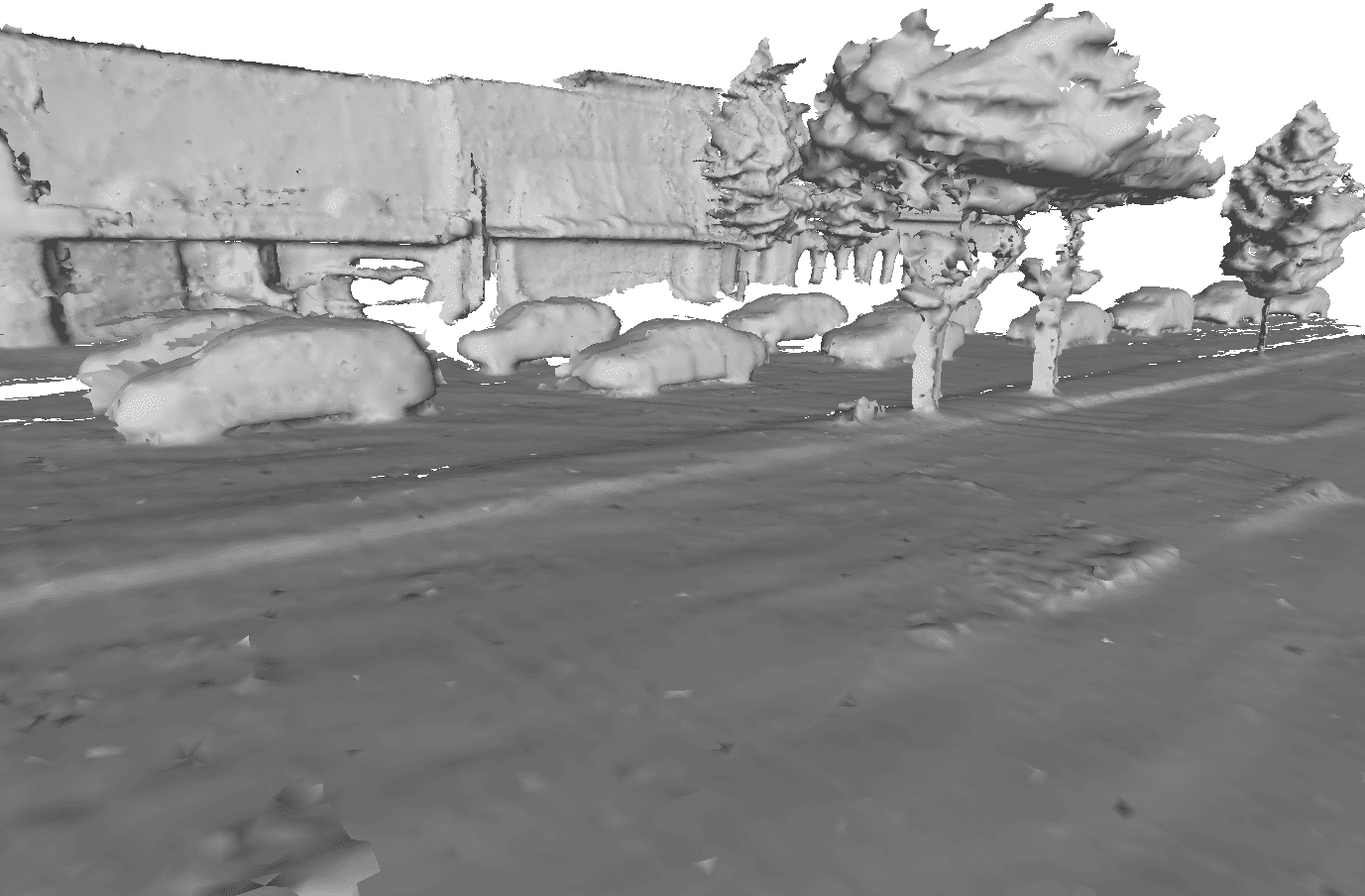}

\includegraphics[width=0.24\textwidth]{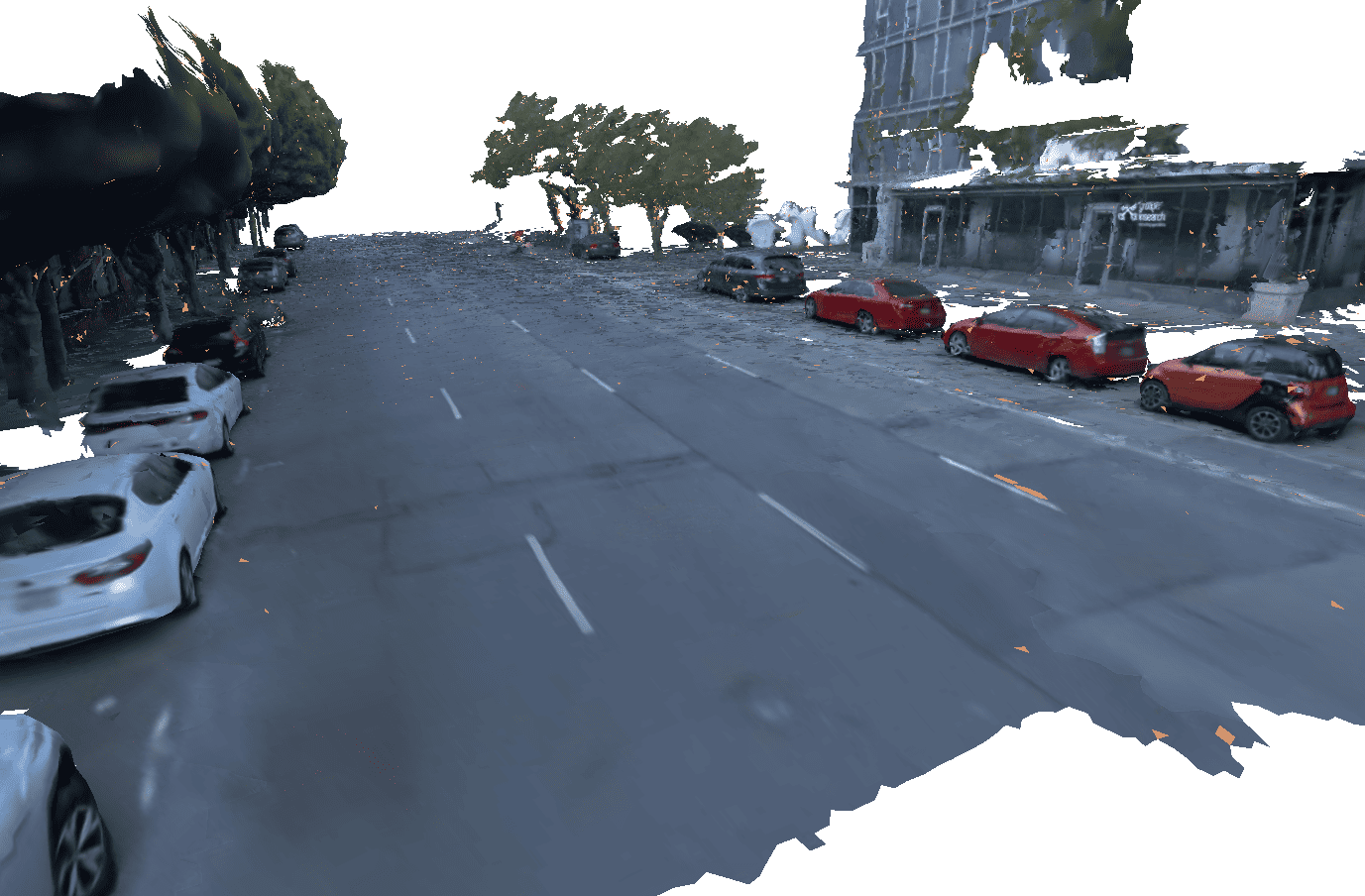}
\includegraphics[width=0.24\textwidth]{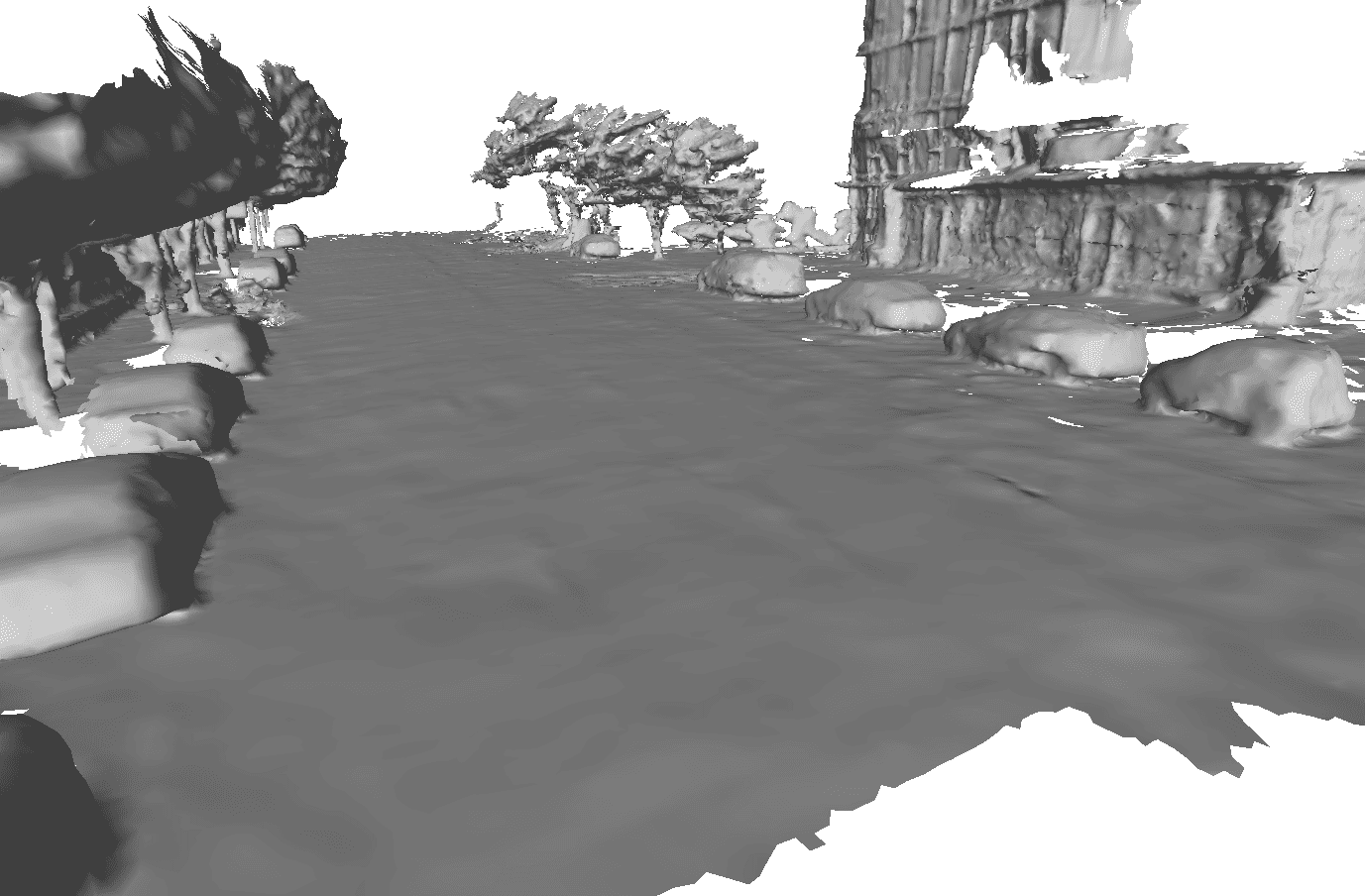}
\includegraphics[width=0.24\textwidth]{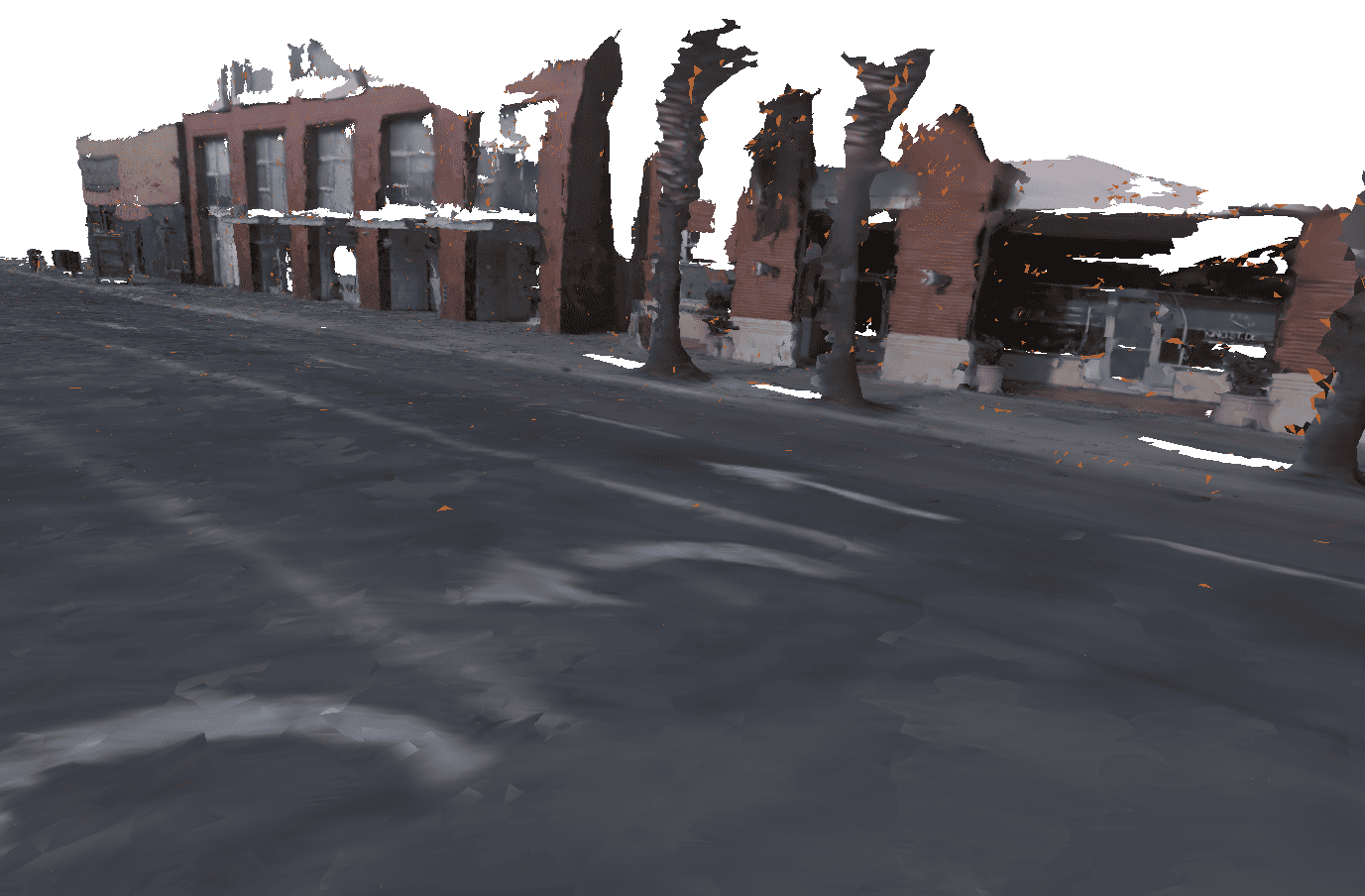}
\includegraphics[width=0.24\textwidth]{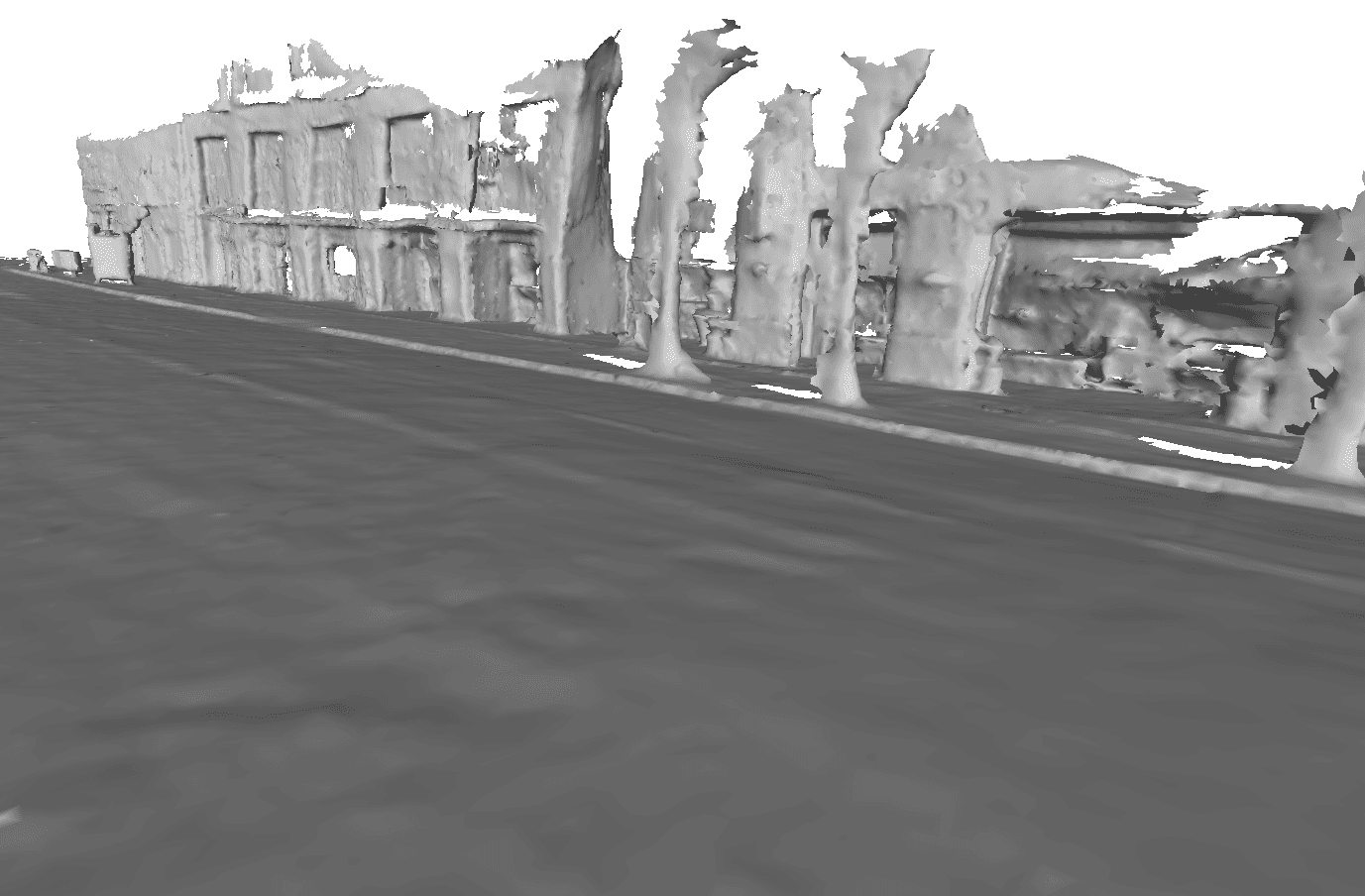}
\caption{\label{fig:3d_textured_mesh} \textbf{Textured meshes after data processing.} Top line shows 2 scenes from Pandaset~\cite{pandaset}, bottom are 2 scenes from WOD~\cite{waymo}. \textcolor{orange}{Orange triangles} in the colorized mesh denote faces not textured during the procedure.}
\end{figure}
\begin{figure}[t]
\centering
\begin{overpic}[width=0.8\linewidth]{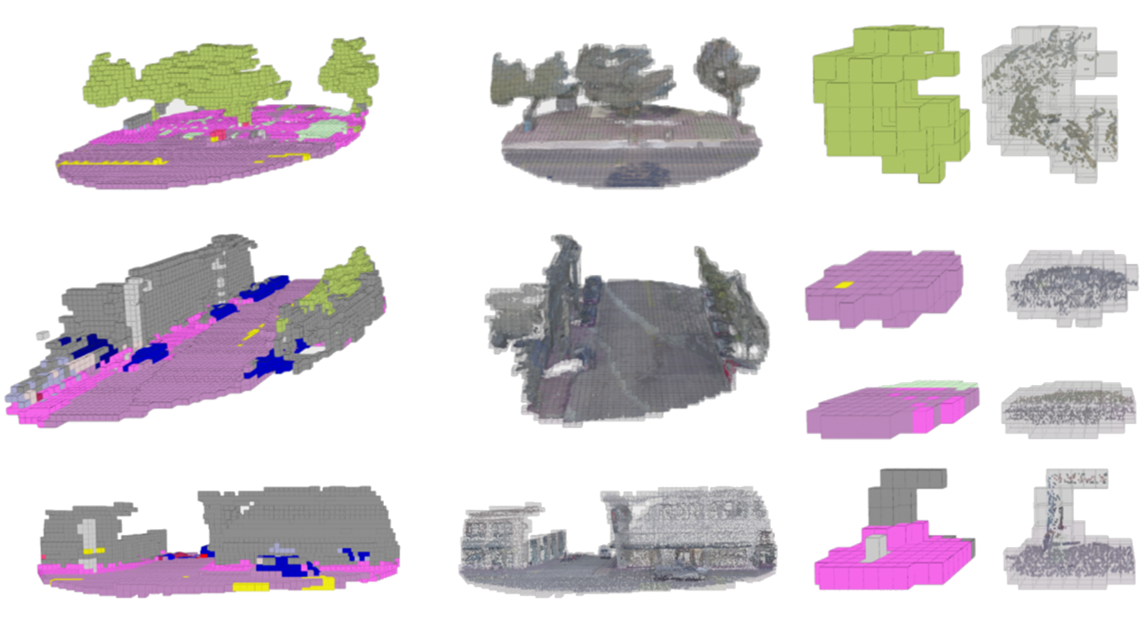}
    \put(30,-6){\small (a)}
    \put(83,-6){\small (b)}
\end{overpic}
\vspace{10pt}
\caption{\textbf{Visualization of $\Sigma$-VoxFields.} 
(a) Large-scene $\Sigma$-VoxFields with their corresponding semantic layouts. 
(b) Examples of $\Sigma$-VoxFields training samples, each containing 50--150 neighboring $\Sigma$-VoxFields used to train our model.}
\label{fig:voxelfields}
\end{figure}

The construction of $\Sigma$-Voxfield grids from raw multi-view driving logs follows a systematic pipeline designed to transform unstructured sensor data into a semantically-aware, discretized volumetric representation. We utilize 60 daytime sequences from the Waymo Open Dataset \cite{waymo} and 60 daytime sequences from PandaSet \cite{pandaset}. To preserve geometric fidelity over long trajectories, each Waymo sequence is partitioned into two distinct sub-reconstructions during the optimization stage, with a maximum of 100 timesteps per reconstruction.
\paragraph{Geometric and Appearance Reconstruction.}
To isolate the static environment, we employ \textbf{OmniRe} \cite{omnire}, a 3DGS-based framework that decouples dynamic actors from the background. We incorporate monocular normal priors from \textbf{DepthAnything} \cite{depthanything} to regularize the geometry of textureless surfaces, such as asphalt and glass.  We use all the available cameras to train the dynamic 3DGS reconstruction (6 for Pandaset and 5 for Waymo). The resulting background is represented as a global Signed Distance Function (SDF) volume, which we convert into a manifold surface mesh via the \textbf{Marching Cubes} \cite{lorensen1987marchingcubes} algorithm. Photorealistic appearance is integrated by texturing 
 with \textbf{OpenMVS} \cite{openmvs}, which aggregates multi-view RGB observations while performing graph-cut-based seam leveling to ensure radiometric consistency across the camera rig. We show in \Cref{fig:3d_textured_mesh} some examples of the textured background mesh used in our work.
\paragraph{3D Semantic labeling.}
We generate per-frame semantic conditioning through a hybrid segmentation approach. General scene parsing is performed via \textbf{SegFormer} \cite{segformer} using a \textbf{Cityscapes}-pretrained~\cite{cityscapes} backbone. To specifically address the structural importance of road topology, we augment this with PriorLane~\cite{priorlane}, a transformer-based lane detection method that leverages prior knowledge to recover thin lane boundaries.
Our final taxonomy consists of 20 semantic classes:
\textit{road, sidewalk, building, wall, fence, pole, traffic light, traffic sign, vegetation, terrain, sky, person, rider, car, truck, bus, train, motorcycle, bicycle,} and a dedicated \textit{road-lane} class. To ensure visual clarity and consistency across all figures in this work, we adopt the standard Cityscapes color palette for the first 19 classes and introduce yellow (RGB: $255, 255, 0$) to denote the road-lane category. The 2D semantic labels are fused into the 3D volume using the \textbf{Scalable TSDF Volume Integration} system in \textbf{Open3D}~\cite{open3d}. We resolve view-dependent inconsistencies and segmentation noises through a majority-voting consensus within the TSDF volume, yielding a 3D semantic layout that is spatially coincident with the geometry.
\paragraph{$\Sigma$-Voxfield Conversion.} The final $\Sigma$-Voxfield grid is obtained by voxelizing the scene with a voxel size of 0.6m. We discard empty voxels and, for each remaining cell, uniformly sample points from the textured mesh surface lying within the voxel boundaries. Neighboring sets of $\Sigma$-Voxfield (from 50 to 150) are used as training samples for our $\Sigma$-Voxfield diffusion model. In total, we use 450K training examples extracted jointly from the PandaSet\cite{pandaset} and WOD\cite{waymo} processed scenes. We show examples of $\Sigma$-Voxfield grid and training sample in \Cref{fig:voxelfields}.

\begin{figure}[ht]
    \centering    
    \includegraphics[width=0.75\linewidth]{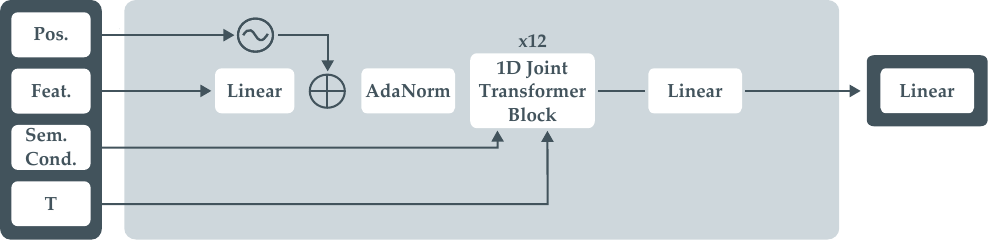}
    \caption{Architecture of the $\Sigma$-Voxfield diffusion model.}
    \label{fig:voxelfield_diff}
\end{figure}

\section{Model architectures and training details}

\subsection{$\Sigma$-Voxfield diffusion model}

\noindent
% \begin{minipage}[t]{0.74\linewidth}
% \vspace{0pt}

% \begin{figure}[h]
%     \centering
%     \scriptsize
%     \setlength{\tabcolsep}{0pt}
%     \renewcommand{\arraystretch}{1.0}
%     \includegraphics[width=\textwidth]{figures/model_f2_v2.pdf}
%     \caption{Architecture of the $\Sigma$-Voxfield diffusion model.}
%     \label{fig:voxelfield_diff}
% \end{figure}

The denoising network is implemented as a transformer over local sets of $\Sigma$-Voxfields, illustrated in \Cref{fig:voxelfield_diff}. Each $\Sigma$-Voxfield token is represented by a $120$-dimensional feature vector obtained by stacking the coordinates and RGB values of $N=20$ surface samples ($6N=120$). The input features are linearly projected to the transformer dimension and processed by $12$ joint transformer layers. Each layer uses multi-head attention with $8$ heads of dimension $128$, resulting in an internal feature dimension of $1024$. Semantic conditioning embeddings are processed jointly with the voxel features, enabling each token to attend to both geometric and semantic context within the local set. Timestep conditioning is injected through adaptive normalization layers. The final transformer features are projected back to dimension $120$ to predict the denoised $\Sigma$-Voxfield representation.

% \end{minipage}\hfill%
% \begin{minipage}[t]{0.24\linewidth}
% \vspace{0pt}
% \centering
% \input{figures/voxelfield_diff}
% \captionof{figure}{Architecture of the $\Sigma$-Voxfield diffusion model.}
% \label{fig:voxelfield_diff}
% \end{minipage}

\subsection{Deferred renderers}
\begin{figure}[tbh]
\centering
\scriptsize
\setlength{\tabcolsep}{0pt}
\renewcommand{\arraystretch}{1.0}
\includegraphics[width=\textwidth]{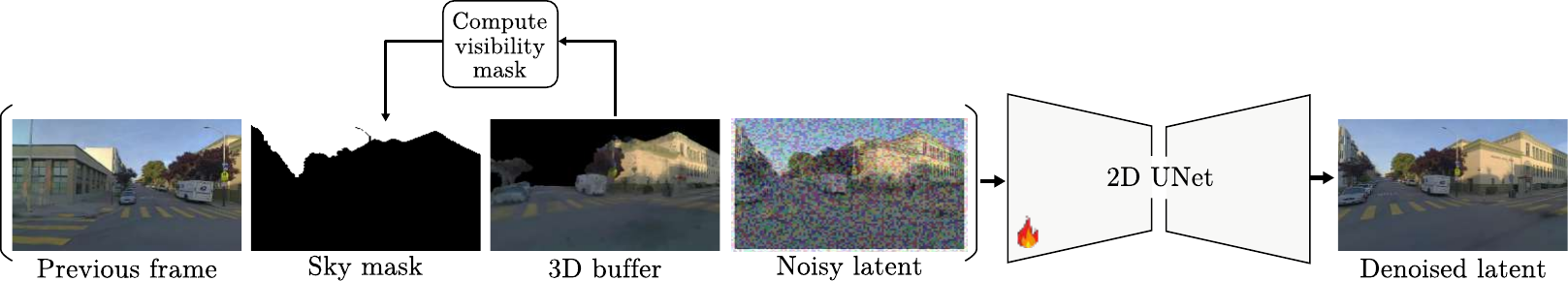}
(a) Autoregressive Stable Diffusion - ASD
\includegraphics[width=\textwidth]{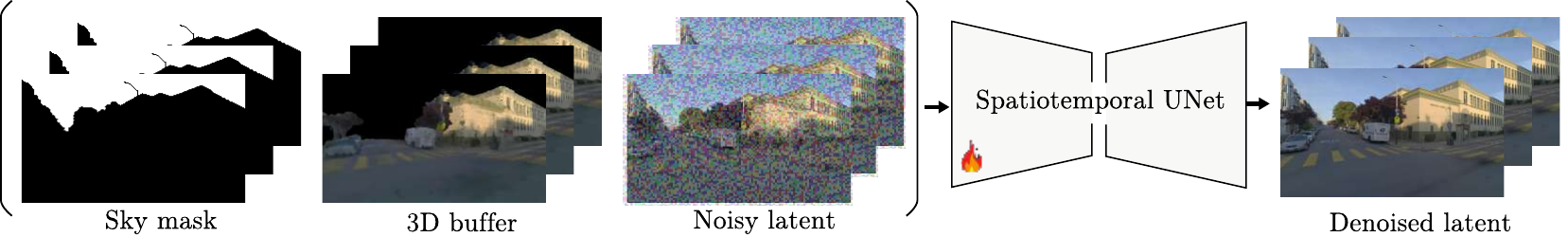}
(b) Video Stable Diffusion - VSD
\caption{\label{fig:deferred_rendering} \textbf{Deferred rendering architectures.} (a) Autoregressive Stable Diffusion: our renderer based on SD model, with additional conditioning for autoregressive generation (previous frame) and consistent 3D generation (3D buffer from $\Sigma$-Voxfield rendering and sky mask);  (b) Video Stable Diffusion: our renderer based on VSD model, with additional conditioning from 3D buffers and sky masks.}
\end{figure}
\paragraph{Autoregressive Stable Diffusion (ASD).} We adapt SD 1.5~\cite{stablediffusion} image diffusion model for our ASD, concatenating along the feature dimension the previous generated frame, the 3D buffer obtained from the rendering of the $\Sigma$-Voxfield grid and the sky mask. Similar to~\cite{gamengen}, we add Gaussian noise to the conditioning of the previous frame (with scale factor going from $0.3$ to $0.7$) to avoid rollout divergence. We train our ASD with images of size $424\times616$, batch size of 4 and $30\%$ of CFG probability. When applying CFG, we mask both the previous frame and the sky mask.

\paragraph{Video Stable Diffusion (VSD).} We fine-tune a VSD~\cite{blattmann2023stable} model, replacing the image conditioning by our 3D buffers and concatenating to the input of the model the sky masks, similar to our ASD. Architectures of both deferred renderers can be found in~\Cref{fig:deferred_rendering}. We train our VSD with images of size $384\times576$, sequence length of 12 images, batch size of 4 and $30\%$ of CFG probability. We train our VSD model on 1 high-end GPU with 156GB of memory for 1 week.

% \section{Training and evaluation Details}

\section{Spatial Outpainting Strategy}

Our diffusion model operates on bounded local sets of $\Sigma$-Voxfields. While this design keeps the computational cost of each denoising step constant, it limits the spatial extent that can be generated in a single pass. To synthesize large scenes, we therefore employ a progressive spatial outpainting strategy that iteratively expands the generated region while conditioning on previously synthesized neighborhoods.

\paragraph{Region extraction for progressive outpainting.}
To apply Repaint~\cite{repaint} based outpainting on large scenes, we first decompose the full set of $\Sigma$-Voxfields into overlapping local regions of bounded size. Each region defines a local set $\mathcal{X}_{\xi}$ on which diffusion is applied, while overlaps with previously generated regions provide the \emph{known} conditioning context used during denoising.

We construct these regions using the distance-guided extraction procedure in Algorithm~\ref{alg:region_growing}. Let $\mathcal{P}$ be the full set of $\Sigma$-Voxfields, $\mathcal{R}$ the extracted regions, and $\mathcal{U}\subseteq\mathcal{P}$ the set of uncovered voxfields. Starting from an initial valid region, the algorithm iteratively selects the uncovered point closest to the existing regions as a seed, then forms a candidate region from its $K$ nearest neighbors in $\mathcal{P}$. If this candidate contains at least $T$ uncovered voxfields, it is added to $\mathcal{R}$ and removed from $\mathcal{U}$. Repeating this process progressively expands coverage while maintaining overlap between neighboring regions.

This strategy is well suited for progressive outpainting. Since each new seed is chosen near already extracted regions, newly generated local sets remain spatially adjacent to previously synthesized content, naturally creating overlaps. These overlaps provide the \emph{known} conditioning tokens required by the Repaint scheduler, allowing geometry and appearance to propagate consistently across diffusion steps. Meanwhile, the $K$-nearest-neighbor construction keeps each region bounded to a fixed size, so every diffusion pass operates on a constant-size local set regardless of overall scene scale.

\begin{figure}[htbp]
    \centering
    \offinterlineskip
    \def\imglist{164001,164003,164005,164007,
                 164025,164027,164031,164033,
                 164036,164037,164039,164048}

    \foreach \n [count=\idx from 1] in \imglist {%
        \begin{minipage}{0.25\textwidth}
            \centering
            \begin{tikzpicture}
                \node[inner sep=0] (img) {
                    \adjustbox{width=\linewidth, height=\linewidth, clip,
                    trim={0.25\width} {0.25\height} {0.25\width} {0.25\height}}{%
                        \includegraphics{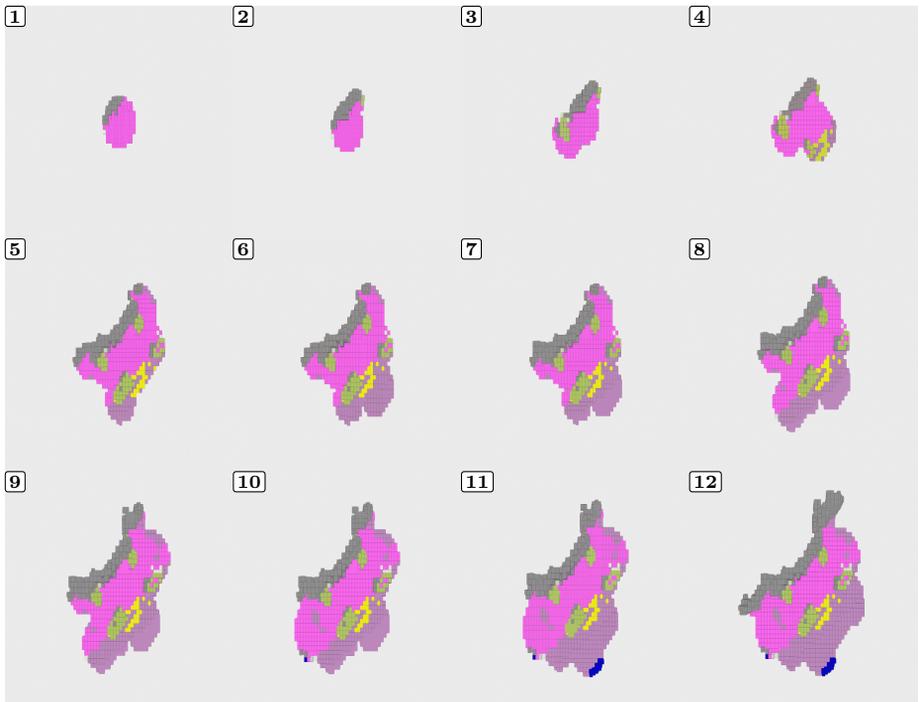}%
                    }
                };
                \node[anchor=north west, fill=white, draw=black, rounded corners=1pt,
                inner sep=1.5pt, font=\scriptsize] at (img.north west) {\textbf{\idx}};
            \end{tikzpicture}
        \end{minipage}%
        \ifnum\numexpr\idx-\idx/4*4\relax=0\par\fi
    }
    \caption{Top-view visualization of progressive region selection on a semantic voxel grid. Each panel shows the region selected at one iteration of Algorithm~\ref{alg:region_growing}, with the overlaid number indicating the selection order. The extraction expands outward from already covered regions, producing overlapping local sets suitable for conditional outpainting.}
    \label{fig:region_extraction}
\end{figure}

Figure~\ref{fig:region_extraction} visualizes the region extraction process from a top-view perspective of the semantic $\Sigma$-Voxfield grid. The panels illustrate how regions are progressively selected and expanded outward from already covered areas, forming spatially adjacent local neighborhoods that overlap with previously extracted regions.

\begin{algorithm}[t]
\caption{Distance-Guided Region Extraction for Spatial Outpainting}
\label{alg:region_growing}
\KwIn{Point set $\mathcal{P}$, region size $K$, coverage threshold $T$, initial extracted regions $\mathcal{R}$}
\KwOut{A set of overlapping local regions $\mathcal{R}$}

$\mathcal{U} \leftarrow \mathcal{P} \setminus \bigcup_{\mathcal{N} \in \mathcal{R}} \mathcal{N}$ \tcp*{Uncovered points}

\BlankLine
\While{$\mathcal{U} \neq \emptyset$}{
    $s \leftarrow \arg\min_{p \in \mathcal{U}} \mathrm{dist}(p,\mathcal{R})$ \tcp*{Seed closest to existing regions}
    $\mathcal{N} \leftarrow \mathrm{kNN}(s,\mathcal{P},K)$ \tcp*{Candidate region}

    \If{$|\mathcal{N} \cap \mathcal{U}| \geq T$}{
        $\mathcal{R} \leftarrow \mathcal{R} \cup \{\mathcal{N}\}$\;
        $\mathcal{U} \leftarrow \mathcal{U} \setminus \mathcal{N}$\;
        Update $\mathrm{dist}(p,\mathcal{R})$ for all $p \in \mathcal{P}$\;
    }
    \Else{
        break \tcp*{Stop when no sufficiently new region can be formed}
    }
}
\BlankLine
\textbf{Distance definition:} $\mathrm{dist}(p,\mathcal{R}) = \min_{\mathcal{N}\in\mathcal{R}} \min_{q\in\mathcal{N}} \|p-q\|_2$\;
\end{algorithm}

%\BlankLine
%\NP{You don't need pass 2 according to your $%$while criteria in pass 1}
%\tcp{Pass 2: attach leftover points to ensure full coverage}
%\ForEach{leftover point $s$}{
%    \If{$s \in \mathcal{U}$}{
%        $\mathcal{N} \leftarrow \mathrm{kNN}(s,\mathcal{P},K)$\;
%        $\mathcal{R} \leftarrow \mathcal{R} \cup \{\mathcal{N}\}$\;
%        $\mathcal{U} \leftarrow \mathcal{U} \setminus \mathcal{N}$\;
%    }
%}

%\end{algorithm}

\begin{figure}[t]
    \centering

    % Row 1
    \begin{subfigure}{0.19\textwidth}
        \centering
        \includegraphics[width=\linewidth,height=\linewidth]{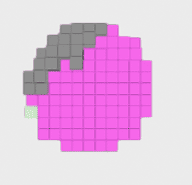}
    \end{subfigure}
    \begin{subfigure}{0.19\textwidth}
        \centering
        \includegraphics[width=\linewidth,height=\linewidth]{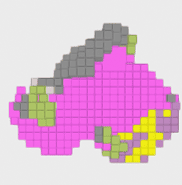}
    \end{subfigure}
    \begin{subfigure}{0.19\textwidth}
        \centering
        \includegraphics[width=\linewidth,height=\linewidth]{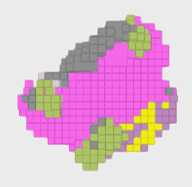}
    \end{subfigure}
    \begin{subfigure}{0.19\textwidth}
        \centering
        \includegraphics[width=\linewidth,height=\linewidth]{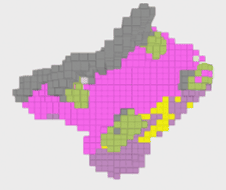}
    \end{subfigure}
    \begin{subfigure}{0.19\textwidth}
        \centering
        \includegraphics[width=\linewidth,height=\linewidth]{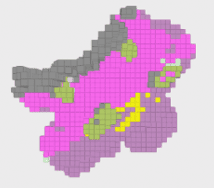}
    \end{subfigure}

    \vspace{0.4em}

    % Row 2
    \begin{subfigure}{0.19\textwidth}
        \centering
        \includegraphics[width=\linewidth,height=\linewidth]{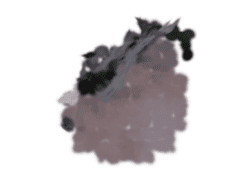}
    \end{subfigure}
    \begin{subfigure}{0.19\textwidth}
        \centering
        \includegraphics[width=\linewidth,height=\linewidth]{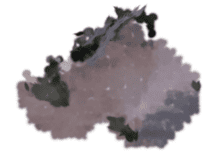}
    \end{subfigure}
    \begin{subfigure}{0.19\textwidth}
        \centering
        \includegraphics[width=\linewidth,height=\linewidth]{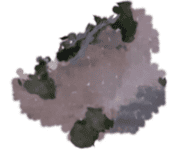}
    \end{subfigure}
    \begin{subfigure}{0.19\textwidth}
        \centering
        \includegraphics[width=\linewidth,height=\linewidth]{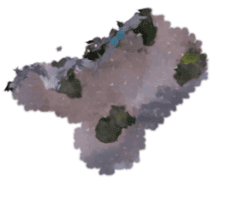}
    \end{subfigure}
    \begin{subfigure}{0.19\textwidth}
        \centering
        \includegraphics[width=\linewidth,height=\linewidth]{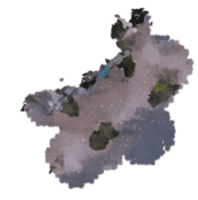}
    \end{subfigure}

    \caption{Top-view visualization of progressive outpainting. The first row shows the semantic voxel grid layout conditioning, while the second row shows the corresponding generated 3D $\Sigma$-Voxfield. The panels correspond to different stages of region expansion and illustrate that our Repaint~\cite{repaint} based outpainting preserves spatial coherence across neighboring local generations.}
    \label{fig:inpainting_gen_grid}
\end{figure}

\paragraph{Outpainting coherence.}
Figure~\ref{fig:inpainting_gen_grid} illustrates the spatial coherence obtained during progressive outpainting. The first row shows top-view visualizations of the semantic $\Sigma$-Voxfield layout, while the second row shows the corresponding generated 3D Voxelfield appearance for the same regions. Although each region is generated independently within a bounded local neighborhood, the overlap between neighboring regions allows the Repaint scheduler to propagate geometry and appearance across successive generations. As a result, the synthesized 3D Voxelfield remains coherent across region boundaries. Examples of the generated 3D buffers are shown in \Cref{fig:3d_buffers}.

\subsection{Ablation on the number of sampled points per voxel}

\noindent
\begin{minipage}[t]{0.58\linewidth}
\vspace{0pt}
\paragraph{Number of samples per voxel.}
We use $N=20$ surface samples per voxel. This choice is motivated by a geometric fidelity study on one mesh from one scene of the Waymo Open Dataset, where we evaluate the Chamfer Distance between the sampled voxel representation and the ground-truth surface as a function of $N$ for a fixed voxel size of $0.6\,\mathrm{m}$. Figure~\ref{fig:ablations} shows that the error decreases rapidly and saturates around $N=20$, reaching approximately $0.025\,\mathrm{m}$, which is below the rendering splat radius $r=0.04\,\mathrm{m}$. Larger values of $N$ provide only marginal gains while increasing the input dimensionality of each $\Sigma$-Voxfield linearly. We therefore adopt $N=20$ in all experiments.
\end{minipage}\hfill%
\begin{minipage}[t]{0.4\linewidth}
\vspace{0pt}
\centering
\centering
\includegraphics[width=\linewidth]{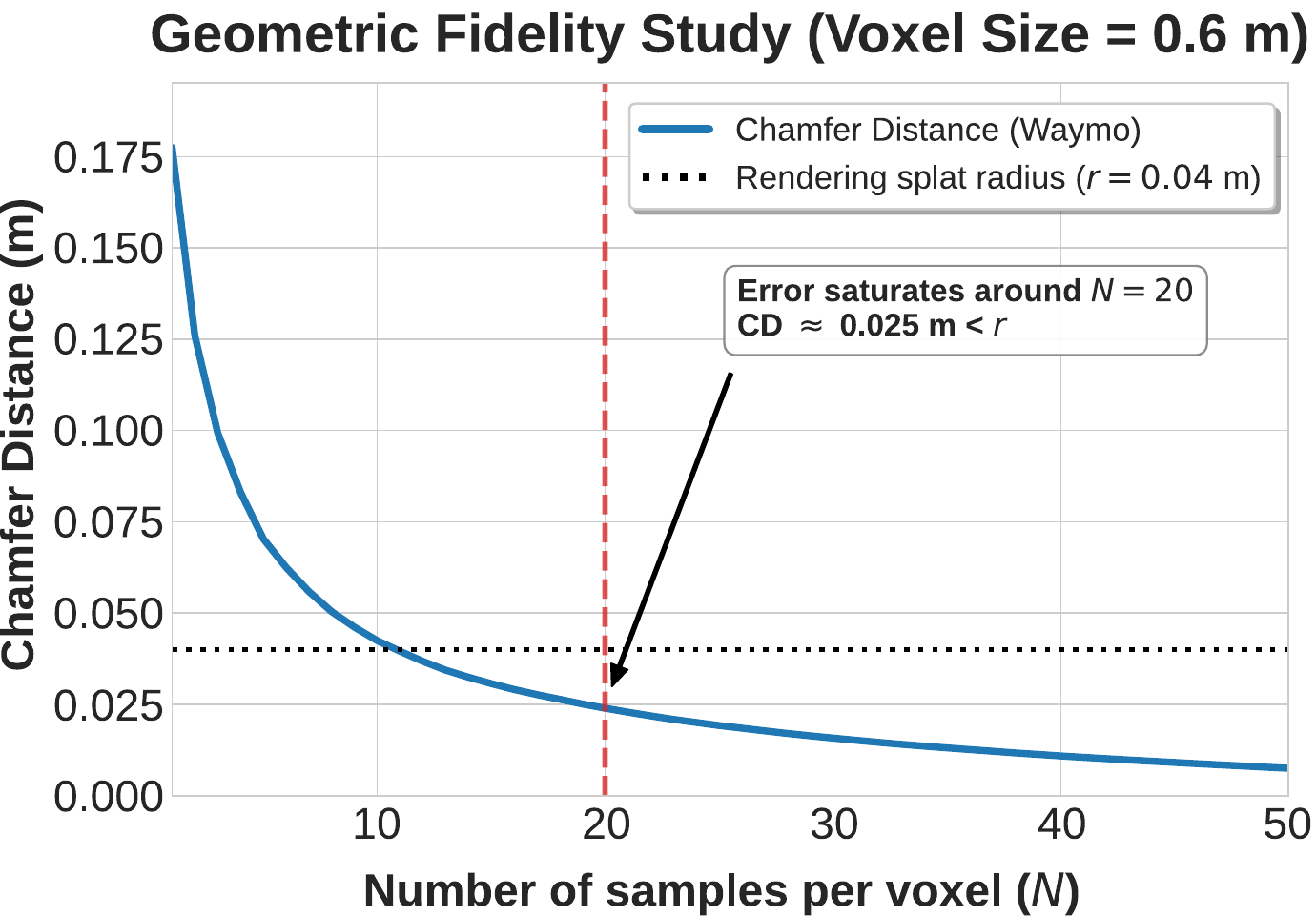}
\captionsetup{hypcap=false}
\captionof{figure}{Chamfer Distance between sampled voxel points and the ground-truth scene mesh as a function of the number of surface samples per voxel. The error saturates around $N=20$.}
\label{fig:ablations}
\end{minipage}

\section{Additional Results}
\begin{figure*}[tbh]
\centering
\scriptsize
\setlength{\tabcolsep}{0pt}
\renewcommand{\arraystretch}{1.0}

% ---- knobs ----
\newcommand{\imgW}{0.17\textwidth}  % slightly smaller to make room for left label
\newcommand{\pairgap}{1pt}
\newcommand{\tightwithin}{-2pt}

% left vertical label (spans the two rows)
\newcommand{\vscenelabel}[1]{%
  \multirow[t]{2}{*}{\rotatebox{0}{\textbf{#1}}\hspace{6pt}}%
}

% 5 images touching in one row
\newcommand{\tilefive}[5]{%
\includegraphics[width=\imgW]{#1}%
\includegraphics[width=\imgW]{#2}%
\includegraphics[width=\imgW]{#3}%
\includegraphics[width=\imgW]{#4}%
\includegraphics[width=\imgW]{#5}%
}

\begin{tabular}{@{}c@{}c@{}}
% col1 = vertical label, col2 = the 5-image strip

% ===================== Scene 1 =====================
\vscenelabel{(a)} &
\tilefive
{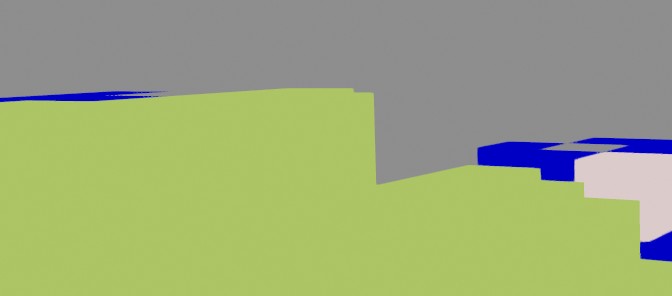}
{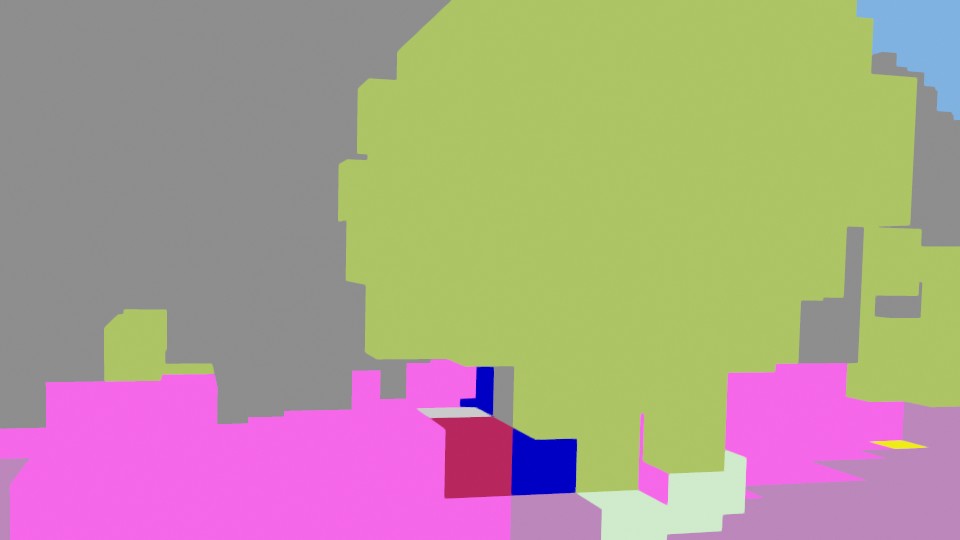}
{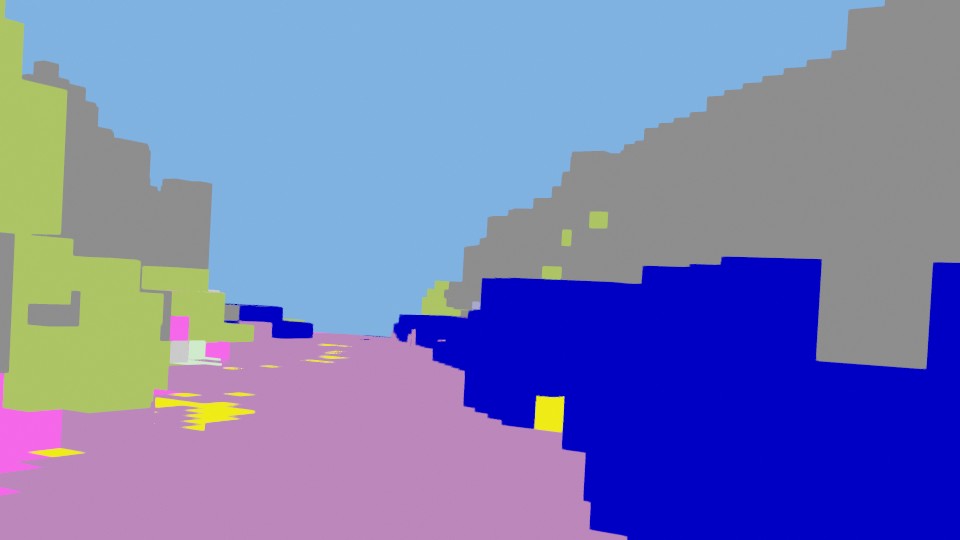}
{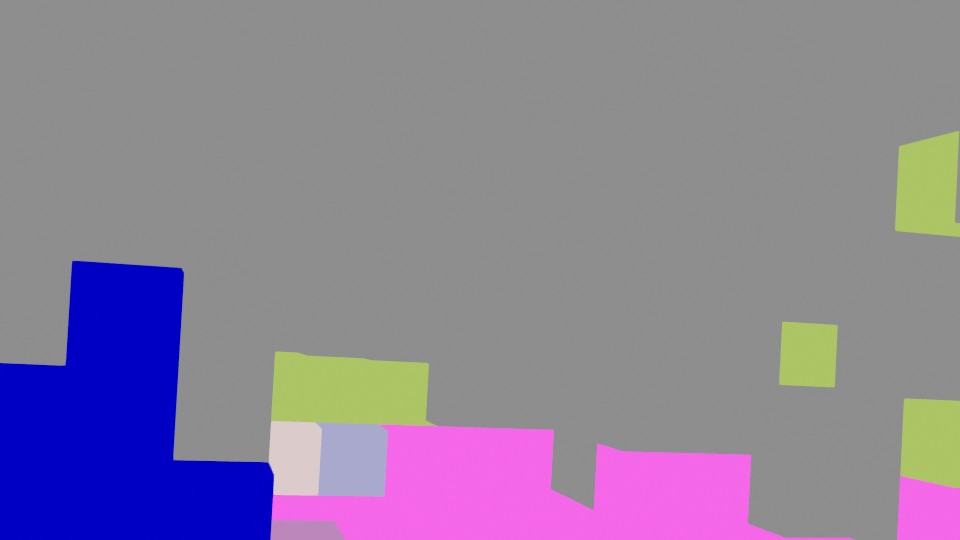}
{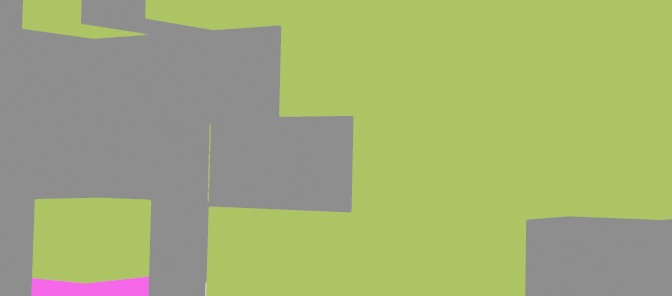}
\\[\tightwithin]
&
\tilefive
{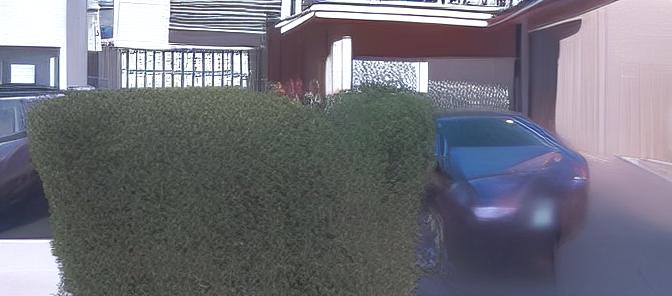}
{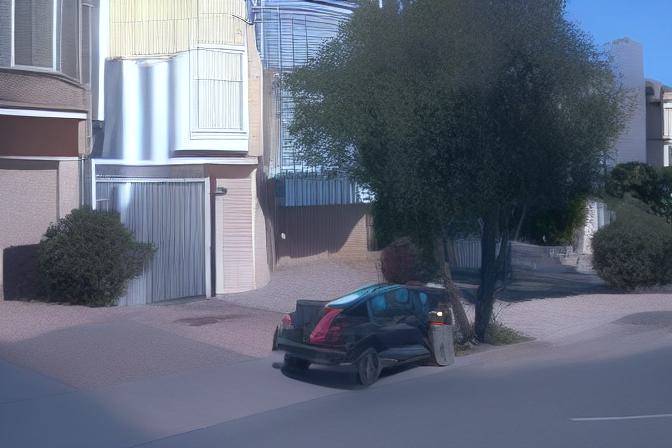}
{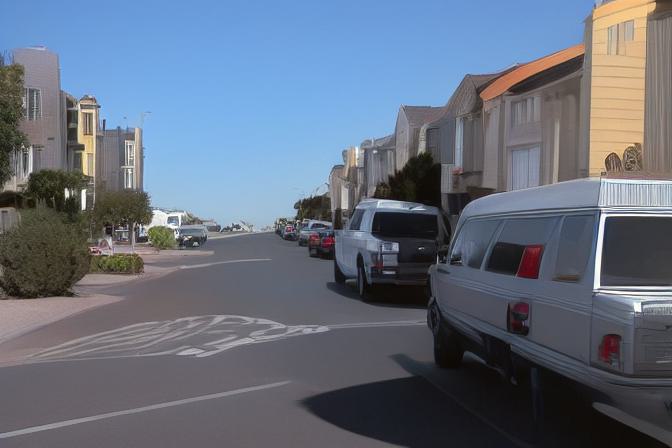}
{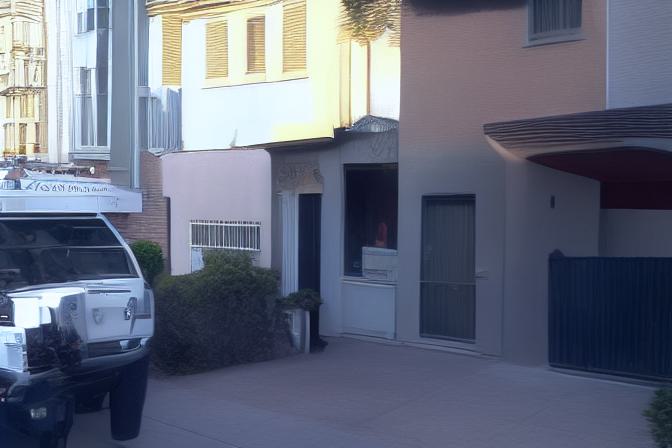}
{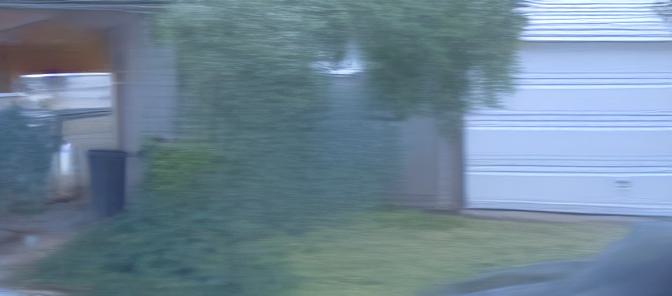}
\\[\pairgap]

% ===================== Scene 1 =====================
\vscenelabel{(b)} &
\tilefive
{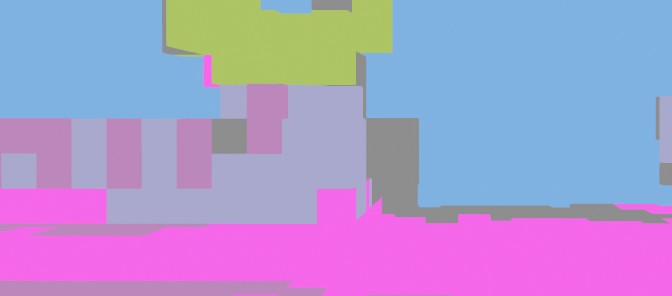}
{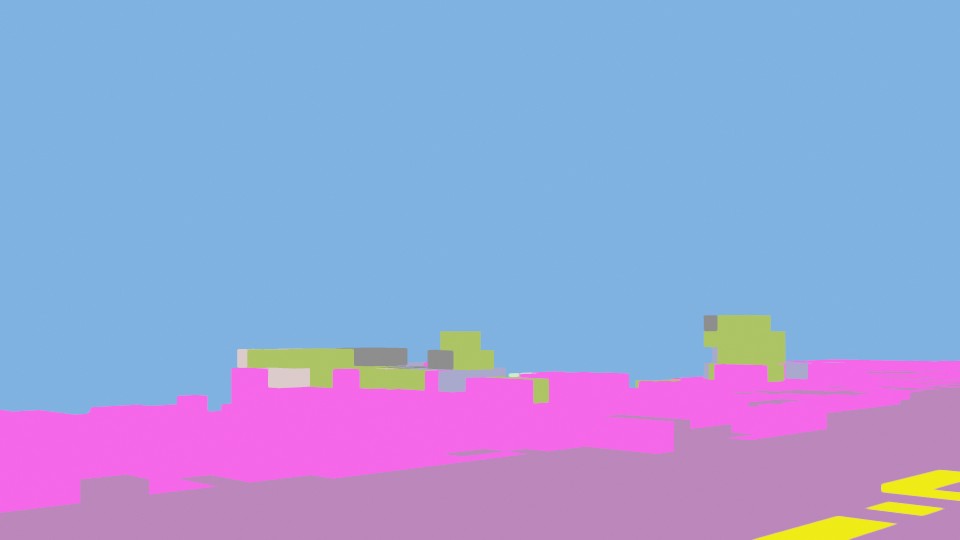}
{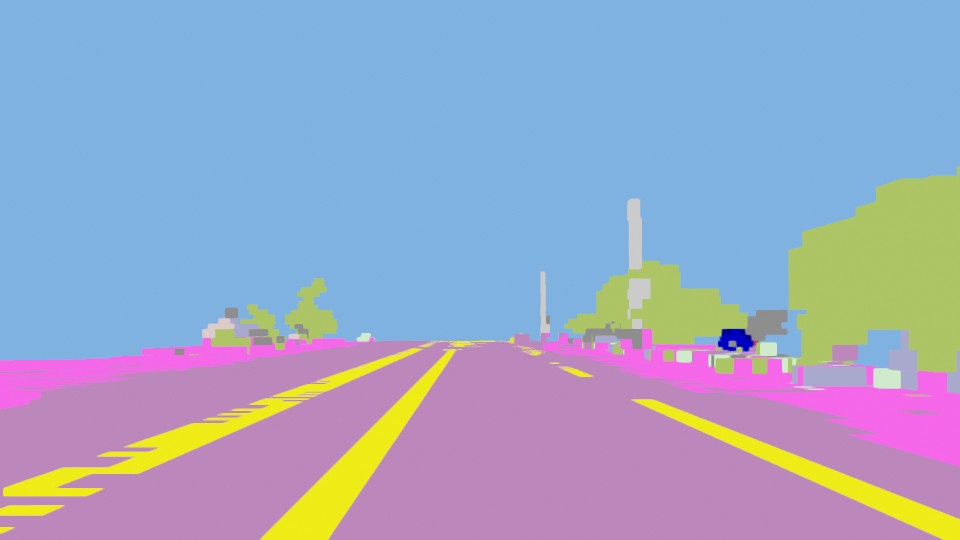}
{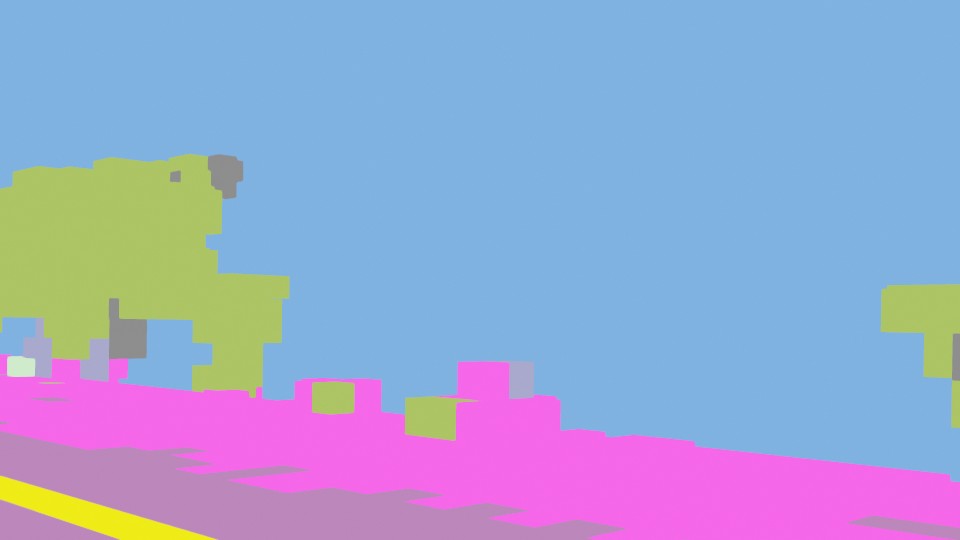}
{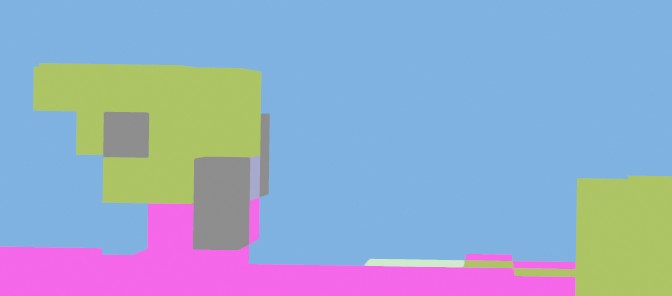}
\\[\tightwithin]
&
\tilefive
{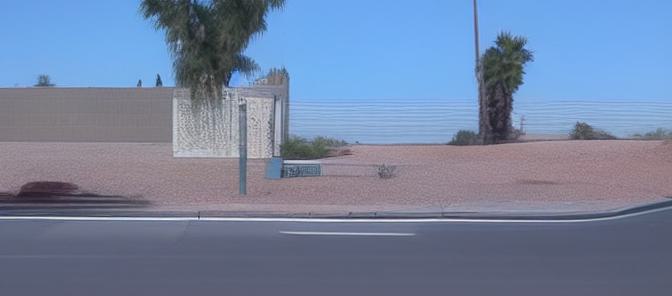}
{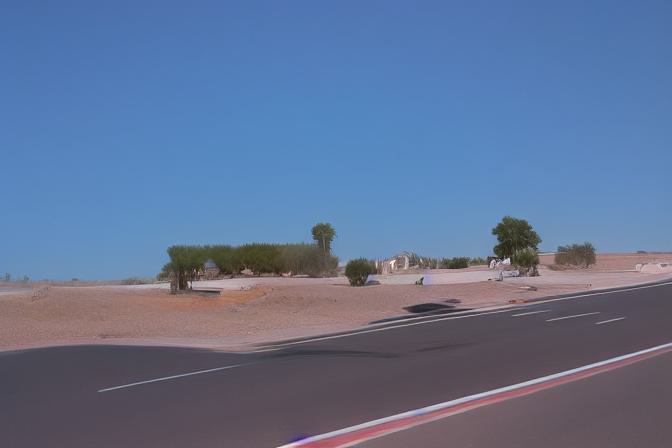}
{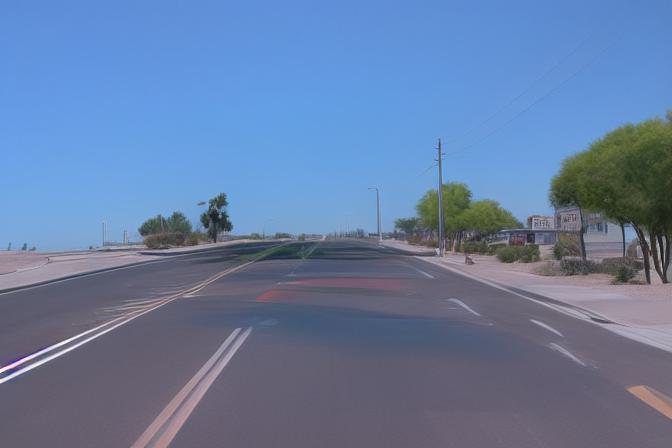}
{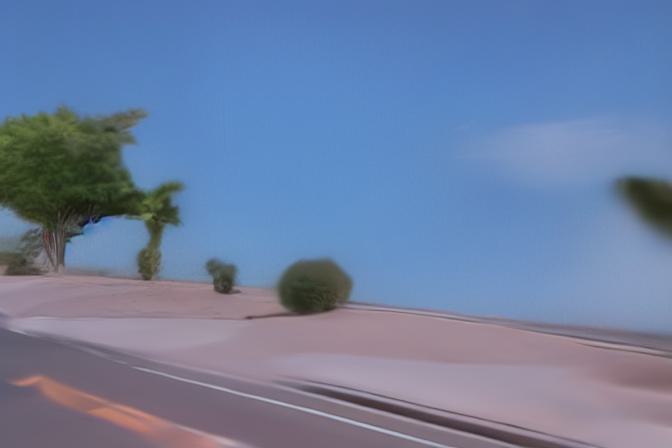}
{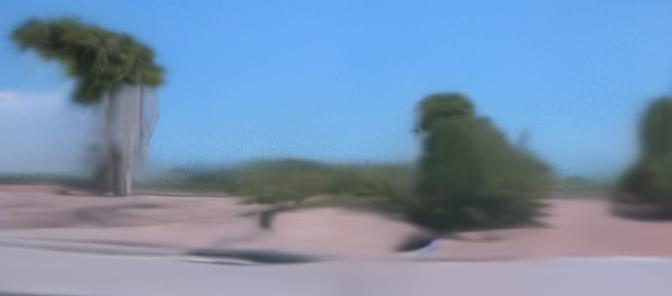}
\\[\pairgap]

% ===================== Scene 1 =====================
\vscenelabel{(c)} &
\tilefive
{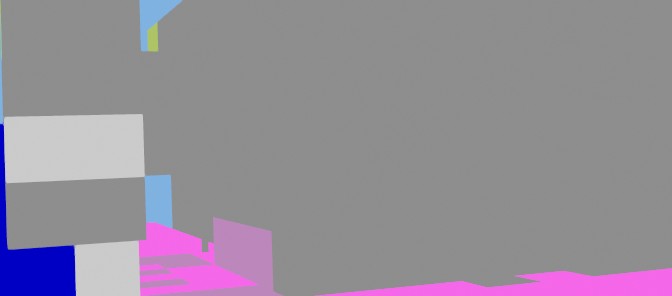}
{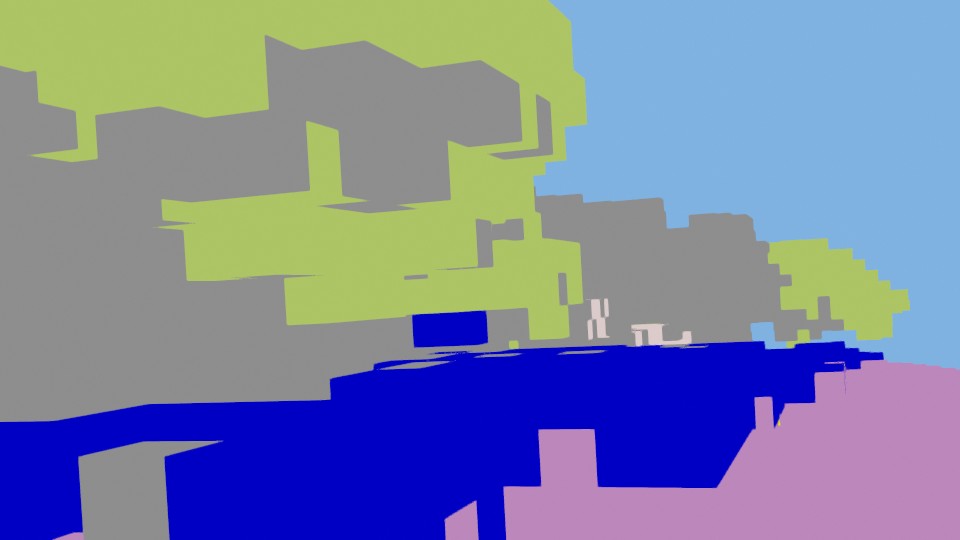}
{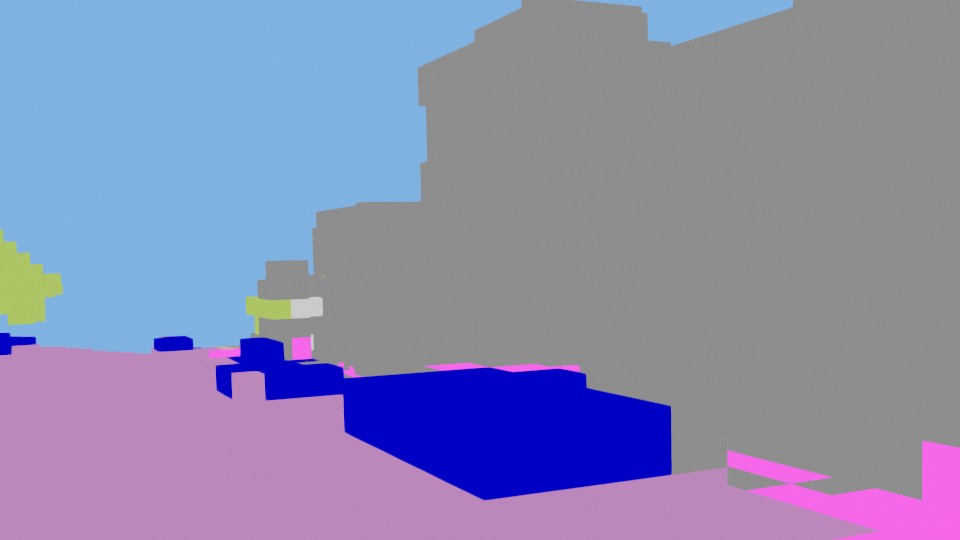}
{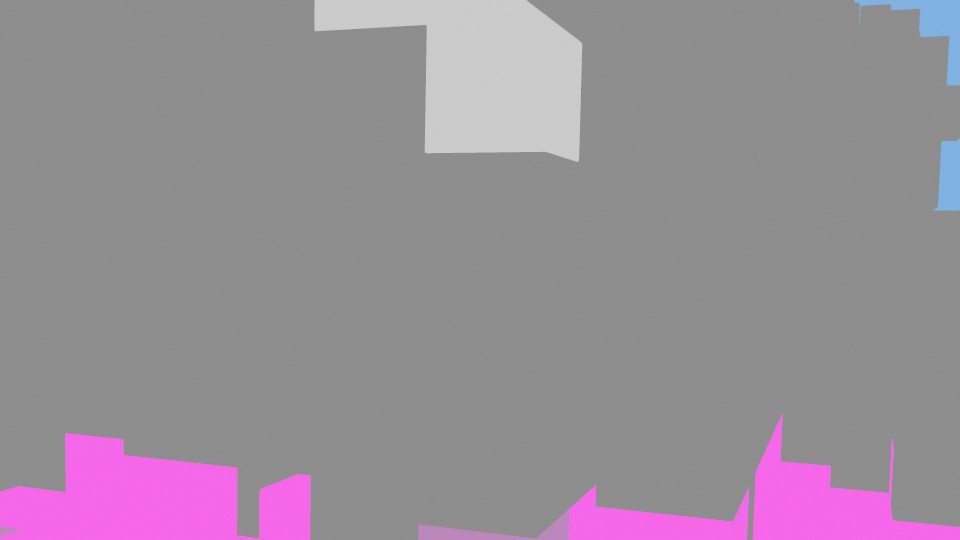}
{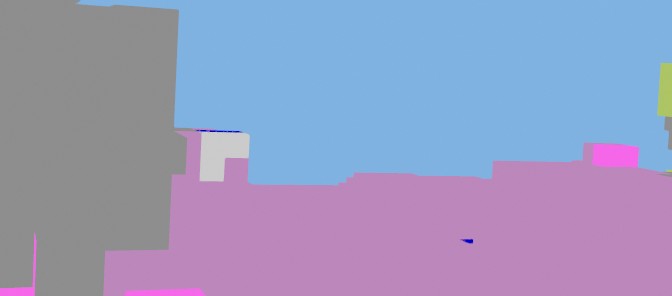}
\\[\tightwithin]
&
\tilefive
{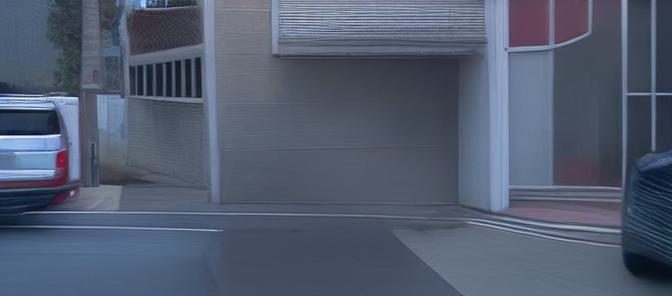}
{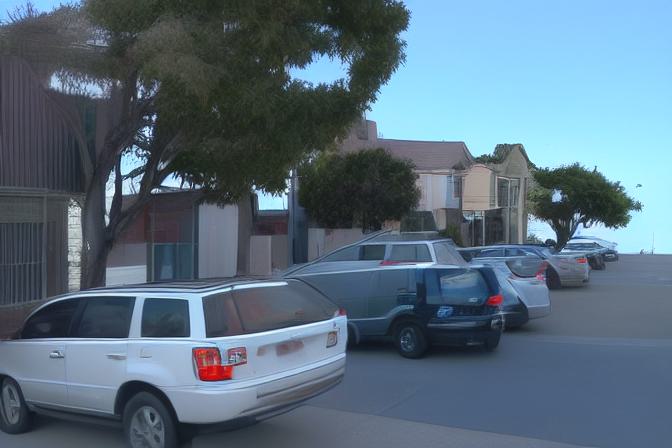}
{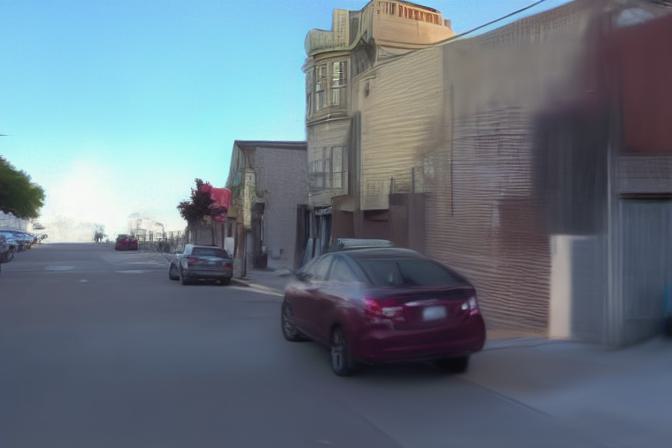}
{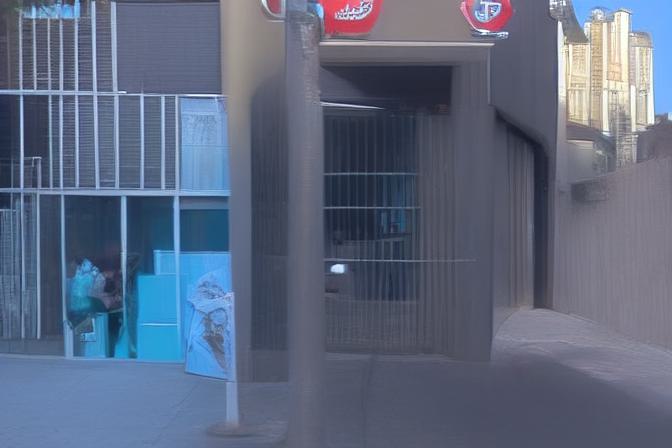}
{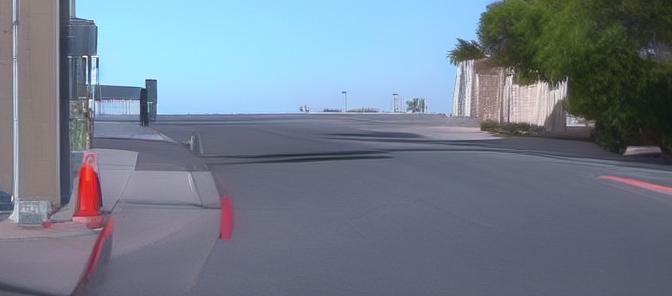}
\\[\pairgap]

% ===================== Scene 4 =====================
% ===================== Scene 1 =====================
\vscenelabel{(d)} &
\tilefive
{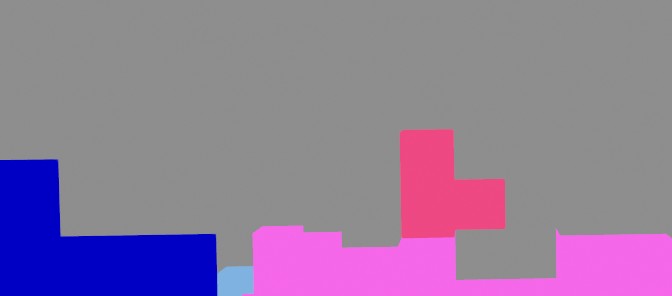}
{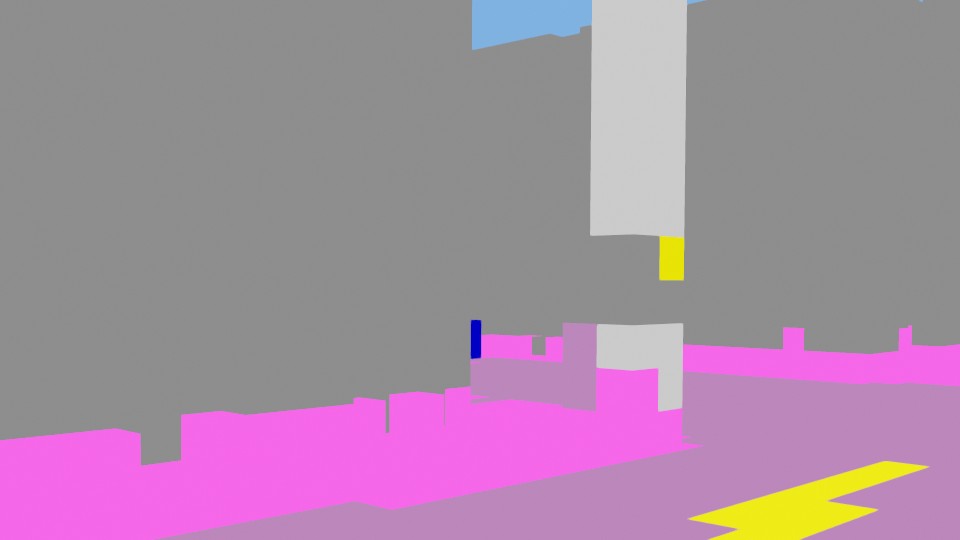}
{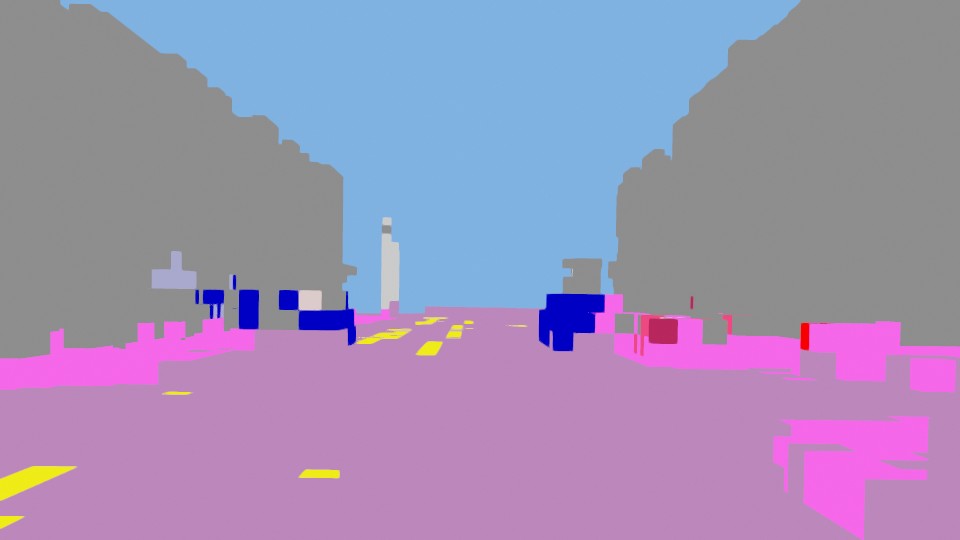}
{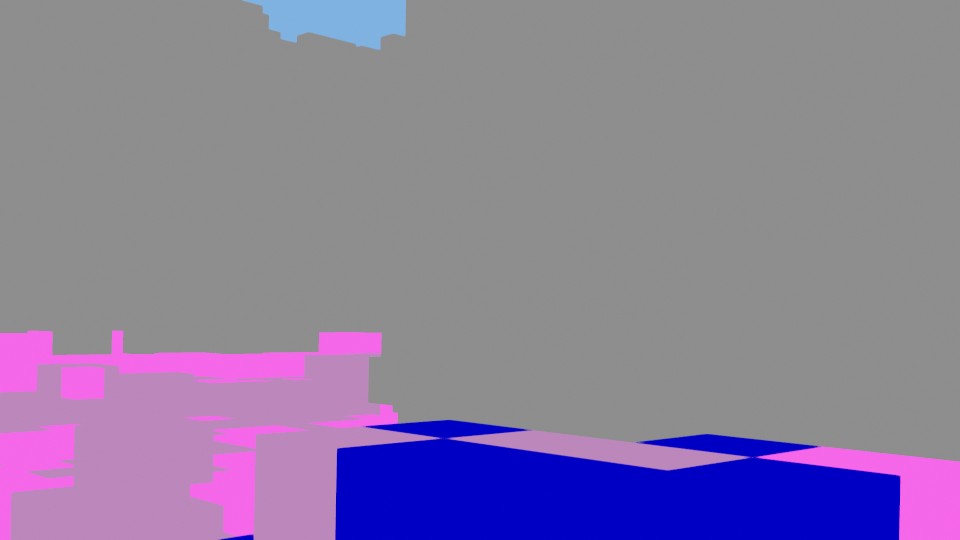}
{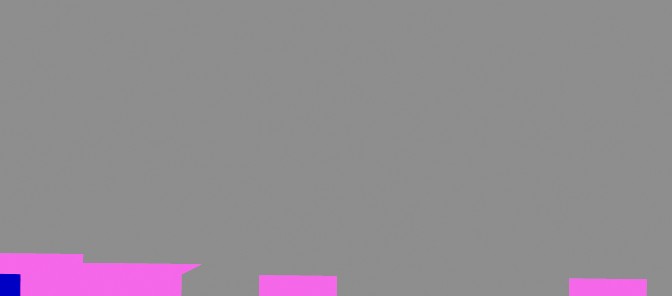}
\\[\tightwithin]
&
\tilefive
{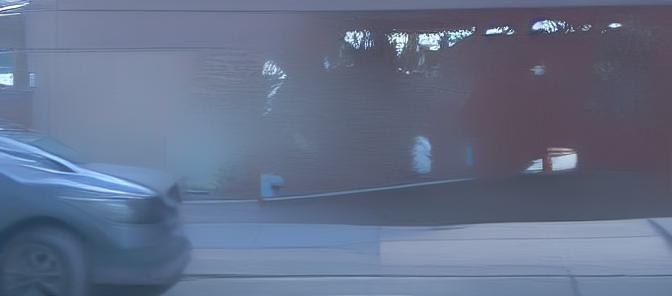}
{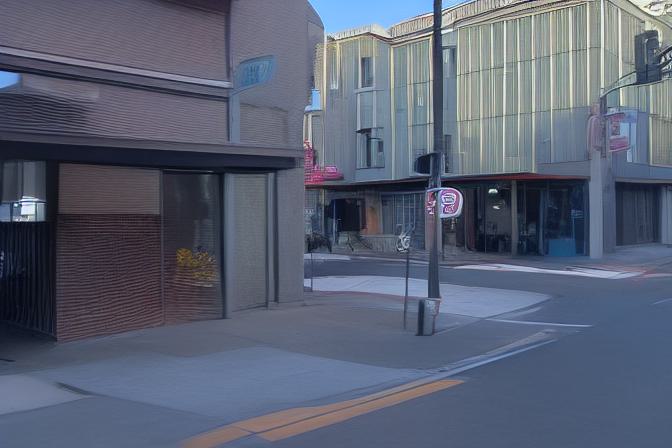}
{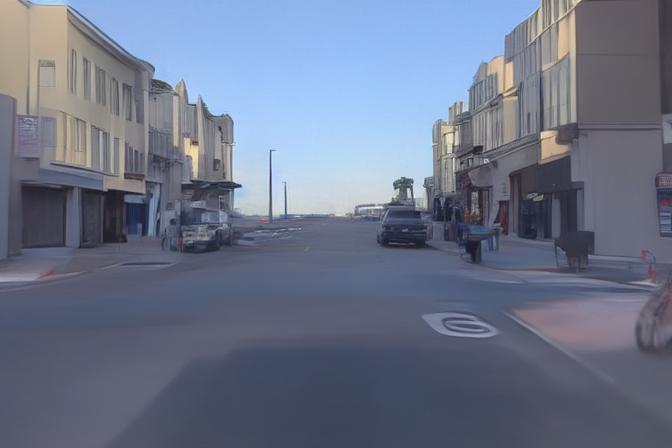}
{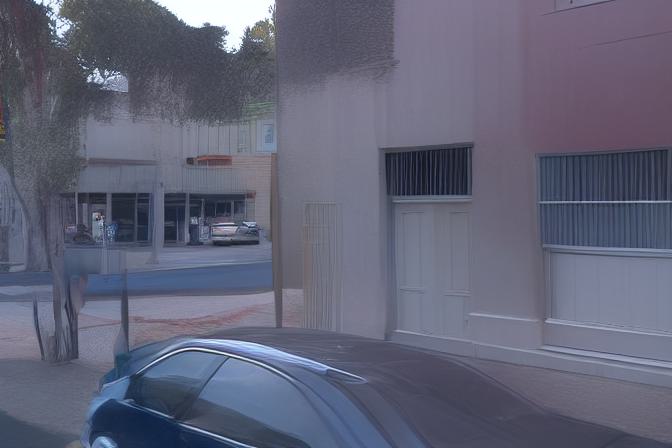}
{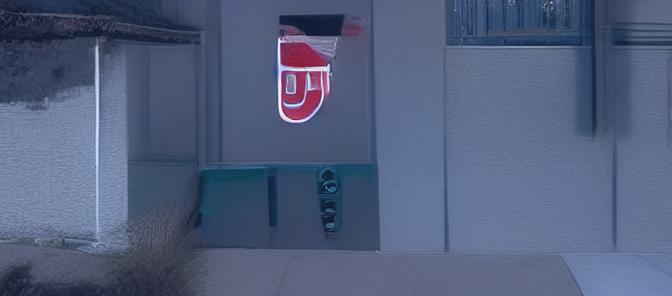}
\\[\pairgap]

% ===================== Scene 1 =====================
\vscenelabel{(e)} &
\tilefive
{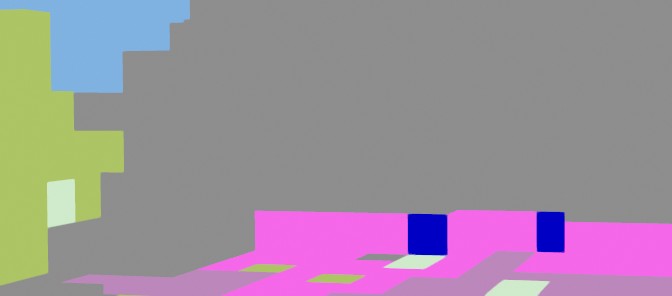}
{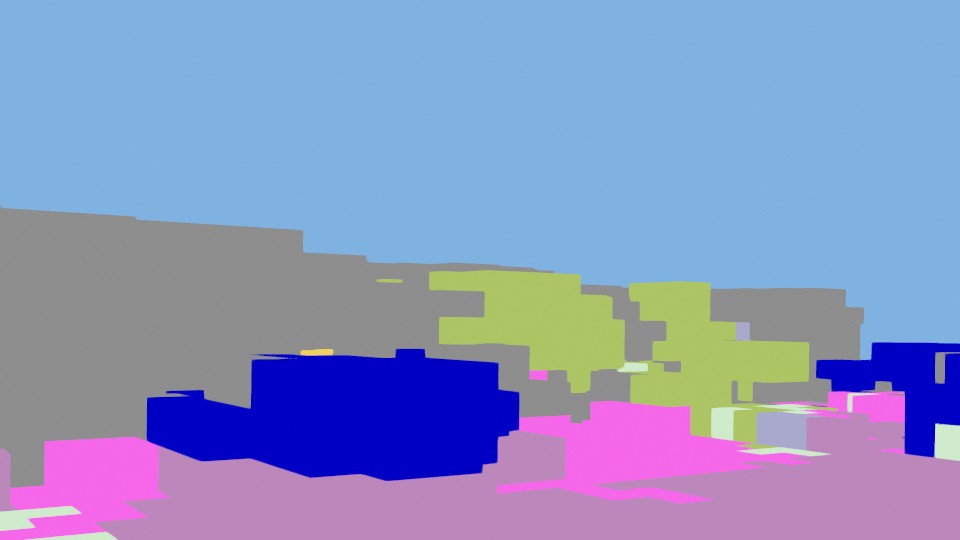}
{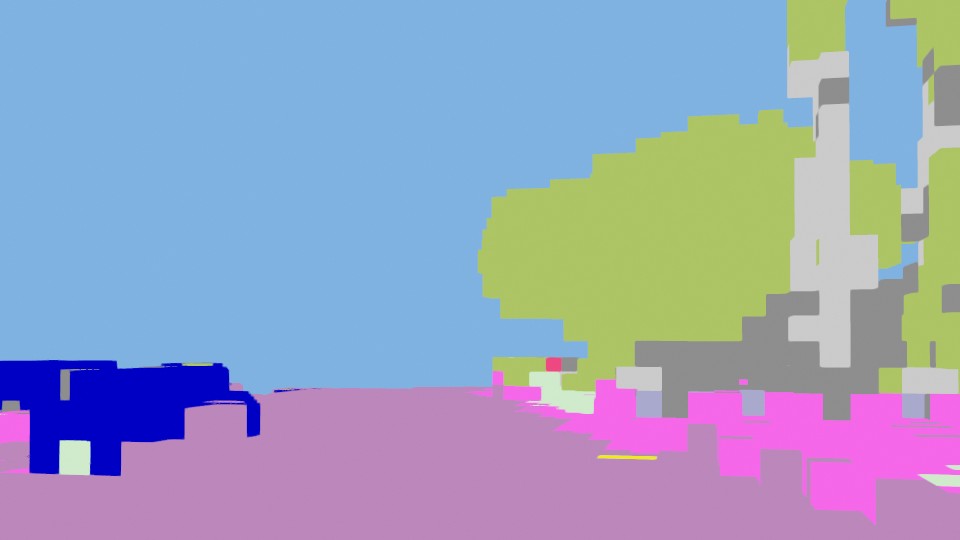}
{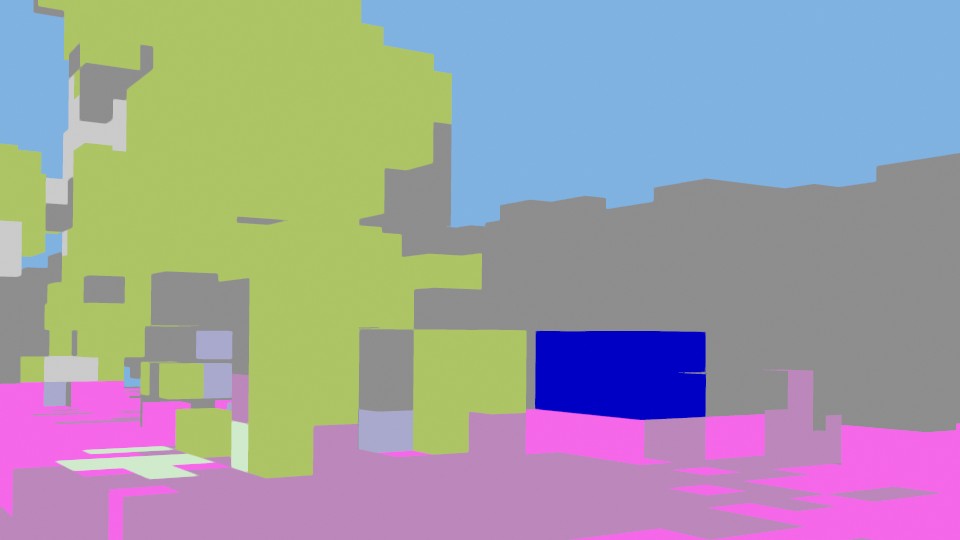}
{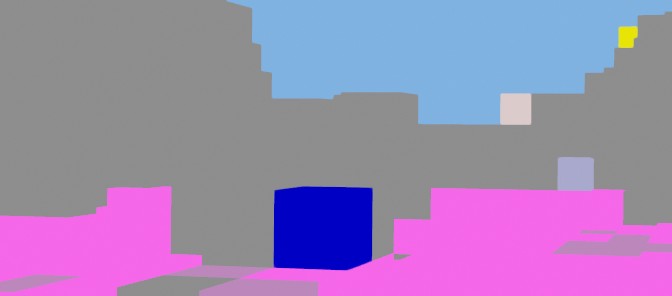}
\\[\tightwithin]
&
\tilefive
{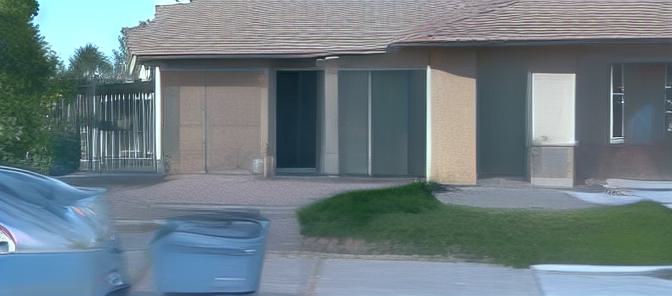}
{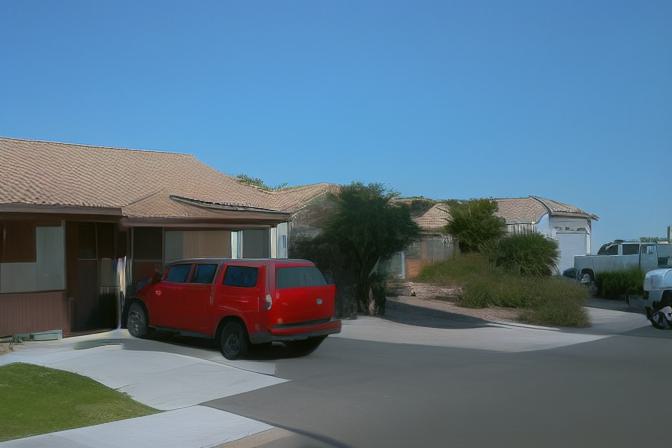}
{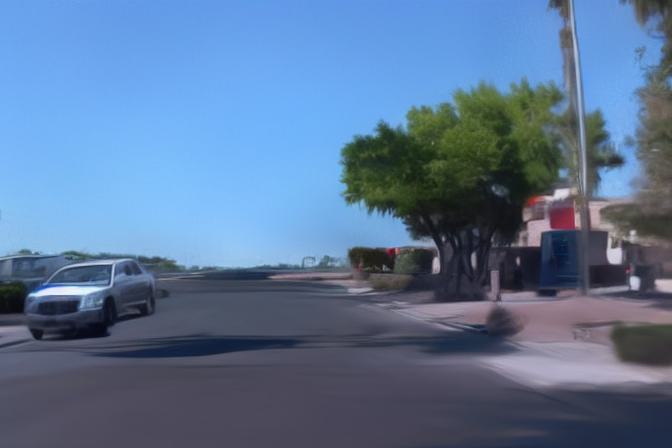}
{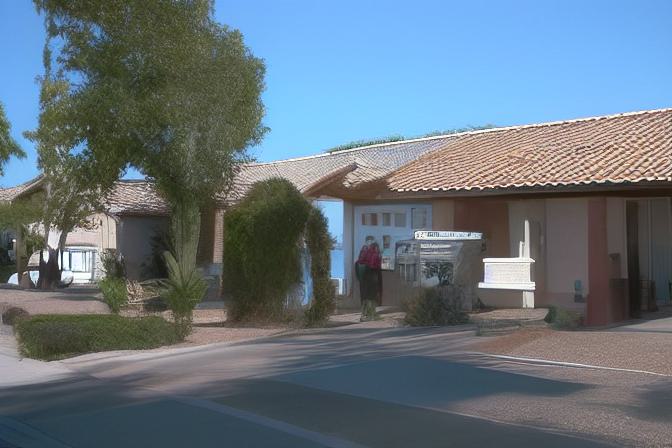}
{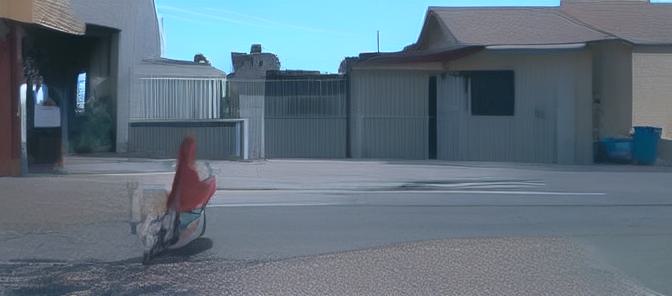}
\\[\pairgap]
% ===================== Scene 1 =====================
\vscenelabel{(f)} &
\tilefive
{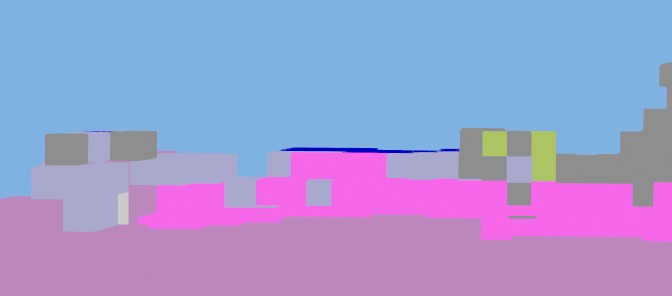}
{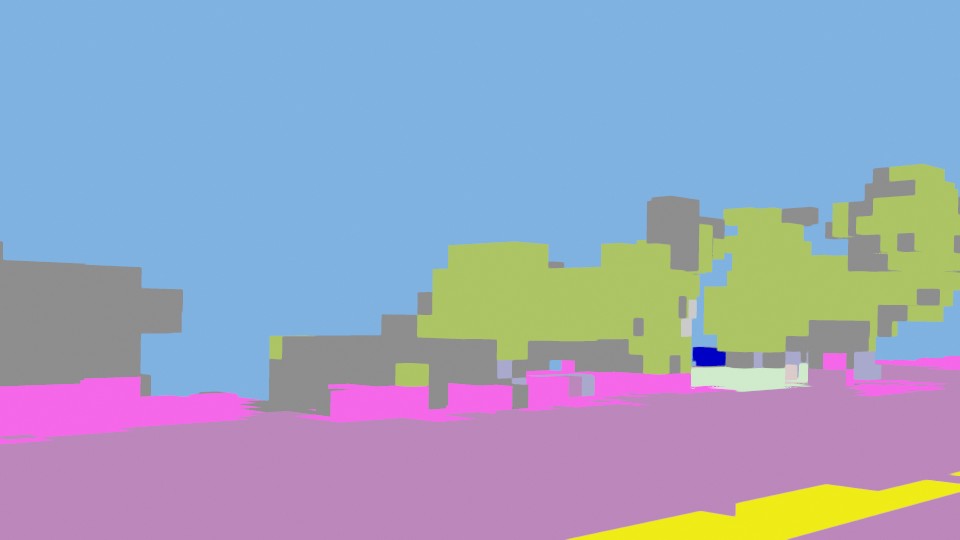}
{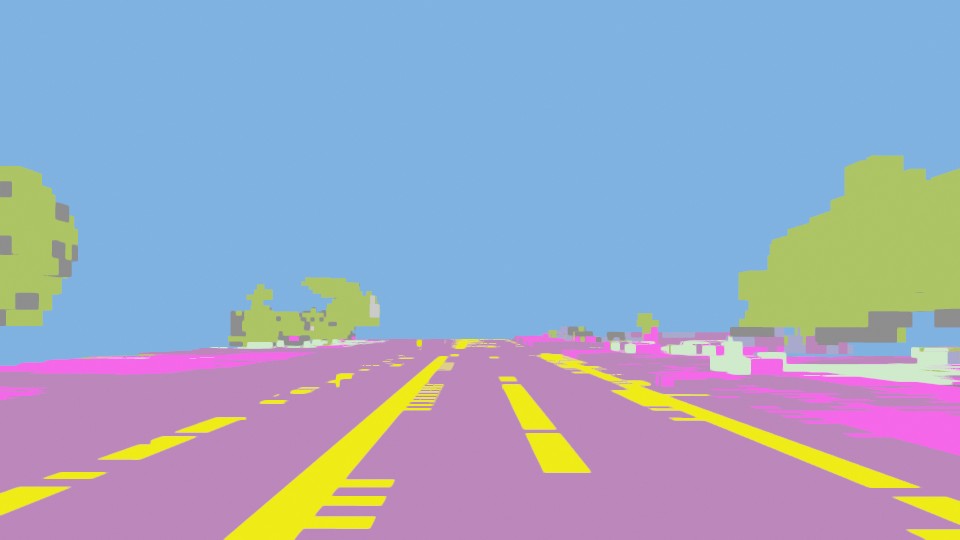}
{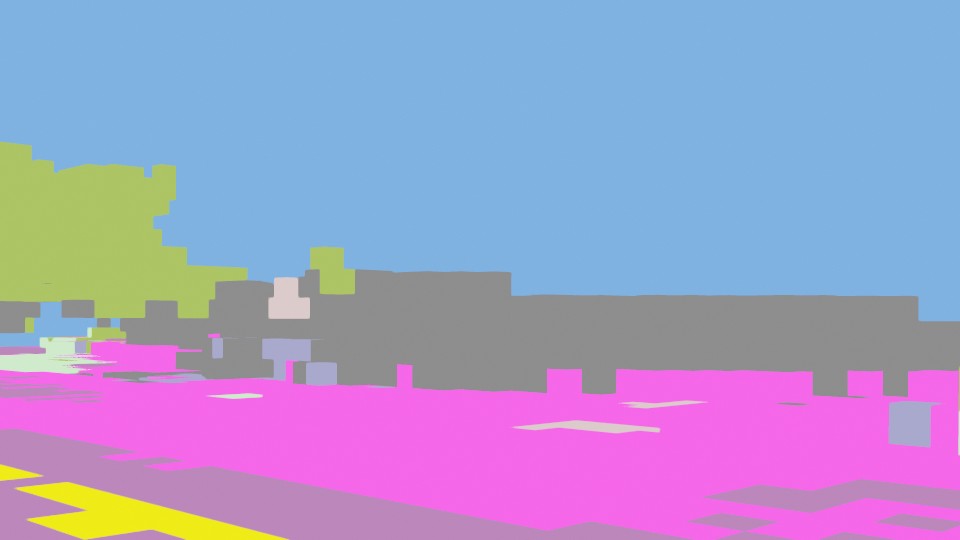}
{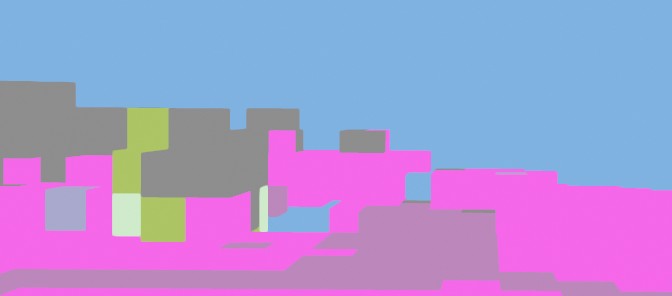}
\\[\tightwithin]
&
\tilefive
{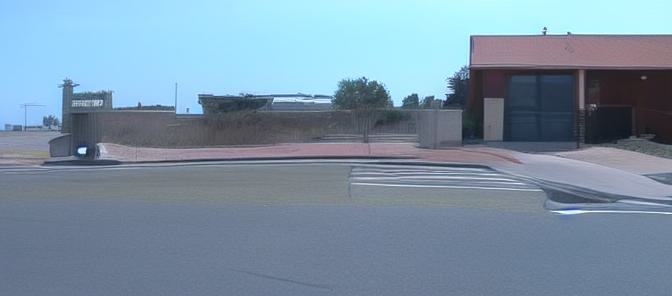}
{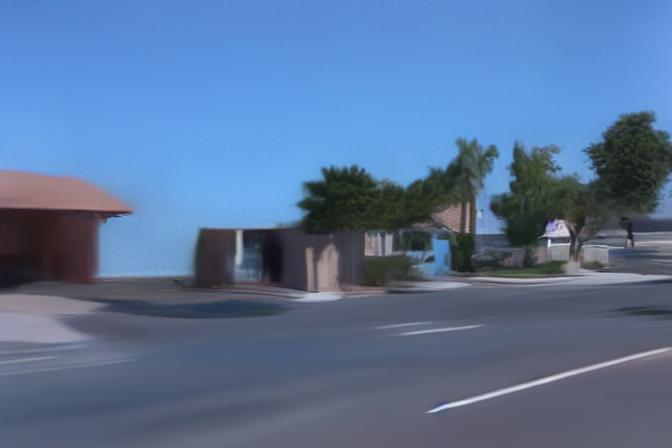}
{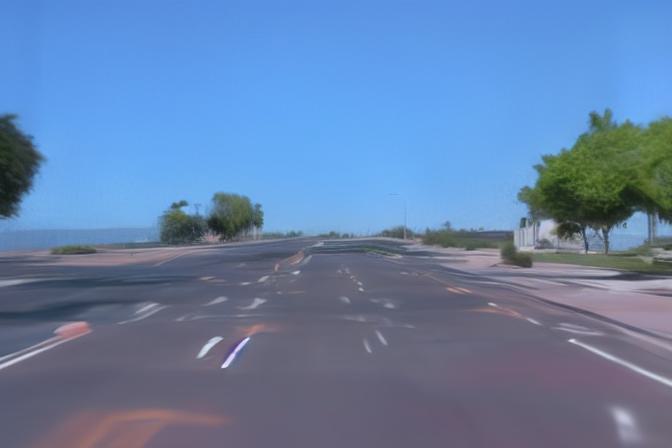}
{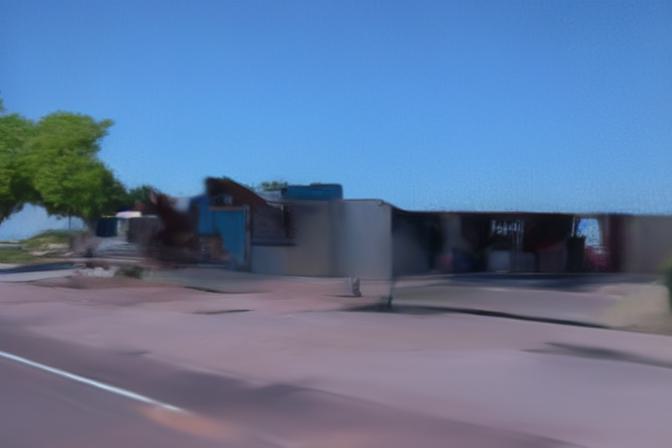}
{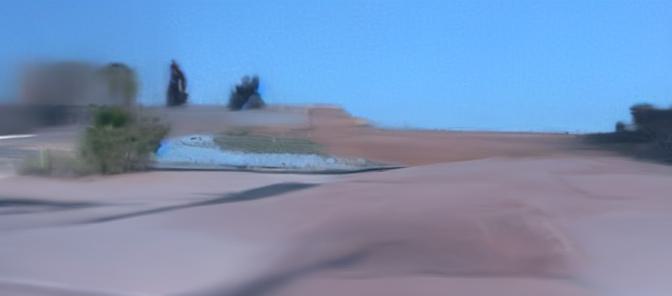}
\\[\pairgap]

\end{tabular}

\caption{\textbf{Additional Qualitative results on WOD~\cite{waymo}.} We show six WOD scenes (a--f). For each scene, the \textbf{top} strip visualizes the semantic voxel rendering used for conditioning, and the \textbf{bottom} strip shows the corresponding generated scene from 5 camera views.}
\label{fig:results_waymo}
\end{figure*}

\begin{figure*}[tbh]
\centering
\scriptsize
\setlength{\tabcolsep}{0pt}
\renewcommand{\arraystretch}{1.0}

% ---- knobs ----
\newcommand{\imgW}{0.17\textwidth}  % slightly smaller to make room for left label
\newcommand{\pairgap}{1pt}
\newcommand{\tightwithin}{-2pt}

% left vertical label (spans the two rows)
\newcommand{\vscenelabel}[1]{%
  \multirow[t]{2}{*}{\rotatebox{0}{\textbf{#1}}\hspace{6pt}}%
}

% 5 images touching in one row
\newcommand{\tilefive}[5]{%
\includegraphics[width=\imgW]{#1}%
\includegraphics[width=\imgW]{#2}%
\includegraphics[width=\imgW]{#3}%
\includegraphics[width=\imgW]{#4}%
\includegraphics[width=\imgW]{#5}%
}

\begin{tabular}{@{}c@{}c@{}}
% col1 = vertical label, col2 = the 5-image strip

% ===================== Scene 1 =====================
\vscenelabel{(a)} &
\tilefive
{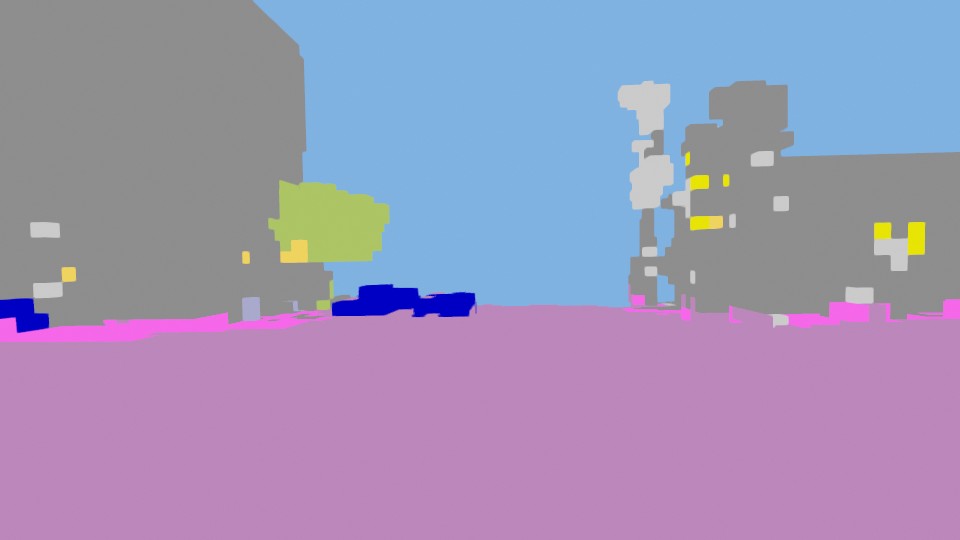}
{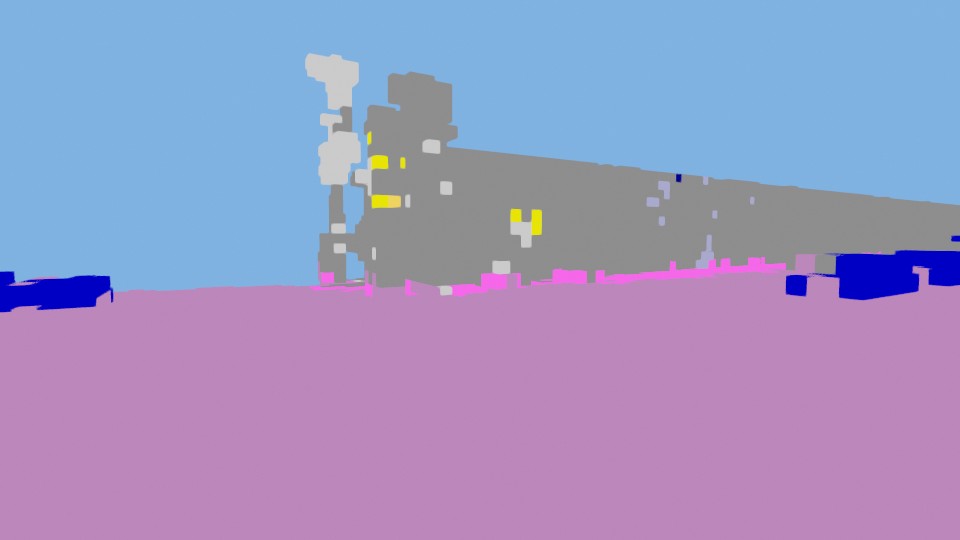}
{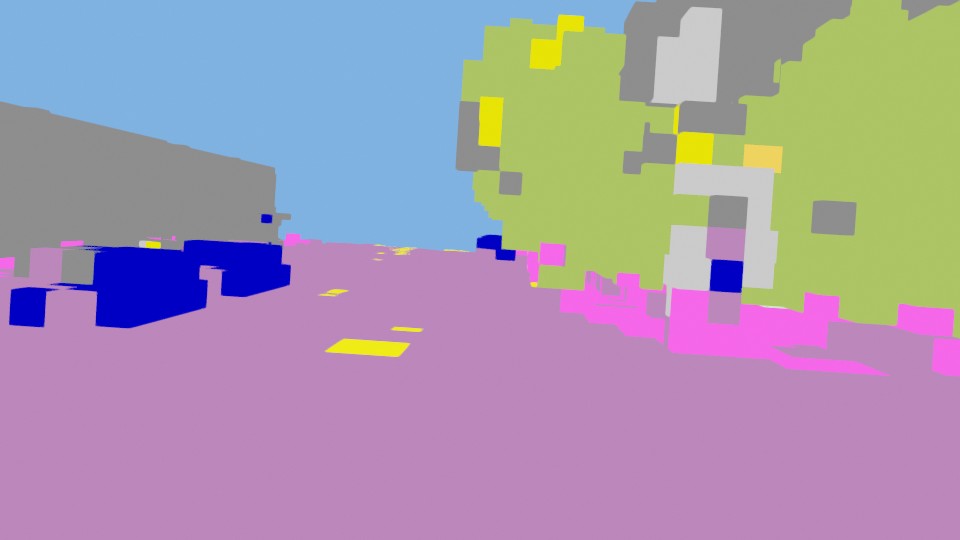}
{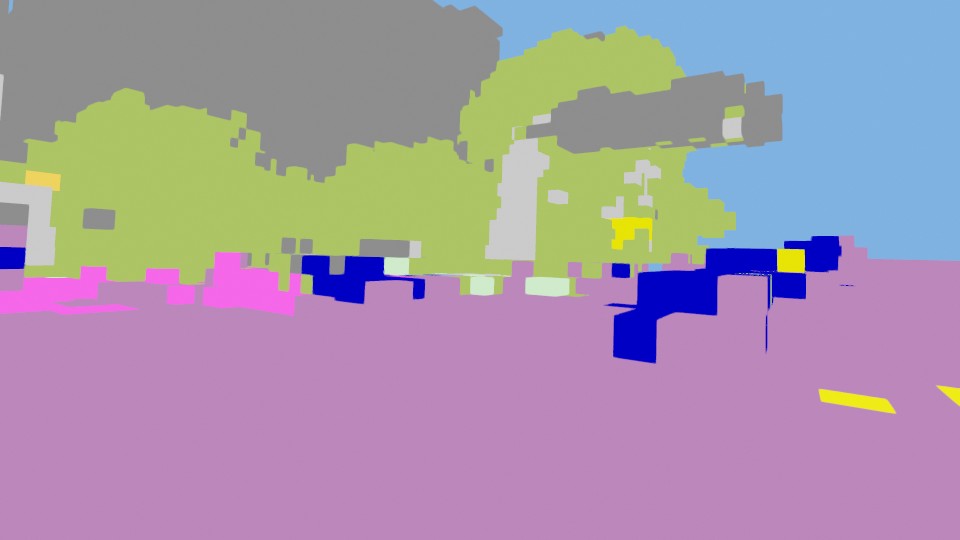}
{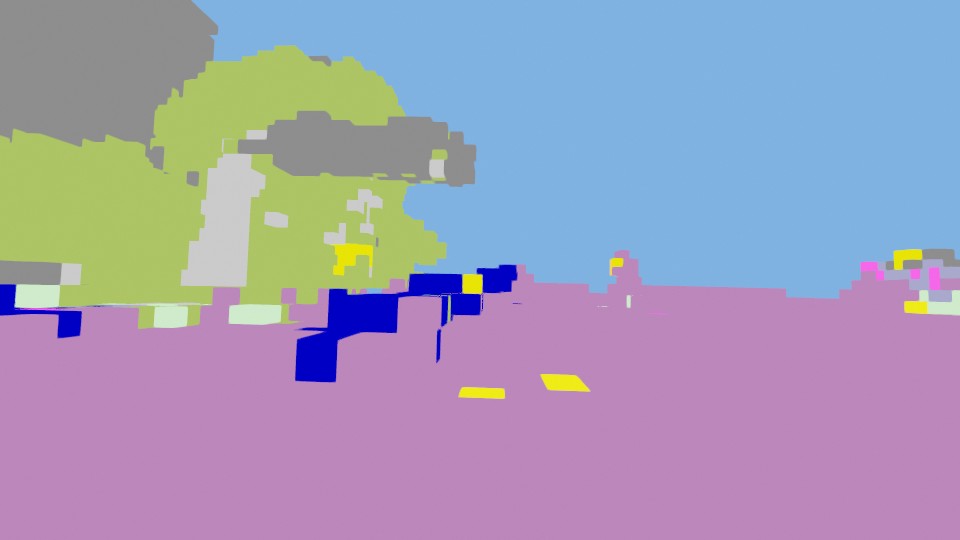}
\\[\tightwithin]
&
\tilefive
{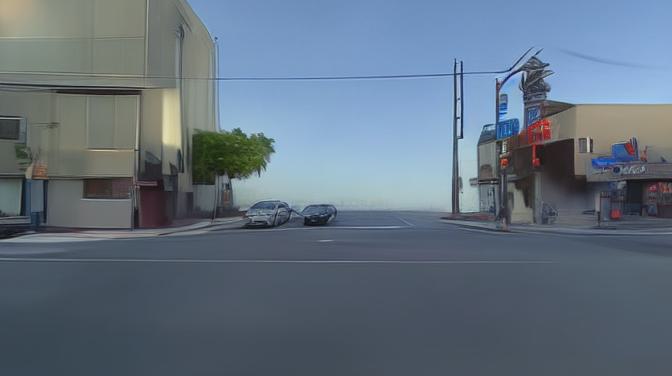}
{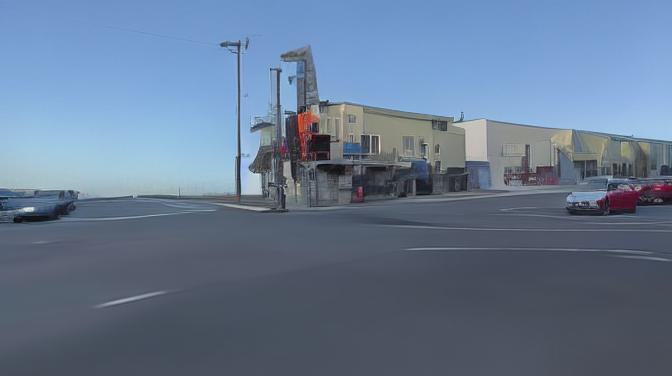}
{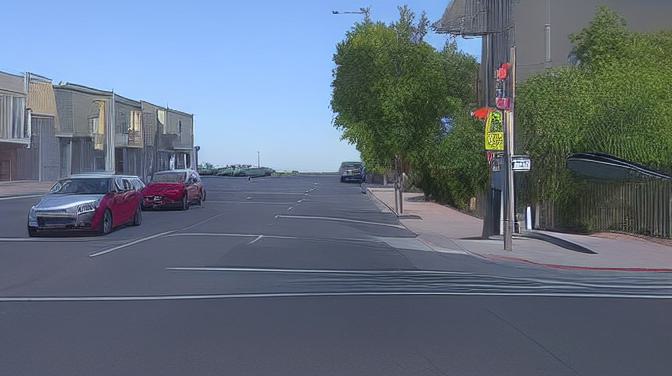}
{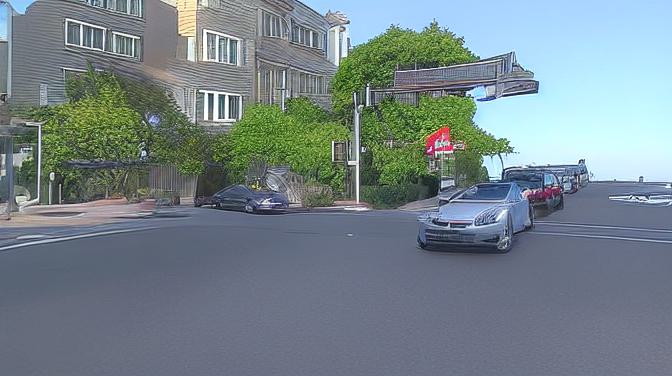}
{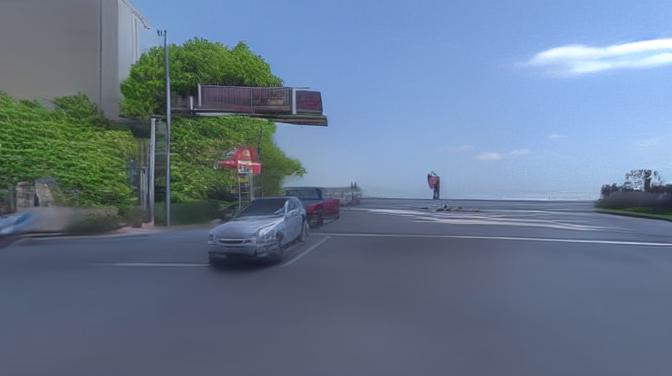}
\\[\pairgap]

% ===================== Scene 1 =====================
\vscenelabel{(b)} &
\tilefive
{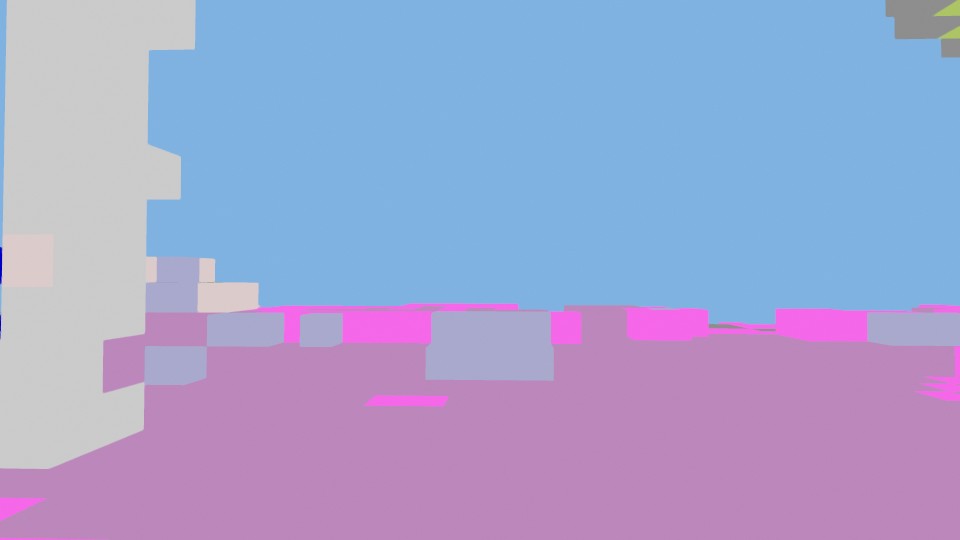}
{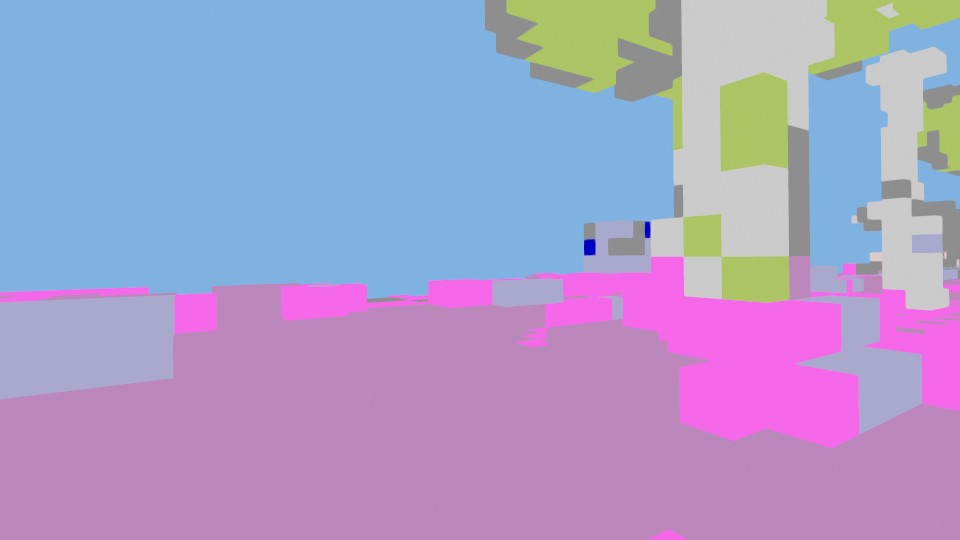}
{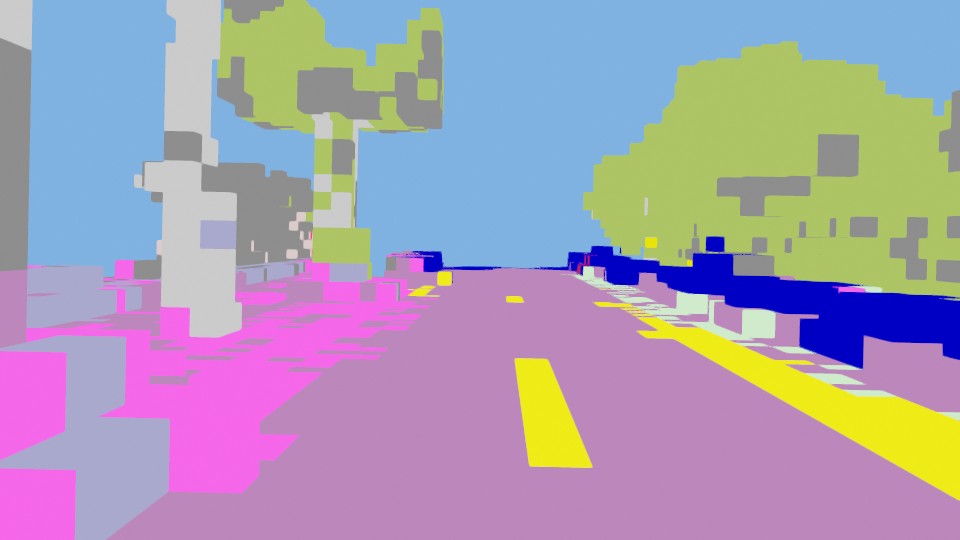}
{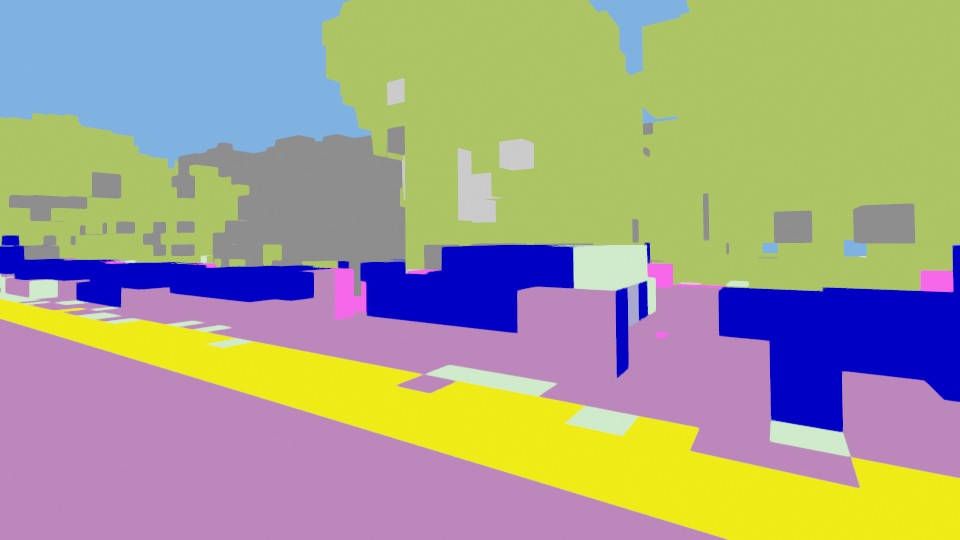}
{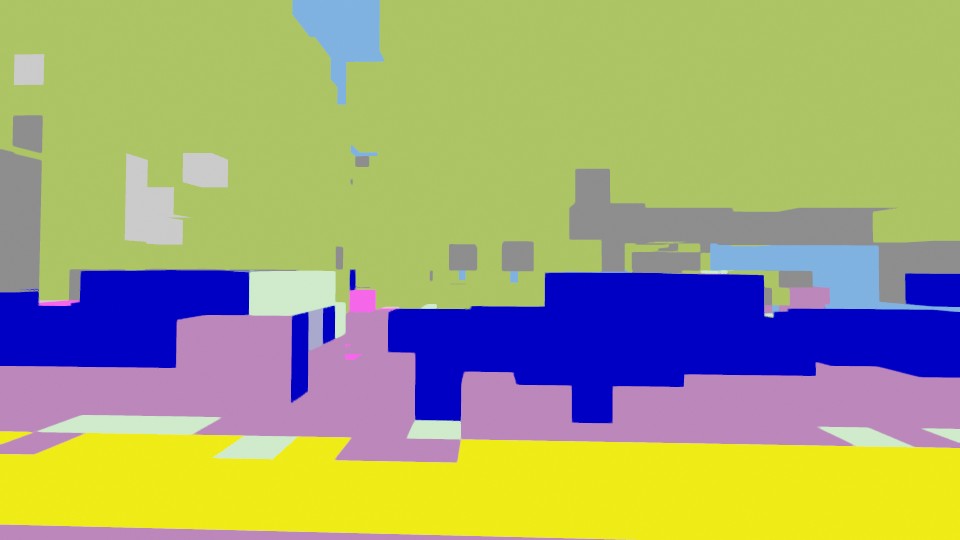}
\\[\tightwithin]
&
\tilefive
{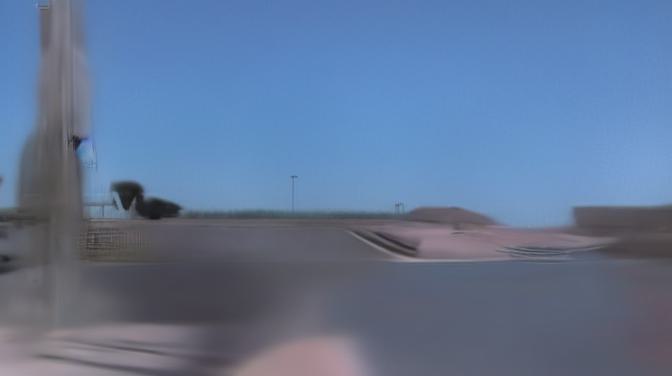}
{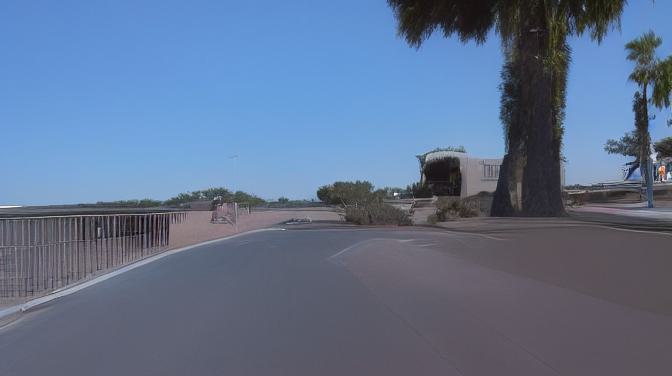}
{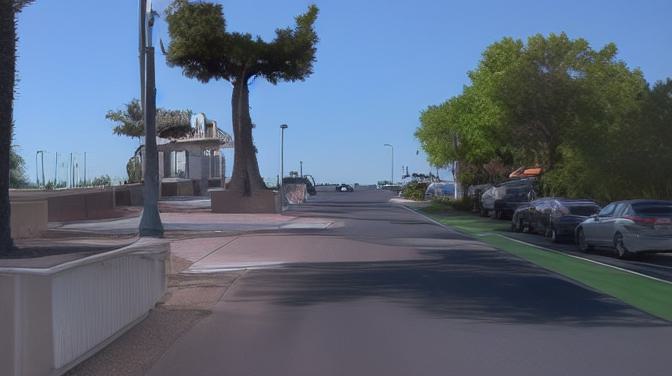}
{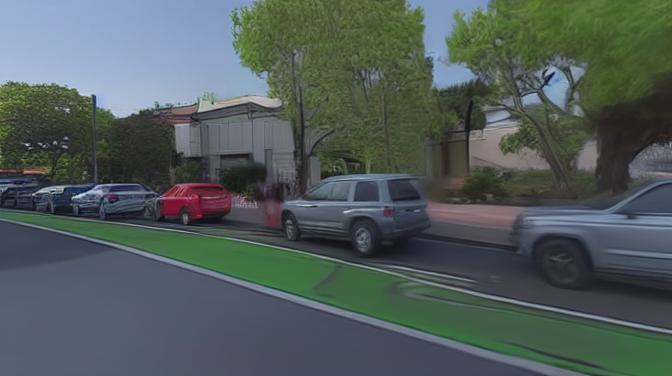}
{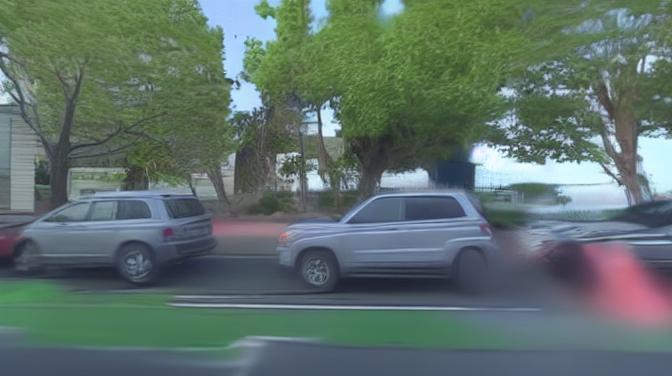}
\\[\pairgap]

% ===================== Scene 1 =====================
\vscenelabel{(c)} &
\tilefive
{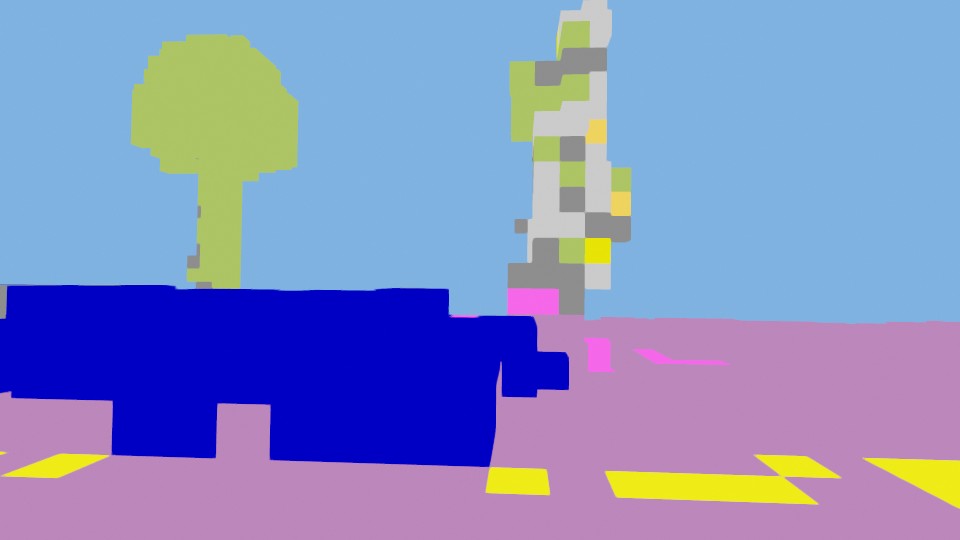}
{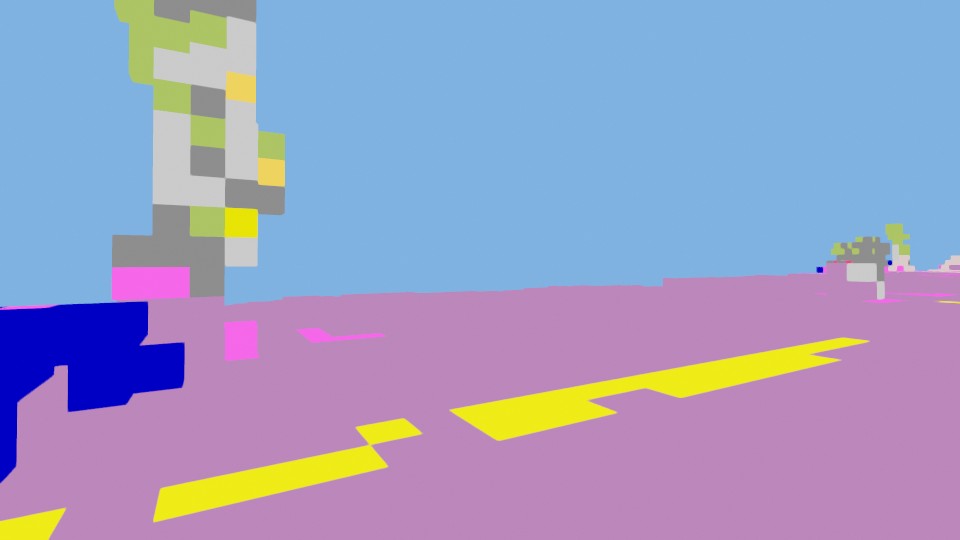}
{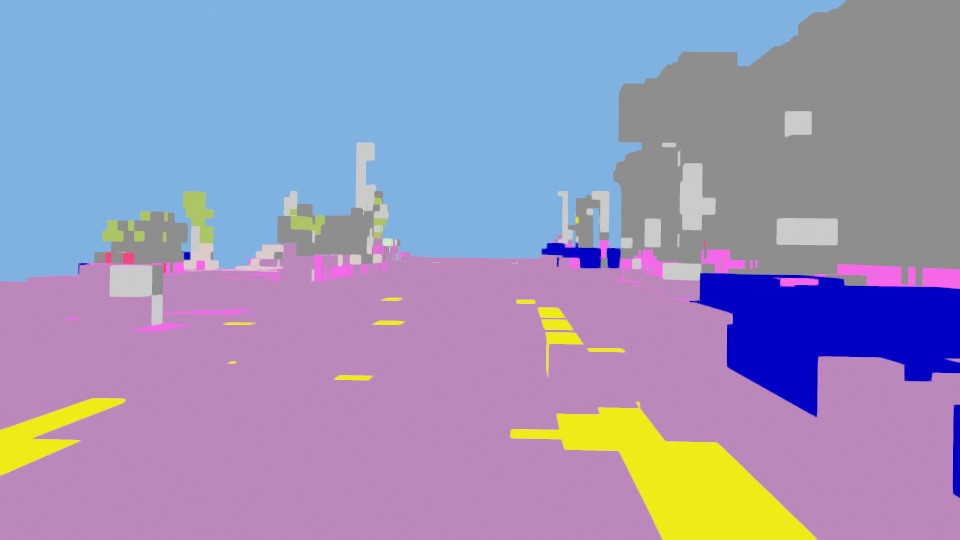}
{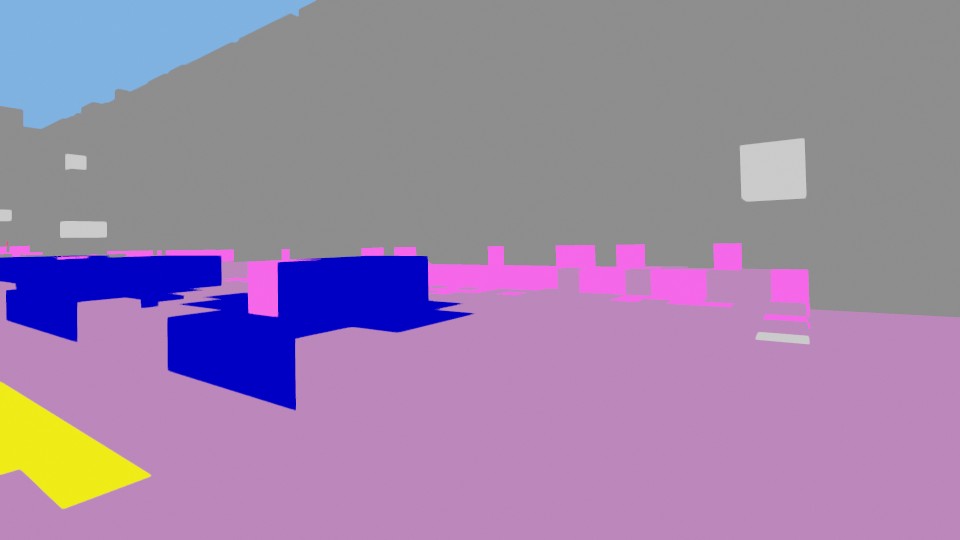}
{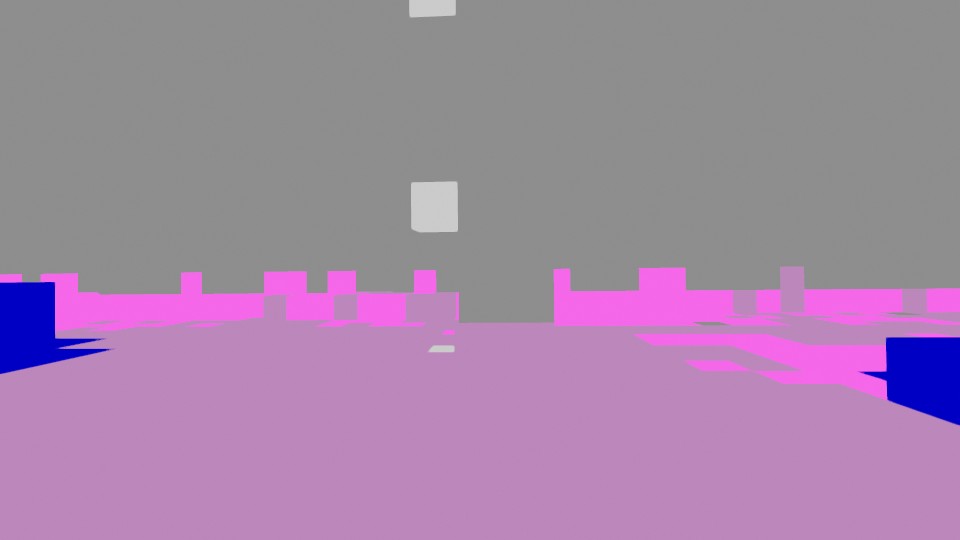}
\\[\tightwithin]
&
\tilefive
{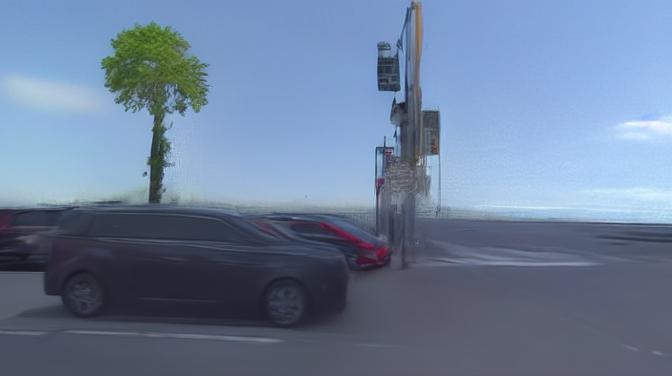}
{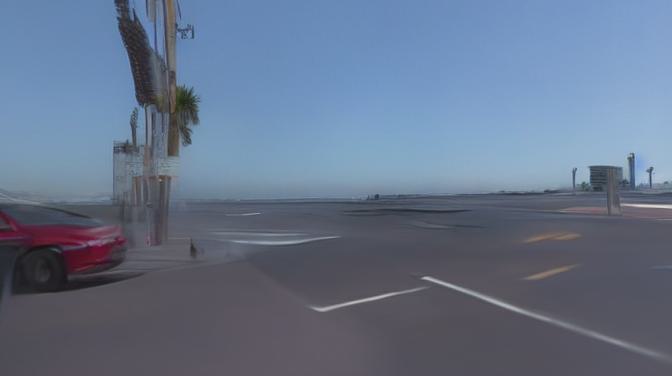}
{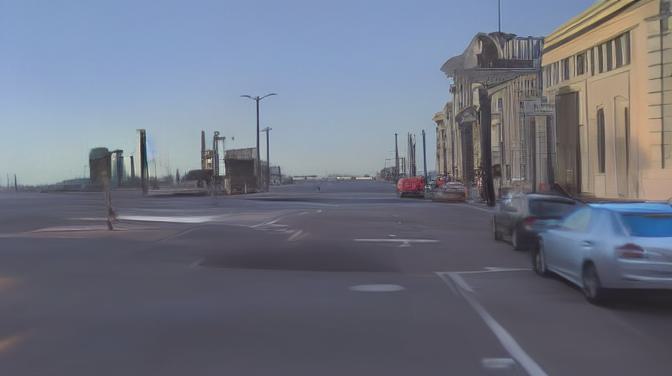}
{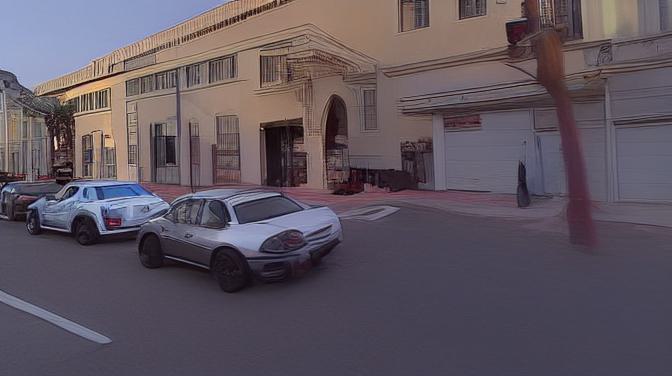}
{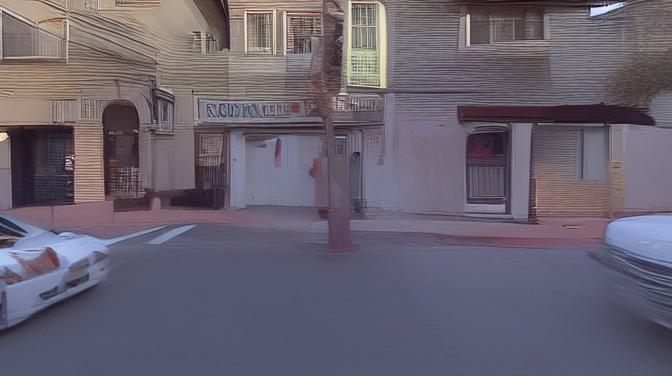}
\\[\pairgap]

% ===================== Scene 4 =====================
% ===================== Scene 1 =====================
\vscenelabel{(d)} &
\tilefive
{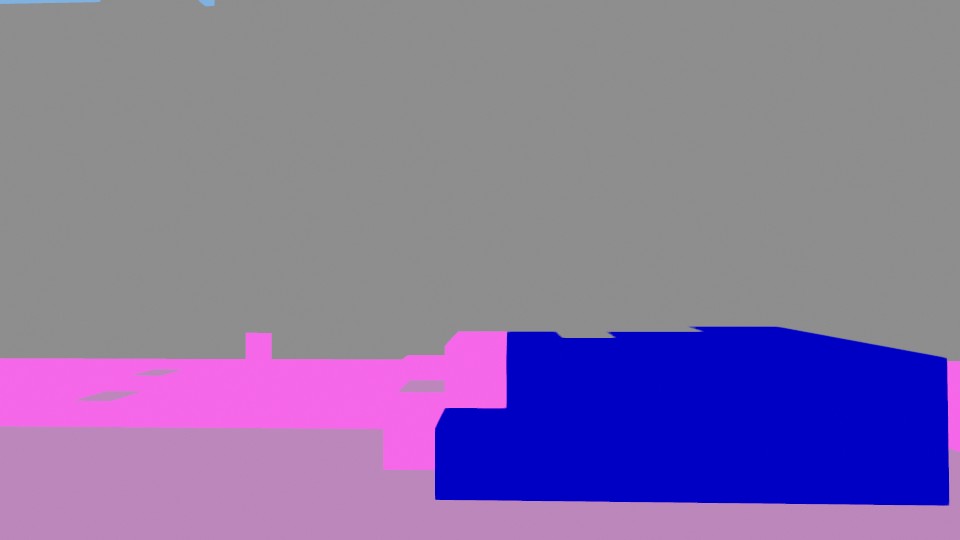}
{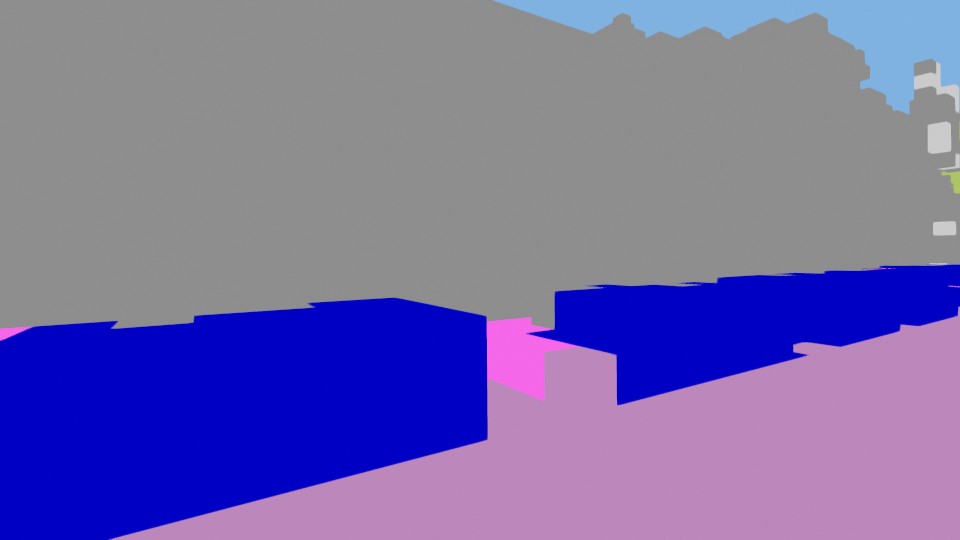}
{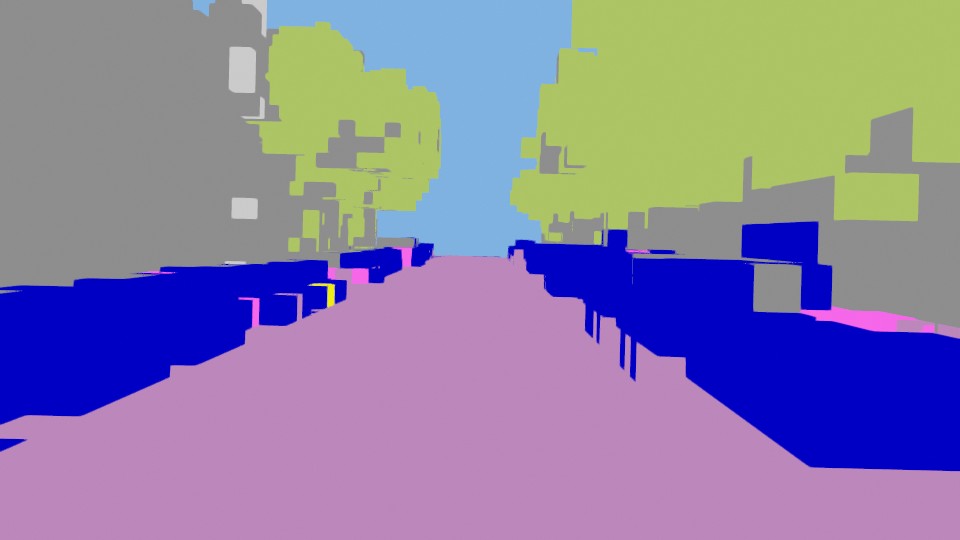}
{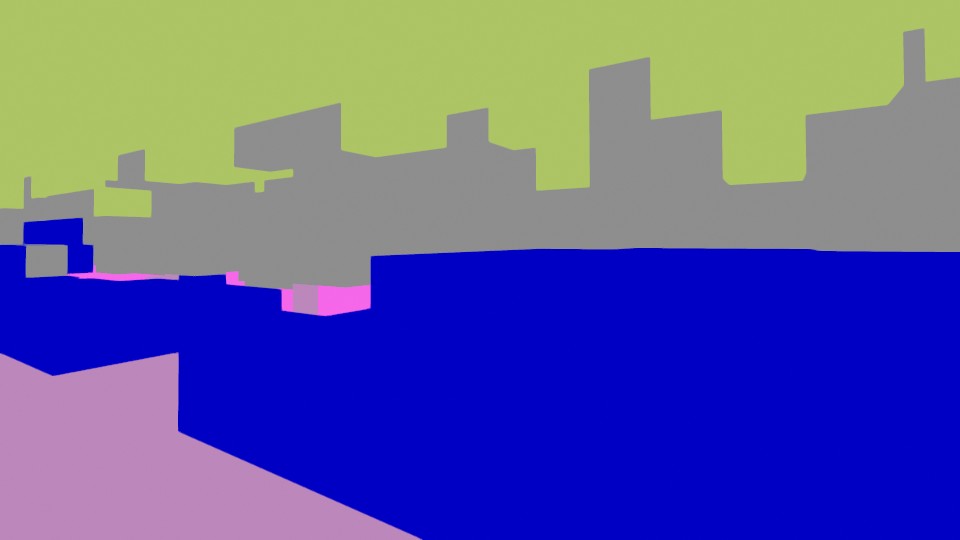}
{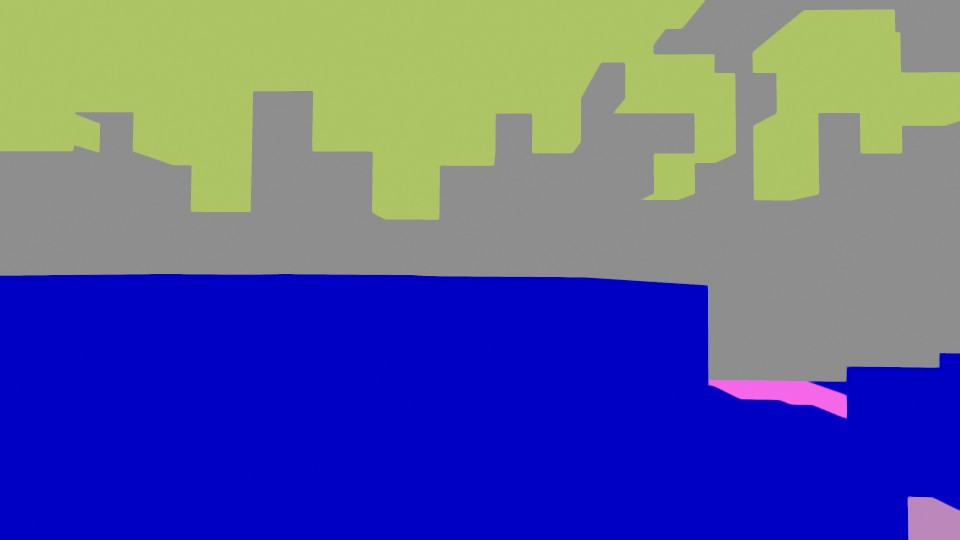}
\\[\tightwithin]
&
\tilefive
{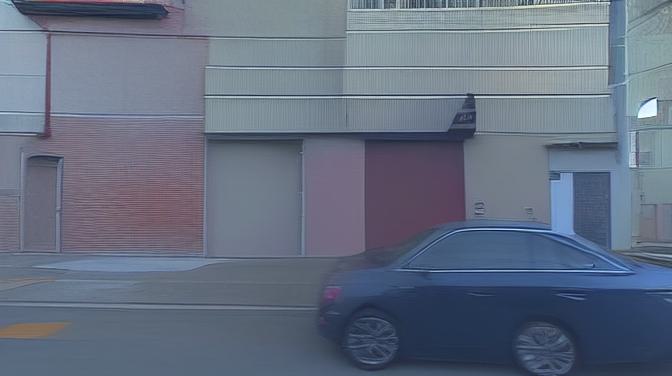}
{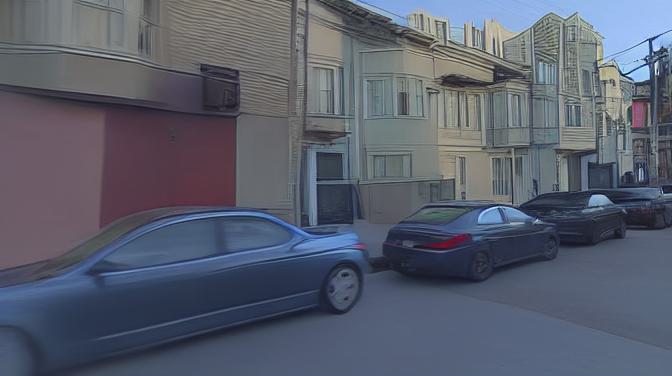}
{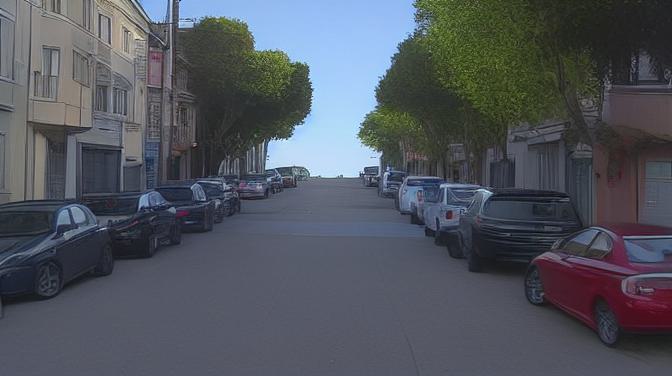}
{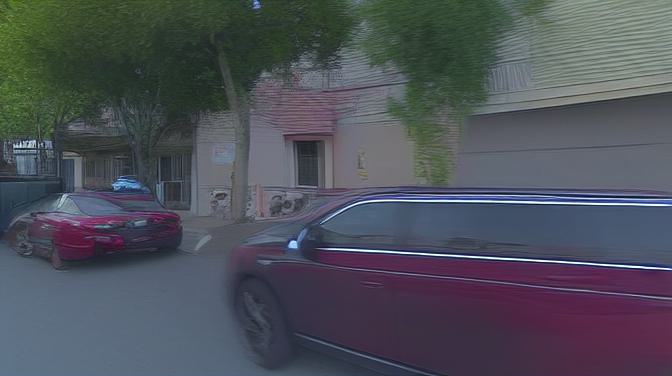}
{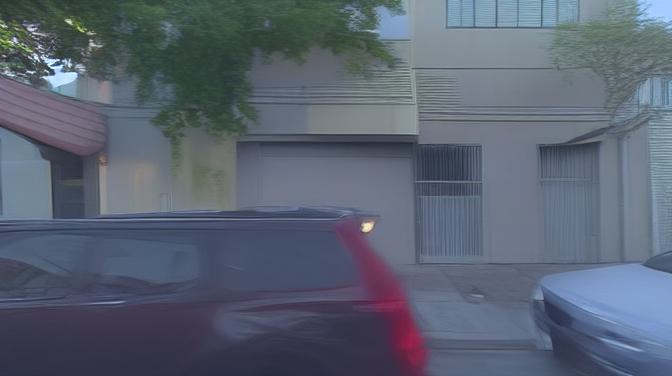}
\\[\pairgap]

% ===================== Scene 1 =====================
\vscenelabel{(e)} &
\tilefive
{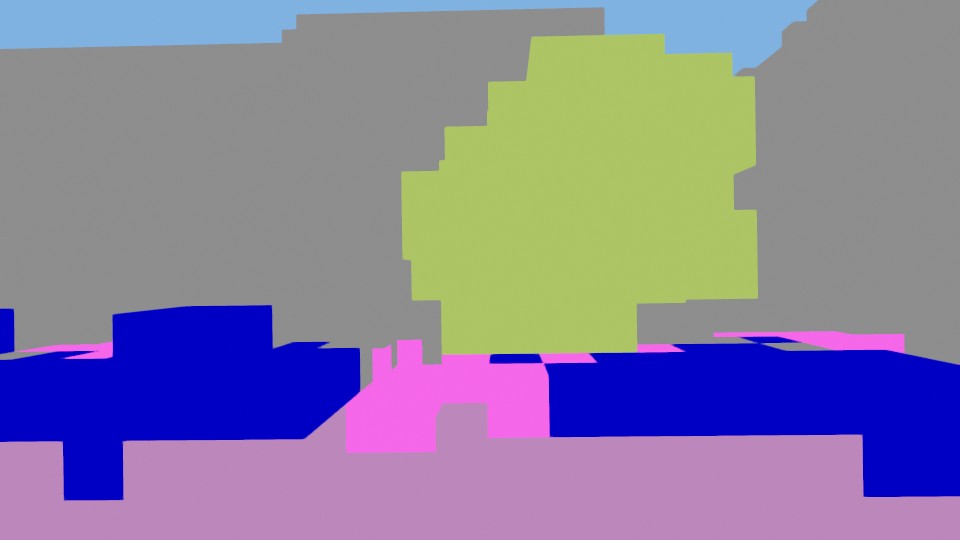}
{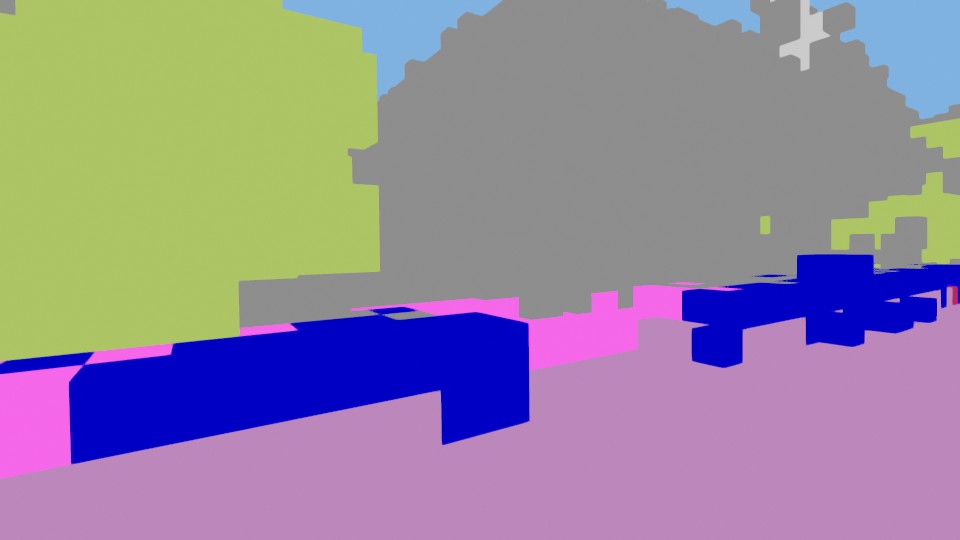}
{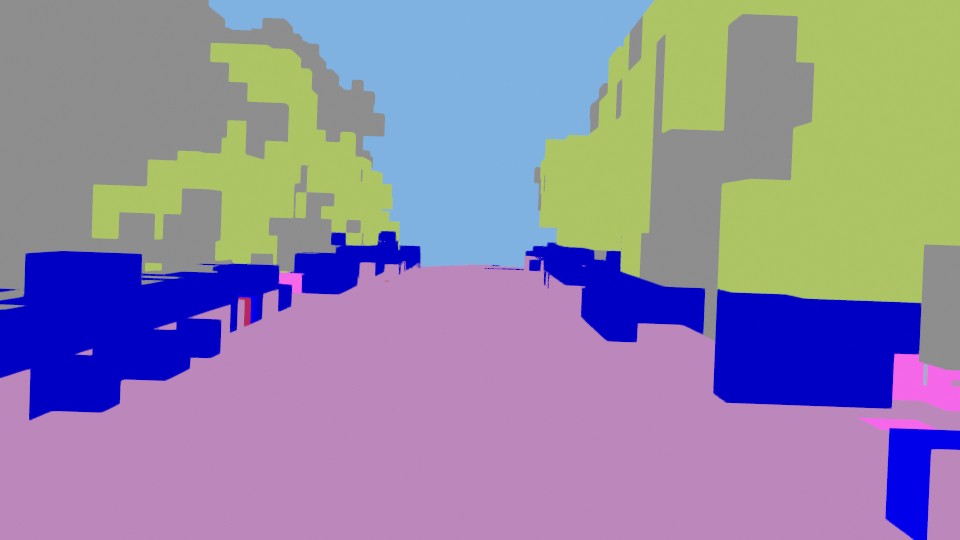}
{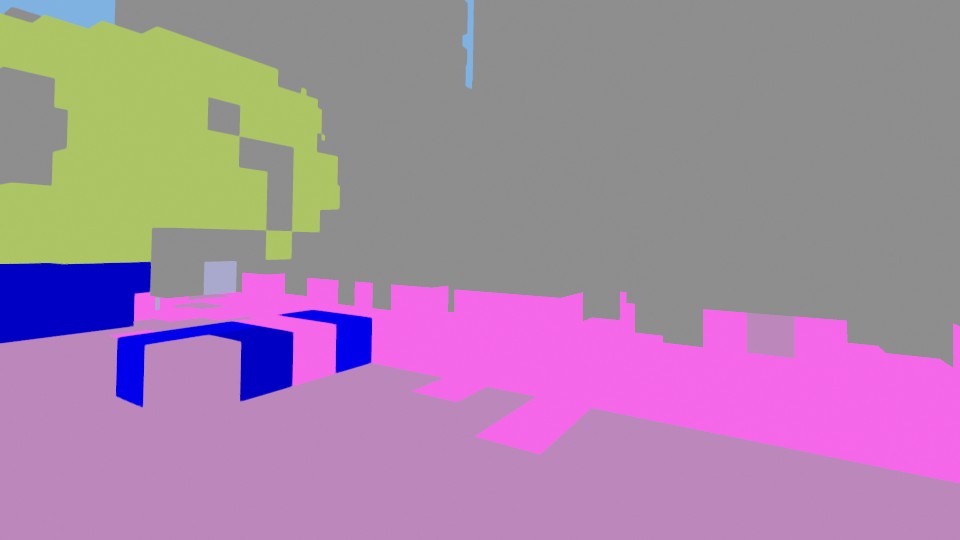}
{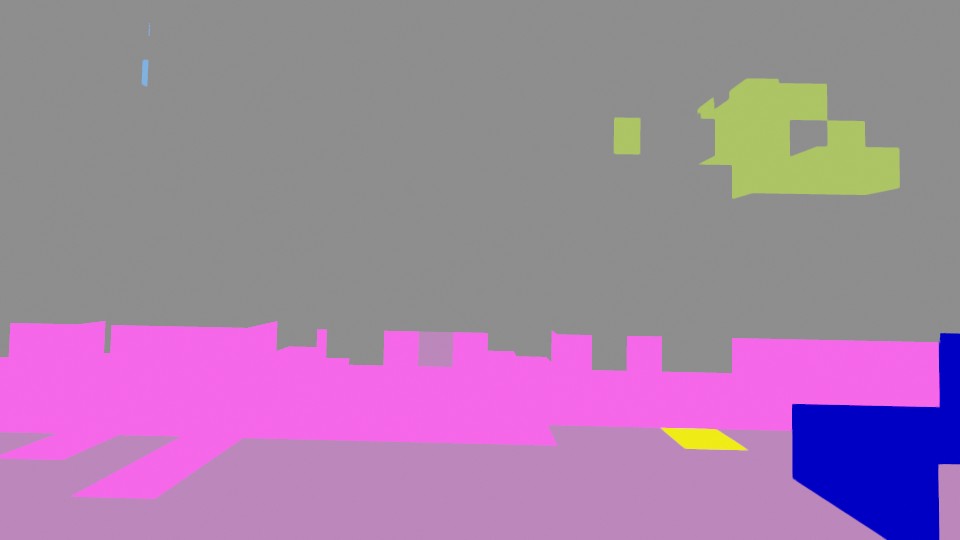}
\\[\tightwithin]
&
\tilefive
{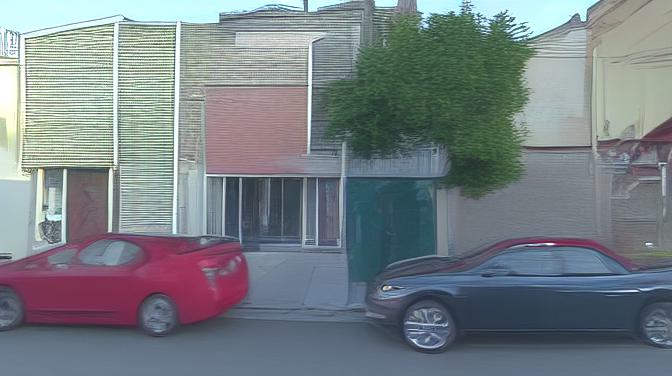}
{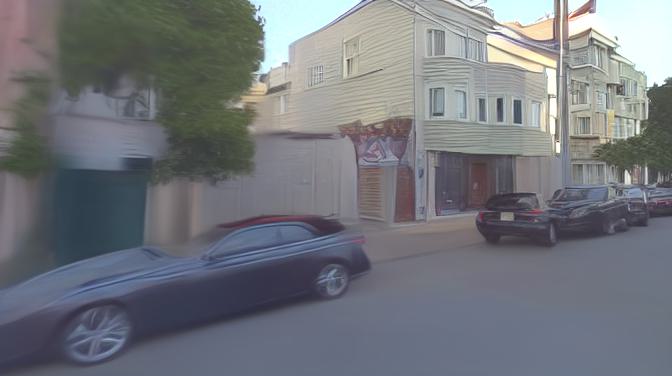}
{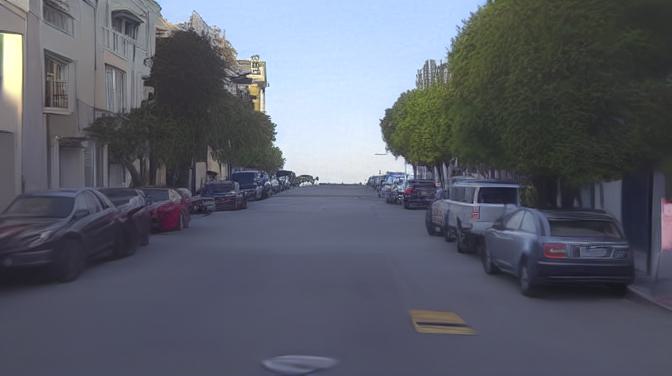}
{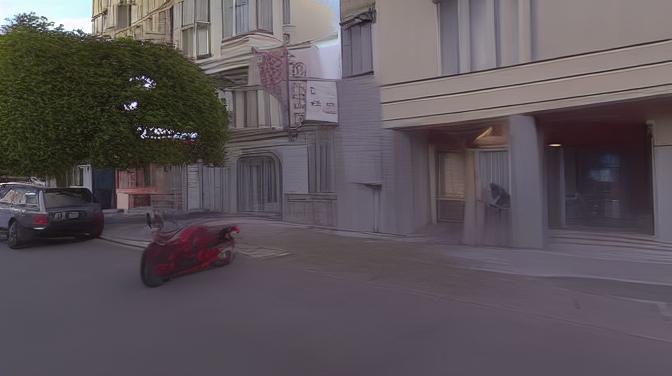}
{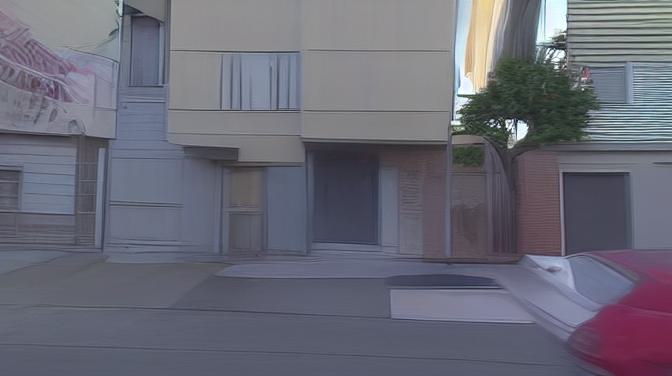}
\\[\pairgap]
% ===================== Scene 1 =====================
\vscenelabel{(f)} &
\tilefive
{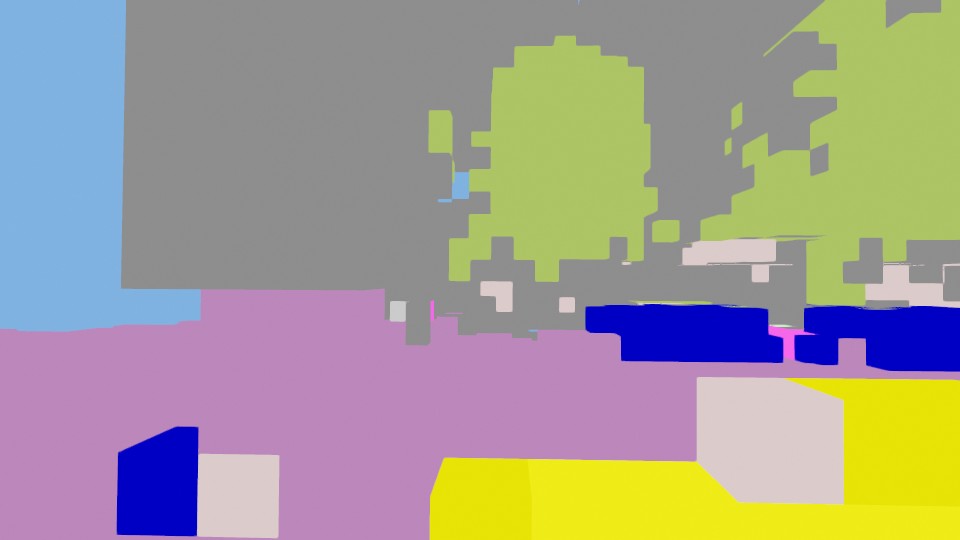}
{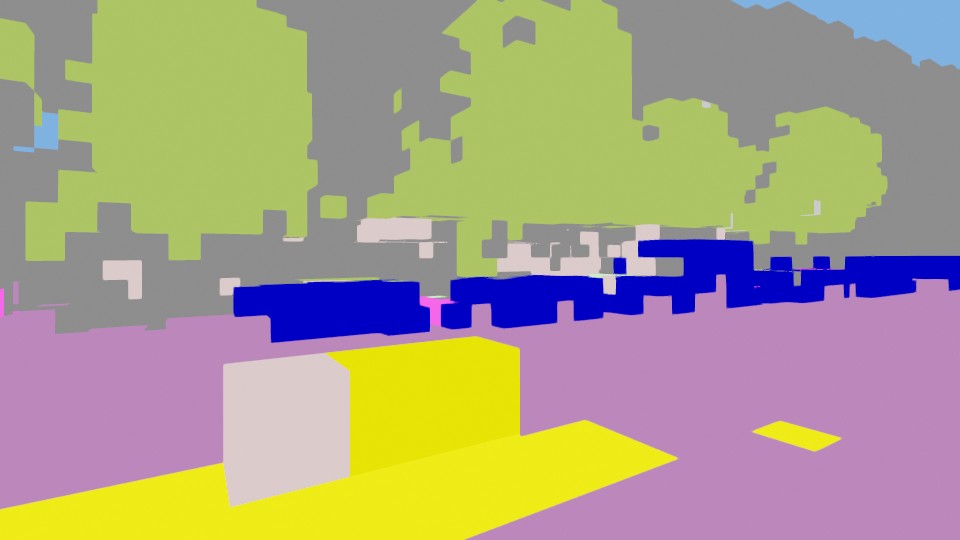}
{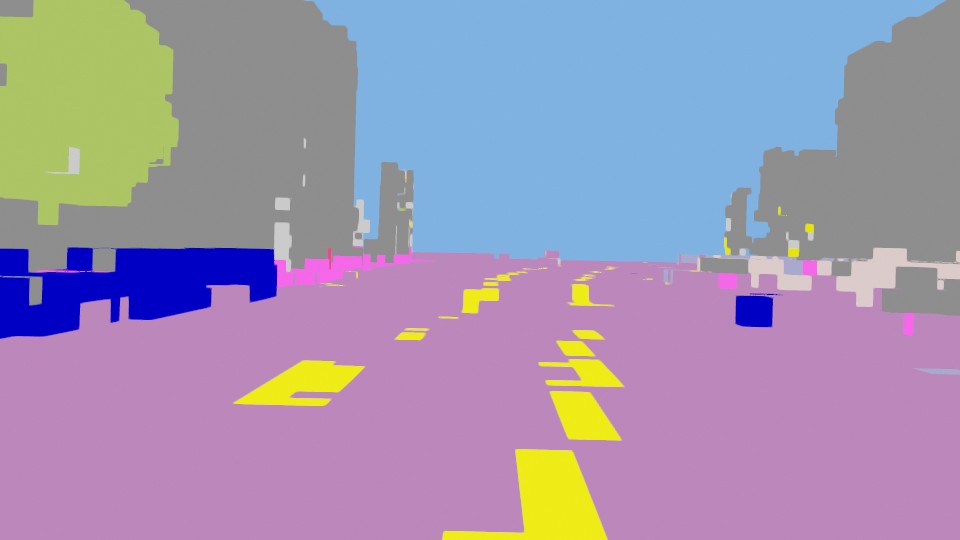}
{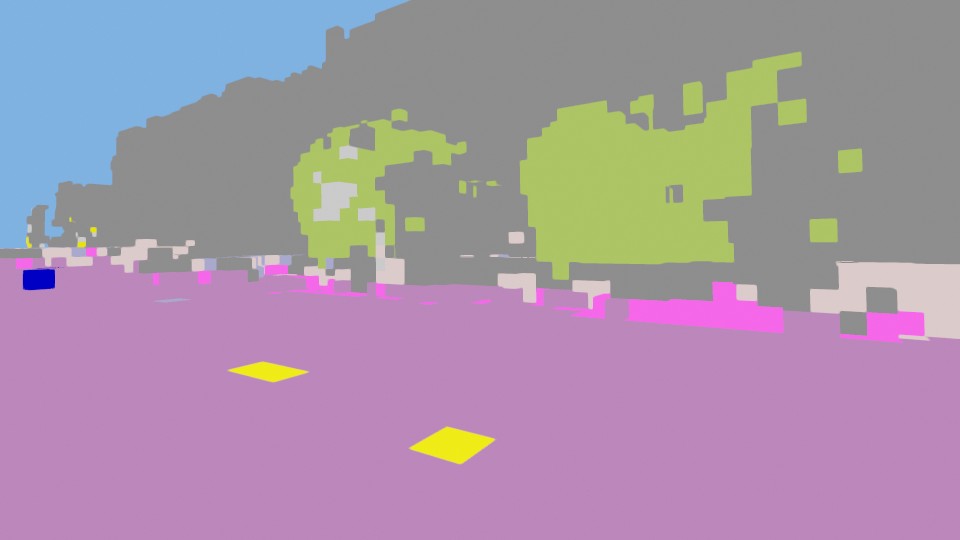}
{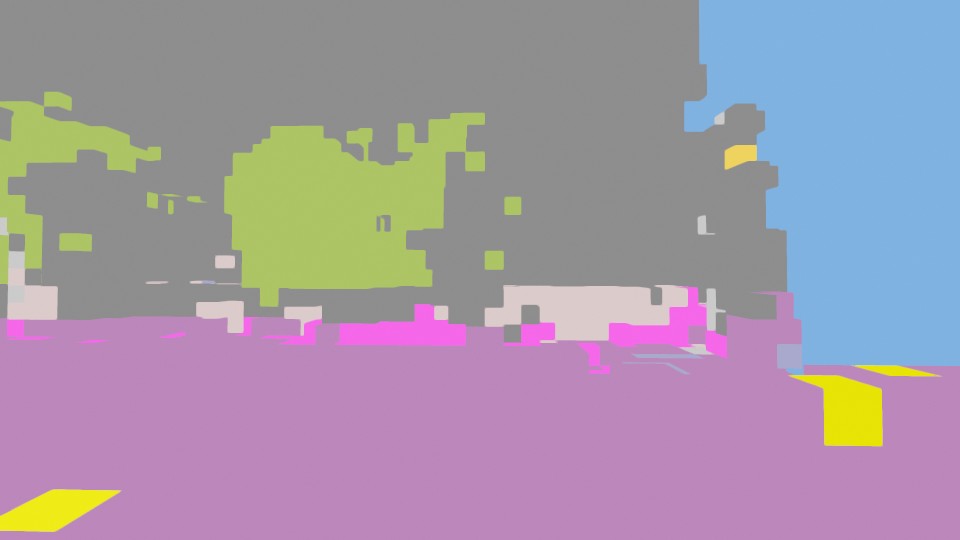}
\\[\tightwithin]
&
\tilefive
{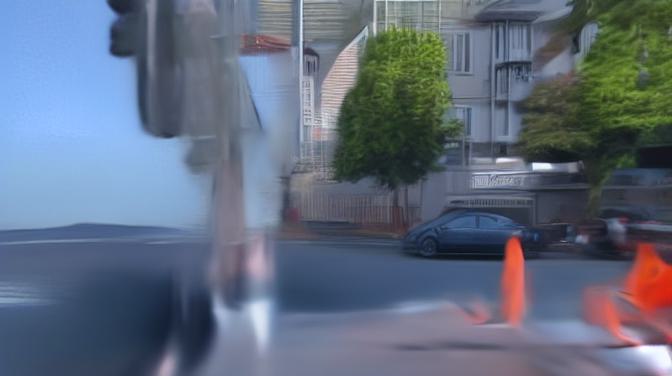}
{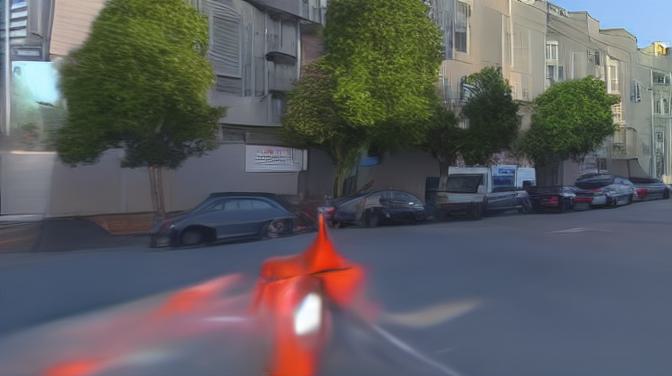}
{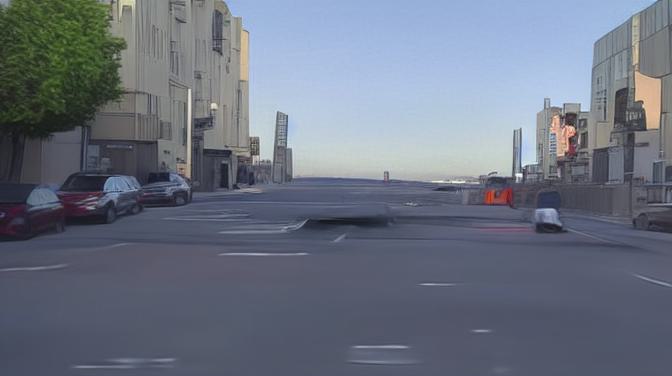}
{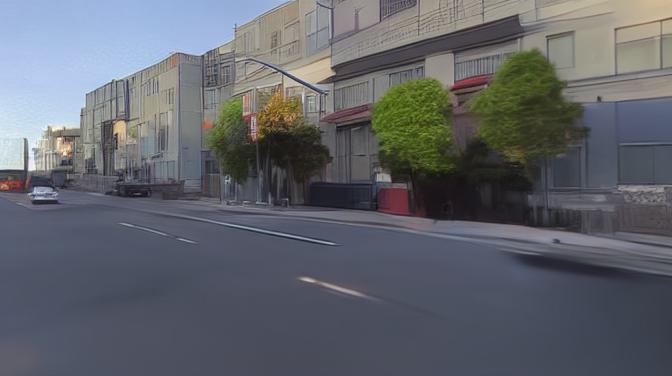}
{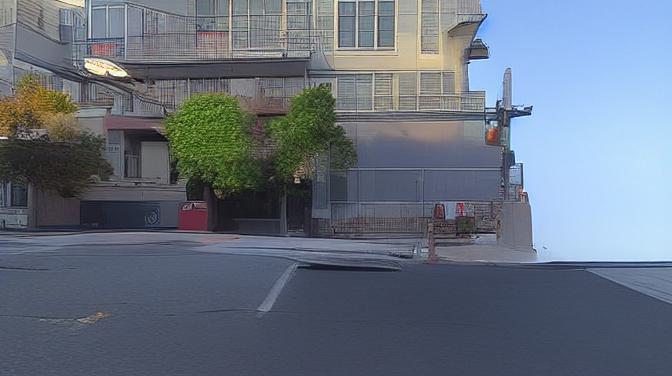}
\\[\pairgap]

\end{tabular}

\caption{\textbf{Additional Qualitative results on Pandaset~\cite{pandaset}.} We show six Pandaset scenes (a--f). For each scene, the \textbf{top} strip visualizes the semantic voxel rendering used for conditioning, and the \textbf{bottom} strip shows the corresponding generated scene from 5 camera views.}
\label{fig:results_pandaset}
\end{figure*}

\begin{figure*}[t]
\centering
\scriptsize
\setlength{\tabcolsep}{0pt}
\renewcommand{\arraystretch}{1.0}

% ---- knobs ----
\newcommand{\imgW}{0.31\textwidth}
\newcommand{\imgH}{0.20\textwidth}
\newcommand{\rowgap}{2pt}

\begin{tabular}{@{}c@{}ccc@{}}
& \multicolumn{1}{c}{\textbf{GEN3C\cite{gen3c}}}
& \multicolumn{1}{c}{\textbf{Infinicube\cite{infinicube}}}
& \multicolumn{1}{c@{}}{\textbf{Ours}} \\[2pt]

\textbf{(a)} &
\includegraphics[width=\imgW,height=\imgH]{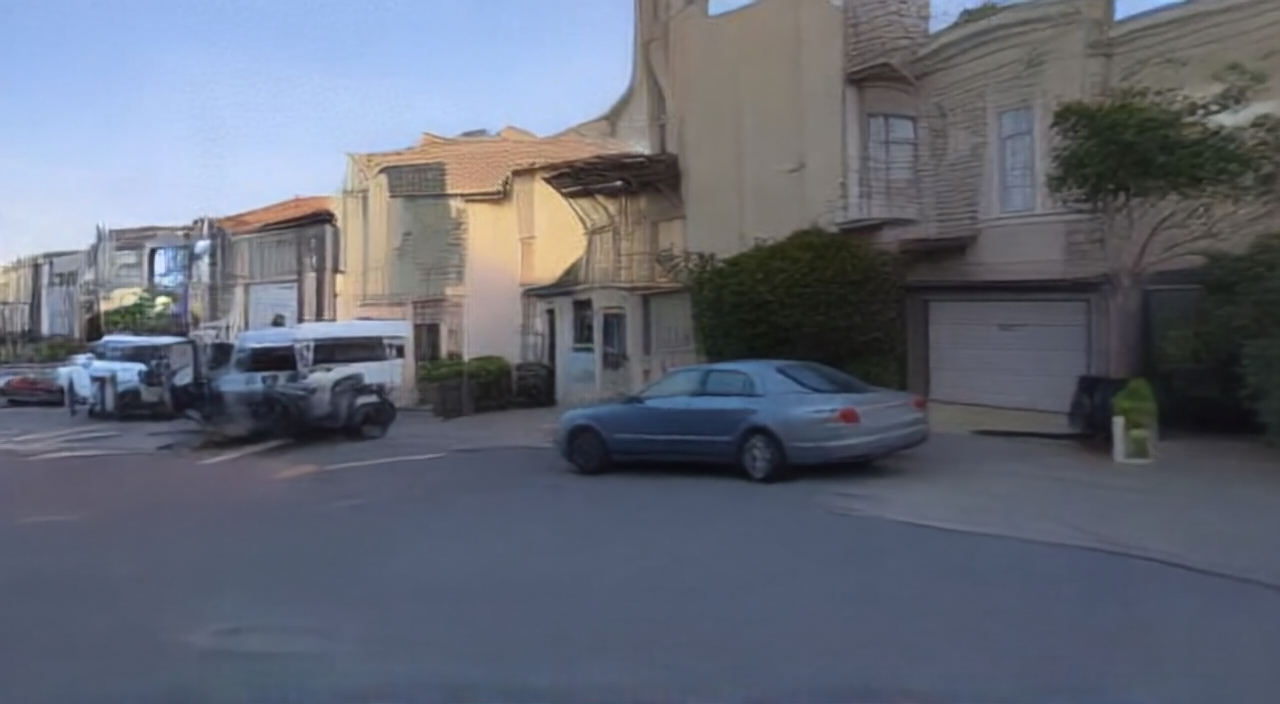} &
\includegraphics[width=\imgW,height=\imgH]{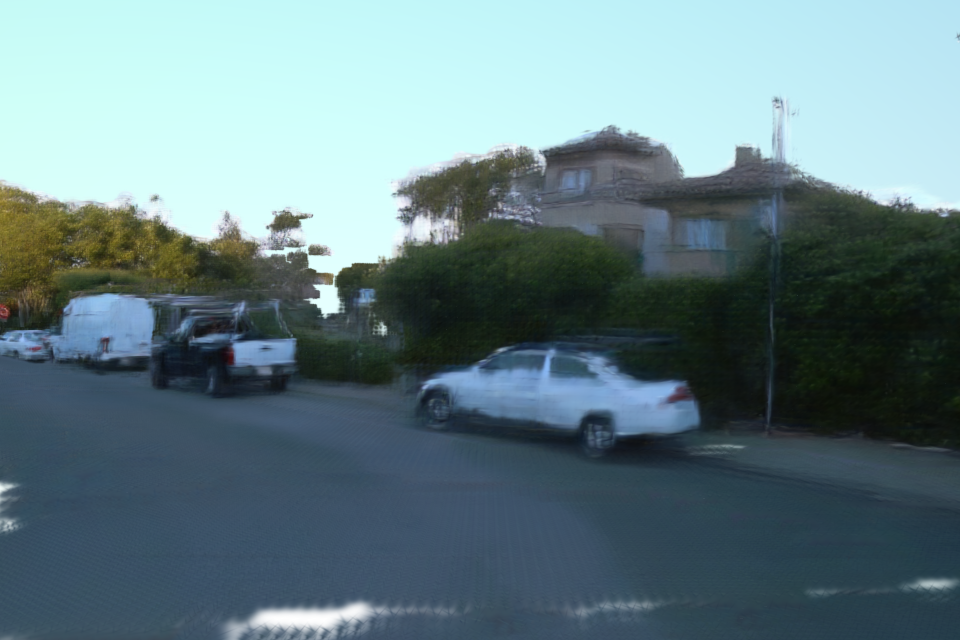} &
\includegraphics[width=\imgW,height=\imgH]{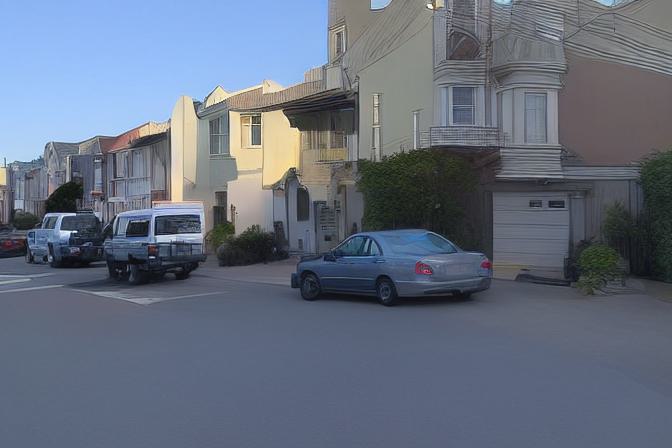}
\\[\rowgap]

\textbf{(b)} &
\includegraphics[width=\imgW,height=\imgH]{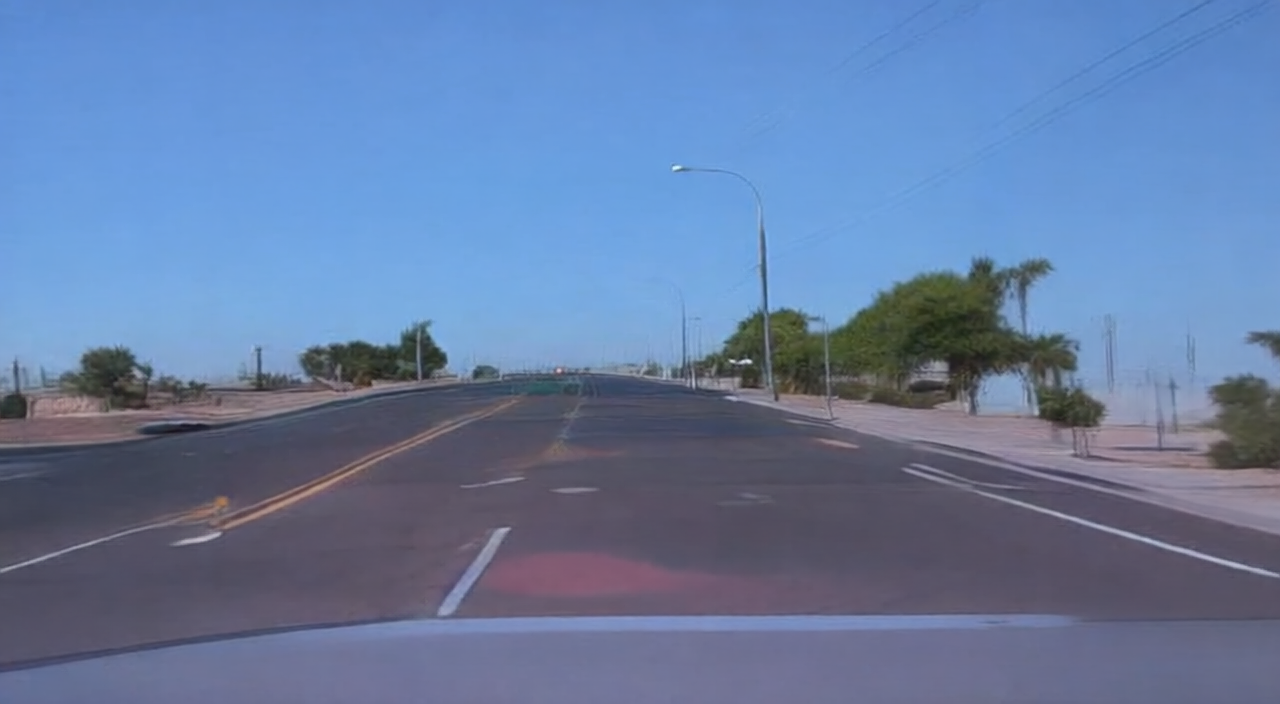} &
\includegraphics[width=\imgW,height=\imgH]{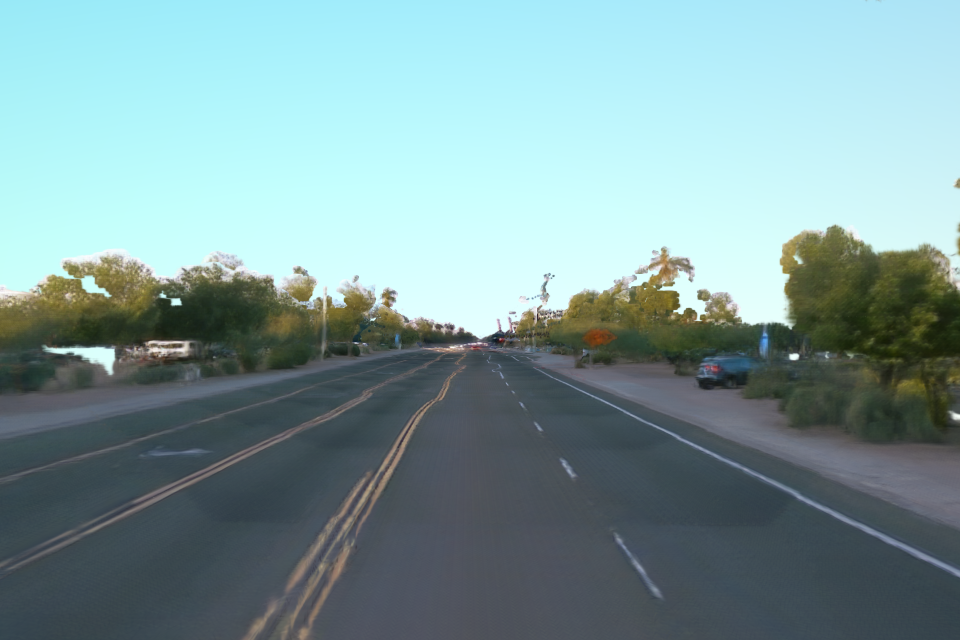} &
\includegraphics[width=\imgW,height=\imgH]{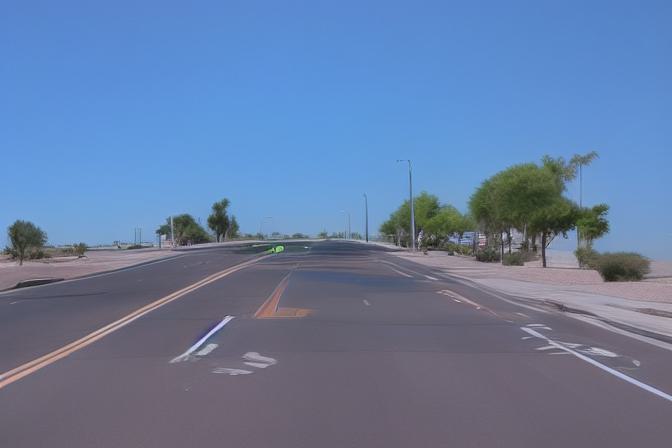}
\\[\rowgap]

\textbf{(c)} &
\includegraphics[width=\imgW,height=\imgH]{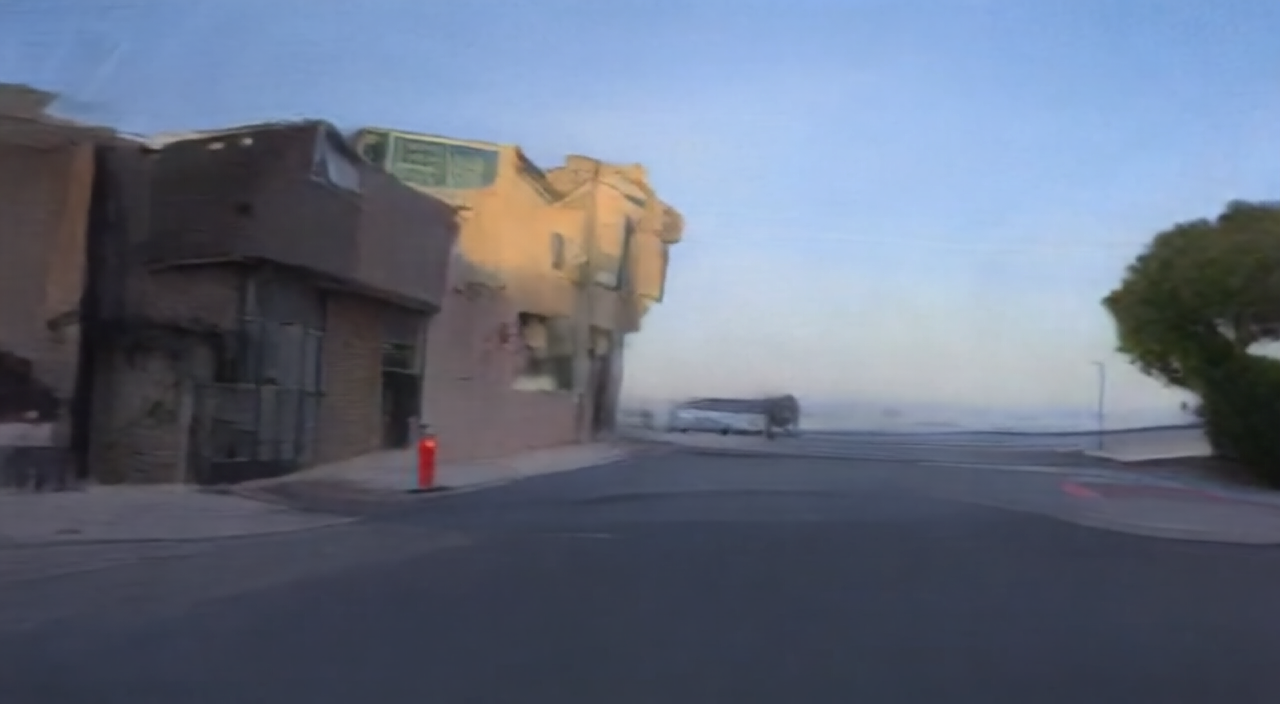} &
\includegraphics[width=\imgW,height=\imgH]{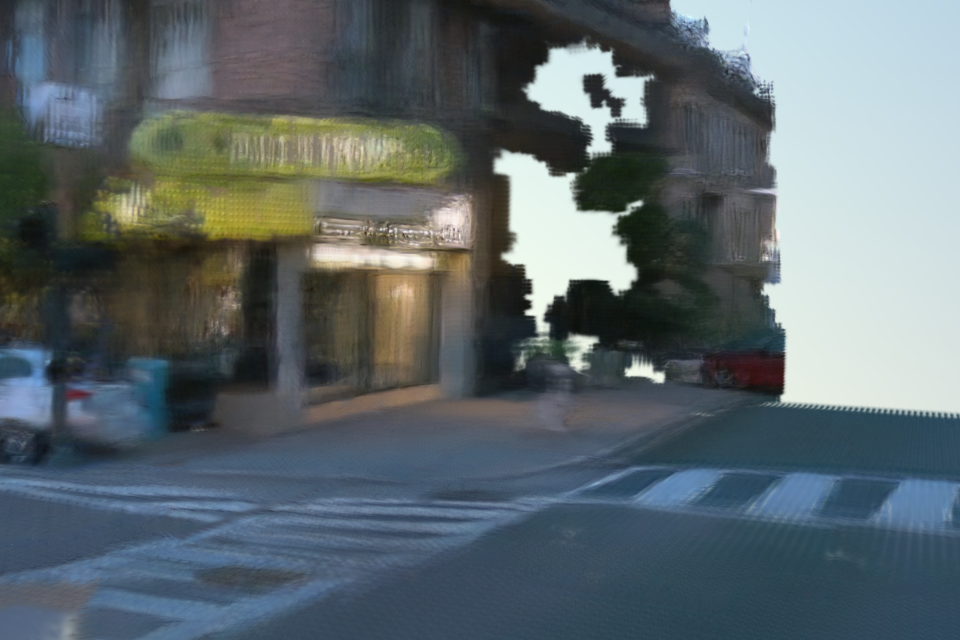} &
\includegraphics[width=\imgW,height=\imgH]{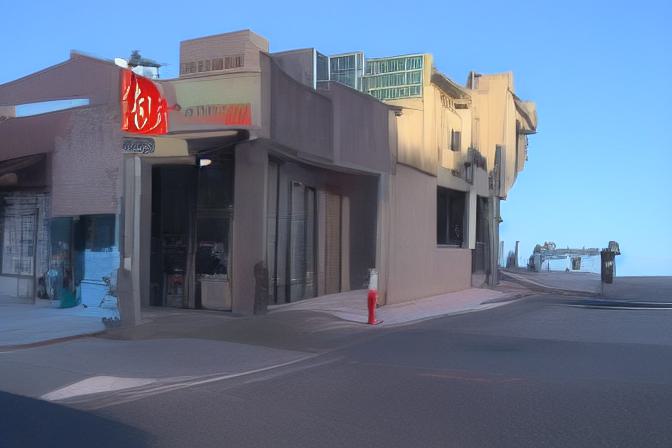}
\\[\rowgap]

\textbf{(d)} &
\includegraphics[width=\imgW,height=\imgH]{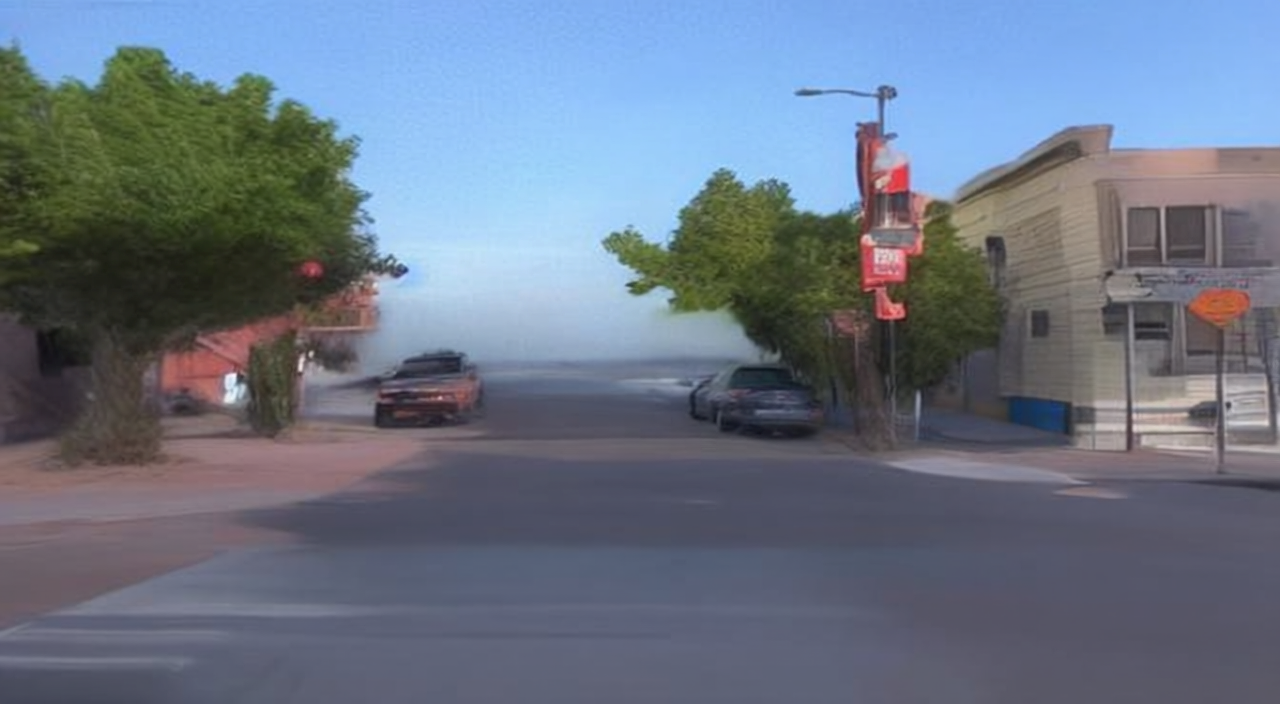} &
\includegraphics[width=\imgW,height=\imgH]{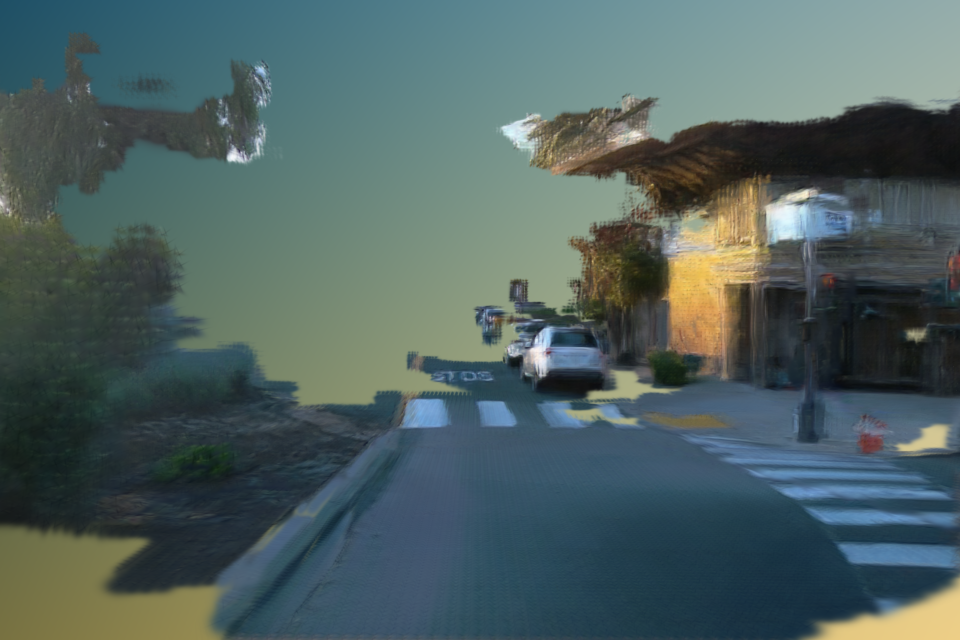} &
\includegraphics[width=\imgW,height=\imgH]{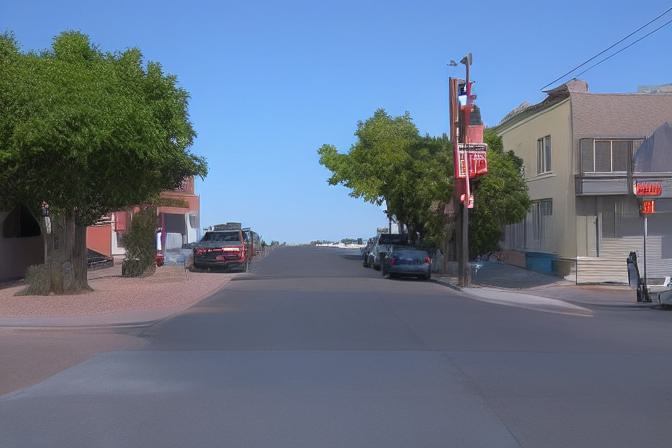}
\\[\rowgap]

\end{tabular}

\caption{\textbf{Comparison with baselines.} We show four scenes (a--d), each rendered from the front camera comparing our method to Infinicube\cite{infinicube} and GEN3C\cite{gen3c}}
\label{fig:baselines}
\end{figure*}

\begin{figure}[tbh]
\centering

\begin{minipage}[t]{0.32\linewidth}
    \centering
    \includegraphics[width=\linewidth,height=3.2cm,keepaspectratio]{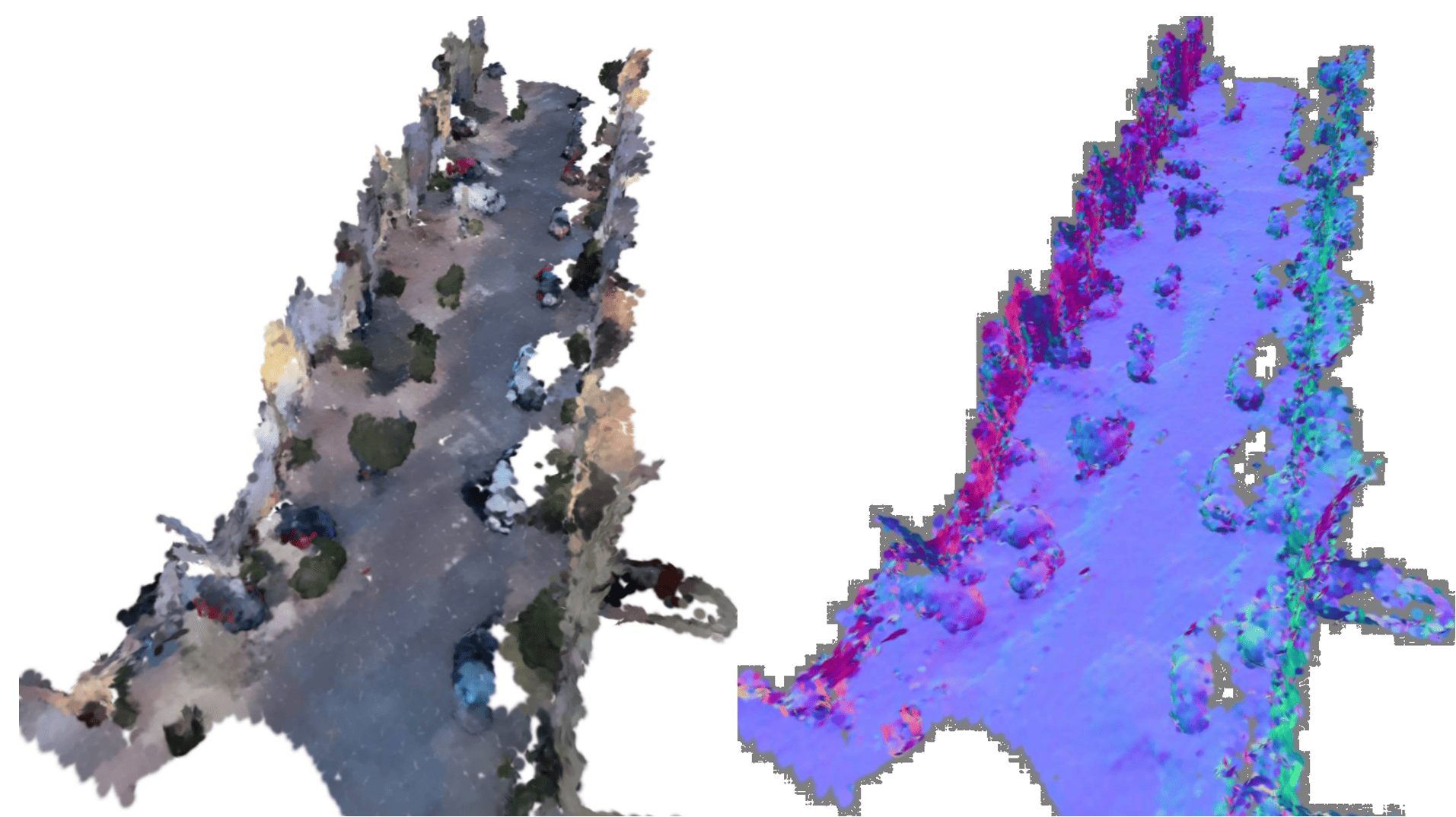}
\end{minipage}\hfill
\begin{minipage}[t]{0.32\linewidth}
    \centering
    \includegraphics[width=\linewidth,height=3.2cm,keepaspectratio]{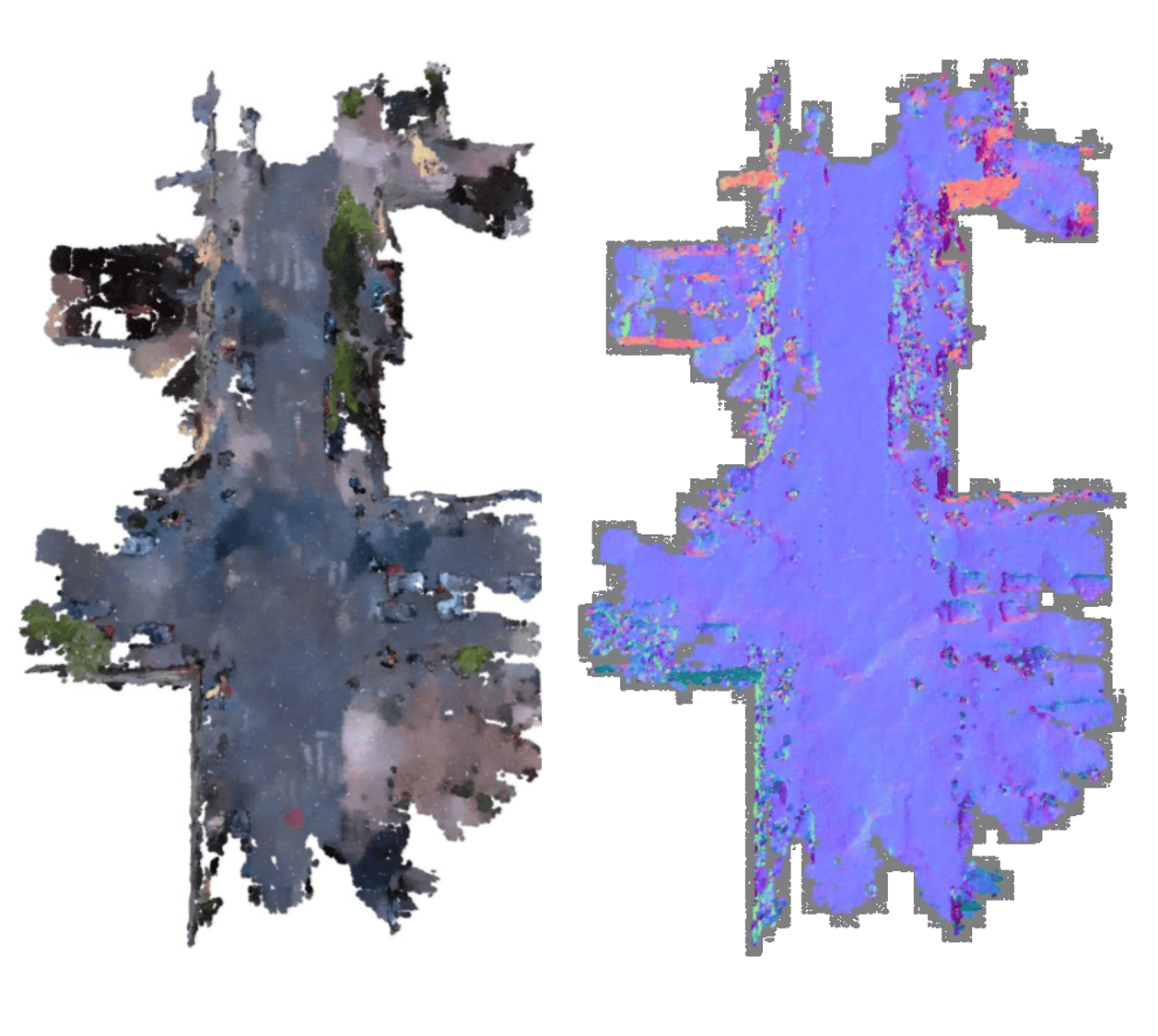}
\end{minipage}\hfill
\begin{minipage}[t]{0.32\linewidth}
    \centering
    \includegraphics[width=\linewidth,height=3.2cm,keepaspectratio]{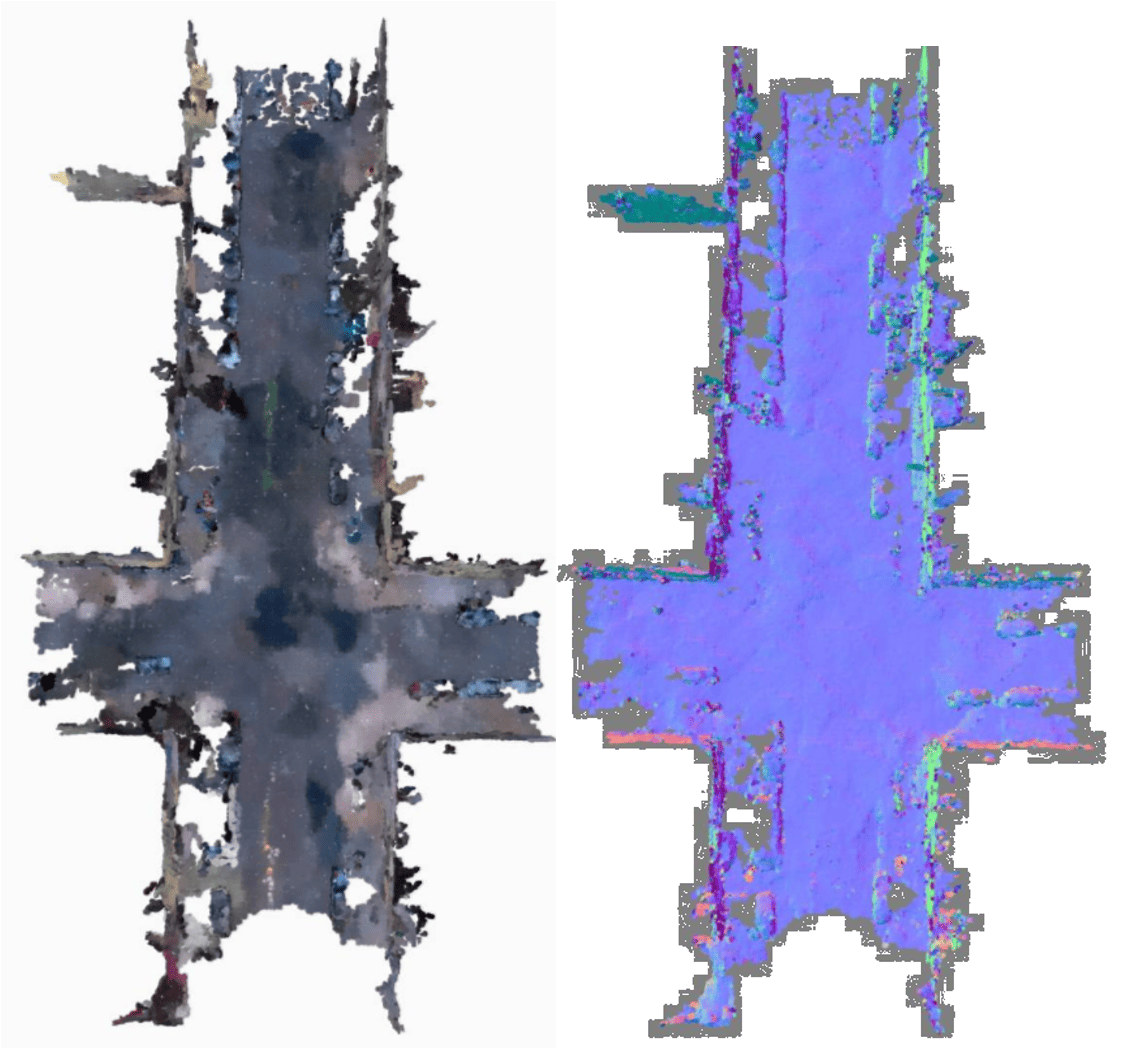}
\end{minipage}

\caption{\textbf{3D Buffers.} Rendering of our generated $\Sigma$-Voxelfield grids along with the generated scene geometry as normal map.}
\label{fig:3d_buffers}
\end{figure}

\begin{figure}[tbh]
\centering
\includegraphics[width=0.8\linewidth]{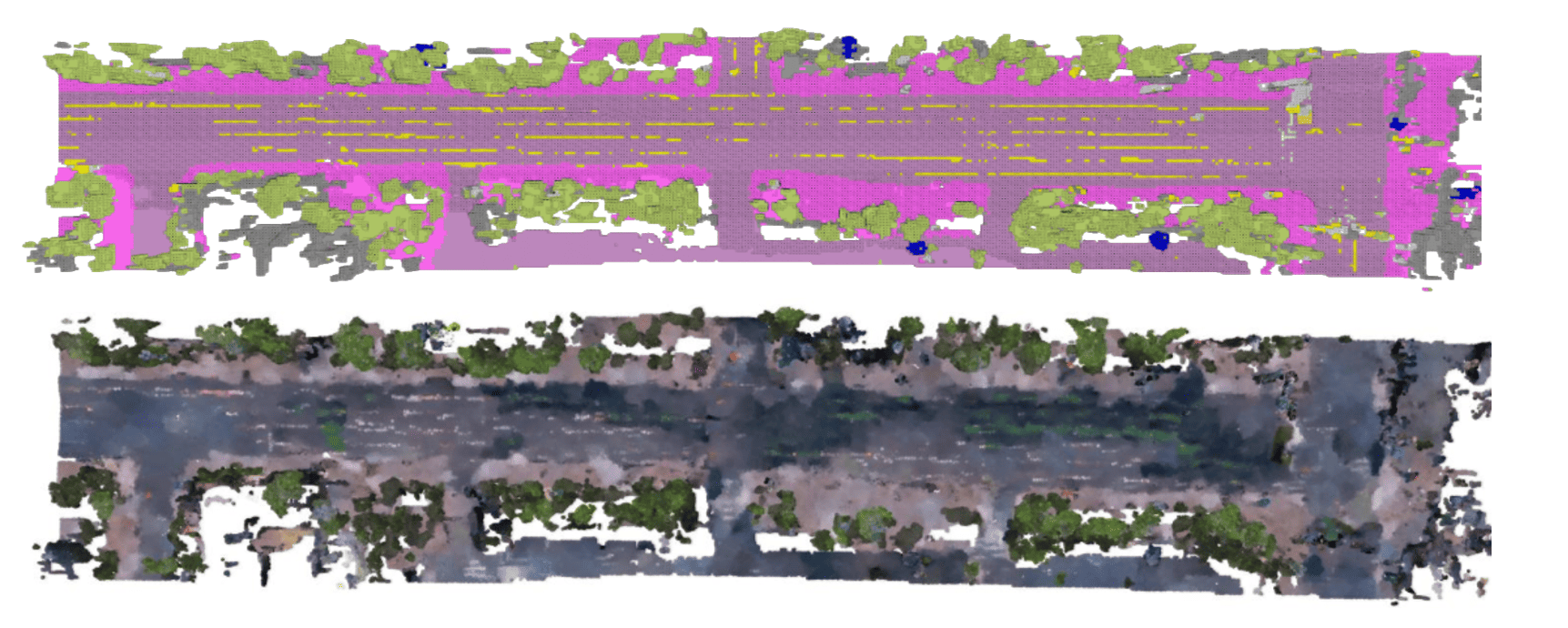}
\\
\includegraphics[width=0.8\linewidth]{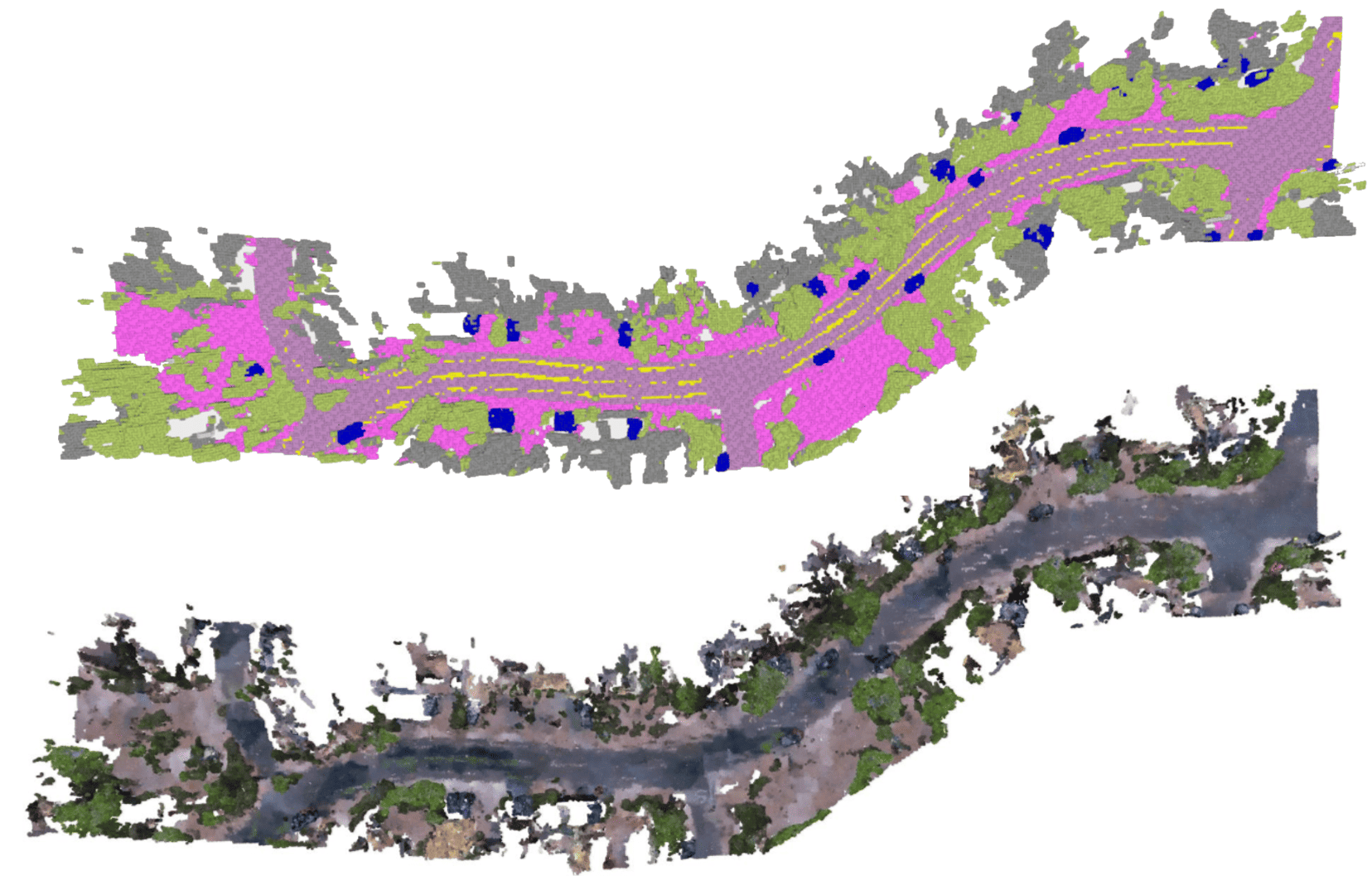}
\caption{\textbf{Large scene generation.} The semantic voxel grid exhibit the conditioning used to create a large driving sequence. We also show a top view rendering of the generated $\Sigma$-Voxfield grid.}
\label{fig:infinite}
\end{figure}

We show additional results of our method on WOD in \Cref{fig:results_waymo} and in \Cref{fig:results_pandaset} for Pandaset.
Additional comparisons with baselines are presented in \Cref{fig:baselines}. 
We also provide some examples of our generated 3D buffers, granting consistent geometry and appearance to our final rendering in \Cref{fig:3d_buffers}.
Our method enables large scene generation spanning over $100^2\,\mathrm{m}$. These large generations can be observed in \Cref{fig:infinite}.

%\paragraph{Supplementary videos.} We encourage the reader to refer to our supplementary videos to fully appreciate the results of our method, including the overall pipeline (\texttt{pipeline.mp4}) and the 3D consistency during car motion (\texttt{results.mp4}).
%\\
%We also provide a video highlighting the generative capabilities of our method (\texttt{variance.mp4}) and a comparison with baselines (\texttt{baselines.mp4}). 
%\\

%Additional videos demonstrate downstream applications, including scene inpainting (\texttt{inpaint.mp4}) and extreme camera motion (\texttt{trajectory.mp4}).

%\clearpage  % TODO FINAL: This \clearpage needs to be removed from both review and camera-ready versions.

\bibliographystyle{splncs04}
\bibliography{main}

\end{document}